\documentclass[10pt,twocolumn,letterpaper]{article}

\usepackage[pagenumbers]{cvpr}

\makeatletter\@namedef{ver@cuted.sty}{}\makeatother

\usepackage{cuted} %

\usepackage{currfile} %

\usepackage{caption} %

\usepackage{array}

\usepackage{multirow}

\usepackage{colortbl}
\definecolor{HGHeader}{HTML}{EFF4F1}
\definecolor{HGBody}{HTML}{EFF4F1}
\definecolor{HGBest}{HTML}{FFF4CC}
\definecolor{HGRule}{HTML}{A6ADB3}
\definecolor{HGFind1}{HTML}{FFF4CC}
\definecolor{HGFind2}{HTML}{E4F0FB}
\definecolor{HGFind3}{HTML}{F0E7F8}
\definecolor{HGFind4}{HTML}{FBE9DC}
\definecolor{HGFind5}{HTML}{F9E7EE}
\usepackage{hhline}
\definecolor{HGRuleStrong}{HTML}{69757C}
\usepackage{tikz}
\newcommand{\HGSwatch}[1]{\raisebox{-0.5pt}{\tikz\path[fill=#1,draw=#1!75!black,line width=0.25pt,rounded corners=1pt] (0,0) rectangle (6pt,6pt);}}

\usepackage{booktabs}
\usepackage{amsmath,amssymb}
\usepackage{tcolorbox}

\definecolor{cvprblue}{rgb}{0.21,0.49,0.74}
\usepackage[pagebackref,breaklinks,colorlinks,allcolors=cvprblue]{hyperref}

\def\confName{CVPR}
\def\confYear{2027}

\definecolor{FPBoxBack}{HTML}{EFF7FB}
\definecolor{FPBoxFrame}{HTML}{8FB3C9}
\definecolor{FPRule}{HTML}{45657F}
\definecolor{FPInk}{HTML}{1F2D3A}

\newcommand{\FPaff}[1]{\textsuperscript{#1}}
\newlength{\FPsealheight}

\makeatletter
\def\@maketitle{%
  \newpage
  \null
  \vspace*{-0.47in}%
  \noindent\begin{minipage}[b]{\textwidth}
    \raisebox{0.012in}{\includegraphics[height=0.265in]{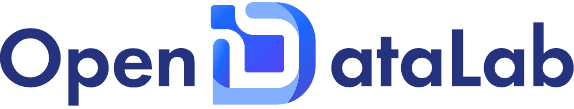}}%
    \hfill
    \raisebox{0.036in}{\includegraphics[height=0.214in]{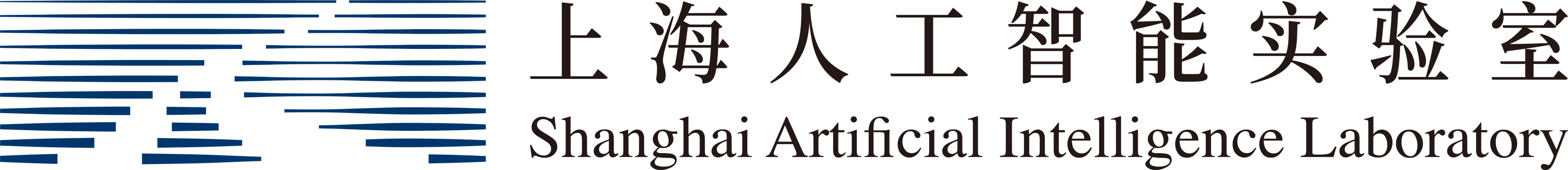}}%
  \end{minipage}\par
  \vskip 5pt
  {\color{FPRule}\hrule height 0.6pt}
  \vskip 0.21in
  \begin{center}
    {\Large\bfseries \@title\par}
    \vskip 12pt
    {\fontsize{11}{13.6}\selectfont
     Chenhao Dang\FPaff{1,2}\quad
     Jiahe Song\FPaff{1,2}\quad
     Tonghan Wang\FPaff{3}\quad
     Xiaoce Wang\FPaff{4,3}\quad
     Li Meng\FPaff{5,3}\quad
     Chuang Wang\FPaff{1,2}\par
     \vskip 5pt
     Hengrui Kang\FPaff{1,2}\quad
     Zhutao Lv\FPaff{6}\quad
     Junyan Ye\FPaff{6}\quad
     Conghui He\FPaff{2}\quad
     Weijia Li\FPaff{3,2,\dag}\par}
    \vskip 9pt
    {\normalsize
     \FPaff{1}Shanghai Jiao Tong University\qquad
     \FPaff{2}Shanghai Artificial Intelligence Laboratory\par
     \vskip 2pt
     \FPaff{3}Shenzhen International Graduate School, Tsinghua University\qquad
     \FPaff{4}Tsinghua University\par
     \vskip 2pt
     \FPaff{5}Sichuan Agricultural University\qquad
     \FPaff{6}Sun Yat-sen University\par}
    \vskip 7pt
    {\small
     \FPaff{\dag}Corresponding author\par
     \vskip 3pt
     {\tt dangchenhao@pjlab.org.cn}\qquad {\tt liweijia@sz.tsinghua.edu.cn}\par}
  \end{center}
  \vskip 0.07in
  \begin{tcolorbox}[width=\textwidth, colback=FPBoxBack, colframe=FPBoxFrame,
      boxrule=0.6pt, arc=7pt, left=15pt, right=15pt, top=8pt, bottom=9pt, boxsep=0pt]
    \noindent
    \raisebox{0.115in}{\fontsize{13.5}{16}\selectfont\bfseries Abstract}%
    \hfill
    \includegraphics[height=\FPsealheight]{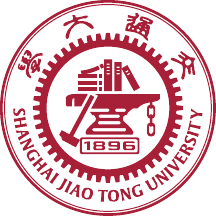}%
    \hspace{8pt}%
    \includegraphics[height=\FPsealheight]{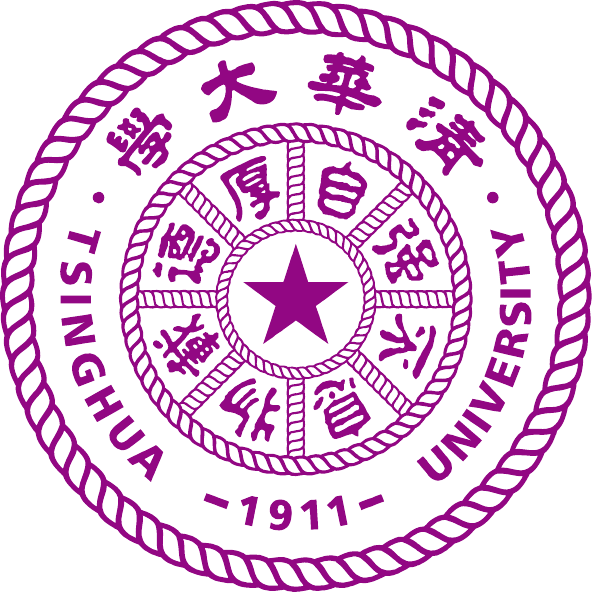}\par
    \vskip 5pt
    \color{FPInk}\normalsize\noindent\ignorespaces
    GUI tasks require repeated visual grounding, whose efficiency is key to completion time. Efficiency relative to humans therefore matters for agents' practical value and adoption. High-performing agents often pursue accuracy through extended reasoning, zooming, and cropping at substantial latency costs, yet this trade-off lacks systematic human-referenced evaluation. We introduce ScreenSpot-Efficiency (SS-E), the most comprehensive GUI grounding efficiency benchmark to date. Its 9,160 tasks from real screen recordings span 36 scenario types across diverse devices, covering the vast majority of everyday work and life scenarios. These tasks cover routine grounding for fast localization in common operations and time-sensitive grounding for timely responses to interface changes and other participants' actions. We annotate every task in SS-E and 4 influential grounding benchmarks with human expert completion times. We propose Human-referenced Grounding Accuracy–Time Efficiency (H-GATE), a metric jointly evaluating accuracy and latency against per-task references. Our evaluation spans 5 benchmarks, 7 GPU types, 14 model families, 38 models, and 87 inference configurations. State-of-the-art agents approach human expert efficiency on simple tasks, yet all evaluated agents lag substantially behind on tasks involving small targets, complex interfaces, and stringent response-time requirements, as in SS-E. Extended reasoning and zooming generally reduce efficiency. Moderate reductions in visual-token budgets improve efficiency for some models. However, some reduced budgets increase latency relative to the default setting, while aggressive reductions sharply lower grounding accuracy. The most efficient model varies across hardware, and mixture-of-experts models show strong efficiency potential. Overall, SS-E and H-GATE support human-referenced efficiency evaluation and guide the development of accurate, responsive GUI agents.%
\par
    \vskip 4pt
    \noindent{\bfseries\color{FPRule}Project Page:}\hspace{0.5em}\href{https://dang-ai.github.io/ScreenSpot-Efficiency}{\texttt{https://dang-ai.github.io/ScreenSpot-Efficiency}}
  \end{tcolorbox}
  \vskip 0.15in
}
\makeatother

\AtEndPreamble{%
  \crefname{appendix}{Appendix}{Appendices}%
  \Crefname{appendix}{Appendix}{Appendices}%
}

\title{Time to Ground: Can GUI Grounding Match Human Efficiency?}

\begin{document}
\maketitle
\begin{figure}
  \centering
  \includegraphics[width=\linewidth]{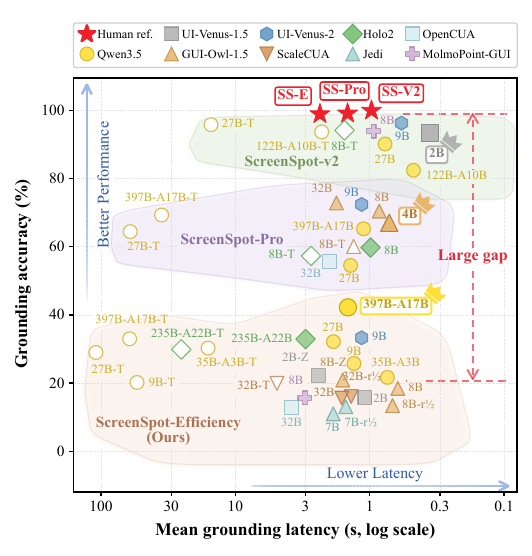}
  \caption{Grounding accuracy and mean latency of representative models and human experts on three ScreenSpot (SS) benchmarks, with model latency measured on H200 GPUs. The suffixes \texttt{-T}, \texttt{-Z}, and \texttt{-r}\textonehalf\ denote thinking mode, local zooming, and half the default visual budget, respectively. Crowns mark the model configuration with the highest H-GATE on each benchmark.}
  \label{fig:teaser}
\end{figure}

\section{Introduction}
\label{sec:intro}

Visual grounding is an atomic capability of GUI agents, repeatedly invoked to map action intent to interface targets~\cite{seeclick,osatlas,agents2}.
Latency accumulates across these operations, while errors can trigger retries and prolong task completion~\cite{osworldhuman}.
When users must wait, supervise, or interact frequently, efficiency relative to humans therefore matters for agents' practical value and adoption, alongside task quality and reliability~\cite{agentsmatter,production,osworldhuman,cuaspeedrun}.

High-performing agents often pursue grounding accuracy through extended reasoning and local zooming or cropping~\cite{visualtts,costdynamic}, which can substantially increase latency, as illustrated in \cref{fig:teaser}.
With this additional time cost, higher accuracy does not necessarily mean higher grounding efficiency.
Prior work explores low-latency grounding~\cite{tinyclick,goclick,innerzoom}, analyzes accuracy--latency trade-offs~\cite{guidiffusion,igg}, and compares agent and human completion times~\cite{componentbench,osworldhuman}.
Yet systematic human-referenced evaluation across diverse grounding tasks, models, and hardware remains lacking.

\begin{figure*}
  \centering
  \includegraphics[width=\linewidth]{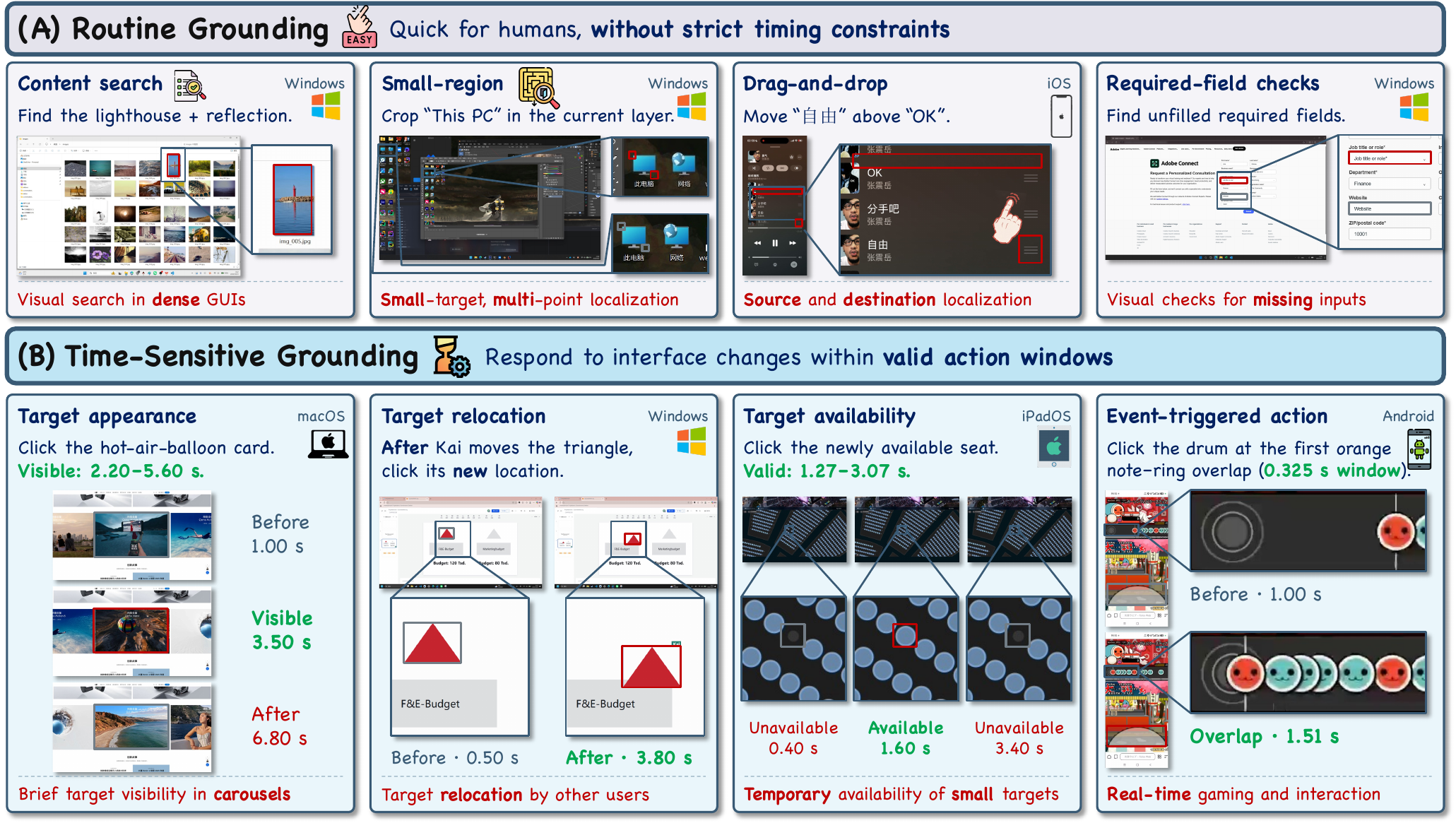}
  \caption{\textbf{Overview of ScreenSpot-Efficiency.} (a) Routine grounding covers common operations that humans can complete quickly without strict timing constraints. (b) Time-sensitive grounding requires responses to interface changes and other participants’ actions within valid action windows.}
  \label{fig:sse_overview}
\end{figure*}

To address this gap, we introduce \textbf{S}creen\textbf{S}pot-\textbf{E}fficiency \textbf{(SS-E)}, the most comprehensive GUI grounding efficiency benchmark to date.
Its 9,160 tasks from real screen recordings span 36 scenario types across diverse devices, covering the vast majority of everyday work and life scenarios.
As shown in \cref{fig:sse_overview}, these tasks cover \textit{routine} grounding for fast localization in common operations without strict timing constraints and \textit{time-sensitive} grounding for timely responses to interface changes and other participants' actions.
Routine tasks include dense visual search, small-target localization, and matching multiple action locations.
In time-sensitive settings such as carousel and collaborative interfaces, targets can appear, move, or become unavailable during inference, so correct localization must also be timely.

\begin{table*}
\centering
\caption{\textbf{Comparison of GUI grounding benchmarks.} Lang. denotes language. VisualWebBench, UI-Vision, and MMBench-GUI entries refer to their grounding subsets.}
\label{tab:benchmark_comparison}
\setlength{\tabcolsep}{2.5pt}
\renewcommand{\arraystretch}{1.13}
\footnotesize
\begin{tabular*}{\textwidth}{@{\extracolsep{\fill}}>{\raggedright\arraybackslash}p{3.0cm}r c c c c c c l c c@{}}
\toprule
Benchmark & \#Tasks & Real-world & \#UI Lang. & \#Platforms & \#Software & PC & Mobile & Task focus & \#Human-timed & Latency \\
\midrule
ScreenSpot-v2~\cite{seeclick,osatlas} & 1,272 & \textcolor{green!50!black}{$\checkmark$} & 2 & 5 & 165 & \textcolor{green!50!black}{$\checkmark$} & \textcolor{green!50!black}{$\checkmark$} & Text/icon & \textcolor{red}{$\times$} & \textcolor{red}{$\times$} \\
ScreenSpot-Pro~\cite{screenspotpro} & 1,581 & \textcolor{green!50!black}{$\checkmark$} & 2 & 3 & 23 & \textcolor{green!50!black}{$\checkmark$} & \textcolor{red}{$\times$} & Professional UI & \textcolor{red}{$\times$} & \textcolor{red}{$\times$} \\
VisualWebBench~\cite{visualwebbench} & 516 & \textcolor{green!50!black}{$\checkmark$} & 2 & 1 & 136 & \textcolor{green!50!black}{$\checkmark$} & \textcolor{red}{$\times$} & Element/action & \textcolor{red}{$\times$} & \textcolor{red}{$\times$} \\
OSWorld-G~\cite{osworldg} & 564 & \textcolor{green!50!black}{$\checkmark$} & 1 & 1 & 40 & \textcolor{green!50!black}{$\checkmark$} & \textcolor{red}{$\times$} & Complex grounding & \textcolor{red}{$\times$} & \textcolor{red}{$\times$} \\
UI-Vision~\cite{uivision} & 5,479 & \textcolor{green!50!black}{$\checkmark$} & 3 & 3 & 83 & \textcolor{green!50!black}{$\checkmark$} & \textcolor{red}{$\times$} & Element grounding & \textcolor{red}{$\times$} & \textcolor{green!50!black}{$\checkmark$} \\
MMBench-GUI~\cite{mmbenchgui} & 3,574 & \textcolor{green!50!black}{$\checkmark$} & 2 & 5 & 138 & \textcolor{green!50!black}{$\checkmark$} & \textcolor{green!50!black}{$\checkmark$} & Basic/advanced & \textcolor{red}{$\times$} & \textcolor{red}{$\times$} \\
Showdown~\cite{showdown} & 557 & \textcolor{green!50!black}{$\checkmark$} & 1 & 1 & 132 & \textcolor{green!50!black}{$\checkmark$} & \textcolor{red}{$\times$} & Single click & \textcolor{red}{$\times$} & \textcolor{green!50!black}{$\checkmark$} \\
IGG~\cite{igg} & 601 & \textcolor{red}{$\times$} & 1 & 3 & 16 & \textcolor{green!50!black}{$\checkmark$} & \textcolor{red}{$\times$} & Reveal \& click & 21 & \textcolor{green!50!black}{$\checkmark$} \\
\midrule
\textbf{ScreenSpot-Efficiency} & \textbf{9,160} & \textcolor{green!50!black}{$\checkmark$} & \textbf{5} & \textbf{6} & \textbf{238} & \textcolor{green!50!black}{$\checkmark$} & \textcolor{green!50!black}{$\checkmark$} & Efficiency-sensitive & \textbf{9,160} & \textcolor{green!50!black}{$\checkmark$} \\
\bottomrule
\end{tabular*}
\end{table*}

Across these tasks, overall accuracy and mean latency obscure the pairing of correctness and time cost at the sample level.
These aggregates alone do not resolve comparisons between slower, more accurate models and faster, less accurate ones.
Human completion times also vary across tasks, so identical model latency can imply different time gains or losses relative to experts.
We therefore annotate every task in SS-E and four influential grounding benchmarks with the mean completion time of 11 human experts.
Using these annotations, we propose \textit{\textbf{H}uman-referenced \textbf{G}rounding \textbf{A}ccuracy--\textbf{T}ime \textbf{E}fficiency \textbf{(H-GATE)}}, which jointly scores correctness and latency against per-task human references and averages over the full task set, including failures.

Using vLLM~\cite{vllm}, we evaluate GUI grounding efficiency across five benchmarks, 7 GPU types, 14 model families, 38 models, and 87 inference configurations.
Our results reveal persistent human--agent gaps on difficult tasks.
On SS-E, extended reasoning and local zooming generally reduce efficiency, while visual-token reduction yields efficiency peaks at intermediate budgets for some models but can also increase latency.
Metric diagnostics establish that the resulting model rankings depend on per-task human references and failed-attempt latency.
Under this evaluation, the most efficient model varies with hardware and task demands.
Within dense model families, increasing size improves accuracy more consistently than efficiency, whereas mixture-of-experts models show strong efficiency potential.

Our main contributions are as follows:
\begin{itemize}
  \item We introduce SS-E, the most comprehensive GUI grounding efficiency benchmark to date, with 9,160 tasks from real screen recordings spanning 36 everyday scenario types and both routine and time-sensitive grounding.
  \item We establish a human-referenced efficiency evaluation framework, providing per-task mean completion times from 11 experts across SS-E and four existing benchmarks. Through the H-GATE metric, this framework jointly assesses correctness and latency against these baselines across all attempts.
  \item We conduct the most extensive GUI grounding efficiency evaluation to date across five benchmarks and validate H-GATE's diagnostic value.
  Our experiments yield four main findings on human--agent gaps, inference trade-offs, task-level efficiency information, and hardware-dependent model selection and mixture-of-experts potential, alongside several additional findings.
  This evidence supports tracking progress toward human efficiency and designing accurate, responsive GUI agents.
\end{itemize}

\section{Related Work}
\label{sec:related_work}

\subsection{GUI Grounding Benchmarks}
ScreenSpot and ScreenSpot-v2 establish cross-platform screenshot grounding~\cite{seeclick,osatlas}, while ScreenSpot-Pro and UI-Vision emphasize professional and desktop interfaces~\cite{screenspotpro,uivision}.
Across desktop, web, and mobile interfaces, AgentStudio's GroundUI provides a public 1,000-task evaluation split~\cite{groundui}, while UI-I2E-Bench tests explicit and implicit instructions on 1,477 annotated samples~\cite{uii2e}.
WinDeskGround evaluates controlled multi-window configurations constructed from real application windows~\cite{windeskground}.
Action coverage expands through ScreenDrag's 5,333 text-selection tasks~\cite{screendrag} and CUActSpot's click, drag, and draw tasks across GUI, text, table, canvas, and natural-image targets~\cite{cuactspot}.
PointerBench probes fine-grained grounding with 1,500 synthetic spreadsheet, text, and professional-interface tasks~\cite{pointerbench}, while VenusBench-GD covers element, spatial, visual, reasoning, functional, and refusal grounding across desktop, web, and mobile interfaces~\cite{venusgd}.
Phi-Ground additionally introduces the proprietary 1,934-task Gold set and 211-task Gold-S subset, neither released as an independent benchmark~\cite{phiground}.

Beyond visible screenshot targets, IGG evaluates targets revealed through scrolling or waiting, reporting accuracy and completion time across target revelation and clicking~\cite{igg}.
SS-E broadens this evaluation to routine and time-sensitive grounding in real screen recordings, pairing joint spatial and temporal correctness with per-task expert completion times.
Its 36 everyday work and life scenario types span PC and mobile devices, multiple operating systems, web interfaces, and five UI languages.
\Cref{tab:benchmark_comparison} compares benchmark coverage and timing evaluation.

\subsection{Efficiency Evaluation for GUI Agents}
R-VLM and RegionFocus refine localization through zooming, with R-VLM reporting higher accuracy and latency~\cite{rvlm,visualtts}.
FastFocus proposes regions through low-resolution attention and verifies high-resolution local views in parallel~\cite{fastfocus}, while RankGround selects crops with a lightweight reranker~\cite{rankground}.
GUI-Actor proposes action regions through an attention-based head in one forward pass~\cite{guiactor}, InnerZoom reuses internal evidence within one pass~\cite{innerzoom}, and DeCoPatch adds little decoding overhead~\cite{decopatch}.
ShowUI selects visual tokens~\cite{showui} and SimpAgent prunes elements and compresses history~\cite{simpagent}, whereas GUI-G1 favors concise outputs~\cite{guig1} and GoClick uses a compact model~\cite{goclick}.
Token and FLOP reductions alone do not establish latency gains.
Kumbhar et al. report latency growth and accuracy saturation with larger diffusion budgets in their setting~\cite{guidiffusion}.
UI-Vision reports Basic element-grounding latency~\cite{uivision}, while Showdown-clicks-dev~\cite{showdown} and Phi-Ground's comparisons on it report grounding accuracy and latency~\cite{phiground}.

At the task level, MMBench-GUI measures step efficiency~\cite{mmbenchgui}, while OSWorld-Human profiles latency and uses human action references for step-based efficiency scoring across 369 OSWorld tasks~\cite{osworldhuman}.
cua-speedrun compares end-to-end success, time, and model cost on 166 tasks from existing benchmarks~\cite{cuaspeedrun}.
ComponentBench evaluates success and latency on 2,910 interactive web-component tasks, comparing completion times with human references on tasks each agent completes successfully~\cite{componentbench}.
To bring human-referenced comparison to systematic grounding evaluation, we provide per-task mean times from 11 experts across five benchmarks and use H-GATE to score accuracy and latency over all tasks, including failures.
\Cref{app:related_work} reviews additional related work.

\section{Benchmarks and Metrics}
\label{sec:benchmarks_metrics}

We first introduce the construction and composition of ScreenSpot-Efficiency in \cref{sec:sse_construction,sec:sse_statistics}.
We then describe the human expert completion-time annotations for individual tasks in \cref{sec:human_timing}.
Finally, \cref{sec:hgate} defines H-GATE to jointly measure grounding correctness and latency against these times.

\subsection{Benchmark Construction}
\label{sec:sse_construction}

\begin{figure*}
  \centering
  \includegraphics[width=\linewidth]{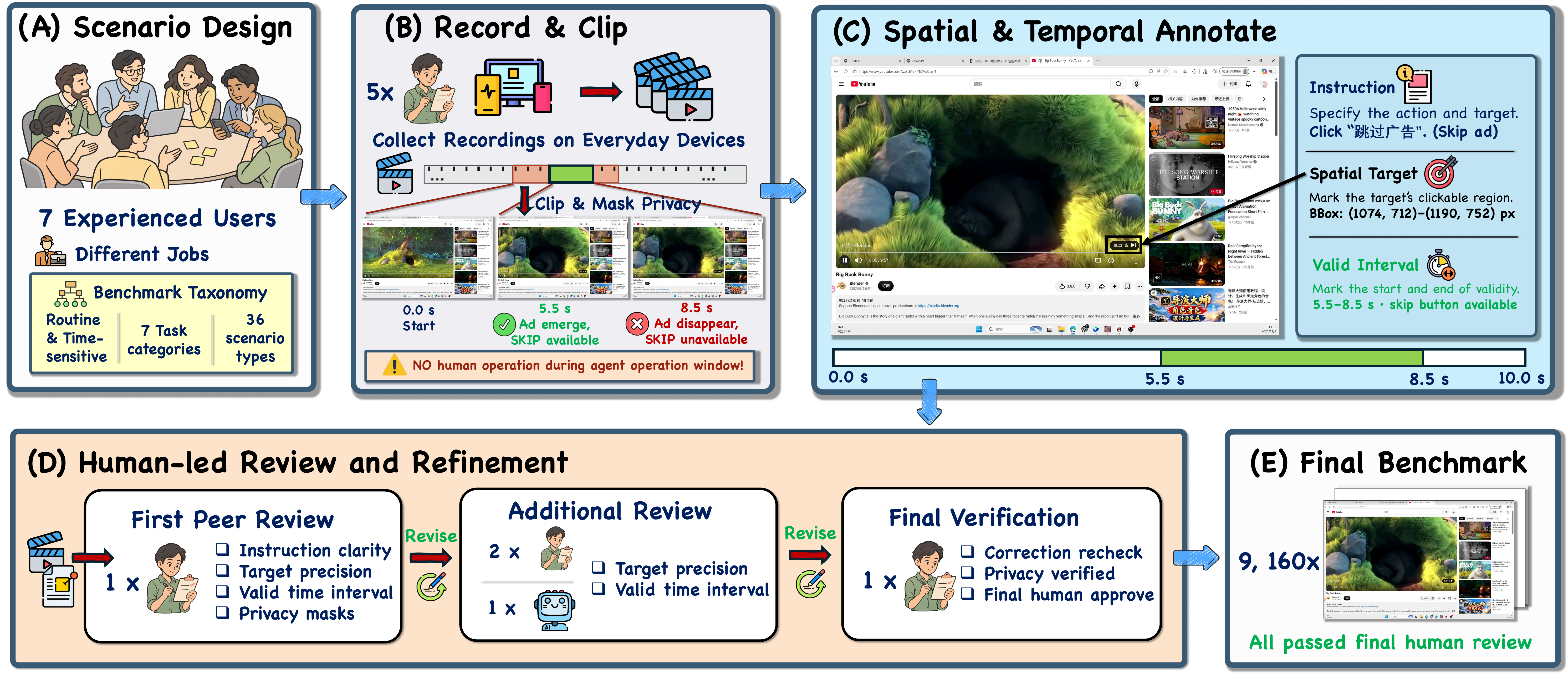}
  \caption{\textbf{Construction and annotation of ScreenSpot-Efficiency}. (A–C) We design scenarios, collect screen recordings, and annotate instructions, spatial targets, and valid time intervals. (D–E) Peer review and revision, supported by AI checks, produce a final set of 9,160 tasks, each verified by human reviewers.}
  \label{fig:sse_construction}
\end{figure*}

\paragraph{Scenario design and recording.}
To evaluate grounding efficiency in everyday work and life, we construct SS-E from real screen recordings, as illustrated in \cref{fig:sse_construction}.
Seven experienced GUI users from different jobs jointly identify 36 scenario types across six platforms: Windows, macOS, Ubuntu, iOS, iPadOS, and Android.
Five collectors use their regular devices to record natural work and everyday activities in actual applications.
At candidate task states, collectors temporarily defer further interaction to leave time for agent responses.
The trimmed clips retain the original playback speed and contain no state transitions caused by the recorder executing the target action.
Besides, we mask small regions containing private information without obscuring task-relevant content and recheck these masks during review.
\Cref{app:ethics} provides our ethics statement.

\paragraph{Spatiotemporal annotation.}
Annotators write a task instruction and mark the required target regions or points for each clip.
Temporal annotations specify the intervals in which these targets satisfy the instruction, defining when a spatially correct response is valid.
For example, in \cref{fig:sse_construction}, the valid interval for an ad-skipping task begins when the skip control becomes available and ends when the advertisement disappears.
\Cref{app:action_space} details the action space.

\paragraph{Human-led quality control.}
Quality control separates creation from review: each task is checked by the four other members of the five-person collection team.
The first reviewer checks instruction clarity, spatial annotation precision, and temporal validity.
After the first review, the creator revises the annotations or re-records the clip as needed before two additional team members review the revised task.
Codex with GPT-6 Astra High, the AI check in \cref{fig:sse_construction}, examines all video frames and annotations without a time limit to identify spatial or temporal inconsistencies.
After additional review, the creator revises the task again based on the combined human and AI feedback. Finally, the remaining team member performs final verification and every task in the final benchmark has passed this stage.

\subsection{Dataset Statistics}
\label{sec:sse_statistics}

The final SS-E dataset comprises 9,160 tasks organized into a three-level taxonomy of two grounding modes, seven task categories, and 36 scenario types, as shown in \cref{fig:sse_distribution}.
Routine grounding comprises 5,149 tasks, or 56.2\%, across search and selection, editing and dragging, sequential operations, and visual checks.
time-sensitive grounding comprises 4,011 tasks, or 43.8\%, across event-driven interaction, collaboration and competition, and real-time gameplay.
Editing and dragging is the largest category with 2,634 tasks, including multi-point localization for cropping and contour outlining and paired-target localization for dragging.
time-sensitive tasks include ad skipping and expiring undo, where target validity changes over time, and collaborative editing, where other participants can alter the target.
Additional dataset statistics, full scenario definitions, and per-scenario task counts are provided in \cref{sec:appendix-b-coverage}, and \cref{app:examples} presents samples and evaluation cases.

\subsection{Human Expert Completion-Time Annotation}
\label{sec:human_timing}

To establish a human reference for efficiency evaluation, we collect completion times from 11 experts for every task in SS-E and four existing grounding benchmarks~\cite{osatlas,screenspotpro,osworldg,visualwebbench}.
The experts are highly familiar with the required operations and are instructed to remain focused and complete each task as quickly as possible.
We use the arithmetic mean of these 11 completion times as the human completion-time reference for each task.
This expert reference time is deliberately demanding, and \cref{app:expert_times} details the protocol.
Because experts are extremely familiar with the applications, scenarios, and operations, their times are clearly shorter than those expected of typical users and establish a best-case human reference under our experimental conditions.
We adopt this design because we aim for future GUI grounding to be faster than extremely proficient humans and equally accurate, which gives GUI agents an explicit and quantifiable efficiency target.
The expert reference time also gives every task a common time scale, so evaluation targets the trade-off between accuracy and response time across model configurations rather than a controlled comparison between humans and agents.

\subsection{Human-referenced Grounding Accuracy--Time Efficiency}
\label{sec:hgate}

H-GATE combines grounding correctness with time relative to human experts, drawing on joint accuracy--time evaluation~\cite{kasahara2024throughput} and expert-referenced usability measurement~\cite{macdorman2011improved}.
For each of $N$ tasks, let $c_i\in\{0,1\}$ denote correctness, $t_i>0$ the model's complete grounding latency, and $h_i>0$ the expert reference time from \cref{sec:human_timing}.

Earned duration management values completed work by multiplying reference duration by completion fraction~\cite{khamooshi2014edm}.
Accordingly, $c_i h_i$ expresses correct completion in units of expert time, while $t_i$ records actual time spent.
Symmetric comparisons in economics and finance~\cite{decker2020changing,morariupatrichi2022state} and shared time fractions in human--robot interaction~\cite{amine2023vahr} motivate a common time scale.
Normalizing the expert and model times into shares with $p_i+q_i=1$ and $p_i/q_i=h_i/t_i$ gives
\begin{equation}
\label{eq:hgate_time_shares}
p_i=\frac{h_i}{h_i+t_i},
\qquad
q_i=\frac{t_i}{h_i+t_i}.
\end{equation}
The normalized time balance is
\begin{equation}
\label{eq:hgate_time_balance}
B_i=c_i p_i-q_i=\frac{c_i h_i-t_i}{h_i+t_i}.
\end{equation}
This balance credits correct completion and accounts for time spent on every attempt.

\begin{figure}
  \centering
  \includegraphics[width=\linewidth]{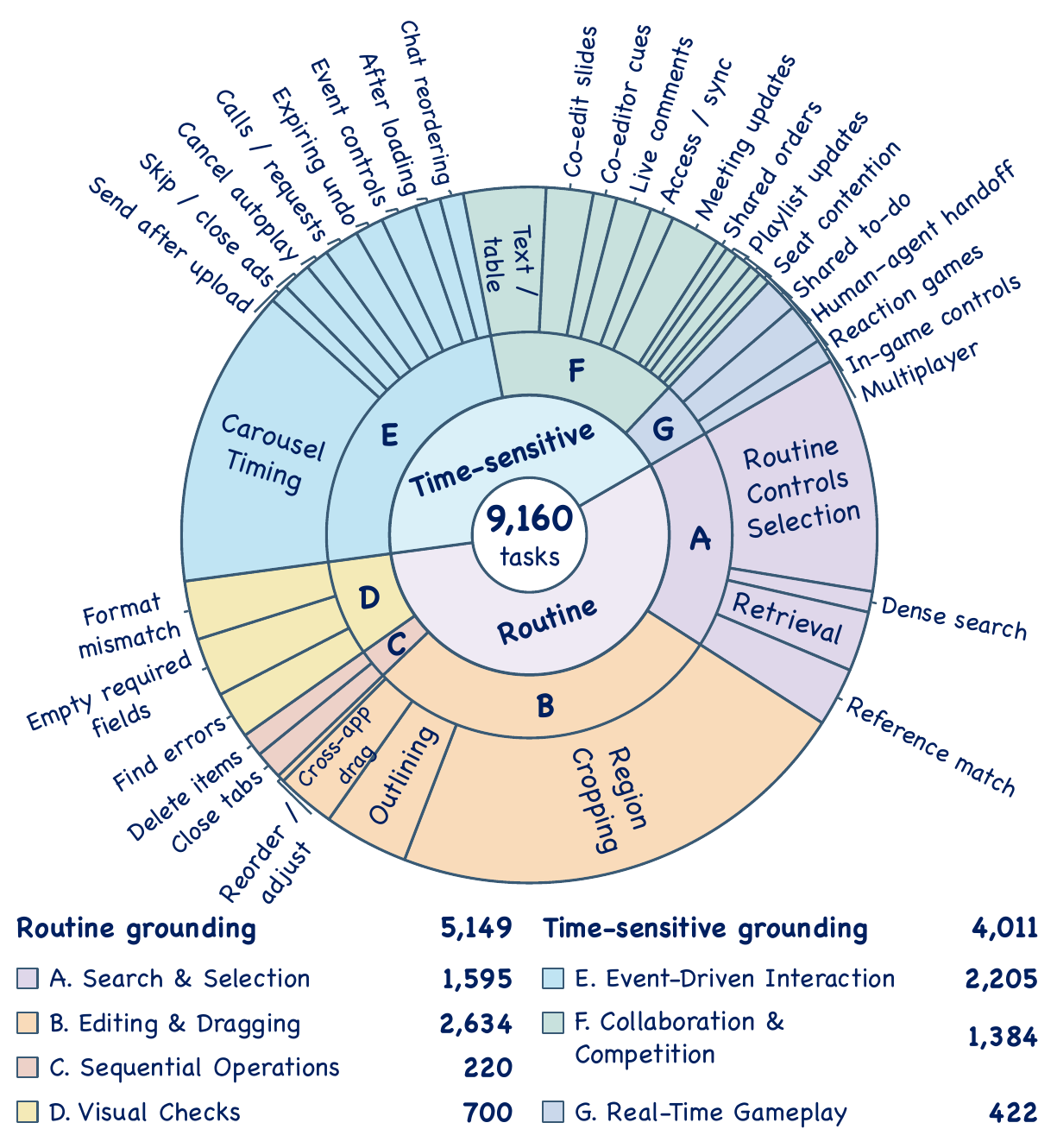}
  \caption{Task distribution in ScreenSpot-Efficiency. From inner to outer, the rings represent two grounding modes, seven task categories, and 36 scenario types. Full definitions and per-scenario counts are provided in \cref{sec:appendix-b-coverage}.}
  \label{fig:sse_distribution}
\end{figure}

We combine correctness and time balance through an affine intermediate score $u_i=\kappa+\alpha c_i+\beta B_i$.
We set its limiting value to zero for failure as $t_i/h_i\to0$ and for success as $t_i/h_i\to\infty$, and assign one to a correct response at $t_i=h_i$.
These conditions give $\kappa=0$ and $\alpha=\beta=1$, hence $u_i=c_i+B_i$.
We linearly map the resulting score from its limiting range of $-1$ to $2$ onto a 0--100 scale and average over tasks:
\begin{equation}
\label{eq:hgate}
\mathrm{H\text{-}GATE}=\frac{100}{3N}\sum_{i=1}^{N}\left(u_i+1\right).
\end{equation}
Scoring before averaging preserves the correspondence between correctness and latency across the full task set, including failures. Under this normalization, 33.33 is the utility-neutral reference where credit for correct completions offsets failed-attempt time penalties, and 66.67 is the human reference for completing every task correctly in its annotated time. See \cref{app:hgate} for detailed theoretical analysis and metric properties.

\section{Experiments and Results}
\label{sec:experiments}

\subsection{Experimental Setup}
\label{sec:experimental_setup}
We evaluate 38 open-weight models from 14 mainstream families in 87 inference configurations, covering thinking and non-thinking modes, multiple levels of visual-token budget reduction, and zoom-in.
All evaluations use vLLM~\cite{vllm} across H200, A100, 910C, RTX 3090, RTX 4090, RTX 5090, and C550 GPUs.
Each mode is evaluated five times with one concurrent request per deployment replica to minimize contention and measure grounding efficiency under single-request conditions.
We report accuracy (Acc), latency, and H-GATE on all five benchmarks, evaluating SS-E with real-time playback, where Acc requires correct localization within applicable action windows and the additional spatial accuracy (Acc-S) disregards timing.
Detailed settings are provided in \cref{app:setup_analysis}.

\begin{table*}
\centering
\caption{\textbf{Representative H200 results across five benchmarks.} Acc and Acc-S denote full and spatial accuracy (\%). SS-E Acc additionally requires temporal validity. Time is mean latency in seconds over all attempts. Higher is better except for Time. -T denotes thinking or reasoned-action configurations. Bold and underlined values mark the best displayed H-GATE and Acc per benchmark, respectively. Full results are provided in \cref{sec:app_full_gpu_results}.}
\label{tab:hgate_main}
\begingroup
\footnotesize
\setlength{\tabcolsep}{1.32pt}
\renewcommand{\arraystretch}{1.11}
\arrayrulecolor{HGRule}
\begin{tabular}{c|c!{\color{HGRuleStrong}\vrule width 0.75pt}cccc@{\hspace{7pt}}ccc@{\hspace{5pt}}ccc@{\hspace{5pt}}ccc@{\hspace{5pt}}ccc}
\noalign{\color{HGRuleStrong}\hrule height 0.8pt}
\multirow{2}{*}{Family} & \multirow{2}{*}{\shortstack{Model\\ /mode}} & \multicolumn{4}{c@{\hspace{7pt}}}{\cellcolor{HGHeader}\textbf{ScreenSpot-Efficiency}} & \multicolumn{3}{c}{ScreenSpot-Pro} & \multicolumn{3}{c}{ScreenSpot-v2} & \multicolumn{3}{c}{OSWorld-G} & \multicolumn{3}{c}{VisualWebBench} \\
\noalign{\global\arrayrulewidth=0.7pt}\arrayrulecolor{HGRuleStrong}\hhline{~~*{16}{-}}\arrayrulecolor{HGRule}\noalign{\global\arrayrulewidth=0.4pt}
 & & \cellcolor{HGHeader}Acc-S & \cellcolor{HGHeader}Acc & \cellcolor{HGHeader}Time & \cellcolor{HGHeader}H-GATE & Acc & Time & H-GATE & Acc & Time & H-GATE & Acc & Time & H-GATE & Acc & Time & H-GATE \\
\noalign{\color{HGRuleStrong}\hrule height 0.7pt}
\textbf{Human} & \textemdash
 & \cellcolor{HGBody}100.0 & \cellcolor{HGBody}99.0 & \cellcolor{HGBody}2.35 & \cellcolor{HGBody}66.2
 & 99.2 & 1.46 & 66.2
 & 100.0 & 0.97 & 66.7
 & 99.9 & 1.11 & 66.6
 & 100.0 & 1.06 & 66.7 \\
\hline
\multirow{1}{*}{UI-TARS-1.5~\cite{exp_ui_tars_1_5}} & 7B & \cellcolor{HGBody}24.2 & \cellcolor{HGBody}18.4 & \cellcolor{HGBody}1.54 & \cellcolor{HGBody}30.6 & 41.7 & 1.34 & 39.9 & 89.5 & 0.67 & 68.1 & 53.4 & 0.48 & 53.7 & 88.0 & 0.41 & 74.7 \\
\hline
\multirow{1}{*}{Jedi~\cite{osworldg}} & 7B & \cellcolor{HGBody}17.5 & \cellcolor{HGBody}10.9 & \cellcolor{HGBody}1.87 & \cellcolor{HGBody}24.5 & 34.2 & 0.69 & 41.8 & 91.2 & 0.58 & 70.9 & 40.1 & 1.49 & 37.6 & 88.6 & 0.74 & 66.5 \\
\hline
\multirow{1}{*}{SE-GUI~\cite{exp_segui}} & 7B & \cellcolor{HGBody}20.5 & \cellcolor{HGBody}13.8 & \cellcolor{HGBody}2.28 & \cellcolor{HGBody}23.2 & 46.5 & 1.52 & 39.7 & 89.9 & 1.05 & 60.6 & 47.7 & 0.90 & 43.4 & 89.1 & 0.80 & 65.6 \\
\hline
\multirow{2}{*}{OpenCUA~\cite{exp_opencua}} & 7B & \cellcolor{HGBody}20.8 & \cellcolor{HGBody}13.7 & \cellcolor{HGBody}1.44 & \cellcolor{HGBody}27.0 & 46.1 & 1.35 & 43.5 & 93.3 & 0.55 & 72.8 & 55.9 & 0.41 & 56.6 & 91.5 & 0.35 & 78.5 \\
 & 72B & \cellcolor{HGBody}26.5 & \cellcolor{HGFind3}18.2 & \cellcolor{HGFind3}3.14 & \cellcolor{HGFind3}23.3 & 60.9 & 2.57 & 41.7 & 87.8 & 1.91 & 54.6 & 60.3 & 2.31 & 40.0 & 11.4 & 4.95 & 17.4 \\
\hline
\multirow{3}{*}{ScaleCUA~\cite{exp_scalecua}} & 7B & \cellcolor{HGBody}21.3 & \cellcolor{HGFind2}13.9 & \cellcolor{HGFind2}0.89 & \cellcolor{HGFind2}30.5 & 47.8 & 0.59 & 51.0 & 92.6 & 0.51 & 73.8 & 55.5 & 0.44 & 55.5 & 91.3 & 0.38 & 77.1 \\
 & 7B-T & \cellcolor{HGBody}24.2 & \cellcolor{HGFind2}18.7 & \cellcolor{HGFind2}3.28 & \cellcolor{HGFind2}22.8 & 29.3 & 1.72 & 30.2 & 85.5 & 1.55 & 53.4 & 44.5 & 1.67 & 35.2 & 89.9 & 1.35 & 58.0 \\
 & 32B & \cellcolor{HGBody}22.4 & \cellcolor{HGBody}15.6 & \cellcolor{HGBody}1.61 & \cellcolor{HGBody}26.7 & 58.7 & 0.99 & 51.0 & 95.1 & 0.90 & 66.1 & 60.8 & 1.34 & 50.9 & 90.3 & 0.73 & 67.7 \\
\hline
\multirow{1}{*}{GroundNext~\cite{exp_groundnext}} & 7B & \cellcolor{HGBody}18.3 & \cellcolor{HGBody}11.8 & \cellcolor{HGBody}2.63 & \cellcolor{HGBody}20.8 & 53.1 & 1.47 & 43.7 & 90.5 & 0.93 & 62.9 & 56.4 & 0.77 & 49.4 & 90.7 & 0.70 & 68.4 \\
\hline
\multirow{3}{*}{Holo2~\cite{exp_holo2}} & 4B-T & \cellcolor{HGBody}25.0 & \cellcolor{HGFind3}18.0 & \cellcolor{HGFind3}3.32 & \cellcolor{HGFind3}33.5 & 57.7 & 0.84 & 53.7 & 93.2 & 0.43 & 75.8 & 61.9 & 0.29 & 63.2 & 90.9 & 0.27 & 81.0 \\
 & 8B & \cellcolor{HGBody}28.5 & \cellcolor{HGBody}20.9 & \cellcolor{HGFind2}1.37 & \cellcolor{HGFind2}35.8 & 59.8 & 0.99 & 53.5 & 94.3 & 0.47 & 75.3 & 63.3 & 0.31 & 63.5 & 92.2 & 0.27 & 81.6 \\
 & 8B-T & \cellcolor{HGBody}26.8 & \cellcolor{HGBody}16.2 & \cellcolor{HGFind2}13.86 & \cellcolor{HGFind2}20.3 & 57.3 & 2.73 & 42.9 & 94.2 & 1.54 & 66.5 & 60.8 & 2.55 & 44.3 & 91.1 & 1.36 & 65.8 \\
\hline
\multirow{1}{*}{POINTS-GUI~\cite{exp_points_gui}} & 8B & \cellcolor{HGBody}31.7 & \cellcolor{HGBody}23.2 & \cellcolor{HGBody}0.99 & \cellcolor{HGBody}34.2 & 59.2 & 0.86 & 53.1 & 95.2 & 0.59 & 72.7 & 66.0 & 0.82 & 56.9 & 92.1 & 0.34 & 79.2 \\
\hline
\multirow{3}{*}{UI-Venus-1.5~\cite{exp_ui_venus_1_5}} & 2B & \cellcolor{HGBody}29.2 & \cellcolor{HGFind1}15.7 & \cellcolor{HGFind1}1.09 & \cellcolor{HGFind1}31.6 & 58.8 & 0.79 & 55.6 & \cellcolor{HGFind1}93.3 & \cellcolor{HGFind1}0.36 & \cellcolor{HGFind1}\textbf{79.8} & 58.3 & 0.25 & 63.7 & \cellcolor{HGFind1}90.1 & \cellcolor{HGFind1}0.22 & \cellcolor{HGFind1}\textbf{83.9} \\
 & 8B & \cellcolor{HGBody}37.6 & \cellcolor{HGBody}26.3 & \cellcolor{HGBody}0.93 & \cellcolor{HGBody}38.1 & 68.2 & 0.91 & 58.9 & 96.2 & 0.41 & 78.8 & 69.3 & 0.28 & 68.5 & 91.5 & 0.25 & 82.8 \\
 & 30B-A3B & \cellcolor{HGBody}40.3 & \cellcolor{HGBody}29.5 & \cellcolor{HGBody}1.77 & \cellcolor{HGBody}36.0 & 69.1 & 0.87 & 60.0 & \underline{96.5} & 0.40 & 79.5 & 71.5 & 0.28 & \cellcolor{HGFind1}\textbf{69.8} & \underline{92.6} & 0.25 & 83.6 \\
\hline
\multirow{3}{*}{GUI-Owl-1.5~\cite{exp_gui_owl}} & 4B & \cellcolor{HGBody}21.8 & \cellcolor{HGBody}14.6 & \cellcolor{HGBody}0.49 & \cellcolor{HGBody}34.9 & 67.0 & 0.71 & \cellcolor{HGFind1}\textbf{60.3} & 93.1 & 0.39 & 77.3 & 63.8 & 0.28 & 64.7 & 91.1 & 0.25 & 81.7 \\
 & 8B & \cellcolor{HGBody}26.3 & \cellcolor{HGBody}18.4 & \cellcolor{HGBody}0.62 & \cellcolor{HGBody}35.8 & 70.5 & 0.85 & 60.0 & 94.2 & 0.48 & 75.0 & 66.8 & 0.35 & 64.4 & 90.7 & 0.32 & 78.8 \\
 & 32B & \cellcolor{HGBody}26.9 & \cellcolor{HGBody}18.7 & \cellcolor{HGBody}1.39 & \cellcolor{HGBody}29.5 & \underline{72.9} & 1.77 & 51.0 & 94.9 & 1.16 & 61.5 & 66.0 & 0.94 & 51.7 & 91.1 & 0.89 & 64.9 \\
\noalign{\color{HGRuleStrong}\hrule height 0.8pt}
\multirow{5}{*}{Qwen3.5~\cite{exp_qwen3_5}} & 0.8B & \cellcolor{HGBody}17.4 & \cellcolor{HGBody}11.7 & \cellcolor{HGBody}1.17 & \cellcolor{HGBody}29.6 & 29.7 & 0.45 & 43.4 & 77.0 & 0.22 & 73.6 & 32.3 & 0.19 & 48.5 & 76.9 & 0.16 & 76.9 \\
 & 9B & \cellcolor{HGFind2}28.6 & \cellcolor{HGFind2}21.7 & \cellcolor{HGFind2}0.74 & \cellcolor{HGFind2}36.4 & 49.8 & 0.79 & 50.1 & 91.2 & 0.37 & 77.0 & 54.3 & 0.25 & 59.9 & 84.7 & 0.21 & 79.4 \\
 & 9B-T & \cellcolor{HGFind2}34.5 & \cellcolor{HGFind2}20.2 & \cellcolor{HGFind2}53.90 & \cellcolor{HGFind2}11.4 & 47.1 & 30.73 & 21.3 & 90.2 & 13.51 & 39.5 & 57.3 & 28.78 & 25.0 & 83.1 & 9.55 & 44.6 \\
 & 27B & \cellcolor{HGBody}38.0 & \cellcolor{HGFind4}32.2 & \cellcolor{HGFind4}1.86 & \cellcolor{HGFind4}35.1 & 54.5 & 1.39 & 45.7 & 90.2 & 0.77 & 65.7 & 58.5 & 0.60 & 53.7 & 90.3 & 0.53 & 72.4 \\
 & 397B-A17B & \cellcolor{HGFind1}50.8 & \cellcolor{HGFind1}\underline{42.2} & \cellcolor{HGFind1}1.45 & \cellcolor{HGFind1}\textbf{43.0} & 65.3 & 1.11 & 54.1 & 95.4 & 0.66 & 70.9 & 69.7 & 0.57 & 60.8 & 91.5 & 0.48 & 74.2 \\
\hline
\multirow{1}{*}{MolmoPoint-GUI~\cite{exp_molmopoint}} & 8B & \cellcolor{HGBody}22.6 & \cellcolor{HGBody}15.6 & \cellcolor{HGBody}3.03 & \cellcolor{HGBody}23.6 & 61.2 & 1.30 & 48.9 & 93.9 & 0.93 & 65.5 & 63.3 & 0.67 & 55.2 & 83.3 & 0.67 & 65.2 \\
\hline
\multirow{1}{*}{VISTA~\cite{exp_vista}} & 9B & \cellcolor{HGBody}32.3 & \cellcolor{HGBody}23.4 & \cellcolor{HGBody}1.04 & \cellcolor{HGBody}34.9 & 67.9 & 1.01 & 58.1 & 96.0 & 0.55 & 74.6 & 67.2 & 0.42 & 62.7 & 92.1 & 0.40 & 77.2 \\
\hline
\multirow{1}{*}{UI-Venus-2~\cite{exp_ui_venus_2}} & 9B & \cellcolor{HGBody}44.7 & \cellcolor{HGBody}33.3 & \cellcolor{HGBody}1.15 & \cellcolor{HGBody}41.4 & 72.4 & 1.15 & 58.4 & 96.3 & 0.58 & 74.0 & \underline{75.4} & 0.35 & 69.6 & 90.5 & 0.41 & 76.2 \\
\noalign{\global\arrayrulewidth=0.8pt}\arrayrulecolor{HGRuleStrong}\hline
\end{tabular}
\global\arrayrulewidth=0.4pt\arrayrulecolor{HGRule}
\par\smallskip
\begin{minipage}{\textwidth}\footnotesize
\HGSwatch{HGBody} identifies our benchmark (SS-E). \HGSwatch{HGFind1}, \HGSwatch{HGFind2}, \HGSwatch{HGFind3}, and \HGSwatch{HGFind4} highlight data primarily used in the analyses in Secs.~\ref{sec:overall_results}, \ref{sec:inference_budgets}, \ref{sec:metric_diagnostics}, and \ref{sec:hardware_architecture}, respectively.
\end{minipage}
\endgroup
\end{table*}

\subsection{Grounding Performance Relative to Humans}
\label{sec:overall_results}
While current state-of-the-art models exhibit near-saturation performance on traditional static interfaces, their capabilities severely degrade under the dual constraints of interface complexity and time sensitivity intrinsic to SS-E as illustrated by the representative results in \cref{tab:hgate_main}.
Across all evaluated configurations, median accuracy and H-GATE scores plummet when transitioning from simpler benchmarks to SS-E, highlighting a fundamental limitation: architectural paradigms optimized for static comprehension struggle to generalize to reactive real-world environments.
The expert comparisons in \cref{fig:teaser} place this cross-benchmark disparity in context, demonstrating that progress on simpler tasks leaves a substantial efficiency gap between agents and human experts on demanding interfaces.
On SS-E, the highest displayed model Acc and H-GATE in \cref{tab:hgate_main} are 42.2\% and 43.0 for Qwen3.5-397B-A17B, against 99.0\% and 66.2 in the Human row, whereas the best displayed H-GATE across the four existing benchmarks spans 60.3 to 83.9.

Crucially, analyzing spatial accuracy (Acc-S) against strict valid-window accuracy (Acc) reveals a persistent temporal bottleneck.
Every configuration in \cref{tab:hgate_main} drops 5.5 to 14.3 points from Acc-S to Acc.
In time-sensitive tasks, models often find the correct target yet answer before the valid interval opens or after it closes.
This consistent performance drop explicitly exposes a requirement that spatial precision alone cannot capture: measuring grounding capabilities in a vacuum artificially inflates perceived utility.
A correct target must also be selected while the action remains valid, underscoring the necessity of evaluating GUI agents through the lens of human-referenced temporal dynamics.

Under direct grounding, the failure breakdown of eight H200 configurations in \cref{tab:app-e-failure-mechanisms} shows timing losses arise mainly at observation, not during inference.
Localization error is the largest mechanism for every configuration at 33.87\% to 67.45\% of failures, so localization remains the central difficulty.
Among the six configurations without thinking, Insufficient evidence and Premature answers together account for 29.20\% to 39.55\% of failures, whereas Overdue answers account for at most 1.69\%.
Enabling thinking reverses this ordering, as Premature answers fall from 1.32\% to 0.03\% and Overdue answers climb from 1.12\% to 14.89\% from Qwen3.5-27B to Qwen3.5-27B-T.
The temporal bottleneck of direct grounding therefore centers on when to observe and commit, while long-latency configurations also lose answers after the valid interval closes.
SS-E failures thus reflect three capabilities: locating the target, deciding when the interface supports an action, and answering before that support expires.

\subsection{Inference Computation and Visual Budgets}
\label{sec:inference_budgets}
\begin{figure*}
\centering
\includegraphics[width=\textwidth]{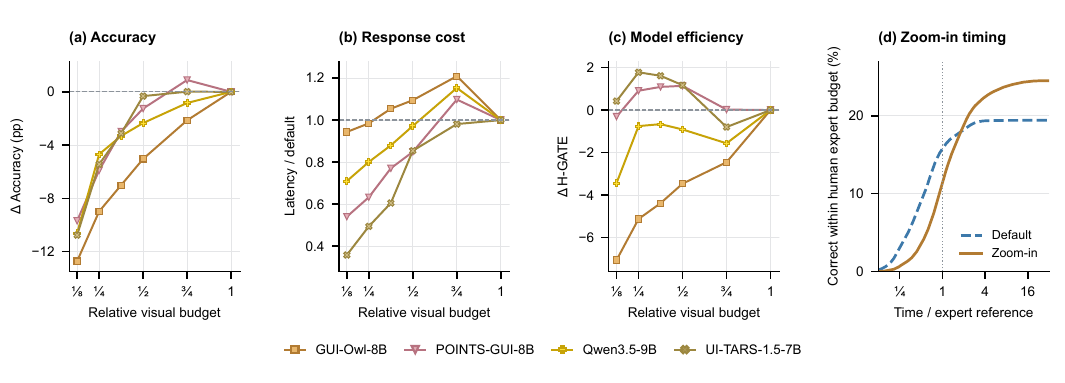}
\caption{Visual-budget reduction and local zooming on SS-E using H200. (a--c) Four models with complete budget sweeps, showing changes in accuracy, mean latency, and H-GATE relative to their default configurations. Markers denote measured settings. (d) Mean cumulative correct completion across nine paired default and zoom-in configurations, weighted equally and evaluated at multiples of each task's expert reference time.}
\label{fig:visual_budget}
\end{figure*}

Strategies designed to bolster accuracy, including extended reasoning with thinking modes and high-resolution local zooming, frequently yield a net negative impact on SS-E efficiency as illustrated in \cref{tab:hgate_main}.
Although these techniques afford marginal gains in spatial precision, the disproportionate latency inflation actively penalizes performance on time-sensitive tasks, driving down H-GATE scores across nearly all evaluated reasoning configurations.
In \cref{tab:hgate_main}, thinking lowers SS-E H-GATE in all three displayed pairs by 7.7 to 25.0 points, and for Qwen3.5-9B it lifts Acc-S from 28.6\% to 34.5\% while Acc drops from 21.7\% to 20.2\% as mean latency grows from 0.74 to 53.90 seconds.
The cumulative completion curves in \cref{fig:visual_budget}d explicitly expose the cost of this trade-off: while zooming pushes absolute success rates higher, it significantly diminishes the fraction of tasks completed correctly within the human expert reference times.
Across nine default and zoom-in pairs, zoom-in lifts Acc by 2.65 to 9.75 points yet lowers H-GATE in every pair by 1.23 to 5.52 points.
Ultimately, such methods trade essential temporal validity for diminishing spatial returns.
Detailed per-model results appear in \cref{app:reasoning_zoom}.

Conversely, manipulating visual budgets yields non-monotonic effects on both latency and localization capability.
The complete sweeps in \cref{fig:visual_budget}a--c demonstrate that intermediate token reductions, which involve retaining 25\% to 50\% of the default budget, can optimize H-GATE for specific architectures by trimming latency whereas aggressive compression inevitably destroys spatial perception.
Paradoxically, fewer visual tokens do not guarantee faster execution because reduced budgets can inadvertently inflate inference latency due to architectural or system-level overheads in several cases.
These findings indicate that budget selection and efficiency optimizations must be grounded in empirical accuracy-latency profiling rather than theoretical token-count assumptions.
Results for all interventions across evaluated models appear in \cref{app:visual_budget}.

Whether added computation improves efficiency depends on the temporal structure of the task.
An intervention on delayed targets in \cref{app:explicit_waiting} separates two ways of spending time, and on 80 send-after-upload tasks, where every evaluated configuration reaches 100\% Acc-S, enabling an explicit wait response lifts UI-Venus-2-9B from 0.0\% to 48.8\% Acc and from 30.2 to 46.3 H-GATE as mean latency rises from 0.5 to 3.5 seconds.
Qwen3.5-9B-T, which waits incidentally through reasoning, attains 43.8\% Acc without the wait option at a 12.7-second mean latency and 39.9 H-GATE.
On these tasks, timing an action through interface observation is therefore cheaper than timing it through inference duration, and H-GATE rewards a slower response only when the added time turns a failure into a success on that task.
On 390 short-window tasks, whose valid intervals open at clip onset and last at most 3 seconds, enabling waiting shifts Acc by at most 0.7 points, and thinking configurations still fail most of these tasks because their mean latencies approach or exceed that bound.
Observation timing and response speed after a valid observation are thus separate optimization targets, and shorter latency yields higher grounding efficiency only when matched to the task state.

\begin{table*}
\centering
\caption{SS-E accuracy and efficiency optima within each GPU deployment's tested configurations, with results across all five benchmarks. Bold and underlined values mark the best displayed H-GATE and Acc per benchmark, respectively.}
\label{tab:hardware_choices}
\begingroup
\footnotesize
\setlength{\tabcolsep}{1.32pt}
\renewcommand{\arraystretch}{1.11}
\arrayrulecolor{black}
\begin{tabular}{c!{\color{HGRule}\vrule width 0.4pt}l!{\color{black}\vrule width 0.75pt}cccc@{\hspace{7pt}}ccc@{\hspace{5pt}}ccc@{\hspace{5pt}}ccc@{\hspace{5pt}}ccc}
\noalign{\color{black}\hrule height 0.8pt}
\multirow{2}{*}{GPU} & \multirow{2}{*}{Model} & \multicolumn{4}{c@{\hspace{7pt}}}{\textbf{ScreenSpot-Efficiency}} & \multicolumn{3}{c}{ScreenSpot-Pro} & \multicolumn{3}{c}{ScreenSpot-v2} & \multicolumn{3}{c}{OSWorld-G} & \multicolumn{3}{c}{VisualWebBench} \\
\noalign{\global\arrayrulewidth=0.7pt}\hhline{>{\arrayrulecolor{HGRule}}~|>{\arrayrulecolor{black}}~|*{16}{-}}\noalign{\global\arrayrulewidth=0.4pt}
 & & Acc-S & Acc & Time & H-GATE & Acc & Time & H-GATE & Acc & Time & H-GATE & Acc & Time & H-GATE & Acc & Time & H-GATE \\
\noalign{\color{black}\hrule height 0.7pt}
\multirow{2}{*}{A100} & UI-Venus-1.5-30B-A3B & 40.3 & 32.2 & 3.85 & 34.1 & \underline{69.1} & 1.91 & 49.1 & \underline{96.5} & 0.99 & 64.9 & \underline{71.3} & 0.78 & 57.2 & \underline{92.6} & 0.68 & 69.8 \\
 & UI-Venus-1.5-8B & 37.7 & 28.4 & 1.45 & 35.4 & 68.2 & 1.65 & 51.6 & 96.2 & 0.66 & 71.7 & 69.3 & 0.47 & 62.8 & 91.5 & 0.38 & 77.2 \\
\noalign{\color{black}\hrule height 0.4pt}
\multirow{2}{*}{RTX 3090} & UI-Venus-1.5-8B & 37.7 & 25.5 & 6.44 & 21.2 & 68.2 & 6.63 & 35.0 & 96.2 & 3.20 & 48.6 & 69.3 & 2.28 & 42.1 & 91.5 & 1.91 & 53.3 \\
 & GUI-Owl-1.5-2B & 21.8 & 15.0 & 1.16 & 30.4 & 58.1 & 1.98 & 43.4 & 90.0 & 0.85 & 64.6 & 52.7 & 0.59 & 51.1 & 87.6 & 0.51 & 71.4 \\
\noalign{\color{black}\hrule height 0.4pt}
\multirow{1}{*}{RTX 4090} & UI-Venus-1.5-8B & 37.6 & 29.2 & 1.99 & 32.9 & 68.2 & 2.19 & 47.7 & 96.2 & 0.92 & 66.5 & 69.3 & 0.62 & 59.3 & 91.5 & 0.52 & 73.1 \\
\noalign{\color{black}\hrule height 0.4pt}
\multirow{1}{*}{H200} & Qwen3.5-397B-A17B & 50.8 & \underline{42.2} & 1.45 & \textbf{43.0} & 65.3 & 1.11 & \textbf{54.1} & 95.4 & 0.66 & 70.9 & 69.7 & 0.57 & 60.8 & 91.5 & 0.48 & 74.2 \\
\noalign{\color{black}\hrule height 0.4pt}
\multirow{1}{*}{C550} & UI-Venus-1.5-8B & 37.7 & 28.6 & 2.43 & 31.0 & 68.2 & 2.71 & 45.0 & 96.2 & 1.18 & 62.5 & 69.3 & 0.80 & 55.8 & 91.5 & 0.69 & 69.0 \\
\noalign{\color{black}\hrule height 0.4pt}
\multirow{1}{*}{RTX 5090} & UI-Venus-1.5-8B & 37.7 & 28.3 & 1.37 & 36.0 & 68.2 & 1.52 & 52.7 & 96.2 & 0.62 & \textbf{72.7} & 69.3 & 0.41 & \textbf{64.2} & 91.5 & 0.34 & \textbf{78.9} \\
\noalign{\color{black}\hrule height 0.4pt}
\multirow{1}{*}{910C} & UI-Venus-1.5-8B & 37.6 & 29.0 & 2.05 & 32.8 & 68.2 & 2.41 & 46.5 & 96.2 & 0.96 & 65.4 & 69.3 & 0.70 & 57.7 & 91.5 & 0.60 & 71.1 \\
\noalign{\global\arrayrulewidth=0.8pt}\hline
\end{tabular}
\global\arrayrulewidth=0.4pt
\endgroup
\end{table*}

\subsection{What H-GATE Captures}
\label{sec:metric_diagnostics}
\begin{figure}
\centering
\includegraphics[width=\linewidth]{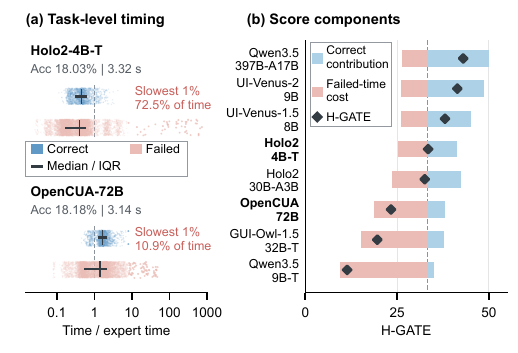}
\caption{SS-E efficiency diagnostics on H200. (a) Per-task model-to-expert time ratios. The slowest 1\% of tasks are selected by each model's raw latency, with percentages denoting their share of its total time. (b) H-GATE decomposition for eight models into a utility-neutral reference (33.33) plus correct-response contributions minus failed-attempt time penalties.}
\label{fig:metric_diagnostics}
\end{figure}

Aggregate accuracy and mean latency conceal the pairing of correctness and per-task response time.
In \cref{fig:metric_diagnostics}a, Holo2-4B-T and OpenCUA-72B both achieve accuracy near 18\% and mean latency near 3.2\,s.
Yet their slowest 1\% of tasks, ranked separately by raw model latency, account for 72.5\% and 10.9\% of their respective total grounding time.
Despite its heavier tail, Holo2 completes 16.38\% of all tasks correctly within their expert reference times, compared with 2.45\% for OpenCUA.
Rare slow responses can dominate mean latency even when most correct responses are timely, so the mean alone cannot characterize the joint outcome distribution.
The eight-model decomposition in \cref{fig:metric_diagnostics}b explains their 10.13-point H-GATE gap through 3.47 points of additional correct-response contributions and 6.66 points of lower failed-attempt time penalties.
These contributions are combined around a utility-neutral reference of 33.33, whose derivation and interpretation are provided in \cref{app:hgate}.
Beyond this pair, ablations across 51 H200 configurations on SS-E reverse 18 of 1,275 pairwise rankings when per-task expert times are replaced with the benchmark mean and 128 when failed-attempt time is omitted.
Per-task references and failed-attempt costs therefore affect efficiency comparisons beyond aggregate accuracy and latency.

Human preference supports both design components, as detailed in \cref{app:human_agreement}.
In a pairwise study where raters compare two configurations' responses on the same six SS-E tasks, H-GATE reaches 84.00\% agreement with the majority vote on 200 random samples, compared with 77.00\% for mean latency and 60.75\% for Acc.
On samples where a comparison metric orders the pair opposite to H-GATE, the majority selects the higher-H-GATE configuration in 66.5\% against Acc, 69.0\% against mean latency, 64.0\% against the benchmark-mean reference, and 65.0\% against the variant without failed-attempt time.
A second study compares H-GATE with Acc@Human, the share of tasks answered correctly within the expert reference time, and across 24 configuration pairs ordered oppositely by the two metrics the majority selects the higher-H-GATE configuration in 92.5\% of samples.
Within the raters and configurations of these studies, continuous per-task time scoring therefore tracks distinctions people draw in model selection.

\subsection{Hardware Adaptation and MoE Efficiency}
\label{sec:hardware_architecture}
Stronger hardware deployments broaden the spectrum of model sizes that remain competitive in both accuracy and efficiency as shown in \cref{tab:hardware_choices}.
Analyzing cross-hardware performance reveals that optimal model capacities scale consistently with successive GPU generations, although deployable size remains inherently constrained by available memory limits.
When evaluated solely on spatial precision, larger models naturally dominate server-grade GPUs, but these accuracy gains frequently compromise efficiency due to increased execution time.
UI-Venus-1.5-8B keeps an SS-E Acc-S of 37.6\% to 37.7\% across RTX 3090, RTX 4090, and RTX 5090, yet mean latency drops from 6.44 to 1.99 and 1.37 seconds and H-GATE rises from 21.2 to 32.9 and 36.0, underscoring the importance of inference speed in interactive GUI environments.
Accuracy and efficiency optima diverge on A100, where UI-Venus-1.5-30B-A3B leads Acc at 32.2\% while the 8B model leads H-GATE at 35.4.

Server-grade hardware unequivocally dominates overall efficiency evaluations by accommodating advanced architectures that consumer GPUs cannot support.
As server deployments gain computing capacity and aggregate memory, they uniquely enable mixture-of-experts architectures to store massive overall weights while sparse activation effectively limits active parameters and per-token computation.
Evaluations on flagship hardware demonstrate that these sparse models maximize both accuracy and efficiency simultaneously, achieving superior precision and faster mean latency compared to their dense counterparts.
On H200, the mixture-of-experts Qwen3.5-397B-A17B achieves 42.2\% Acc and 43.0 H-GATE at a 1.45-second mean latency, compared with 32.2\%, 35.1, and 1.86 seconds for the dense Qwen3.5-27B.
By decoupling total parameter capacity from active inference costs, mixture-of-experts architectures bypass the traditional accuracy-efficiency bottleneck.
These findings highlight the substantial potential of sparse modeling paradigms and motivate the continuous scaling of GUI agents alongside expanding hardware capabilities.
\section{Conclusion}
\label{sec:conclusion}
We introduce ScreenSpot-Efficiency (SS-E), featuring 9,160 real-world tasks across 36 everyday scenarios, alongside a human-referenced framework that uses the H-GATE metric to benchmark accuracy and latency against per-task completion times from 11 experts. Strong performance on simpler benchmarks leaves complex, time-sensitive grounding far from human expert efficiency. Extended reasoning and zooming generally reduce SS-E efficiency. Visual-budget sweeps reveal intermediate efficiency peaks for some models, latency increases despite fewer tokens, and accuracy losses under excessive compression. Task-level diagnostics show how H-GATE distinguishes outcome and timing distributions hidden by aggregate accuracy and mean latency. Deployment comparisons identify promising larger sparse models as hardware capacity expands. A breakdown of failures and an explicit waiting intervention show that deciding when to observe and act is a capability distinct from localization and response speed. \Cref{app:limitations} discusses limitations. By exposing efficiency gaps relative to humans and the trade-offs behind them, our benchmark, metric, and findings give GUI grounding research a concrete basis for addressing the demands of practical applications.

{
    \small
    \bibliographystyle{ieeenat_fullname}
    \bibliography{main,appendix}
}

\clearpage
\appendix
\crefalias{section}{appendix}
\crefalias{subsection}{appendix}
\crefalias{subsubsection}{appendix}
\makeatletter
\renewcommand{\thetable}{\thesection\arabic{table}}
\renewcommand{\thefigure}{\thesection\arabic{figure}}
\renewcommand{\theequation}{\thesection\arabic{equation}}
\@addtoreset{table}{section}
\@addtoreset{figure}{section}
\@addtoreset{equation}{section}
\makeatother
\twocolumn[\centering\Large\textbf{Appendix}\vspace{1.0em}]

\section{Action Space}
\label{app:action_space}

This section details the GUI action space behind the spatiotemporal annotation protocol in \cref{sec:sse_construction}.
SS-E grounds everyday actions from activating a single control to transferring content between applications and specifying regions for editing or inspection.
\Cref{tab:app_action_space} organizes these tasks by the requested action and the answer geometry required for valid grounding.

\begin{table*}
\centering
\caption{\textbf{GUI action space grounded by SS-E.} Each task grounds one action, written in function form, based on its instruction and answer geometry. Platforms are Windows, macOS, Ubuntu, iOS, iPadOS, and Android.}
\label{tab:app_action_space}
\footnotesize
\setlength{\tabcolsep}{4pt}
\renewcommand{\arraystretch}{1.12}
\begin{tabular*}{\textwidth}{@{\extracolsep{\fill}}>{\raggedright\arraybackslash}p{0.15\textwidth}>{\raggedright\arraybackslash}p{0.145\textwidth}>{\raggedright\arraybackslash}p{0.2\textwidth}>{\raggedright\arraybackslash}p{0.225\textwidth}>{\raggedright\arraybackslash}p{0.125\textwidth}@{}}
\toprule
Action & Answer geometry & Function & Example task & Platforms \\
\midrule
\texttt{click(point)} & One region per valid interval & Activate a single control or item & Skip an ad once the countdown ends, admit a participant from the waiting room & All six \\
\texttt{click\_all(points)} & One region per required target & Activate every target specified by the instruction & Locate both playback controls, click all cyan spheres & iOS, Windows \\
\texttt{click\_seq(points)} & Ordered target regions & Activate targets sequentially in the stated order & Close three browser tabs in a given order & Windows \\
\texttt{drag(src, dst)} & Ordered source and destination regions & Move an item from source to destination & Drag a file into an email draft, reorder a shared playlist & Windows, Android, iOS, iPadOS \\
\texttt{crop(p1, p2)} & Two opposite corner points & Select a rectangle using two opposite corners & Crop an icon in an image editor & Windows \\
\texttt{outline(points)} & Polygon of 3 to 10 vertices & Enclose a target object with a polygon & Outline a shape on an image layer & Windows \\
\texttt{mark(boxes)} & One or more complete regions & Mark every region violating a stated rule & Mark spreadsheet cells with errors, find incomplete required fields & Windows \\
\texttt{swipe(path)} & Timestamped target regions & Swipe through targets while visible on screen & Slice fruit as it appears on screen & Android \\
\bottomrule
\end{tabular*}
\end{table*}

\paragraph{From control selection to spatial manipulation.}
Single-target activation, such as selecting a menu item, carousel card, or meeting control, is the same type of interaction evaluated by prior screenshot benchmarks.
Other actions require richer geometric outputs.
A drag specifies the source region followed by the destination region, cropping specifies two opposite corners, outlining traces a polygon, and visual checks mark every region violating a formatting or completeness rule.
These structured targets extend grounding from isolated controls to the content that users edit, transfer, and inspect across office documents, image editors, file managers, email clients, and web forms.

\paragraph{Order and timing.}
Ordered activation, such as closing browser tabs in a prescribed sequence, requires tracking targets as the interface updates after each click.
The temporal annotation separately specifies when a target is valid, so a spatially correct response is accepted only while the target satisfies the instruction.
In 439 tasks, the target changes position across successive valid intervals, for example as collaborators edit a shared document or a meeting roster changes.
Together, the geometric and temporal annotations cover pointer actions common in GUI agent action spaces, namely clicking, ordered clicking, dragging, rectangle and polygon selection, and swiping, under both routine and time-sensitive conditions.

\paragraph{Scope.}
SS-E evaluates where and when an agent must act. Keyboard entry and scrolling appear as task preconditions or as actions of other participants.
For drags that require a modifier key, such as holding Shift while dragging an image into a prompt composer, SS-E scores the required drag endpoints.

\section{Scenario and Application Coverage}
\label{sec:appendix-b-coverage}

\begin{figure}
  \centering
  \includegraphics[width=\linewidth]{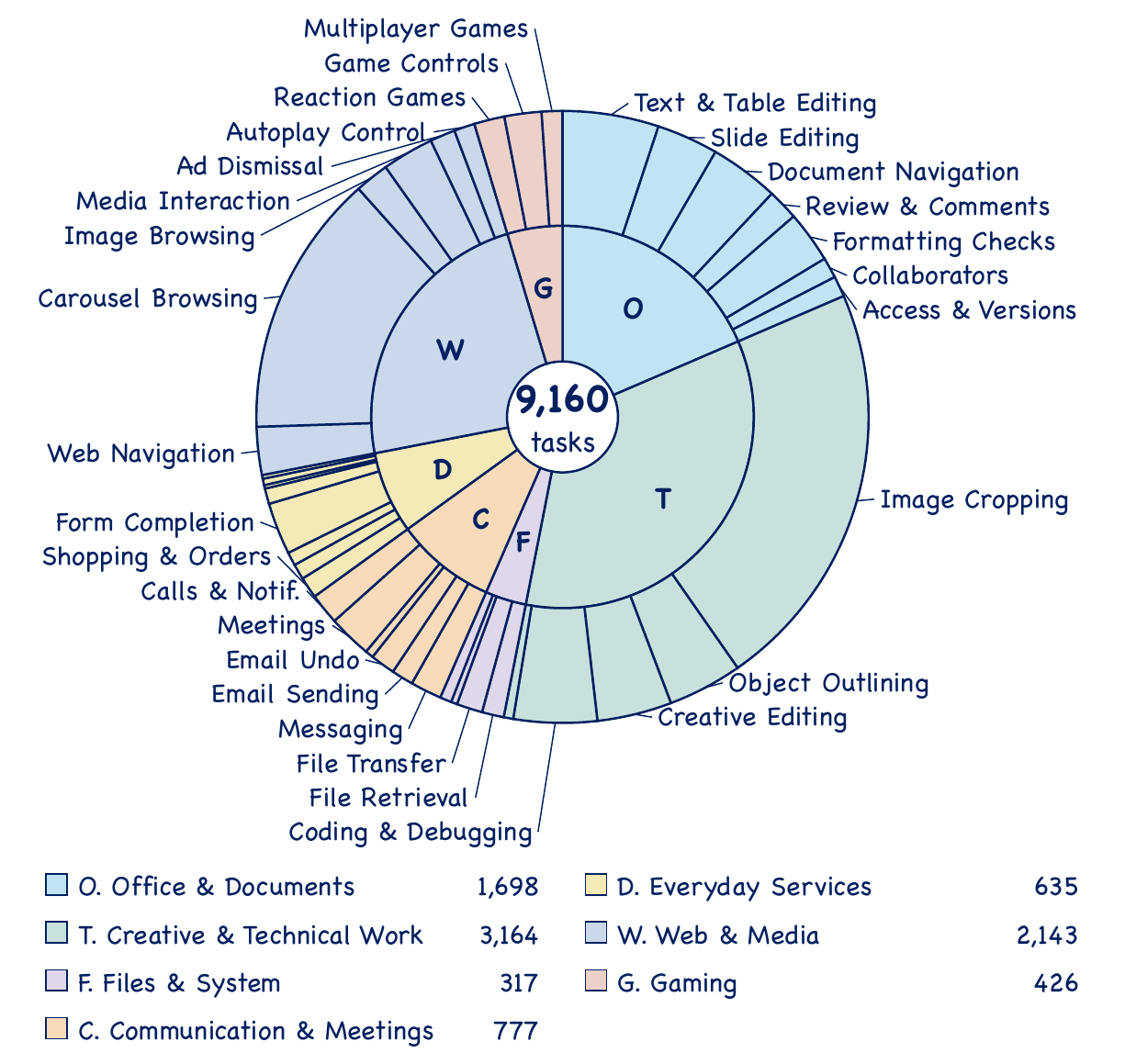}
  \caption{SS-E task distribution by application domain. The inner ring represents the seven domains and the outer ring their 39 application areas, with full definitions and counts in \cref{tab:appendix-b-domains-a} and \cref{tab:appendix-b-domains-b}.}
  \label{fig:app_domains}
\end{figure}
\begin{table*}
\centering
\caption{\textbf{Grounding modes and categories in SS-E.} Scenario and task counts are grouped by mode and category. Share is the percentage of all 9,160 tasks.}
\label{tab:appendix-b-levels}
\footnotesize
\setlength{\tabcolsep}{4pt}
\renewcommand{\arraystretch}{1.07}
\begin{tabular}{@{}>{\raggedright\arraybackslash}p{.15\textwidth}>{\raggedright\arraybackslash}p{.17\textwidth}>{\raggedright\arraybackslash}p{.40\textwidth}rrr@{}}
\toprule
Mode & Category & Definition & \#Scenarios & \#Tasks & Share \\
\midrule
\textbf{Routine grounding} &  &  & \textbf{13} & \textbf{5,149} & \textbf{56.21\%} \\
 & A. Search \& Selection & Find a control or match a target to a text or visual reference. & 4 & 1,595 & 17.41\% \\
 & B. Editing \& Dragging & Locate points and regions for editing, dragging, or reordering. & 4 & 2,634 & 28.76\% \\
 & C. Sequential Operations & Locate successive targets as actions update the interface. & 2 & 220 & 2.40\% \\
 & D. Visual Checks & Identify visible errors, omissions, or inconsistent formatting. & 3 & 700 & 7.64\% \\
\addlinespace[3pt]
\textbf{Time-sensitive grounding} &  &  & \textbf{23} & \textbf{4,011} & \textbf{43.79\%} \\
 & E. Event-Driven Interaction & Locate targets whose availability or position changes with an interface event. & 9 & 2,205 & 24.07\% \\
 & F. Collaboration \& Competition & Locate targets whose identity, position, or availability depends on another participant. & 11 & 1,384 & 15.11\% \\
 & G. Real-Time Gameplay & Locate controls or targets in real-time game interactions. & 3 & 422 & 4.61\% \\
\bottomrule
\end{tabular}
\end{table*}

\begin{table*}
\centering
\renewcommand{\thetable}{\thesection\arabic{table}a}
\providecommand{\theHtable}{\arabic{table}}
\renewcommand{\theHtable}{\thesection.\arabic{table}.a}
\caption{\textbf{Routine grounding scenarios in SS-E.} Each row defines the grounding requirement for one scenario. Share is the percentage of all 9,160 tasks.}
\label{tab:appendix-b-scenarios-a}
\footnotesize
\setlength{\tabcolsep}{4pt}
\renewcommand{\arraystretch}{1.07}
\begin{tabular}{@{}l>{\raggedright\arraybackslash}p{.25\textwidth}>{\raggedright\arraybackslash}p{.52\textwidth}rr@{}}
\toprule
ID & Scenario & Definition & \#Tasks & Share \\
\midrule
A1 & Routine Control Selection & Locate a control for a routine selection or navigation action. & 1,002 & 10.94\% \\
A2 & Dense Content Search & Disambiguate a target among dense text, table entries, or visually similar icons. & 88 & 0.96\% \\
A3 & Content-Based Retrieval & Find the image, page, or slide thumbnail that matches a text description. & 255 & 2.78\% \\
A4 & Reference Image Matching & Match the target to a supplied reference image. & 250 & 2.73\% \\
\addlinespace[3pt]
B1 & Region Cropping & Locate the corners of a rectangular crop. & 1,994 & 21.77\% \\
B2 & Contour Outlining & Locate multiple points that trace an object boundary. & 360 & 3.93\% \\
B3 & Cross-App Dragging & Locate both the source file and the drop region across application windows. & 250 & 2.73\% \\
B4 & Reordering \& Adjustment & Locate the item and destination for reordering or the control for an adjustment. & 30 & 0.33\% \\
\addlinespace[3pt]
C1 & Sequential Tab Closing & Locate the next tab to close in a sequence of tab closures. & 120 & 1.31\% \\
C2 & Sequential Item Deletion & Locate successive items for deletion as the list updates. & 100 & 1.09\% \\
\addlinespace[3pt]
D1 & Explicit Error Localization & Locate a visibly flagged error in a cell, an upload list, or a form field. & 200 & 2.18\% \\
D2 & Missing Required Fields & Locate an unfinished field by combining its required marker with its completion state. & 250 & 2.73\% \\
D3 & Formatting Inconsistencies & Locate text or cells whose font, size, or spacing differs from neighboring content. & 250 & 2.73\% \\
\bottomrule
\end{tabular}
\end{table*}
\addtocounter{table}{-1}

\begin{table*}
\centering
\renewcommand{\thetable}{\thesection\arabic{table}b}
\providecommand{\theHtable}{\arabic{table}}
\renewcommand{\theHtable}{\thesection.\arabic{table}.b}
\caption{\textbf{Time-sensitive grounding scenarios in SS-E.} Each row defines the grounding requirement for one scenario. Share is the percentage of all 9,160 tasks.}
\label{tab:appendix-b-scenarios-b}
\footnotesize
\setlength{\tabcolsep}{4pt}
\renewcommand{\arraystretch}{1.07}
\begin{tabular}{@{}l>{\raggedright\arraybackslash}p{.25\textwidth}>{\raggedright\arraybackslash}p{.52\textwidth}rr@{}}
\toprule
ID & Scenario & Definition & \#Tasks & Share \\
\midrule
E1 & Carousel Timing & Locate a carousel target while its card is visible. & 1,274 & 13.91\% \\
E2 & Send After Upload & Locate the send control once the upload completes. & 80 & 0.87\% \\
E3 & Ad Skip \& Close & Locate the skip or close control once the advertisement permits the action. & 122 & 1.33\% \\
E4 & Autoplay Cancellation & Locate the cancel control before the next item starts playing automatically. & 100 & 1.09\% \\
E5 & Incoming Requests & Locate the response control associated with an incoming call, invitation, notification, or transfer request. & 150 & 1.64\% \\
E6 & Expiring Undo & Locate the temporary undo control after sending or archiving and before the control expires. & 120 & 1.31\% \\
E7 & Event-Triggered Controls & Locate a media or editing control when the specified content or timing condition is met. & 152 & 1.66\% \\
E8 & Post-Load Localization & Locate the requested target after the interface finishes loading. & 107 & 1.17\% \\
E9 & Chat List Reordering & Track a conversation by identity after a new message changes its list position. & 100 & 1.09\% \\
\addlinespace[3pt]
F1 & Text \& Table Co-editing & Locate text or cells whose identity or position depends on edits by another participant. & 350 & 3.82\% \\
F2 & Slide Co-editing & Locate a slide or object after a collaborator moves, modifies, or reorders it. & 200 & 2.18\% \\
F3 & Collaborator Cues & Use a collaborator's cursor, selection, or avatar to identify the requested target. & 100 & 1.09\% \\
F4 & Live Review \& Comments & Locate the active comment, reply, or document anchor as the review state changes. & 150 & 1.64\% \\
F5 & Permissions \& Sync & Locate a target for editing, synchronization, or version recovery as access or document state changes. & 100 & 1.09\% \\
F6 & Live Meeting Management & Locate a meeting participant or waiting room control as attendance or participant state changes. & 210 & 2.29\% \\
F7 & Joint Ordering & Locate the relevant order item or control as other participants submit, revise, or lock the order. & 49 & 0.53\% \\
F8 & Shared Playlist Updates & Track the requested track or control as other participants change the shared queue. & 50 & 0.55\% \\
F9 & Seat Contention & Locate the requested seat while it is available amid other participants' reservations and releases. & 75 & 0.82\% \\
F10 & Shared To-do Updates & Locate the relevant task after another participant completes, removes, or revises an item in the shared list. & 50 & 0.55\% \\
F11 & Human–Agent Handoffs & Locate the next target after a person changes the view and hands control to the agent. & 50 & 0.55\% \\
\addlinespace[3pt]
G1 & Reaction Games & Locate a moving or briefly visible game target at the required moment. & 147 & 1.60\% \\
G2 & Complex Game Controls & Locate a control for a local action in a complex game interface. & 175 & 1.91\% \\
G3 & Multiplayer Games & Locate a target for a local action during multiplayer competition or cooperation. & 100 & 1.09\% \\
\bottomrule
\end{tabular}
\end{table*}

\subsection{Scenario Coverage}
\label{app:scenario_coverage}
This section provides the full scenario definitions and counts behind \cref{fig:sse_distribution} alongside a complementary breakdown of SS-E by application domain.
Across everyday work and life, SS-E captures the primary ways grounding speed shapes interaction outcomes by varying the three properties that govern delay and localization costs.
The taxonomy in \cref{tab:appendix-b-levels} structures all 36 scenarios around what must be localized, how targets evolve, and when an action stays valid.
Routine grounding spans locating a single control or matching a reference, specifying multiple points for edits and drags, tracking sequential targets after an action, and detecting visual errors or omissions.
For example, identifying a required field demands judging live form state, whereas cross-app dragging requires localizing both source and destination endpoints.

Time-sensitive grounding comprises 4,011 tasks and extends these localization demands to dynamic interfaces that evolve during inference.
As detailed in \cref{tab:appendix-b-scenarios-a} and \cref{tab:appendix-b-scenarios-b}, action validity hinges on an interface event, another participant's input, or a shifting game state.
An undo control expires, a collaborator's edit relocates the target, or another user reserves a seat before the agent responds.
Delay therefore imposes distinct operational penalties across personal and professional contexts.
A slow agent loses access to the control, clicks an outdated coordinate, or forfeits a transient opportunity.
Combined with the eight action families in \cref{app:action_space}, these two modes cover the answer-geometry and timing regimes left untested by prior screenshot benchmarks, which restrict evaluation to static single-target tasks as summarized in \cref{tab:benchmark_comparison}.

\begin{table*}
\centering
\renewcommand{\thetable}{\thesection\arabic{table}a}
\providecommand{\theHtable}{\arabic{table}}
\renewcommand{\theHtable}{\thesection.\arabic{table}.a}
\caption{\textbf{Application domains and areas in SS-E, part 1 of 2.} Bold rows give domain totals and their shares of all 9,160 tasks.}
\label{tab:appendix-b-domains-a}
\footnotesize
\setlength{\tabcolsep}{4pt}
\renewcommand{\arraystretch}{1.07}
\begin{tabular}{@{}l>{\raggedright\arraybackslash}p{.24\textwidth}>{\raggedright\arraybackslash}p{.53\textwidth}rr@{}}
\toprule
ID & Domain/Area & Definition & \#Tasks & Share \\
\midrule
\textbf{O} & \textbf{Office \& Documents} &  & \textbf{1,698} & \textbf{18.54\%} \\
O1 & Text \& Table Editing & Locate document text, cells, formulas, or controls for editing and error checks. & 465 & 5.08\% \\
O2 & Slide Editing & Locate slide content, objects, or controls for insertion, editing, and reordering. & 298 & 3.25\% \\
O3 & Document Navigation & Find document content, views, or thumbnails and locate controls for managing document tabs. & 335 & 3.66\% \\
O4 & Review \& Comments & Locate document comments, review controls, or reply targets. & 150 & 1.64\% \\
O5 & Formatting Checks & Identify text whose font, size, or spacing differs from its surroundings. & 250 & 2.73\% \\
O6 & Following Collaborators & Locate the cell or page indicated by a collaborator's cursor, selection, or avatar. & 100 & 1.09\% \\
O7 & Access \& Versions & Locate document controls for access changes, synchronization, reloading, or version recovery. & 100 & 1.09\% \\
\addlinespace[3pt]
\textbf{T} & \textbf{Creative \& Technical Work} &  & \textbf{3,164} & \textbf{34.54\%} \\
T1 & Image Cropping & Locate rectangular crop corners in an image editing interface. & 1,994 & 21.77\% \\
T2 & Object Outlining & Locate the vertices of a polygon that traces an object's contour. & 360 & 3.93\% \\
T3 & Creative Editing & Locate menus, panels, objects, or settings in visual, 3D, and audiovisual creation tools. & 361 & 3.94\% \\
T4 & Coding \& Debugging & Locate development controls or content such as code, logs, debugger entries, and version history. & 403 & 4.40\% \\
T5 & Engineering \& Analysis & Locate controls in design, engineering, statistical analysis, or scientific plotting tools. & 46 & 0.50\% \\
\addlinespace[3pt]
\textbf{F} & \textbf{Files \& System} &  & \textbf{317} & \textbf{3.46\%} \\
F1 & File Retrieval & Locate a target file in a file manager, cloud drive, or file card. & 106 & 1.16\% \\
F2 & File Transfer & Locate controls or file entries for uploads, cloud transfers, attachments, or upload failures. & 125 & 1.36\% \\
F3 & File Management & Locate controls for archives, trash, folder sorting, download history, or disk space. & 30 & 0.33\% \\
F4 & Apps \& Settings & Locate controls for application installation, launch, system settings, or virtual machines. & 56 & 0.61\% \\
\addlinespace[3pt]
\textbf{C} & \textbf{Communication \& Meetings} &  & \textbf{777} & \textbf{8.48\%} \\
C1 & Messaging & Locate a conversation, chat attachment, or control for messaging. & 151 & 1.65\% \\
C2 & Email Sending & Locate controls for drafting an email, adding an attachment, or sending after upload. & 106 & 1.16\% \\
C3 & Email Undo & Locate the control that reverses an email send or archive action. & 120 & 1.31\% \\
C4 & Inbox Cleanup & Locate the next email to delete as the inbox updates. & 40 & 0.44\% \\
C5 & Meeting Management & Locate meeting participants or controls as hand raises, mute state, or waiting room membership changes. & 210 & 2.29\% \\
C6 & Calls \& Notifications & Locate a response control for a call, invitation, notification, or transfer request. & 150 & 1.64\% \\
\bottomrule
\end{tabular}
\end{table*}
\addtocounter{table}{-1}

\begin{table*}
\centering
\renewcommand{\thetable}{\thesection\arabic{table}b}
\providecommand{\theHtable}{\arabic{table}}
\renewcommand{\theHtable}{\thesection.\arabic{table}.b}
\caption{\textbf{Application domains and areas in SS-E, part 2 of 2.} Columns follow \cref{tab:appendix-b-domains-a}.}
\label{tab:appendix-b-domains-b}
\footnotesize
\setlength{\tabcolsep}{4pt}
\renewcommand{\arraystretch}{1.07}
\begin{tabular}{@{}l>{\raggedright\arraybackslash}p{.24\textwidth}>{\raggedright\arraybackslash}p{.53\textwidth}rr@{}}
\toprule
ID & Domain/Area & Definition & \#Tasks & Share \\
\midrule
\textbf{D} & \textbf{Everyday Services} &  & \textbf{635} & \textbf{6.93\%} \\
D1 & Shopping \& Orders & Locate items or controls for shopping, cart management, shared orders, or order handoffs. & 103 & 1.12\% \\
D2 & Seat Booking & Locate a seat for booking as availability changes. & 75 & 0.82\% \\
D3 & Shared To-dos & Locate entries for updates, completion, assignment, or handoffs in shared lists. & 65 & 0.71\% \\
D4 & Form Completion & Locate an unfilled required field or validation control in a service form. & 251 & 2.74\% \\
D5 & Form Correction & Locate explicitly flagged errors in account, contact, address, booking, or order forms. & 75 & 0.82\% \\
D6 & Maps \& Travel & Locate controls for maps, routes, locations, rides, accommodation, or public transit. & 17 & 0.19\% \\
D7 & Health \& Fitness & Locate health information, activity records, or workout controls. & 31 & 0.34\% \\
D8 & Weather & Locate cities, forecasts, units, or settings in a weather interface. & 18 & 0.20\% \\
\addlinespace[3pt]
\textbf{W} & \textbf{Web \& Media} &  & \textbf{2,143} & \textbf{23.40\%} \\
W1 & Web Navigation & Locate web search targets, links, browser tabs, bookmarks, or newly loaded content. & 232 & 2.53\% \\
W2 & Carousel Browsing & Locate a requested card or banner in a carousel. & 1,274 & 13.91\% \\
W3 & Image Browsing & Find an image in an album, gallery, or visual asset grid. & 165 & 1.80\% \\
W4 & Media Interaction & Locate playback, queue, social media, or editing targets under the relevant content or timing conditions. & 250 & 2.73\% \\
W5 & Ad Dismissal & Locate an advertisement's dismiss control once it becomes available. & 122 & 1.33\% \\
W6 & Autoplay Control & Locate the control that cancels automatic playback of the next item. & 100 & 1.09\% \\
\addlinespace[3pt]
\textbf{G} & \textbf{Gaming} &  & \textbf{426} & \textbf{4.65\%} \\
G1 & Reaction Games & Locate a moving or brief target at the required moment in a game. & 147 & 1.60\% \\
G2 & Game Controls & Locate controls in complex game interfaces or routine game menus. & 179 & 1.95\% \\
G3 & Multiplayer Games & Locate targets for local actions in multiplayer competition or cooperation. & 100 & 1.09\% \\
\bottomrule
\end{tabular}
\end{table*}

\begin{figure}
  \centering
  \includegraphics[width=\linewidth]{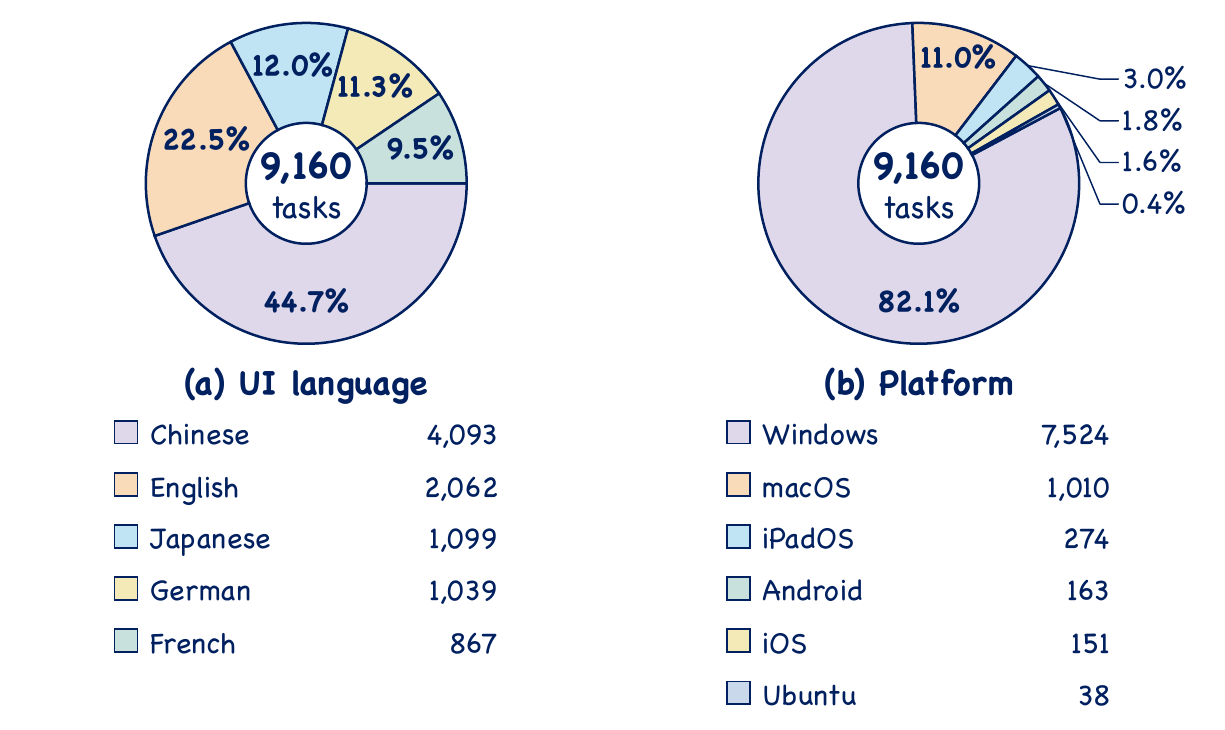}
  \caption{SS-E task distribution by (a) UI language and (b) platform.}
  \label{fig:app_lang_platform}
\end{figure}
\subsection{Application Coverage}
\label{app:application_coverage}
The application taxonomy in \cref{tab:appendix-b-domains-a} and \cref{tab:appendix-b-domains-b}, visualized in \cref{fig:app_domains}, covers 39 areas across seven domains.
Work tasks range from reviewing shared document comments and admitting meeting participants to locating debugging controls and editing images.
Everyday tasks span web forms, shared lists, seat booking, and travel services alongside browsing, media playback, and games.
File and system operations add locating documents, transferring files, and adjusting system settings.
Together, these domains show that the same grounding requirements recur across distinct applications.
For example, live interface updates shift the relevant target across a shared order, a meeting roster, and a media queue alike.

\subsection{Language and Platform}
\label{app:language_platform}
Across its 9,160 tasks, SS-E spans five UI languages and six platforms, as shown in \cref{fig:app_lang_platform}.
Chinese accounts for 44.68\% of these tasks, while the remainder spans English, Japanese, German, and French.
Windows represents 82.14\% of the tasks, and the remaining tasks cover macOS, Ubuntu, iPadOS, iOS, and Android to broaden evaluation across desktop, tablet, and mobile interfaces.

\subsection{Resolution Distribution}
\label{app:resolution}
\Cref{tab:appendix-b-resolution} categorizes all 9,160 SS-E clips by display class.
Every clip retains its recorded resolution, and the dataset spans 45 distinct sizes ranging by pixel count from $1280\times584$ to $3840\times2160$.

\begin{table}
\centering
\caption{\textbf{Display resolutions in SS-E.} Desktop classes partition clips by long-edge length into HD below 1,500 pixels, Full HD from 1,500 to 2,299, QHD from 2,300 to 2,999, and 4K at 3,000 or above. QHD is further grouped by aspect ratio. Resolution reports the most frequent size in each class, and Share gives the percentage of all 9,160 tasks.}
\label{tab:appendix-b-resolution}
\footnotesize
\setlength{\tabcolsep}{4pt}
\begin{tabular}{@{}llrr@{}}
\toprule
Display class & Resolution & \#Tasks & Share \\
\midrule
\textbf{Desktop and laptop} & & \textbf{8,572} & \textbf{93.58\%} \\
\hspace{0.6em}HD & $1276\times720$ & 428 & 4.67\% \\
\hspace{0.6em}Full HD & $1728\times1080$ & 231 & 2.52\% \\
\hspace{0.6em}QHD, 16:9 or wider & $2560\times1440$ & 7,178 & 78.36\% \\
\hspace{0.6em}QHD, 16:10 & $2560\times1600$ & 728 & 7.95\% \\
\hspace{0.6em}4K & $3840\times2160$ & 7 & 0.08\% \\
\textbf{Tablet and phone} & & \textbf{588} & \textbf{6.42\%} \\
\hspace{0.6em}Landscape & $1600\times1000$ & 297 & 3.24\% \\
\hspace{0.6em}Portrait & $1180\times2556$ & 291 & 3.18\% \\
\midrule
Total & & 9,160 & 100.00\% \\
\bottomrule
\end{tabular}
\end{table}

\paragraph{Coverage of everyday displays.}
SS-E spans the display classes used across daily work and life.
Desktop and laptop recordings cover HD and Full HD laptop screens, QHD monitors in 16:9 and 16:10 layouts, and 4K displays.
Tablet and phone recordings contribute landscape screens alongside portrait screens at the 9:19.5 and 9:20 aspect ratios of current phones.
Across all clips, those classified as 16:9 or 16:10 account for 96.22\% of tasks, while portrait clips account for 3.18\%.
QHD constitutes the largest class and 86.39\% of tasks are recorded at QHD or above, so most tasks present the high-resolution visual inputs whose token cost \cref{app:visual_budget} analyzes.
An agent evaluated on SS-E therefore grounds targets from compact phone and laptop displays to high-resolution monitors, which are the display types found on everyday devices.

\section{Derivation and Properties of H-GATE}
\label{app:hgate}

This section derives H-GATE from the per-task quantities in \cref{sec:hgate} and clarifies the physical meaning of each intermediate term.
It then establishes the reference values and algebraic properties that govern how H-GATE scores individual grounding attempts before averaging across the benchmark.

\subsection{Task-Level Construction}
\label{app:hgate_construction}

\paragraph{Completion credit and elapsed time.}
Consider a benchmark of $N\geq1$ tasks.
For each task $i$, let $c_i\in\{0,1\}$ denote full correctness, $t_i>0$ the complete grounding latency, and $h_i>0$ the mean completion time across $K$ human experts.
On SS-E, $c_i=1$ requires both spatial correctness and a response within the valid time interval.
Every configuration evaluated on task $i$ shares the same reference $h_i$, and the latency ratio $r_i=t_i/h_i$ expresses model latency in expert-time units, where $r_i=1$ marks expert speed.
Both $t_i$ and $h_i$ represent end-to-end durations, and $h_i$ serves as the efficiency target an agent must meet to replace a proficient human at this grounding step, as detailed in \cref{app:expert_times}.

Earned duration management values completed work by the reference duration it represents~\cite{khamooshi2014edm}.
Adopting expert time as that reference yields the completion credit $W_i$ and the time difference $\Delta_i$:
\begin{equation}
\label{eq:app_earned}
W_i=c_ih_i,\qquad \Delta_i=W_i-t_i.
\end{equation}
A correct grounding earns $h_i$ seconds of reference credit, whereas every attempt, whether correct or failed, incurs its full measured latency.

\paragraph{Common time scale.}
Raw time differences in seconds are unbounded and give greater weight to long tasks.
Following symmetric comparisons in economics and finance~\cite{decker2020changing,morariupatrichi2022state} and shared time fractions in human--robot interaction~\cite{amine2023vahr}, we normalize both durations into complementary shares that sum to one and preserve their ratio:
\begin{gather}
p_i+q_i=1,\qquad \frac{p_i}{q_i}=\frac{h_i}{t_i},
\label{eq:app_share_constraints}\\[3pt]
\begin{aligned}
p_i&=\frac{h_i}{h_i+t_i}=\frac{1}{1+r_i},\\[3pt]
q_i&=\frac{t_i}{h_i+t_i}=\frac{r_i}{1+r_i}.
\end{aligned}
\label{eq:app_shares}
\end{gather}
The expert share $p_i$ equals $1/2$ at expert speed, exceeds $1/2$ when the model is faster, and tends to zero as model latency grows without bound.

\paragraph{Normalized time balance.}
On this bounded scale, a correct response earns $c_ip_i$ while every attempt expends $q_i$, which yields the normalized time balance
\begin{equation}
\label{eq:app_balance}
B_i=c_ip_i-q_i=\frac{\Delta_i}{h_i+t_i}=\frac{c_i-r_i}{1+r_i}.
\end{equation}
For a correct response, $B_i=(1-r_i)/(1+r_i)$ lies in $(-1,1)$, is positive exactly when the model is faster than the expert, and reverses sign when $r_i$ is replaced by $1/r_i$.
For a failed attempt, $B_i=-q_i$ lies in $(-1,0)$ and directly records the time cost incurred without completion credit.

\paragraph{Calibration of the intermediate score.}
We combine correctness and time balance through an affine intermediate score $u_i=\kappa+\alpha c_i+\beta B_i$ with baseline offset $\kappa$, completion weight $\alpha$, and balance weight $\beta$.
Three calibration conditions uniquely determine these parameters:
\begin{align}
\lim_{r_i\to0^+}u_i\big|_{c_i=0}&=\kappa=0,
\label{eq:app_cal_fail}\\
u_i\big|_{c_i=1,\,r_i=1}&=\kappa+\alpha=1,
\label{eq:app_cal_expert}\\
\lim_{r_i\to\infty}u_i\big|_{c_i=1}&=\kappa+\alpha-\beta=0.
\label{eq:app_cal_slow}
\end{align}
\Cref{eq:app_cal_fail} requires that an instantaneous failure neither earns credit nor wastes time.
\Cref{eq:app_cal_expert} anchors unit utility to a correct response delivered at exact expert speed, where the time balance vanishes.
\Cref{eq:app_cal_slow} ensures that a correct response after an unbounded delay retains zero utility.
Solving these conditions yields $\kappa=0$ and $\alpha=\beta=1$, which gives
\begin{align}
u_i&=c_i+B_i=c_i-1+\left(1+c_i\right)p_i
\label{eq:app_u}\\
&=\begin{cases}
2p_i=\dfrac{2}{1+r_i}, & c_i=1,\\[6pt]
-q_i=-\dfrac{r_i}{1+r_i}, & c_i=0.
\end{cases}
\label{eq:app_branches}
\end{align}
Thus, a correct response earns twice the expert share, whereas a failure incurs a penalty equal to the model share.

\subsection{Normalization and Aggregation}
\label{app:hgate_normalization}
For finite positive times, \cref{eq:app_branches} places correct responses in $(0,2)$ and failures in $(-1,0)$.
The unique increasing affine transformation mapping the endpoints $-1$ and $2$ to $0$ and $100$ defines the normalized task score
\begin{equation}
\label{eq:app_task_score}
s_i=\frac{100}{3}\left(u_i+1\right)=\frac{100}{3}\Bigl[c_i+\left(1+c_i\right)p_i\Bigr],
\end{equation}
and H-GATE aggregates these task scores with uniform weight across the benchmark:
\begin{equation}
\label{eq:app_hgate}
\mathrm{H\text{-}GATE}=\frac{1}{N}\sum_{i=1}^{N}s_i=\frac{100}{3}\left(\bar u+1\right),\qquad \bar u=\frac{1}{N}\sum_{i=1}^{N}u_i.
\end{equation}
The expanded form of \cref{eq:app_task_score} separates the completion credit from the time-dependent contribution.
A failure receives $\frac{100}{3}p_i$, which scales the expert share to one third of the full range.
A correct response receives $\frac{100}{3}+\frac{200}{3}p_i$, which adds a fixed correctness offset to twice the scaled expert share.
Because scoring precedes aggregation, every failed attempt remains in the denominator and contributes its measured latency.

The unshifted percentage $S_0=100\bar u$ converts to the final H-GATE scale via
\begin{equation}
\label{eq:app_scale_conversion}
\mathrm{H\text{-}GATE}=\frac{S_0+100}{3}.
\end{equation}

\subsection{Reference Values}
\label{app:hgate_references}
\Cref{tab:app_hgate_references} summarizes the four reference anchors across the intermediate mean $\bar u$, the unshifted percentage $S_0$, and the final H-GATE scale.

\begin{table}
\centering
\caption{\textbf{Reference values of H-GATE.} The mean intermediate score $\bar u$, unshifted percentage $S_0=100\bar u$, and final H-GATE score are shown for each reference condition. The latency ratio is $r_i=t_i/h_i$.}
\label{tab:app_hgate_references}
\footnotesize
\setlength{\tabcolsep}{3pt}
\begin{tabular}{@{}lccc@{}}
\toprule
Condition & $\bar u$ & $S_0$ & H-GATE \\
\midrule
All tasks fail, $r_i\to\infty$ & $-1$ & $-100$ & $0$ \\
Correct credit offsets failure time & $0$ & $0$ & $33.33$ \\
All tasks correct, $r_i=1$ & $1$ & $100$ & $66.67$ \\
All tasks correct, $r_i\to0$ & $2$ & $200$ & $100$ \\
\bottomrule
\end{tabular}
\end{table}

\paragraph{0 and 100 as limits.}
H-GATE approaches 0 only when every task fails with $r_i\to\infty$, where unbounded latency yields zero completed work.
It approaches 100 only when every task succeeds with $r_i\to0$, which corresponds to instantaneous correct grounding.
Because finite positive latencies enforce $0<s_i<100$ on every task, every measured configuration satisfies $0<\mathrm{H\text{-}GATE}<100$.

\paragraph{33.33 as the utility-neutral reference.}
The equality $\mathrm{H\text{-}GATE}=100/3$ holds precisely when $\bar u=0$, matching the zero point of $S_0$.
Substituting \cref{eq:app_branches} decomposes the deviation from this neutral baseline as
\begin{equation}
\label{eq:app_neutral}
\mathrm{H\text{-}GATE}-\frac{100}{3}=\frac{100}{3N}\left(\sum_{i:\,c_i=1}2p_i-\sum_{i:\,c_i=0}q_i\right).
\end{equation}
At this reference, displayed as 33.33, credit from correct responses exactly balances the normalized time cost of failed attempts.
This identity directly yields the decomposition in \cref{fig:metric_diagnostics}b into a neutral baseline, positive correct-response contributions, and negative failed-attempt time penalties.
Scores below the utility-neutral reference of approximately 33.33 therefore indicate that the normalized cost of failed attempts outweighs correct-completion credit, even when a model solves a substantial fraction of tasks.

\paragraph{Fast failures approach the neutral reference.}
By \cref{eq:app_task_score}, an agent that fails every task scores $\frac{100}{3}\bar p$, which tends to 33.33 as its latency tends to zero.
The neutral reference is therefore attainable in the limit without a single correct response.
It marks zero net utility under the calibration in \cref{eq:app_cal_fail}, where an instantaneous failure costs nothing, and it is not a lower bound on grounding ability.
H-GATE consequently places such a hypothetical agent above any configuration whose failed-attempt time outweighs its correct-completion credit.
Among the 53 H200 configurations in \cref{tab:d_h200_screenspot,tab:d_h200_screenspot_part2}, 33 display SS-E H-GATE below 33.3 at accuracies from 6.9\% to 36.2\%, and Holo2-235B-A22B displays 33.3 at 36.4\% accuracy.
These configurations ground a share of tasks that the hypothetical agent cannot, yet they spend more expert-normalized time on failures than their correct responses return.
H-GATE alone thus does not certify grounding ability near or below the reference, and every result table reports Acc beside it for this reason.

\paragraph{66.67 as the expert-speed reference.}
When a model answers every task correctly at exact expert speed, $c_i=1$ and $r_i=1$ hold for all $i$, so $u_i=1$, $\bar u=1$, $S_0=100$, and $\mathrm{H\text{-}GATE}=200/3\approx66.67$, where the decimal is the rounded display value.
The reference applies across benchmarks because it fixes correctness and relative speed on every task.
The human row in \cref{tab:hgate_main} uses the experts' observed correctness and completion times. Its five benchmark scores span 66.2 to 66.7 at 99.0\% to 100.0\% accuracy.
On any single task, a correct response exceeds the exact expert-speed reference, displayed as 66.67, precisely when the model is faster than the expert, which directly links per-task scoring to the Acc@Human columns in \cref{sec:app_full_gpu_results}.

\subsection{Properties}
\label{app:hgate_properties}

\paragraph{Correctness separates outcomes.}
On every task, failures lie strictly in $(0,100/3)$ and correct responses in $(100/3,100)$, so any correct grounding outscores any failure regardless of latency.
At a fixed latency ratio, turning a failure into a correct response increases the intermediate score by
\begin{equation}
\label{eq:app_correctness_gain}
u_i\big|_{c_i=1}-u_i\big|_{c_i=0}=1+p_i>0,
\end{equation}
which equals one completion unit plus the expert share.
Denoting benchmark accuracy by $A$, these task-level bounds yield $\frac{100}{3}A<\mathrm{H\text{-}GATE}<\frac{100}{3}(1+2A)$, so reaching the expert-speed reference displayed as 66.67 requires accuracy strictly above 50\% regardless of inference speed.

\paragraph{Latency is always penalized.}
Holding correctness fixed, the partial derivatives of the intermediate score are
\begin{equation}
\label{eq:app_derivatives}
\frac{\partial u_i}{\partial t_i}=-\frac{(1+c_i)h_i}{(h_i+t_i)^2}<0,\qquad
\frac{\partial u_i}{\partial h_i}=\frac{(1+c_i)t_i}{(h_i+t_i)^2}>0.
\end{equation}
Additional latency strictly reduces the score for both correct and failed attempts, whereas a longer expert reference raises it because the model occupies a smaller share of the common scale.
At the same expert reference time and latency ratio, the marginal time penalty for a correct response doubles that of a failure, which captures how delay erodes the utility of an otherwise accurate localization.
On a shared task set, any model that matches or exceeds another in both correctness and speed on every task achieves a strictly higher H-GATE whenever either advantage is strict on at least one task.

\paragraph{Correctness and latency stay paired.}
Let $\bar p$ denote the mean expert share across all $p_i$.
Averaging \cref{eq:app_u} yields
\begin{equation}
\label{eq:app_pairing}
\bar u=A-1+\left(1+A\right)\bar p+\operatorname{Cov}_N(c,p),
\end{equation}
where $\operatorname{Cov}_N$ denotes the empirical covariance with divisor $N$.
At fixed accuracy and fixed time shares, H-GATE increases when correct predictions align with faster responses relative to the expert, a joint dependence that aggregate accuracy and mean latency discard.
For instance, on two tasks with $h_i=1$\,s and model latencies of 0.5\,s and 2\,s, solving only the fast task yields $\mathrm{H\text{-}GATE}=400/9\approx44.44$, whereas solving only the slow task yields $350/9\approx38.89$, despite identical 50\% accuracy and 1.25\,s mean latency.

\paragraph{Invariance to time units and ranking preservation.}
Because each task score depends on $t_i$ and $h_i$ solely through $r_i$, the same rescaling of both durations, including a change of time unit, leaves the score unchanged. For a given correctness outcome, equal relative speeds produce equal scores across tasks.
For any two configurations $X$ and $Y$, \cref{eq:app_hgate} implies $G_X-G_Y=\frac{100}{3}(\bar u_X-\bar u_Y)$, so the linear normalization preserves all rankings and ties relative to $S_0$ while scaling every pairwise margin by a constant factor.

\paragraph{Combination across task subsets.}
Across disjoint task subsets of sizes $N_g$ with subset scores $G_g$, the score on their union satisfies
\begin{equation}
\label{eq:app_pooled}
G_{\mathrm{pool}}=\frac{\sum_gN_gG_g}{\sum_gN_g}.
\end{equation}
Every task receives equal weight in direct and pooled scoring. A complete partition by category, language, or platform recovers the overall SS-E score when pooling unrounded subgroup scores over the same task set. \Cref{app:setup_analysis} reports results by these groupings.
Appending the same set of task scores to two configurations scales their gap by $N/(N+m)$ for $m$ appended tasks and preserves their relative order.
The five-benchmark mean in \cref{tab:e2_posttraining} is a separate summary that weights each benchmark equally.

\subsection{Sensitivity and Relation to Simpler Summaries}
\label{app:hgate_sensitivity}

\paragraph{Explicit failure cost.}
The calibration charges a failed attempt only its elapsed time, whereas a wrong action in deployment can also trigger recovery steps.
A constant cost $\lambda\geq0$ per failure gives the variant $u_i^{\lambda}=u_i-\lambda(1-c_i)$, and averaging on the scale of \cref{eq:app_hgate} yields
\begin{equation}
\label{eq:app_failure_cost}
\mathrm{H\text{-}GATE}_{\lambda}=\mathrm{H\text{-}GATE}-\frac{100}{3}\lambda\left(1-A\right),
\end{equation}
where $A$ denotes benchmark accuracy.
Readers can therefore apply their own failure cost to any reported pair of H-GATE and Acc without rerunning a model.
At $\lambda=1$, an instantaneous failure scores 0, so the fast-failing agent above no longer approaches the neutral reference.
On SS-E with H200, this setting keeps Qwen3.5-397B-A17B, UI-Venus-2-9B, and UI-Venus-1.5-8B as the three highest-ranked configurations, reverses 113 of the 1,378 configuration pairs, and correlates with the H-GATE ordering at Spearman $\rho=0.958$.
The corresponding values are 57 reversals and $\rho=0.984$ at $\lambda=0.5$, and 181 reversals and $\rho=0.896$ at $\lambda=2$, where the two highest-ranked configurations remain unchanged.
The leading SS-E configurations are thus stable under explicit failure costs, while orderings among slower configurations depend on the cost assigned to a wrong action.

\begin{table}
\centering
\caption{\textbf{Rank agreement between H-GATE and simpler summaries on H200.} Spearman correlations are computed over the 53 configurations from the values displayed in \cref{sec:app_full_gpu_results}. Time is mean latency. Reversed counts the configuration pairs, out of 1,378, that H-GATE and Acc@Human order strictly oppositely.}
\label{tab:app_hgate_rank}
\footnotesize
\setlength{\tabcolsep}{4pt}
\begin{tabular}{@{}lrrrr@{}}
\toprule
Benchmark & Acc & Acc@Human & Time & Reversed \\
\midrule
SS-E & 0.368 & 0.940 & $-0.774$ & 129 \\
ScreenSpot-Pro & 0.628 & 0.935 & $-0.676$ & 132 \\
ScreenSpot-v2 & 0.329 & 0.920 & $-0.928$ & 165 \\
OSWorld-G & 0.439 & 0.938 & $-0.838$ & 152 \\
VisualWebBench & 0.317 & 0.782 & $-0.912$ & 270 \\
\bottomrule
\end{tabular}
\end{table}

\paragraph{Agreement with Acc@Human, Acc, and mean latency.}
\Cref{tab:app_hgate_rank} compares the H-GATE ordering of the 53 H200 configurations with the orderings by Acc, Acc@Human, and mean latency.
These 53 configurations yield 1,378 pairs, whereas the ablations in \cref{sec:metric_diagnostics} cover 51 configurations and 1,275 pairs.
H-GATE correlates with Acc@Human at $\rho$ from 0.782 to 0.940, well above its correlation with Acc at 0.317 to 0.628, because both summaries credit only correct responses and both depend on the per-task expert time.
The two summaries still order 129 of the 1,378 pairs oppositely on SS-E and 270 on VisualWebBench.
Acc@Human applies a binary cutoff at the expert time, so it ignores how early or late a correct response arrives and the time spent on failed attempts, both of which H-GATE scores continuously.
\Cref{sec:app_full_gpu_results} discusses cases of each kind on individual deployments.
Acc@Human remains the more direct answer to how often a configuration is correct at expert speed, and we therefore report both summaries.

\section{Full Results by GPU Deployment}
\label{sec:app_full_gpu_results}

This section extends \cref{tab:hgate_main,tab:hardware_choices} to all base configurations across the seven GPU deployments, which include thinking and reasoned-action modes, while \cref{app:visual_budget} covers visual-budget interventions.
The H200 tables report 53 base configurations, whereas the ranking ablations in \cref{sec:metric_diagnostics} cover 51.
Combined with the 24 reduced-budget configurations in \cref{app:visual_budget} and the 10 zoom-in configurations in \cref{app:reasoning_zoom}, these 53 base configurations constitute the 87 inference configurations in the main text.
Each deployment spans one combined table for SS-E, ScreenSpot-Pro, and ScreenSpot-v2 alongside dedicated tables for OSWorld-G and VisualWebBench.
In addition to main-paper columns, P95 reports 95th-percentile latency across all attempts, while Acc@Human measures the share of tasks solved correctly within their expert reference time to separate final accuracy from timely execution.
Per-deployment analyses diagnose why accuracy and efficiency optima diverge on each platform, and \cref{sec:app_cross_gpu} synthesizes cross-hardware trends.

\subsection{H200}
\begin{table*}
\centering
\caption{\textbf{Full results on H200: SS-E and the ScreenSpot benchmarks, part 1 of 2.} Acc, Acc-S, and Acc@Human are percentages for full correctness, spatial correctness, and correct completion by the expert reference time. SS-E Acc requires temporal validity. Time and P95 are mean and 95th-percentile latency in seconds over all attempts. H-GATE combines correctness and latency relative to expert time. -T denotes thinking or reasoned-action. Bold marks peak model H-GATE and underlining marks peak model Acc across both parts. Dashes indicate unavailable values.}
\label{tab:d_h200_screenspot}
\begingroup
\footnotesize
\setlength{\tabcolsep}{1.32pt}
\renewcommand{\arraystretch}{1.08}
\arrayrulecolor{HGRule}
\begin{tabular}{c|c!{\color{HGRuleStrong}\vrule width 0.75pt}cccccc@{\hspace{7pt}}ccccc@{\hspace{5pt}}ccccc}
\noalign{\color{HGRuleStrong}\hrule height 0.8pt}
\multirow{2}{*}{Family} & \multirow{2}{*}{\shortstack{Model\\ /mode}} & \multicolumn{6}{c@{\hspace{7pt}}}{\cellcolor{HGHeader}\textbf{ScreenSpot-Efficiency}} & \multicolumn{5}{c}{\textbf{ScreenSpot-Pro}} & \multicolumn{5}{c}{\textbf{ScreenSpot-v2}} \\
\noalign{\global\arrayrulewidth=0.7pt}\arrayrulecolor{HGRuleStrong}\hhline{~~*{16}{-}}\arrayrulecolor{HGRule}\noalign{\global\arrayrulewidth=0.4pt}
 &  & \cellcolor{HGHeader}Acc-S & \cellcolor{HGHeader}Acc & \cellcolor{HGHeader}\shortstack{Acc@\\Human} & \cellcolor{HGHeader}Time & \cellcolor{HGHeader}P95 & \cellcolor{HGHeader}H-GATE & Acc & \shortstack{Acc@\\Human} & Time & P95 & H-GATE & Acc & \shortstack{Acc@\\Human} & Time & P95 & H-GATE \\
\noalign{\color{HGRuleStrong}\hrule height 0.7pt}
\textbf{Human} & \textemdash & \cellcolor{HGBody}100.0 & \cellcolor{HGBody}99.0 & \cellcolor{HGBody}\textemdash & \cellcolor{HGBody}2.35 & \cellcolor{HGBody}\textemdash & \cellcolor{HGBody}66.2 & 99.2 & \textemdash & 1.46 & \textemdash & 66.2 & 100.0 & \textemdash & 0.97 & \textemdash & 66.7 \\
\hline
\multirow{1}{*}{UI-TARS-1.5} & 7B & \cellcolor{HGBody}24.2 & \cellcolor{HGBody}18.4 & \cellcolor{HGBody}14.8 & \cellcolor{HGBody}1.54 & \cellcolor{HGBody}6.19 & \cellcolor{HGBody}30.6 & 41.7 & 28.1 & 1.34 & 3.10 & 39.9 & 89.5 & 84.7 & 0.67 & 0.88 & 68.1 \\
\hline
\multirow{2}{*}{Jedi} & 3B & \cellcolor{HGBody}21.1 & \cellcolor{HGBody}13.2 & \cellcolor{HGBody}8.5 & \cellcolor{HGBody}1.90 & \cellcolor{HGBody}6.91 & \cellcolor{HGBody}25.0 & 36.2 & 36.0 & 0.69 & 0.94 & 43.1 & 88.9 & 87.2 & 0.58 & 0.66 & 69.5 \\
 & 7B & \cellcolor{HGBody}17.5 & \cellcolor{HGBody}10.9 & \cellcolor{HGBody}6.7 & \cellcolor{HGBody}1.87 & \cellcolor{HGBody}6.77 & \cellcolor{HGBody}24.5 & 34.2 & 34.0 & 0.69 & 0.95 & 41.8 & 91.2 & 89.3 & 0.58 & 0.66 & 70.9 \\
\hline
\multirow{2}{*}{SE-GUI} & 3B & \cellcolor{HGBody}17.5 & \cellcolor{HGBody}11.9 & \cellcolor{HGBody}2.9 & \cellcolor{HGBody}2.90 & \cellcolor{HGBody}10.52 & \cellcolor{HGBody}21.2 & 34.3 & 14.0 & 1.71 & 2.38 & 32.4 & 86.9 & 11.5 & 1.32 & 1.49 & 56.2 \\
 & 7B & \cellcolor{HGBody}20.5 & \cellcolor{HGBody}13.8 & \cellcolor{HGBody}3.1 & \cellcolor{HGBody}2.28 & \cellcolor{HGBody}4.54 & \cellcolor{HGBody}23.2 & 46.5 & 26.4 & 1.52 & 2.17 & 39.7 & 89.9 & 28.5 & 1.05 & 1.23 & 60.6 \\
\hline
\multirow{3}{*}{OpenCUA} & 7B & \cellcolor{HGBody}20.8 & \cellcolor{HGBody}13.7 & \cellcolor{HGBody}9.3 & \cellcolor{HGBody}1.44 & \cellcolor{HGBody}4.72 & \cellcolor{HGBody}27.0 & 46.1 & 36.4 & 1.35 & 2.92 & 43.5 & 93.3 & 91.7 & 0.55 & 0.73 & 72.8 \\
 & 32B & \cellcolor{HGBody}21.0 & \cellcolor{HGBody}12.8 & \cellcolor{HGBody}1.8 & \cellcolor{HGBody}3.82 & \cellcolor{HGBody}11.54 & \cellcolor{HGBody}21.0 & 55.6 & 18.5 & 1.99 & 4.15 & 41.5 & 93.4 & 21.0 & 1.17 & 1.46 & 61.0 \\
 & 72B & \cellcolor{HGBody}26.5 & \cellcolor{HGBody}18.2 & \cellcolor{HGBody}2.5 & \cellcolor{HGBody}3.14 & \cellcolor{HGBody}8.91 & \cellcolor{HGBody}23.3 & 60.9 & 5.6 & 2.57 & 4.84 & 41.7 & 87.8 & 14.4 & 1.91 & 3.60 & 54.6 \\
\hline
\multirow{6}{*}{ScaleCUA} & 3B & \cellcolor{HGBody}19.1 & \cellcolor{HGBody}12.7 & \cellcolor{HGBody}11.4 & \cellcolor{HGBody}0.98 & \cellcolor{HGBody}1.89 & \cellcolor{HGBody}29.3 & 39.7 & 39.6 & 0.59 & 0.81 & 46.4 & 87.8 & 86.9 & 0.53 & 0.61 & 70.6 \\
 & 3B-T & \cellcolor{HGBody}21.9 & \cellcolor{HGBody}15.7 & \cellcolor{HGBody}2.7 & \cellcolor{HGBody}4.29 & \cellcolor{HGBody}9.21 & \cellcolor{HGBody}19.4 & 29.6 & 2.3 & 2.18 & 4.08 & 28.1 & 79.6 & 0.1 & 2.36 & 4.21 & 45.7 \\
 & 7B & \cellcolor{HGBody}21.3 & \cellcolor{HGBody}13.9 & \cellcolor{HGBody}12.7 & \cellcolor{HGBody}0.89 & \cellcolor{HGBody}1.62 & \cellcolor{HGBody}30.5 & 47.8 & 47.8 & 0.59 & 0.85 & 51.0 & 92.6 & 91.7 & 0.51 & 0.56 & 73.8 \\
 & 7B-T & \cellcolor{HGBody}24.2 & \cellcolor{HGBody}18.7 & \cellcolor{HGBody}4.1 & \cellcolor{HGBody}3.28 & \cellcolor{HGBody}7.82 & \cellcolor{HGBody}22.8 & 29.3 & 10.4 & 1.72 & 2.12 & 30.2 & 85.5 & 3.9 & 1.55 & 1.94 & 53.4 \\
 & 32B & \cellcolor{HGBody}22.4 & \cellcolor{HGBody}15.6 & \cellcolor{HGBody}11.7 & \cellcolor{HGBody}1.61 & \cellcolor{HGBody}3.04 & \cellcolor{HGBody}26.7 & 58.7 & 57.8 & 0.99 & 1.25 & 51.0 & 95.1 & 69.3 & 0.90 & 0.99 & 66.1 \\
 & 32B-T & \cellcolor{HGBody}25.9 & \cellcolor{HGBody}19.9 & \cellcolor{HGBody}3.4 & \cellcolor{HGBody}4.90 & \cellcolor{HGBody}9.60 & \cellcolor{HGBody}19.8 & 43.8 & 0.0 & 3.22 & 3.70 & 30.1 & 86.4 & 0.0 & 3.27 & 5.86 & 44.3 \\
\hline
\multirow{1}{*}{GroundNext} & 7B & \cellcolor{HGBody}18.3 & \cellcolor{HGBody}11.8 & \cellcolor{HGBody}1.4 & \cellcolor{HGBody}2.63 & \cellcolor{HGBody}3.70 & \cellcolor{HGBody}20.8 & 53.1 & 33.1 & 1.47 & 2.85 & 43.7 & 90.5 & 51.0 & 0.93 & 1.11 & 62.9 \\
\hline
\multirow{8}{*}{Holo2} & 4B & \cellcolor{HGBody}24.2 & \cellcolor{HGBody}17.5 & \cellcolor{HGBody}15.9 & \cellcolor{HGBody}2.64 & \cellcolor{HGBody}5.51 & \cellcolor{HGBody}33.4 & 58.2 & 54.2 & 0.79 & 2.50 & 54.7 & 93.5 & 93.4 & 0.40 & 0.53 & 77.1 \\
 & 4B-T & \cellcolor{HGBody}25.0 & \cellcolor{HGBody}18.0 & \cellcolor{HGBody}16.4 & \cellcolor{HGBody}3.32 & \cellcolor{HGBody}5.53 & \cellcolor{HGBody}33.5 & 57.7 & 52.5 & 0.84 & 2.64 & 53.7 & 93.2 & 93.2 & 0.43 & 0.59 & 75.8 \\
 & 8B & \cellcolor{HGBody}28.5 & \cellcolor{HGBody}20.9 & \cellcolor{HGBody}18.9 & \cellcolor{HGBody}1.37 & \cellcolor{HGBody}1.05 & \cellcolor{HGBody}35.8 & 59.8 & 49.8 & 0.99 & 3.22 & 53.5 & 94.3 & 94.2 & 0.47 & 0.64 & 75.3 \\
 & 8B-T & \cellcolor{HGBody}26.8 & \cellcolor{HGBody}16.2 & \cellcolor{HGBody}3.4 & \cellcolor{HGBody}13.86 & \cellcolor{HGBody}14.48 & \cellcolor{HGBody}20.3 & 57.3 & 17.7 & 2.73 & 3.99 & 42.9 & 94.2 & 55.3 & 1.54 & 2.64 & 66.5 \\
 & 30B-A3B & \cellcolor{HGBody}30.8 & \cellcolor{HGBody}23.3 & \cellcolor{HGBody}19.5 & \cellcolor{HGBody}3.89 & \cellcolor{HGBody}5.54 & \cellcolor{HGBody}32.5 & 65.0 & 45.8 & 1.20 & 3.42 & 53.0 & 95.8 & 92.0 & 0.71 & 0.87 & 69.9 \\
 & 30B-A3B-T & \cellcolor{HGBody}27.7 & \cellcolor{HGBody}19.7 & \cellcolor{HGBody}12.5 & \cellcolor{HGBody}5.26 & \cellcolor{HGBody}5.58 & \cellcolor{HGBody}27.6 & 63.3 & 38.1 & 1.52 & 3.70 & 49.0 & 95.9 & 69.6 & 0.84 & 1.16 & 67.3 \\
 & 235B-A22B & \cellcolor{HGBody}42.9 & \cellcolor{HGBody}36.4 & \cellcolor{HGBody}8.3 & \cellcolor{HGBody}3.49 & \cellcolor{HGBody}6.44 & \cellcolor{HGBody}33.3 & 69.9 & 3.7 & 2.18 & 3.52 & 46.3 & 96.7 & 1.4 & 1.50 & 1.99 & 58.2 \\
 & 235B-A22B-T & \cellcolor{HGBody}39.5 & \cellcolor{HGBody}33.4 & \cellcolor{HGBody}5.1 & \cellcolor{HGBody}12.27 & \cellcolor{HGBody}18.36 & \cellcolor{HGBody}30.2 & 64.0 & 11.3 & 2.45 & 4.45 & 44.7 & 84.8 & 3.0 & 1.55 & 2.60 & 53.1 \\
\hline
\multirow{1}{*}{POINTS-GUI} & 8B & \cellcolor{HGBody}31.7 & \cellcolor{HGBody}23.2 & \cellcolor{HGBody}20.0 & \cellcolor{HGBody}0.99 & \cellcolor{HGBody}1.61 & \cellcolor{HGBody}34.2 & 59.2 & 58.4 & 0.86 & 1.16 & 53.1 & 95.2 & 94.0 & 0.59 & 0.81 & 72.7 \\
\hline
\multirow{3}{*}{UI-Venus-1.5} & 2B & \cellcolor{HGBody}29.2 & \cellcolor{HGBody}15.7 & \cellcolor{HGBody}14.9 & \cellcolor{HGBody}1.09 & \cellcolor{HGBody}2.93 & \cellcolor{HGBody}31.6 & 58.8 & 54.8 & 0.79 & 2.37 & 55.6 & 93.3 & 91.8 & 0.36 & 0.44 & \cellcolor{HGBest}\textbf{79.8} \\
 & 8B & \cellcolor{HGBody}37.6 & \cellcolor{HGBody}26.3 & \cellcolor{HGBody}24.2 & \cellcolor{HGBody}0.93 & \cellcolor{HGBody}2.89 & \cellcolor{HGBody}38.1 & 68.2 & 60.3 & 0.91 & 2.79 & 58.9 & 96.2 & 95.3 & 0.41 & 0.53 & 78.8 \\
 & 30B-A3B & \cellcolor{HGBody}40.3 & \cellcolor{HGBody}29.5 & \cellcolor{HGBody}23.1 & \cellcolor{HGBody}1.77 & \cellcolor{HGBody}4.26 & \cellcolor{HGBody}36.0 & 69.1 & 62.3 & 0.87 & 2.61 & 60.0 & 96.5 & 95.6 & 0.40 & 0.50 & 79.5 \\
\noalign{\global\arrayrulewidth=0.8pt}\arrayrulecolor{HGRuleStrong}\hline
\end{tabular}
\global\arrayrulewidth=0.4pt\arrayrulecolor{HGRule}
\endgroup
\end{table*}

\begin{table*}
\centering
\caption{\textbf{Full results on H200: SS-E and the ScreenSpot benchmarks, part 2 of 2.} Columns follow Table~\ref{tab:d_h200_screenspot}. Bold marks peak model H-GATE and underlining marks peak model Acc across both parts.}
\label{tab:d_h200_screenspot_part2}
\begingroup
\footnotesize
\setlength{\tabcolsep}{1.0pt}
\renewcommand{\arraystretch}{1.08}
\arrayrulecolor{HGRule}
\begin{tabular}{c|c!{\color{HGRuleStrong}\vrule width 0.75pt}cccccc@{\hspace{7pt}}ccccc@{\hspace{5pt}}ccccc}
\noalign{\color{HGRuleStrong}\hrule height 0.8pt}
\multirow{2}{*}{Family} & \multirow{2}{*}{\shortstack{Model\\ /mode}} & \multicolumn{6}{c@{\hspace{7pt}}}{\cellcolor{HGHeader}\textbf{ScreenSpot-Efficiency}} & \multicolumn{5}{c}{\textbf{ScreenSpot-Pro}} & \multicolumn{5}{c}{\textbf{ScreenSpot-v2}} \\
\noalign{\global\arrayrulewidth=0.7pt}\arrayrulecolor{HGRuleStrong}\hhline{~~*{16}{-}}\arrayrulecolor{HGRule}\noalign{\global\arrayrulewidth=0.4pt}
 &  & \cellcolor{HGHeader}Acc-S & \cellcolor{HGHeader}Acc & \cellcolor{HGHeader}\shortstack{Acc@\\Human} & \cellcolor{HGHeader}Time & \cellcolor{HGHeader}P95 & \cellcolor{HGHeader}H-GATE & Acc & \shortstack{Acc@\\Human} & Time & P95 & H-GATE & Acc & \shortstack{Acc@\\Human} & Time & P95 & H-GATE \\
\noalign{\color{HGRuleStrong}\hrule height 0.7pt}
\textbf{Human} & \textemdash & \cellcolor{HGBody}100.0 & \cellcolor{HGBody}99.0 & \cellcolor{HGBody}\textemdash & \cellcolor{HGBody}2.35 & \cellcolor{HGBody}\textemdash & \cellcolor{HGBody}66.2 & 99.2 & \textemdash & 1.46 & \textemdash & 66.2 & 100.0 & \textemdash & 0.97 & \textemdash & 66.7 \\
\hline
\multirow{6}{*}{GUI-Owl-1.5} & 2B & \cellcolor{HGBody}21.8 & \cellcolor{HGBody}15.1 & \cellcolor{HGBody}14.9 & \cellcolor{HGBody}0.47 & \cellcolor{HGBody}0.85 & \cellcolor{HGBody}36.5 & 58.2 & 57.9 & 0.59 & 1.09 & 57.4 & 90.2 & 89.2 & 0.33 & 0.40 & 78.4 \\
 & 4B & \cellcolor{HGBody}21.8 & \cellcolor{HGBody}14.6 & \cellcolor{HGBody}14.3 & \cellcolor{HGBody}0.49 & \cellcolor{HGBody}1.10 & \cellcolor{HGBody}34.9 & 67.0 & 64.2 & 0.71 & 1.32 & \cellcolor{HGBest}\textbf{60.3} & 93.1 & 91.9 & 0.39 & 0.47 & 77.3 \\
 & 8B & \cellcolor{HGBody}26.3 & \cellcolor{HGBody}18.4 & \cellcolor{HGBody}18.0 & \cellcolor{HGBody}0.62 & \cellcolor{HGBody}1.61 & \cellcolor{HGBody}35.8 & 70.5 & 64.8 & 0.85 & 1.53 & 60.0 & 94.2 & 93.2 & 0.48 & 0.59 & 75.0 \\
 & 8B-T & \cellcolor{HGBody}26.0 & \cellcolor{HGBody}18.2 & \cellcolor{HGBody}14.6 & \cellcolor{HGBody}1.44 & \cellcolor{HGBody}2.78 & \cellcolor{HGBody}29.3 & 60.1 & 40.3 & 1.32 & 2.03 & 48.5 & 92.6 & 61.7 & 0.90 & 1.21 & 64.7 \\
 & 32B & \cellcolor{HGBody}26.9 & \cellcolor{HGBody}18.7 & \cellcolor{HGBody}15.1 & \cellcolor{HGBody}1.39 & \cellcolor{HGBody}2.70 & \cellcolor{HGBody}29.5 & \underline{72.9} & 31.1 & 1.77 & 2.86 & 51.0 & 94.9 & 19.7 & 1.16 & 1.36 & 61.5 \\
 & 32B-T & \cellcolor{HGBody}24.0 & \cellcolor{HGBody}18.0 & \cellcolor{HGBody}2.6 & \cellcolor{HGBody}4.58 & \cellcolor{HGBody}8.32 & \cellcolor{HGBody}19.6 & 55.6 & 0.0 & 3.55 & 4.89 & 34.0 & 92.6 & 0.0 & 2.61 & 3.40 & 48.6 \\
\noalign{\color{HGRuleStrong}\hrule height 0.8pt}
\multirow{16}{*}{Qwen3.5} & 0.8B & \cellcolor{HGBody}17.4 & \cellcolor{HGBody}11.7 & \cellcolor{HGBody}11.1 & \cellcolor{HGBody}1.17 & \cellcolor{HGBody}4.07 & \cellcolor{HGBody}29.6 & 29.7 & 29.5 & 0.45 & 1.13 & 43.4 & 77.0 & 77.0 & 0.22 & 0.33 & 73.6 \\
 & 0.8B-T & \cellcolor{HGBody}13.1 & \cellcolor{HGBody}6.9 & \cellcolor{HGBody}1.9 & \cellcolor{HGBody}8.32 & \cellcolor{HGBody}32.34 & \cellcolor{HGBody}14.8 & 14.0 & 10.9 & 7.63 & 36.25 & 20.4 & 66.7 & 52.3 & 1.66 & 5.39 & 54.5 \\
 & 2B & \cellcolor{HGBody}23.7 & \cellcolor{HGBody}16.1 & \cellcolor{HGBody}15.3 & \cellcolor{HGBody}0.54 & \cellcolor{HGBody}0.90 & \cellcolor{HGBody}35.1 & 42.8 & 41.0 & 0.66 & 1.82 & 48.1 & 81.4 & 81.4 & 0.31 & 0.45 & 73.2 \\
 & 2B-T & \cellcolor{HGBody}27.3 & \cellcolor{HGBody}18.6 & \cellcolor{HGBody}4.2 & \cellcolor{HGBody}8.16 & \cellcolor{HGBody}30.61 & \cellcolor{HGBody}20.2 & 37.4 & 7.6 & 7.39 & 34.77 & 27.4 & 80.7 & 33.9 & 2.97 & 10.68 & 52.1 \\
 & 4B & \cellcolor{HGBody}25.6 & \cellcolor{HGBody}19.0 & \cellcolor{HGBody}17.3 & \cellcolor{HGBody}1.01 & \cellcolor{HGBody}1.42 & \cellcolor{HGBody}35.1 & 47.4 & 45.0 & 0.72 & 1.88 & 49.5 & 89.0 & 88.9 & 0.37 & 0.51 & 75.3 \\
 & 4B-T & \cellcolor{HGBody}30.1 & \cellcolor{HGBody}17.7 & \cellcolor{HGBody}0.9 & \cellcolor{HGBody}36.08 & \cellcolor{HGBody}110.52 & \cellcolor{HGBody}11.6 & 40.5 & 1.6 & 23.72 & 62.92 & 20.6 & 86.8 & 4.2 & 10.99 & 44.32 & 40.3 \\
 & 9B & \cellcolor{HGBody}28.6 & \cellcolor{HGBody}21.7 & \cellcolor{HGBody}20.1 & \cellcolor{HGBody}0.74 & \cellcolor{HGBody}1.37 & \cellcolor{HGBody}36.4 & 49.8 & 46.7 & 0.79 & 2.28 & 50.1 & 91.2 & 90.7 & 0.37 & 0.53 & 77.0 \\
 & 9B-T & \cellcolor{HGBody}34.5 & \cellcolor{HGBody}20.2 & \cellcolor{HGBody}0.7 & \cellcolor{HGBody}53.90 & \cellcolor{HGBody}202.05 & \cellcolor{HGBody}11.4 & 47.1 & 0.2 & 30.73 & 104.38 & 21.3 & 90.2 & 3.0 & 13.51 & 51.61 & 39.5 \\
 & 27B & \cellcolor{HGBody}38.0 & \cellcolor{HGBody}32.2 & \cellcolor{HGBody}21.5 & \cellcolor{HGBody}1.86 & \cellcolor{HGBody}4.92 & \cellcolor{HGBody}35.1 & 54.5 & 35.2 & 1.39 & 3.38 & 45.7 & 90.2 & 76.7 & 0.77 & 0.98 & 65.7 \\
 & 27B-T & \cellcolor{HGBody}46.3 & \cellcolor{HGBody}29.0 & \cellcolor{HGBody}0.1 & \cellcolor{HGBody}108.93 & \cellcolor{HGBody}481.93 & \cellcolor{HGBody}12.9 & 64.5 & 0.0 & 60.50 & 244.29 & 25.5 & 95.8 & 0.0 & 15.18 & 50.07 & 41.4 \\
 & 35B-A3B & \cellcolor{HGBody}31.7 & \cellcolor{HGBody}25.7 & \cellcolor{HGBody}19.4 & \cellcolor{HGBody}1.31 & \cellcolor{HGBody}3.30 & \cellcolor{HGBody}35.6 & 28.9 & 26.9 & 0.84 & 2.24 & 37.9 & 68.3 & 66.7 & 0.43 & 0.65 & 62.3 \\
 & 35B-A3B-T & \cellcolor{HGBody}41.8 & \cellcolor{HGBody}30.2 & \cellcolor{HGBody}9.3 & \cellcolor{HGBody}15.95 & \cellcolor{HGBody}64.79 & \cellcolor{HGBody}23.5 & 54.4 & 29.2 & 3.69 & 12.23 & 41.7 & 88.4 & 63.7 & 0.98 & 1.54 & 62.2 \\
 & 122B-A10B & \cellcolor{HGBody}33.2 & \cellcolor{HGBody}26.0 & \cellcolor{HGBody}22.7 & \cellcolor{HGBody}1.06 & \cellcolor{HGBody}2.26 & \cellcolor{HGBody}36.9 & 44.3 & 39.5 & 0.92 & 2.18 & 45.5 & 82.5 & 81.2 & 0.47 & 0.65 & 69.0 \\
 & 122B-A10B-T & \cellcolor{HGBody}47.3 & \cellcolor{HGBody}36.2 & \cellcolor{HGBody}8.8 & \cellcolor{HGBody}21.47 & \cellcolor{HGBody}98.72 & \cellcolor{HGBody}26.1 & 59.5 & 20.2 & 9.41 & 43.53 & 38.2 & 93.7 & 37.8 & 2.28 & 8.03 & 59.6 \\
 & 397B-A17B & \cellcolor{HGBody}50.8 & \cellcolor{HGBody}\underline{42.2} & \cellcolor{HGBody}35.9 & \cellcolor{HGBody}1.45 & \cellcolor{HGBody}3.52 & \cellcolor{HGBest}\textbf{43.0} & 65.3 & 52.9 & 1.11 & 2.51 & 54.1 & 95.4 & 93.6 & 0.66 & 0.83 & 70.9 \\
 & 397B-A17B-T & \cellcolor{HGBody}49.3 & \cellcolor{HGBody}33.0 & \cellcolor{HGBody}1.3 & \cellcolor{HGBody}61.17 & \cellcolor{HGBody}194.17 & \cellcolor{HGBody}16.8 & 69.3 & 1.3 & 35.37 & 125.97 & 31.4 & \underline{96.8} & 1.2 & 9.37 & 41.18 & 48.0 \\
\hline
\multirow{1}{*}{MolmoPoint-GUI} & 8B & \cellcolor{HGBody}22.6 & \cellcolor{HGBody}15.6 & \cellcolor{HGBody}7.2 & \cellcolor{HGBody}3.03 & \cellcolor{HGBody}6.86 & \cellcolor{HGBody}23.6 & 61.2 & 43.6 & 1.30 & 1.65 & 48.9 & 93.9 & 49.9 & 0.93 & 1.23 & 65.5 \\
\hline
\multirow{2}{*}{VISTA} & 4B & \cellcolor{HGBody}31.5 & \cellcolor{HGBody}23.2 & \cellcolor{HGBody}21.1 & \cellcolor{HGBody}0.85 & \cellcolor{HGBody}1.60 & \cellcolor{HGBody}35.8 & 63.0 & 59.3 & 0.78 & 1.70 & 57.0 & 94.7 & 93.8 & 0.58 & 0.57 & 74.8 \\
 & 9B & \cellcolor{HGBody}32.3 & \cellcolor{HGBody}23.4 & \cellcolor{HGBody}20.6 & \cellcolor{HGBody}1.04 & \cellcolor{HGBody}2.10 & \cellcolor{HGBody}34.9 & 67.9 & 61.6 & 1.01 & 2.15 & 58.1 & 96.0 & 95.0 & 0.55 & 0.65 & 74.6 \\
\hline
\multirow{1}{*}{UI-Venus-2} & 9B & \cellcolor{HGBody}44.7 & \cellcolor{HGBody}33.3 & \cellcolor{HGBody}30.9 & \cellcolor{HGBody}1.15 & \cellcolor{HGBody}3.40 & \cellcolor{HGBody}41.4 & 72.4 & 55.0 & 1.15 & 2.98 & 58.4 & 96.3 & 89.4 & 0.58 & 1.13 & 74.0 \\
\noalign{\global\arrayrulewidth=0.8pt}\arrayrulecolor{HGRuleStrong}\hline
\end{tabular}
\global\arrayrulewidth=0.4pt\arrayrulecolor{HGRule}
\endgroup
\end{table*}

\begin{table}
\centering
\caption{\textbf{Full results on H200: OSWorld-G, part 1 of 2.} Columns follow Table~\ref{tab:d_h200_screenspot}. Bold marks peak model H-GATE and underlining marks peak model Acc across both parts.}
\label{tab:d_h200_osworldg}
\begingroup
\footnotesize
\setlength{\tabcolsep}{1.32pt}
\renewcommand{\arraystretch}{1.08}
\arrayrulecolor{HGRule}
\begin{tabular}{c|c!{\color{HGRuleStrong}\vrule width 0.75pt}ccccc}
\noalign{\color{HGRuleStrong}\hrule height 0.8pt}
\multirow{2}{*}{Family} & \multirow{2}{*}{\shortstack{Model\\ /mode}} & \multicolumn{5}{c}{\textbf{OSWorld-G}} \\
\noalign{\global\arrayrulewidth=0.7pt}\arrayrulecolor{HGRuleStrong}\hhline{~~*{5}{-}}\arrayrulecolor{HGRule}\noalign{\global\arrayrulewidth=0.4pt}
 &  & Acc & \shortstack{Acc@\\Human} & Time & P95 & H-GATE \\
\noalign{\color{HGRuleStrong}\hrule height 0.7pt}
\textbf{Human} & \textemdash & 99.9 & \textemdash & 1.11 & \textemdash & 66.6 \\
\hline
\multirow{1}{*}{UI-TARS-1.5} & 7B & 53.4 & 53.2 & 0.48 & 0.55 & 53.7 \\
\hline
\multirow{2}{*}{Jedi} & 3B & 49.8 & 16.0 & 3.86 & 7.71 & 36.4 \\
 & 7B & 40.1 & 28.9 & 1.49 & 2.64 & 37.6 \\
\hline
\multirow{2}{*}{SE-GUI} & 3B & 37.6 & 23.6 & 1.06 & 1.15 & 35.9 \\
 & 7B & 47.7 & 43.4 & 0.90 & 0.94 & 43.4 \\
\hline
\multirow{3}{*}{OpenCUA} & 7B & 55.9 & 55.9 & 0.41 & 0.47 & 56.6 \\
 & 32B & 58.5 & 35.1 & 1.16 & 1.97 & 45.8 \\
 & 72B & 60.3 & 13.5 & 2.31 & 2.84 & 40.0 \\
\hline
\multirow{6}{*}{ScaleCUA} & 3B & 51.2 & 51.2 & 0.47 & 0.56 & 52.5 \\
 & 3B-T & 39.4 & 0.7 & 2.18 & 6.31 & 30.2 \\
 & 7B & 55.5 & 55.5 & 0.44 & 0.52 & 55.5 \\
 & 7B-T & 44.5 & 6.6 & 1.67 & 3.80 & 35.2 \\
 & 32B & 60.8 & 54.4 & 1.34 & 0.95 & 50.9 \\
 & 32B-T & 53.2 & 0.0 & 3.13 & 3.77 & 31.7 \\
\hline
\multirow{1}{*}{GroundNext} & 7B & 56.4 & 53.7 & 0.77 & 0.86 & 49.4 \\
\hline
\multirow{8}{*}{Holo2} & 4B & 62.4 & 62.4 & 0.27 & 0.30 & 64.3 \\
 & 4B-T & 61.9 & 61.9 & 0.29 & 0.34 & 63.2 \\
 & 8B & 63.3 & 63.3 & 0.31 & 0.36 & 63.5 \\
 & 8B-T & 60.8 & 19.3 & 2.55 & 2.52 & 44.3 \\
 & 30B-A3B & 67.0 & 66.8 & 0.56 & 0.61 & 59.1 \\
 & 30B-A3B-T & 66.5 & 62.4 & 0.85 & 0.91 & 53.6 \\
 & 235B-A22B & 71.1 & 10.6 & 1.48 & 1.75 & 48.2 \\
 & 235B-A22B-T & 67.4 & 18.8 & 1.36 & 1.58 & 48.3 \\
\hline
\multirow{1}{*}{POINTS-GUI} & 8B & 66.0 & 65.4 & 0.82 & 0.76 & 56.9 \\
\hline
\multirow{3}{*}{UI-Venus-1.5} & 2B & 58.3 & 58.0 & 0.25 & 0.25 & 63.7 \\
 & 8B & 69.3 & 68.8 & 0.28 & 0.32 & 68.5 \\
 & 30B-A3B & 71.5 & 70.7 & 0.28 & 0.30 & \cellcolor{HGBest}\textbf{69.8} \\
\noalign{\global\arrayrulewidth=0.8pt}\arrayrulecolor{HGRuleStrong}\hline
\end{tabular}
\global\arrayrulewidth=0.4pt\arrayrulecolor{HGRule}
\endgroup
\end{table}

\begin{table}
\centering
\caption{\textbf{Full results on H200: OSWorld-G, part 2 of 2.} Columns follow Table~\ref{tab:d_h200_screenspot}. Bold marks peak model H-GATE and underlining marks peak model Acc across both parts.}
\label{tab:d_h200_osworldg_part2}
\begingroup
\footnotesize
\setlength{\tabcolsep}{1.32pt}
\renewcommand{\arraystretch}{1.08}
\arrayrulecolor{HGRule}
\begin{tabular}{c|c!{\color{HGRuleStrong}\vrule width 0.75pt}ccccc}
\noalign{\color{HGRuleStrong}\hrule height 0.8pt}
\multirow{2}{*}{Family} & \multirow{2}{*}{\shortstack{Model\\ /mode}} & \multicolumn{5}{c}{\textbf{OSWorld-G}} \\
\noalign{\global\arrayrulewidth=0.7pt}\arrayrulecolor{HGRuleStrong}\hhline{~~*{5}{-}}\arrayrulecolor{HGRule}\noalign{\global\arrayrulewidth=0.4pt}
 &  & Acc & \shortstack{Acc@\\Human} & Time & P95 & H-GATE \\
\noalign{\color{HGRuleStrong}\hrule height 0.7pt}
\textbf{Human} & \textemdash & 99.9 & \textemdash & 1.11 & \textemdash & 66.6 \\
\hline
\multirow{6}{*}{GUI-Owl-1.5} & 2B & 52.7 & 52.7 & 0.23 & 0.28 & 59.7 \\
 & 4B & 63.8 & 63.8 & 0.28 & 0.32 & 64.7 \\
 & 8B & 66.8 & 66.8 & 0.35 & 0.40 & 64.4 \\
 & 8B-T & 57.8 & 52.0 & 0.82 & 1.06 & 49.7 \\
 & 32B & 66.0 & 50.5 & 0.94 & 1.03 & 51.7 \\
 & 32B-T & 56.4 & 0.0 & 3.45 & 9.53 & 33.2 \\
\noalign{\color{HGRuleStrong}\hrule height 0.8pt}
\multirow{16}{*}{Qwen3.5} & 0.8B & 32.3 & 32.3 & 0.19 & 0.22 & 48.5 \\
 & 0.8B-T & 11.3 & 2.8 & 6.29 & 24.04 & 15.2 \\
 & 2B & 41.5 & 41.5 & 0.23 & 0.28 & 52.8 \\
 & 2B-T & 41.5 & 18.3 & 4.15 & 15.56 & 33.0 \\
 & 4B & 37.9 & 37.8 & 0.55 & 0.38 & 48.7 \\
 & 4B-T & 45.6 & 1.2 & 22.22 & 67.00 & 21.9 \\
 & 9B & 54.3 & 54.3 & 0.25 & 0.31 & 59.9 \\
 & 9B-T & 57.3 & 0.9 & 28.78 & 94.75 & 25.0 \\
 & 27B & 58.5 & 58.5 & 0.60 & 0.70 & 53.7 \\
 & 27B-T & 71.5 & 0.2 & 49.94 & 217.53 & 29.8 \\
 & 35B-A3B & 35.8 & 35.8 & 0.33 & 0.46 & 46.9 \\
 & 35B-A3B-T & 65.1 & 50.4 & 2.48 & 7.17 & 49.8 \\
 & 122B-A10B & 52.8 & 52.8 & 0.33 & 0.40 & 56.7 \\
 & 122B-A10B-T & 72.2 & 31.0 & 7.78 & 40.13 & 45.5 \\
 & 397B-A17B & 69.7 & 69.3 & 0.57 & 0.68 & 60.8 \\
 & 397B-A17B-T & \underline{77.0} & 0.5 & 51.16 & 180.79 & 31.7 \\
\hline
\multirow{1}{*}{MolmoPoint-GUI} & 8B & 63.3 & 61.5 & 0.67 & 0.84 & 55.2 \\
\hline
\multirow{2}{*}{VISTA} & 4B & 59.6 & 59.4 & 0.40 & 0.47 & 58.6 \\
 & 9B & 67.2 & 67.2 & 0.42 & 0.48 & 62.7 \\
\hline
\multirow{1}{*}{UI-Venus-2} & 9B & 75.4 & 75.0 & 0.35 & 0.43 & 69.6 \\
\noalign{\global\arrayrulewidth=0.8pt}\arrayrulecolor{HGRuleStrong}\hline
\end{tabular}
\global\arrayrulewidth=0.4pt\arrayrulecolor{HGRule}
\endgroup
\end{table}

\begin{table}
\centering
\caption{\textbf{Full results on H200: VisualWebBench, part 1 of 2.} Columns follow Table~\ref{tab:d_h200_screenspot}. Bold marks peak model H-GATE and underlining marks peak model Acc across both parts.}
\label{tab:d_h200_visualweb}
\begingroup
\footnotesize
\setlength{\tabcolsep}{1.32pt}
\renewcommand{\arraystretch}{1.08}
\arrayrulecolor{HGRule}
\begin{tabular}{c|c!{\color{HGRuleStrong}\vrule width 0.75pt}ccccc}
\noalign{\color{HGRuleStrong}\hrule height 0.8pt}
\multirow{2}{*}{Family} & \multirow{2}{*}{\shortstack{Model\\ /mode}} & \multicolumn{5}{c}{\textbf{VisualWebBench}} \\
\noalign{\global\arrayrulewidth=0.7pt}\arrayrulecolor{HGRuleStrong}\hhline{~~*{5}{-}}\arrayrulecolor{HGRule}\noalign{\global\arrayrulewidth=0.4pt}
 &  & Acc & \shortstack{Acc@\\Human} & Time & P95 & H-GATE \\
\noalign{\color{HGRuleStrong}\hrule height 0.7pt}
\textbf{Human} & \textemdash & 100.0 & \textemdash & 1.06 & \textemdash & 66.7 \\
\hline
\multirow{1}{*}{UI-TARS-1.5} & 7B & 88.0 & 87.0 & 0.41 & 0.44 & 74.7 \\
\hline
\multirow{2}{*}{Jedi} & 3B & 89.3 & 88.4 & 0.79 & 0.88 & 65.8 \\
 & 7B & 88.6 & 88.0 & 0.74 & 0.82 & 66.5 \\
\hline
\multirow{2}{*}{SE-GUI} & 3B & 86.4 & 58.1 & 0.99 & 1.06 & 60.8 \\
 & 7B & 89.1 & 89.0 & 0.80 & 0.83 & 65.6 \\
\hline
\multirow{3}{*}{OpenCUA} & 7B & 91.5 & 91.5 & 0.35 & 0.38 & 78.5 \\
 & 32B & 90.5 & 88.6 & 0.84 & 0.91 & 65.6 \\
 & 72B & 11.4 & 6.8 & 4.95 & 21.30 & 17.4 \\
\hline
\multirow{6}{*}{ScaleCUA} & 3B & 89.3 & 89.3 & 0.41 & 0.48 & 75.2 \\
 & 3B-T & 87.0 & 0.4 & 1.90 & 2.43 & 51.9 \\
 & 7B & 91.3 & 91.3 & 0.38 & 0.44 & 77.1 \\
 & 7B-T & 89.9 & 10.9 & 1.35 & 1.74 & 58.0 \\
 & 32B & 90.3 & 90.3 & 0.73 & 0.81 & 67.7 \\
 & 32B-T & 89.0 & 0.0 & 3.36 & 4.21 & 45.3 \\
\hline
\multirow{1}{*}{GroundNext} & 7B & 90.7 & 90.1 & 0.70 & 0.76 & 68.4 \\
\hline
\multirow{8}{*}{Holo2} & 4B & 91.1 & 91.1 & 0.24 & 0.26 & 82.1 \\
 & 4B-T & 90.9 & 90.9 & 0.27 & 0.30 & 81.0 \\
 & 8B & 92.2 & 92.2 & 0.27 & 0.30 & 81.6 \\
 & 8B-T & 91.1 & 51.0 & 1.36 & 2.39 & 65.8 \\
 & 30B-A3B & 93.8 & 93.8 & 0.53 & 0.56 & 74.2 \\
 & 30B-A3B-T & \underline{94.2} & 93.6 & 0.75 & 0.85 & 69.3 \\
 & 235B-A22B & 93.0 & 26.0 & 1.21 & 1.62 & 61.1 \\
 & 235B-A22B-T & 82.4 & 24.4 & 1.28 & 2.02 & 56.1 \\
\hline
\multirow{1}{*}{POINTS-GUI} & 8B & 92.1 & 92.1 & 0.34 & 0.38 & 79.2 \\
\hline
\multirow{3}{*}{UI-Venus-1.5} & 2B & 90.1 & 88.6 & 0.22 & 0.20 & \cellcolor{HGBest}\textbf{83.9} \\
 & 8B & 91.5 & 90.7 & 0.25 & 0.26 & 82.8 \\
 & 30B-A3B & 92.6 & 91.7 & 0.25 & 0.25 & 83.6 \\
\noalign{\global\arrayrulewidth=0.8pt}\arrayrulecolor{HGRuleStrong}\hline
\end{tabular}
\global\arrayrulewidth=0.4pt\arrayrulecolor{HGRule}
\endgroup
\end{table}

\begin{table}
\centering
\caption{\textbf{Full results on H200: VisualWebBench, part 2 of 2.} Columns follow Table~\ref{tab:d_h200_screenspot}. Bold marks peak model H-GATE and underlining marks peak model Acc across both parts.}
\label{tab:d_h200_visualweb_part2}
\begingroup
\footnotesize
\setlength{\tabcolsep}{1.32pt}
\renewcommand{\arraystretch}{1.08}
\arrayrulecolor{HGRule}
\begin{tabular}{c|c!{\color{HGRuleStrong}\vrule width 0.75pt}ccccc}
\noalign{\color{HGRuleStrong}\hrule height 0.8pt}
\multirow{2}{*}{Family} & \multirow{2}{*}{\shortstack{Model\\ /mode}} & \multicolumn{5}{c}{\textbf{VisualWebBench}} \\
\noalign{\global\arrayrulewidth=0.7pt}\arrayrulecolor{HGRuleStrong}\hhline{~~*{5}{-}}\arrayrulecolor{HGRule}\noalign{\global\arrayrulewidth=0.4pt}
 &  & Acc & \shortstack{Acc@\\Human} & Time & P95 & H-GATE \\
\noalign{\color{HGRuleStrong}\hrule height 0.7pt}
\textbf{Human} & \textemdash & 100.0 & \textemdash & 1.06 & \textemdash & 66.7 \\
\hline
\multirow{6}{*}{GUI-Owl-1.5} & 2B & 87.4 & 87.4 & 0.19 & 0.21 & 81.9 \\
 & 4B & 91.1 & 91.1 & 0.25 & 0.27 & 81.7 \\
 & 8B & 90.7 & 90.7 & 0.32 & 0.35 & 78.8 \\
 & 8B-T & 87.8 & 84.5 & 0.79 & 0.99 & 65.3 \\
 & 32B & 91.1 & 84.7 & 0.89 & 0.93 & 64.9 \\
 & 32B-T & 92.6 & 0.0 & 2.77 & 3.41 & 48.8 \\
\noalign{\color{HGRuleStrong}\hrule height 0.8pt}
\multirow{16}{*}{Qwen3.5} & 0.8B & 76.9 & 76.9 & 0.16 & 0.19 & 76.9 \\
 & 0.8B-T & 69.2 & 67.2 & 0.78 & 2.63 & 64.2 \\
 & 2B & 78.3 & 78.3 & 0.19 & 0.21 & 76.4 \\
 & 2B-T & 69.4 & 41.3 & 2.97 & 13.36 & 51.5 \\
 & 4B & 84.5 & 84.5 & 0.25 & 0.29 & 77.7 \\
 & 4B-T & 78.7 & 3.7 & 10.52 & 41.03 & 38.0 \\
 & 9B & 84.7 & 84.7 & 0.21 & 0.24 & 79.4 \\
 & 9B-T & 83.1 & 21.7 & 9.55 & 41.18 & 44.6 \\
 & 27B & 90.3 & 90.3 & 0.53 & 0.57 & 72.4 \\
 & 27B-T & 92.1 & 0.0 & 13.99 & 33.32 & 42.4 \\
 & 35B-A3B & 56.2 & 56.2 & 0.30 & 0.38 & 59.2 \\
 & 35B-A3B-T & 86.6 & 80.4 & 0.81 & 1.21 & 65.1 \\
 & 122B-A10B & 83.1 & 83.1 & 0.26 & 0.29 & 76.6 \\
 & 122B-A10B-T & 91.7 & 82.2 & 1.19 & 1.52 & 65.7 \\
 & 397B-A17B & 91.5 & 91.5 & 0.48 & 0.56 & 74.2 \\
 & 397B-A17B-T & 92.2 & 20.7 & 3.98 & 10.71 & 55.8 \\
\hline
\multirow{1}{*}{MolmoPoint-GUI} & 8B & 83.3 & 83.3 & 0.67 & 0.78 & 65.2 \\
\hline
\multirow{2}{*}{VISTA} & 4B & 91.9 & 91.9 & 0.39 & 0.44 & 77.4 \\
 & 9B & 92.1 & 92.1 & 0.40 & 0.44 & 77.2 \\
\hline
\multirow{1}{*}{UI-Venus-2} & 9B & 90.5 & 89.7 & 0.41 & 0.69 & 76.2 \\
\noalign{\global\arrayrulewidth=0.8pt}\arrayrulecolor{HGRuleStrong}\hline
\end{tabular}
\global\arrayrulewidth=0.4pt\arrayrulecolor{HGRule}
\endgroup
\end{table}

\paragraph{SS-E and ScreenSpot results, part 1.}
On SS-E, the larger Holo2 checkpoint trades timely completion for higher final accuracy in \cref{tab:d_h200_screenspot}.
Holo2-235B-A22B attains 36.4\% Acc yet solves only 8.3\% of tasks within the expert reference time, whereas UI-Venus-1.5-8B completes 24.2\% on time and scores higher on H-GATE.
The larger checkpoint's late successes secure an accuracy edge, while the smaller UI-Venus model's timely responses drive superior efficiency.

\paragraph{SS-E and ScreenSpot results, part 2.}
Qwen3.5-397B-A17B leads both SS-E accuracy and H-GATE across the full H200 sweep, yet its 43.0 H-GATE still trails the human reference by 23.2 points in \cref{tab:d_h200_screenspot_part2}.
ScreenSpot-v2 sharpens this latency divide, where Qwen3.5-397B-A17B-T tops Acc at 96.8\% yet completes only 1.2\% of tasks correctly within expert reference times.
High static accuracy can therefore coexist with severe response delays, while SS-E action windows further invalidate late localizations.

\paragraph{OSWorld-G.}
Response speed separates the top two H-GATE contenders on OSWorld-G in \cref{tab:d_h200_osworldg,tab:d_h200_osworldg_part2}.
UI-Venus-2-9B surpasses UI-Venus-1.5-30B-A3B by 3.9 accuracy points, yet the latter averages 0.28 seconds compared with 0.35 seconds for UI-Venus-2-9B.
This latency edge nearly cancels the accuracy gap and narrows their H-GATE separation to 0.13 points.

\paragraph{VisualWebBench.}
Enabling thinking lowers efficiency even when nearly every solved VisualWebBench task already meets its expert reference in \cref{tab:d_h200_visualweb}.
For Holo2-30B-A3B, thinking adds 0.4 accuracy points yet raises mean latency from 0.53 to 0.75 seconds and cuts H-GATE by 4.9 points.
Even when timely completion remains high, slower responses continuously erode efficiency credit.

\paragraph{Mean latency above P95.}
Mean latency exceeds P95 in seven H200 cells, for example 1.37\,s against 1.05\,s for Holo2-8B on SS-E.
This ordering arises when rare attempts run far longer than the rest.
Because at least 95\% of attempts finish within P95, the slowest 5\% of Holo2-8B attempts on SS-E must average more than 7\,s and contribute more than a quarter of its total latency.
The other six cells are VISTA-4B on ScreenSpot-v2, UI-Venus-1.5-2B on VisualWebBench, and ScaleCUA-32B, Holo2-8B-T, POINTS-GUI-8B, and Qwen3.5-4B on OSWorld-G.
\Cref{sec:metric_diagnostics} documents the same heavy-tail pattern for Holo2-4B-T, which is why the tables report P95 alongside the mean.

\subsection{A100}
\begin{table*}
\centering
\caption{\textbf{Full results on A100: SS-E and the ScreenSpot benchmarks.} Columns follow Table~\ref{tab:d_h200_screenspot}. Bold marks peak model H-GATE and underlining marks peak model Acc.}
\label{tab:d_a100_screenspot}
\begingroup
\footnotesize
\setlength{\tabcolsep}{1.0pt}
\renewcommand{\arraystretch}{1.08}
\arrayrulecolor{HGRule}
\begin{tabular}{c|c!{\color{HGRuleStrong}\vrule width 0.75pt}cccccc@{\hspace{7pt}}ccccc@{\hspace{5pt}}ccccc}
\noalign{\color{HGRuleStrong}\hrule height 0.8pt}
\multirow{2}{*}{Family} & \multirow{2}{*}{\shortstack{Model\\ /mode}} & \multicolumn{6}{c@{\hspace{7pt}}}{\cellcolor{HGHeader}\textbf{ScreenSpot-Efficiency}} & \multicolumn{5}{c}{\textbf{ScreenSpot-Pro}} & \multicolumn{5}{c}{\textbf{ScreenSpot-v2}} \\
\noalign{\global\arrayrulewidth=0.7pt}\arrayrulecolor{HGRuleStrong}\hhline{~~*{16}{-}}\arrayrulecolor{HGRule}\noalign{\global\arrayrulewidth=0.4pt}
 &  & \cellcolor{HGHeader}Acc-S & \cellcolor{HGHeader}Acc & \cellcolor{HGHeader}\shortstack{Acc@\\Human} & \cellcolor{HGHeader}Time & \cellcolor{HGHeader}P95 & \cellcolor{HGHeader}H-GATE & Acc & \shortstack{Acc@\\Human} & Time & P95 & H-GATE & Acc & \shortstack{Acc@\\Human} & Time & P95 & H-GATE \\
\noalign{\color{HGRuleStrong}\hrule height 0.7pt}
\textbf{Human} & \textemdash & \cellcolor{HGBody}100.0 & \cellcolor{HGBody}99.0 & \cellcolor{HGBody}\textemdash & \cellcolor{HGBody}2.35 & \cellcolor{HGBody}\textemdash & \cellcolor{HGBody}66.2 & 99.2 & \textemdash & 1.46 & \textemdash & 66.2 & 100.0 & \textemdash & 0.97 & \textemdash & 66.7 \\
\hline
\multirow{1}{*}{UI-TARS-1.5} & 7B & \cellcolor{HGBody}24.2 & \cellcolor{HGBody}16.2 & \cellcolor{HGBody}3.8 & \cellcolor{HGBody}3.07 & \cellcolor{HGBody}13.13 & \cellcolor{HGBody}24.1 & 41.8 & 8.9 & 2.39 & 5.65 & 33.3 & 89.3 & 21.2 & 1.18 & 1.58 & 59.2 \\
\hline
\multirow{2}{*}{Jedi} & 3B & \cellcolor{HGBody}21.1 & \cellcolor{HGBody}13.9 & \cellcolor{HGBody}4.4 & \cellcolor{HGBody}2.77 & \cellcolor{HGBody}10.24 & \cellcolor{HGBody}22.6 & 36.2 & 35.7 & 0.99 & 1.30 & 39.1 & 88.9 & 70.9 & 0.84 & 0.93 & 63.5 \\
 & 7B & \cellcolor{HGBody}17.5 & \cellcolor{HGBody}11.6 & \cellcolor{HGBody}2.4 & \cellcolor{HGBody}2.65 & \cellcolor{HGBody}9.74 & \cellcolor{HGBody}21.9 & 34.2 & 32.7 & 1.08 & 1.29 & 37.2 & 91.2 & 54.2 & 0.90 & 1.00 & 63.8 \\
\hline
\multirow{2}{*}{SE-GUI} & 3B & \cellcolor{HGBody}17.5 & \cellcolor{HGBody}12.9 & \cellcolor{HGBody}2.6 & \cellcolor{HGBody}4.42 & \cellcolor{HGBody}17.55 & \cellcolor{HGBody}18.8 & 34.3 & 0.3 & 2.37 & 3.08 & 28.8 & 86.9 & 0.6 & 1.65 & 1.98 & 52.2 \\
 & 7B & \cellcolor{HGBody}20.5 & \cellcolor{HGBody}15.0 & \cellcolor{HGBody}3.0 & \cellcolor{HGBody}3.60 & \cellcolor{HGBody}7.57 & \cellcolor{HGBody}20.5 & 46.5 & 0.3 & 2.44 & 3.34 & 34.1 & 89.9 & 5.7 & 1.65 & 1.95 & 53.7 \\
\hline
\multirow{2}{*}{OpenCUA} & 7B & \cellcolor{HGBody}20.8 & \cellcolor{HGBody}13.9 & \cellcolor{HGBody}4.2 & \cellcolor{HGBody}2.08 & \cellcolor{HGBody}7.21 & \cellcolor{HGBody}24.0 & 46.0 & 21.2 & 2.33 & 5.94 & 37.3 & 93.2 & 40.3 & 0.98 & 1.28 & 63.9 \\
 & 32B & \cellcolor{HGBody}21.0 & \cellcolor{HGBody}15.8 & \cellcolor{HGBody}2.8 & \cellcolor{HGBody}8.16 & \cellcolor{HGBody}17.78 & \cellcolor{HGBody}17.8 & 55.7 & 0.0 & 4.30 & 8.81 & 32.6 & 93.4 & 0.2 & 2.55 & 3.17 & 49.6 \\
\hline
\multirow{6}{*}{ScaleCUA} & 3B & \cellcolor{HGBody}19.1 & \cellcolor{HGBody}12.9 & \cellcolor{HGBody}10.1 & \cellcolor{HGBody}1.59 & \cellcolor{HGBody}3.09 & \cellcolor{HGBody}25.8 & 39.8 & 29.6 & 1.32 & 1.64 & 37.7 & 87.9 & 72.1 & 0.83 & 0.91 & 63.2 \\
 & 3B-T & \cellcolor{HGBody}22.0 & \cellcolor{HGBody}15.0 & \cellcolor{HGBody}1.8 & \cellcolor{HGBody}6.62 & \cellcolor{HGBody}14.78 & \cellcolor{HGBody}16.0 & 29.6 & 0.0 & 3.59 & 7.21 & 23.0 & 79.8 & 0.0 & 3.90 & 6.92 & 39.9 \\
 & 7B & \cellcolor{HGBody}21.3 & \cellcolor{HGBody}14.4 & \cellcolor{HGBody}11.7 & \cellcolor{HGBody}1.55 & \cellcolor{HGBody}3.00 & \cellcolor{HGBody}26.5 & 47.8 & 31.9 & 1.37 & 1.71 & 41.3 & 92.6 & 74.5 & 0.83 & 0.91 & 65.7 \\
 & 7B-T & \cellcolor{HGBody}24.2 & \cellcolor{HGBody}17.7 & \cellcolor{HGBody}2.4 & \cellcolor{HGBody}5.04 & \cellcolor{HGBody}11.91 & \cellcolor{HGBody}19.0 & 29.1 & 0.0 & 2.76 & 3.20 & 25.0 & 85.4 & 0.0 & 2.57 & 3.13 & 46.4 \\
 & 32B & \cellcolor{HGBody}22.4 & \cellcolor{HGBody}15.7 & \cellcolor{HGBody}1.9 & \cellcolor{HGBody}3.52 & \cellcolor{HGBody}7.74 & \cellcolor{HGBody}20.5 & 58.7 & 0.0 & 2.93 & 3.29 & 37.1 & 95.1 & 1.1 & 1.88 & 2.05 & 54.1 \\
 & 32B-T & \cellcolor{HGBody}25.9 & \cellcolor{HGBody}16.5 & \cellcolor{HGBody}0.4 & \cellcolor{HGBody}10.54 & \cellcolor{HGBody}20.53 & \cellcolor{HGBody}13.4 & 43.7 & 0.0 & 6.34 & 7.26 & 23.5 & 86.4 & 0.0 & 7.30 & 13.43 & 36.9 \\
\hline
\multirow{1}{*}{GroundNext} & 7B & \cellcolor{HGBody}18.3 & \cellcolor{HGBody}13.2 & \cellcolor{HGBody}2.8 & \cellcolor{HGBody}3.28 & \cellcolor{HGBody}3.57 & \cellcolor{HGBody}19.9 & 53.0 & 6.3 & 2.18 & 4.41 & 38.9 & 90.4 & 13.4 & 1.36 & 1.65 & 57.0 \\
\hline
\multirow{6}{*}{Holo2} & 4B & \cellcolor{HGBody}24.3 & \cellcolor{HGBody}17.7 & \cellcolor{HGBody}16.1 & \cellcolor{HGBody}4.91 & \cellcolor{HGBody}1.36 & \cellcolor{HGBody}33.3 & 57.8 & 40.8 & 1.13 & 3.75 & 50.5 & 93.4 & 93.2 & 0.50 & 0.66 & 74.1 \\
 & 4B-T & \cellcolor{HGBody}24.9 & \cellcolor{HGBody}18.0 & \cellcolor{HGBody}16.3 & \cellcolor{HGBody}6.92 & \cellcolor{HGBody}1.47 & \cellcolor{HGBody}33.4 & 58.0 & 40.5 & 1.15 & 3.72 & 50.4 & 93.3 & 93.2 & 0.51 & 0.67 & 73.8 \\
 & 8B & \cellcolor{HGBody}28.2 & \cellcolor{HGBody}20.8 & \cellcolor{HGBody}17.8 & \cellcolor{HGBody}6.60 & \cellcolor{HGBody}1.36 & \cellcolor{HGBody}32.6 & 60.5 & 36.9 & 1.57 & 5.22 & 48.1 & 94.3 & 83.1 & 0.70 & 0.94 & 69.7 \\
 & 8B-T & \cellcolor{HGBody}27.2 & \cellcolor{HGBody}19.0 & \cellcolor{HGBody}3.1 & \cellcolor{HGBody}21.12 & \cellcolor{HGBody}15.15 & \cellcolor{HGBody}20.5 & 57.4 & 5.4 & 2.64 & 4.73 & 38.8 & 94.0 & 36.2 & 1.44 & 2.94 & 60.8 \\
 & 30B-A3B & \cellcolor{HGBody}30.6 & \cellcolor{HGBody}23.2 & \cellcolor{HGBody}20.3 & \cellcolor{HGBody}2.85 & \cellcolor{HGBody}3.91 & \cellcolor{HGBody}34.0 & 65.2 & 38.5 & 1.55 & 5.35 & 50.7 & 95.8 & 91.1 & 0.64 & 0.88 & 71.8 \\
 & 30B-A3B-T & \cellcolor{HGBody}27.5 & \cellcolor{HGBody}19.8 & \cellcolor{HGBody}16.2 & \cellcolor{HGBody}4.75 & \cellcolor{HGBody}4.07 & \cellcolor{HGBody}29.1 & 63.5 & 36.8 & 1.64 & 5.27 & 48.8 & 95.8 & 87.7 & 0.68 & 0.91 & 70.7 \\
\hline
\multirow{1}{*}{POINTS-GUI} & 8B & \cellcolor{HGBody}31.7 & \cellcolor{HGBody}23.6 & \cellcolor{HGBody}4.5 & \cellcolor{HGBody}2.15 & \cellcolor{HGBody}5.51 & \cellcolor{HGBody}28.2 & 59.3 & 13.2 & 1.71 & 2.25 & 44.3 & 95.2 & 26.7 & 1.15 & 1.56 & 62.7 \\
\hline
\multirow{3}{*}{UI-Venus-1.5} & 2B & \cellcolor{HGBody}29.2 & \cellcolor{HGBody}16.1 & \cellcolor{HGBody}15.1 & \cellcolor{HGBody}1.23 & \cellcolor{HGBody}3.12 & \cellcolor{HGBody}31.2 & 58.8 & 49.3 & 0.98 & 3.39 & \cellcolor{HGBest}\textbf{53.2} & 93.3 & 93.3 & 0.39 & 0.57 & \cellcolor{HGBest}\textbf{77.3} \\
 & 8B & \cellcolor{HGBody}37.7 & \cellcolor{HGBody}28.4 & \cellcolor{HGBody}24.0 & \cellcolor{HGBody}1.45 & \cellcolor{HGBody}4.17 & \cellcolor{HGBest}\textbf{35.4} & 68.2 & 40.7 & 1.65 & 5.74 & 51.6 & 96.2 & 91.9 & 0.66 & 0.87 & 71.7 \\
 & 30B-A3B & \cellcolor{HGBody}40.3 & \cellcolor{HGBody}\underline{32.2} & \cellcolor{HGBody}16.9 & \cellcolor{HGBody}3.85 & \cellcolor{HGBody}6.46 & \cellcolor{HGBody}34.1 & 69.1 & 33.9 & 1.91 & 5.57 & 49.1 & \underline{96.5} & 38.3 & 0.99 & 1.21 & 64.9 \\
\hline
\multirow{6}{*}{GUI-Owl-1.5} & 2B & \cellcolor{HGBody}21.9 & \cellcolor{HGBody}15.0 & \cellcolor{HGBody}14.5 & \cellcolor{HGBody}0.74 & \cellcolor{HGBody}1.54 & \cellcolor{HGBody}33.4 & 58.2 & 46.7 & 0.98 & 1.86 & 51.8 & 90.1 & 89.7 & 0.50 & 0.66 & 72.4 \\
 & 4B & \cellcolor{HGBody}21.8 & \cellcolor{HGBody}14.5 & \cellcolor{HGBody}13.4 & \cellcolor{HGBody}0.96 & \cellcolor{HGBody}2.33 & \cellcolor{HGBody}30.7 & 67.0 & 42.3 & 1.25 & 2.31 & 53.1 & 93.1 & 90.8 & 0.67 & 0.84 & 69.5 \\
 & 8B & \cellcolor{HGBody}26.3 & \cellcolor{HGBody}18.3 & \cellcolor{HGBody}16.0 & \cellcolor{HGBody}1.15 & \cellcolor{HGBody}2.77 & \cellcolor{HGBody}31.1 & 70.5 & 38.8 & 1.71 & 3.09 & 50.8 & 94.2 & 48.6 & 0.93 & 1.15 & 64.9 \\
 & 8B-T & \cellcolor{HGBody}26.0 & \cellcolor{HGBody}18.3 & \cellcolor{HGBody}4.4 & \cellcolor{HGBody}2.97 & \cellcolor{HGBody}5.85 & \cellcolor{HGBody}23.5 & 60.2 & 0.1 & 2.73 & 4.05 & 39.1 & 92.7 & 2.0 & 1.78 & 2.32 & 54.0 \\
 & 32B & \cellcolor{HGBody}27.0 & \cellcolor{HGBody}18.9 & \cellcolor{HGBody}2.8 & \cellcolor{HGBody}3.27 & \cellcolor{HGBody}6.41 & \cellcolor{HGBody}22.7 & \underline{72.7} & 0.0 & 4.05 & 6.53 & 40.2 & 94.9 & 0.0 & 2.61 & 3.07 & 49.5 \\
 & 32B-T & \cellcolor{HGBody}24.0 & \cellcolor{HGBody}15.0 & \cellcolor{HGBody}0.8 & \cellcolor{HGBody}10.47 & \cellcolor{HGBody}19.13 & \cellcolor{HGBody}12.9 & 55.6 & 0.0 & 7.97 & 10.22 & 26.8 & 92.6 & 0.0 & 5.79 & 7.40 & 40.3 \\
\noalign{\color{HGRuleStrong}\hrule height 0.8pt}
\multirow{12}{*}{Qwen3.5} & 0.8B & \cellcolor{HGBody}17.5 & \cellcolor{HGBody}11.7 & \cellcolor{HGBody}10.8 & \cellcolor{HGBody}1.52 & \cellcolor{HGBody}5.49 & \cellcolor{HGBody}27.9 & 29.7 & 28.7 & 0.67 & 1.68 & 40.2 & 77.0 & 77.0 & 0.34 & 0.47 & 69.5 \\
 & 0.8B-T & \cellcolor{HGBody}13.1 & \cellcolor{HGBody}5.9 & \cellcolor{HGBody}1.3 & \cellcolor{HGBody}11.63 & \cellcolor{HGBody}45.62 & \cellcolor{HGBody}12.5 & 14.0 & 1.3 & 6.65 & 14.78 & 13.9 & 66.4 & 39.7 & 1.68 & 4.28 & 47.3 \\
 & 2B & \cellcolor{HGBody}23.7 & \cellcolor{HGBody}16.1 & \cellcolor{HGBody}14.7 & \cellcolor{HGBody}0.78 & \cellcolor{HGBody}1.17 & \cellcolor{HGBody}32.3 & 42.8 & 34.5 & 1.08 & 2.98 & 42.9 & 81.4 & 81.1 & 0.51 & 0.67 & 67.1 \\
 & 2B-T & \cellcolor{HGBody}27.3 & \cellcolor{HGBody}16.4 & \cellcolor{HGBody}1.5 & \cellcolor{HGBody}14.46 & \cellcolor{HGBody}53.98 & \cellcolor{HGBody}15.5 & 37.0 & 1.1 & 8.22 & 16.08 & 21.6 & 80.4 & 1.7 & 4.00 & 7.91 & 41.4 \\
 & 4B & \cellcolor{HGBody}25.5 & \cellcolor{HGBody}19.2 & \cellcolor{HGBody}15.1 & \cellcolor{HGBody}1.74 & \cellcolor{HGBody}2.38 & \cellcolor{HGBody}31.0 & 47.6 & 30.3 & 1.33 & 3.88 & 42.9 & 89.0 & 85.9 & 0.69 & 0.86 & 66.8 \\
 & 4B-T & \cellcolor{HGBody}30.1 & \cellcolor{HGBody}16.8 & \cellcolor{HGBody}0.2 & \cellcolor{HGBody}69.88 & \cellcolor{HGBody}215.91 & \cellcolor{HGBody}8.9 & 40.4 & 0.0 & 56.10 & 56.10 & 14.7 & 86.7 & 0.1 & 11.89 & 22.77 & 34.8 \\
 & 9B & \cellcolor{HGBody}28.6 & \cellcolor{HGBody}21.9 & \cellcolor{HGBody}17.4 & \cellcolor{HGBody}1.36 & \cellcolor{HGBody}2.70 & \cellcolor{HGBody}31.4 & 50.0 & 28.1 & 1.71 & 5.26 & 41.6 & 91.3 & 74.9 & 0.75 & 0.98 & 67.0 \\
 & 9B-T & \cellcolor{HGBody}34.5 & \cellcolor{HGBody}19.4 & \cellcolor{HGBody}0.2 & \cellcolor{HGBody}112.67 & \cellcolor{HGBody}421.31 & \cellcolor{HGBody}9.0 & 46.7 & 0.0 & 55.33 & 140.70 & 17.8 & 89.9 & 0.0 & 23.41 & 44.43 & 33.3 \\
 & 27B & \cellcolor{HGBody}38.0 & \cellcolor{HGBody}24.1 & \cellcolor{HGBody}3.5 & \cellcolor{HGBody}4.07 & \cellcolor{HGBody}10.88 & \cellcolor{HGBody}24.0 & 54.9 & 0.2 & 3.18 & 7.81 & 35.8 & 90.3 & 5.9 & 1.73 & 2.18 & 53.4 \\
 & 27B-T & \cellcolor{HGBody}46.3 & \cellcolor{HGBody}28.4 & \cellcolor{HGBody}0.0 & \cellcolor{HGBody}251.17 & \cellcolor{HGBody}1109.21 & \cellcolor{HGBody}11.0 & 64.5 & 0.0 & 157.95 & 558.40 & 23.0 & 95.8 & 0.0 & 33.34 & 109.85 & 36.8 \\
 & 35B-A3B & \cellcolor{HGBody}31.7 & \cellcolor{HGBody}25.1 & \cellcolor{HGBody}14.6 & \cellcolor{HGBody}2.11 & \cellcolor{HGBody}5.92 & \cellcolor{HGBody}31.1 & 28.8 & 15.9 & 1.67 & 5.52 & 31.3 & 68.2 & 56.5 & 0.76 & 1.07 & 54.8 \\
 & 35B-A3B-T & \cellcolor{HGBody}41.8 & \cellcolor{HGBody}27.1 & \cellcolor{HGBody}1.8 & \cellcolor{HGBody}25.04 & \cellcolor{HGBody}101.38 & \cellcolor{HGBody}19.1 & 54.3 & 1.2 & 5.36 & 16.79 & 34.4 & 88.1 & 6.7 & 1.46 & 1.72 & 54.7 \\
\hline
\multirow{1}{*}{MolmoPoint-GUI} & 8B & \cellcolor{HGBody}22.6 & \cellcolor{HGBody}15.9 & \cellcolor{HGBody}2.7 & \cellcolor{HGBody}4.83 & \cellcolor{HGBody}11.29 & \cellcolor{HGBody}20.4 & 61.2 & 2.2 & 2.12 & 2.52 & 42.4 & 93.9 & 14.9 & 1.51 & 2.02 & 57.6 \\
\hline
\multirow{2}{*}{VISTA} & 4B & \cellcolor{HGBody}31.5 & \cellcolor{HGBody}23.3 & \cellcolor{HGBody}20.0 & \cellcolor{HGBody}1.46 & \cellcolor{HGBody}2.92 & \cellcolor{HGBody}31.7 & 63.0 & 40.2 & 1.33 & 3.33 & 50.5 & 94.8 & 88.1 & 0.73 & 0.90 & 68.9 \\
 & 9B & \cellcolor{HGBody}32.3 & \cellcolor{HGBody}23.7 & \cellcolor{HGBody}17.8 & \cellcolor{HGBody}1.87 & \cellcolor{HGBody}4.21 & \cellcolor{HGBody}30.2 & 67.8 & 39.8 & 1.70 & 4.45 & 50.0 & 96.0 & 62.8 & 0.86 & 1.04 & 67.1 \\
\noalign{\global\arrayrulewidth=0.8pt}\arrayrulecolor{HGRuleStrong}\hline
\end{tabular}
\global\arrayrulewidth=0.4pt\arrayrulecolor{HGRule}
\endgroup
\end{table*}

\begin{table}
\centering
\caption{\textbf{Full results on A100: OSWorld-G.} Columns follow Table~\ref{tab:d_h200_screenspot}. Bold marks peak model H-GATE and underlining marks peak model Acc.}
\label{tab:d_a100_osworldg}
\begingroup
\footnotesize
\setlength{\tabcolsep}{1.32pt}
\renewcommand{\arraystretch}{1.08}
\arrayrulecolor{HGRule}
\begin{tabular}{c|c!{\color{HGRuleStrong}\vrule width 0.75pt}ccccc}
\noalign{\color{HGRuleStrong}\hrule height 0.8pt}
\multirow{2}{*}{Family} & \multirow{2}{*}{\shortstack{Model\\ /mode}} & \multicolumn{5}{c}{\textbf{OSWorld-G}} \\
\noalign{\global\arrayrulewidth=0.7pt}\arrayrulecolor{HGRuleStrong}\hhline{~~*{5}{-}}\arrayrulecolor{HGRule}\noalign{\global\arrayrulewidth=0.4pt}
 &  & Acc & \shortstack{Acc@\\Human} & Time & P95 & H-GATE \\
\noalign{\color{HGRuleStrong}\hrule height 0.7pt}
\textbf{Human} & \textemdash & 99.9 & \textemdash & 1.11 & \textemdash & 66.6 \\
\hline
\multirow{1}{*}{UI-TARS-1.5} & 7B & 53.2 & 45.6 & 0.84 & 0.98 & 47.0 \\
\hline
\multirow{2}{*}{Jedi} & 3B & 49.6 & 39.9 & 2.35 & 2.62 & 42.7 \\
 & 7B & 40.4 & 28.7 & 1.64 & 3.67 & 35.9 \\
\hline
\multirow{2}{*}{SE-GUI} & 3B & 37.8 & 6.4 & 1.41 & 1.50 & 32.8 \\
 & 7B & 47.7 & 10.6 & 1.37 & 1.49 & 38.0 \\
\hline
\multirow{2}{*}{OpenCUA} & 7B & 55.9 & 53.5 & 0.74 & 0.84 & 49.9 \\
 & 32B & 58.5 & 0.4 & 2.05 & 2.27 & 38.2 \\
\hline
\multirow{6}{*}{ScaleCUA} & 3B & 51.2 & 47.5 & 0.79 & 0.90 & 46.5 \\
 & 3B-T & 39.0 & 0.0 & 3.70 & 10.80 & 25.0 \\
 & 7B & 56.2 & 52.3 & 0.79 & 0.91 & 49.3 \\
 & 7B-T & 44.5 & 0.0 & 2.62 & 3.69 & 29.7 \\
 & 32B & 60.6 & 2.0 & 1.85 & 2.08 & 40.5 \\
 & 32B-T & 53.0 & 0.0 & 6.86 & 8.28 & 25.2 \\
\hline
\multirow{1}{*}{GroundNext} & 7B & 56.4 & 25.9 & 1.16 & 1.25 & 44.2 \\
\hline
\multirow{6}{*}{Holo2} & 4B & 62.4 & 62.4 & 0.34 & 0.39 & 62.1 \\
 & 4B-T & 62.6 & 62.6 & 0.36 & 0.40 & 61.8 \\
 & 8B & 63.3 & 63.3 & 0.48 & 0.54 & 59.1 \\
 & 8B-T & 61.9 & 11.0 & 1.74 & 2.86 & 42.7 \\
 & 30B-A3B & 67.0 & 67.0 & 0.45 & 0.52 & 61.9 \\
 & 30B-A3B-T & 66.0 & 66.0 & 0.53 & 0.60 & 59.4 \\
\hline
\multirow{1}{*}{POINTS-GUI} & 8B & 66.0 & 29.6 & 1.45 & 1.28 & 49.0 \\
\hline
\multirow{3}{*}{UI-Venus-1.5} & 2B & 58.3 & 58.3 & 0.26 & 0.30 & 62.1 \\
 & 8B & 69.3 & 69.3 & 0.47 & 0.54 & \cellcolor{HGBest}\textbf{62.8} \\
 & 30B-A3B & \underline{71.3} & 67.7 & 0.78 & 0.86 & 57.2 \\
\hline
\multirow{6}{*}{GUI-Owl-1.5} & 2B & 52.5 & 52.5 & 0.37 & 0.41 & 55.7 \\
 & 4B & 63.8 & 63.8 & 0.51 & 0.57 & 58.5 \\
 & 8B & 66.7 & 65.4 & 0.71 & 0.80 & 56.0 \\
 & 8B-T & 57.6 & 2.7 & 1.67 & 2.09 & 40.2 \\
 & 32B & 66.1 & 0.0 & 2.13 & 2.34 & 40.9 \\
 & 32B-T & 56.7 & 0.0 & 5.87 & 7.40 & 27.5 \\
\noalign{\color{HGRuleStrong}\hrule height 0.8pt}
\multirow{12}{*}{Qwen3.5} & 0.8B & 32.3 & 32.3 & 0.29 & 0.33 & 45.8 \\
 & 0.8B-T & 11.3 & 2.8 & 5.60 & 16.88 & 13.6 \\
 & 2B & 41.7 & 41.7 & 0.39 & 0.46 & 48.7 \\
 & 2B-T & 41.7 & 1.1 & 7.79 & 17.91 & 22.8 \\
 & 4B & 38.7 & 38.5 & 0.57 & 0.70 & 43.3 \\
 & 4B-T & 46.1 & 0.0 & 33.05 & 98.99 & 18.2 \\
 & 9B & 54.1 & 54.1 & 0.54 & 0.63 & 52.5 \\
 & 9B-T & 56.6 & 0.0 & 91.04 & 219.48 & 20.1 \\
 & 27B & 58.5 & 11.2 & 1.38 & 1.59 & 43.1 \\
 & 27B-T & 70.9 & 0.0 & 76.54 & 281.25 & 26.2 \\
 & 35B-A3B & 36.2 & 35.5 & 0.59 & 0.74 & 41.7 \\
 & 35B-A3B-T & 65.4 & 2.0 & 5.91 & 16.85 & 39.9 \\
\hline
\multirow{1}{*}{MolmoPoint-GUI} & 8B & 63.1 & 33.9 & 1.11 & 1.35 & 48.4 \\
\hline
\multirow{2}{*}{VISTA} & 4B & 59.6 & 59.6 & 0.58 & 0.62 & 54.6 \\
 & 9B & 67.2 & 66.7 & 0.65 & 0.72 & 57.4 \\
\noalign{\global\arrayrulewidth=0.8pt}\arrayrulecolor{HGRuleStrong}\hline
\end{tabular}
\global\arrayrulewidth=0.4pt\arrayrulecolor{HGRule}
\endgroup
\end{table}

\begin{table}
\centering
\caption{\textbf{Full results on A100: VisualWebBench.} Columns follow Table~\ref{tab:d_h200_screenspot}. Bold marks peak model H-GATE and underlining marks peak model Acc.}
\label{tab:d_a100_visualweb}
\begingroup
\footnotesize
\setlength{\tabcolsep}{1.32pt}
\renewcommand{\arraystretch}{1.08}
\arrayrulecolor{HGRule}
\begin{tabular}{c|c!{\color{HGRuleStrong}\vrule width 0.75pt}ccccc}
\noalign{\color{HGRuleStrong}\hrule height 0.8pt}
\multirow{2}{*}{Family} & \multirow{2}{*}{\shortstack{Model\\ /mode}} & \multicolumn{5}{c}{\textbf{VisualWebBench}} \\
\noalign{\global\arrayrulewidth=0.7pt}\arrayrulecolor{HGRuleStrong}\hhline{~~*{5}{-}}\arrayrulecolor{HGRule}\noalign{\global\arrayrulewidth=0.4pt}
 &  & Acc & \shortstack{Acc@\\Human} & Time & P95 & H-GATE \\
\noalign{\color{HGRuleStrong}\hrule height 0.7pt}
\textbf{Human} & \textemdash & 100.0 & \textemdash & 1.06 & \textemdash & 66.7 \\
\hline
\multirow{1}{*}{UI-TARS-1.5} & 7B & 88.2 & 88.2 & 0.69 & 0.73 & 67.4 \\
\hline
\multirow{2}{*}{Jedi} & 3B & 89.3 & 11.6 & 1.18 & 1.23 & 59.5 \\
 & 7B & 89.0 & 18.8 & 1.14 & 1.20 & 59.9 \\
\hline
\multirow{2}{*}{SE-GUI} & 3B & 86.4 & 3.7 & 1.30 & 1.36 & 56.6 \\
 & 7B & 89.1 & 7.4 & 1.23 & 1.31 & 58.8 \\
\hline
\multirow{2}{*}{OpenCUA} & 7B & 91.5 & 91.5 & 0.63 & 0.67 & 70.3 \\
 & 32B & 90.5 & 0.0 & 1.83 & 1.96 & 53.4 \\
\hline
\multirow{6}{*}{ScaleCUA} & 3B & 89.1 & 89.1 & 0.68 & 0.72 & 68.0 \\
 & 3B-T & 87.0 & 0.0 & 3.03 & 3.51 & 45.5 \\
 & 7B & 91.3 & 91.3 & 0.65 & 0.70 & 69.8 \\
 & 7B-T & 90.1 & 0.0 & 2.20 & 2.77 & 50.9 \\
 & 32B & 90.1 & 0.4 & 1.57 & 1.68 & 55.5 \\
 & 32B-T & 88.8 & 0.0 & 7.40 & 9.15 & 37.8 \\
\hline
\multirow{1}{*}{GroundNext} & 7B & 90.9 & 41.7 & 1.06 & 1.11 & 62.0 \\
\hline
\multirow{6}{*}{Holo2} & 4B & 91.1 & 91.1 & 0.29 & 0.31 & 80.4 \\
 & 4B-T & 90.7 & 90.7 & 0.30 & 0.33 & 79.5 \\
 & 8B & 92.6 & 92.6 & 0.40 & 0.42 & 77.5 \\
 & 8B-T & 91.3 & 42.1 & 1.18 & 2.44 & 63.2 \\
 & 30B-A3B & 93.6 & 93.6 & 0.36 & 0.37 & 79.5 \\
 & 30B-A3B-T & \underline{94.0} & 94.0 & 0.41 & 0.44 & 77.8 \\
\hline
\multirow{1}{*}{POINTS-GUI} & 8B & 92.1 & 92.1 & 0.63 & 0.67 & 70.8 \\
\hline
\multirow{3}{*}{UI-Venus-1.5} & 2B & 90.1 & 90.1 & 0.21 & 0.22 & \cellcolor{HGBest}\textbf{82.9} \\
 & 8B & 91.5 & 91.5 & 0.38 & 0.41 & 77.2 \\
 & 30B-A3B & 92.6 & 92.6 & 0.68 & 0.71 & 69.8 \\
\hline
\multirow{6}{*}{GUI-Owl-1.5} & 2B & 87.4 & 87.4 & 0.32 & 0.34 & 77.1 \\
 & 4B & 91.1 & 91.1 & 0.46 & 0.48 & 74.7 \\
 & 8B & 90.7 & 90.7 & 0.64 & 0.68 & 69.8 \\
 & 8B-T & 87.8 & 0.2 & 1.59 & 1.95 & 54.4 \\
 & 32B & 91.3 & 0.0 & 2.00 & 2.07 & 52.5 \\
 & 32B-T & 92.6 & 0.0 & 6.19 & 7.26 & 40.3 \\
\noalign{\color{HGRuleStrong}\hrule height 0.8pt}
\multirow{12}{*}{Qwen3.5} & 0.8B & 76.9 & 76.9 & 0.24 & 0.30 & 73.5 \\
 & 0.8B-T & 69.6 & 59.7 & 1.28 & 4.30 & 54.5 \\
 & 2B & 78.5 & 78.5 & 0.33 & 0.34 & 71.6 \\
 & 2B-T & 69.0 & 28.9 & 2.96 & 7.90 & 44.8 \\
 & 4B & 84.9 & 84.9 & 0.48 & 0.55 & 70.5 \\
 & 4B-T & 79.3 & 0.0 & 19.04 & 45.84 & 30.8 \\
 & 9B & 84.9 & 84.9 & 0.45 & 0.49 & 71.5 \\
 & 9B-T & 83.1 & 0.2 & 12.17 & 30.51 & 35.4 \\
 & 27B & 90.5 & 11.6 & 1.21 & 1.29 & 59.7 \\
 & 27B-T & 92.2 & 0.0 & 29.10 & 64.70 & 37.2 \\
 & 35B-A3B & 57.2 & 57.2 & 0.52 & 0.61 & 54.2 \\
 & 35B-A3B-T & 85.9 & 25.2 & 1.22 & 1.40 & 58.0 \\
\hline
\multirow{1}{*}{MolmoPoint-GUI} & 8B & 83.3 & 18.6 & 1.15 & 1.25 & 57.0 \\
\hline
\multirow{2}{*}{VISTA} & 4B & 92.1 & 92.1 & 0.53 & 0.56 & 73.2 \\
 & 9B & 92.1 & 92.1 & 0.58 & 0.61 & 72.0 \\
\noalign{\global\arrayrulewidth=0.8pt}\arrayrulecolor{HGRuleStrong}\hline
\end{tabular}
\global\arrayrulewidth=0.4pt\arrayrulecolor{HGRule}
\endgroup
\end{table}

\paragraph{SS-E and ScreenSpot results.}
On A100, scaling UI-Venus raises SS-E accuracy yet reduces timely completions within the expert reference in \cref{tab:d_a100_screenspot}.
UI-Venus-1.5-30B-A3B gains 3.7 accuracy points over the 8B variant, whereas its Acc@Human drops from 24.0\% to 16.9\%.
Additional valid localizations place the larger checkpoint first in Acc, while the faster 8B model leads H-GATE.

\paragraph{OSWorld-G.}
The same UI-Venus size trade-off separates accuracy from efficiency on OSWorld-G in \cref{tab:d_a100_osworldg}.
UI-Venus-1.5-30B-A3B adds 2.0 accuracy points over the 8B variant, yet mean latency rises from 0.47 to 0.78 seconds and fewer successes meet the expert reference.
Consequently, the 8B model outperforms its larger counterpart by 5.5 H-GATE points even without dynamic validity windows.

\paragraph{VisualWebBench.}
Satisfying the expert reference on every correct attempt still leaves substantial room for efficiency gains on VisualWebBench in \cref{tab:d_a100_visualweb}.
Both UI-Venus-1.5-2B and Holo2-30B-A3B-T match Acc@Human to Acc, yet UI-Venus runs in roughly half the mean latency and leads by 5.1 H-GATE points.
Once binary timing thresholds saturate, H-GATE's continuous latency term continues to reward faster delivery.

\subsection{RTX 5090}
\begin{table*}
\centering
\caption{\textbf{Full results on RTX 5090: SS-E and the ScreenSpot benchmarks.} Columns follow Table~\ref{tab:d_h200_screenspot}. Bold marks peak model H-GATE and underlining marks peak model Acc.}
\label{tab:d_5090_screenspot}
\begingroup
\footnotesize
\setlength{\tabcolsep}{1.32pt}
\renewcommand{\arraystretch}{1.08}
\arrayrulecolor{HGRule}
\begin{tabular}{c|c!{\color{HGRuleStrong}\vrule width 0.75pt}cccccc@{\hspace{7pt}}ccccc@{\hspace{5pt}}ccccc}
\noalign{\color{HGRuleStrong}\hrule height 0.8pt}
\multirow{2}{*}{Family} & \multirow{2}{*}{\shortstack{Model\\ /mode}} & \multicolumn{6}{c@{\hspace{7pt}}}{\cellcolor{HGHeader}\textbf{ScreenSpot-Efficiency}} & \multicolumn{5}{c}{\textbf{ScreenSpot-Pro}} & \multicolumn{5}{c}{\textbf{ScreenSpot-v2}} \\
\noalign{\global\arrayrulewidth=0.7pt}\arrayrulecolor{HGRuleStrong}\hhline{~~*{16}{-}}\arrayrulecolor{HGRule}\noalign{\global\arrayrulewidth=0.4pt}
 &  & \cellcolor{HGHeader}Acc-S & \cellcolor{HGHeader}Acc & \cellcolor{HGHeader}\shortstack{Acc@\\Human} & \cellcolor{HGHeader}Time & \cellcolor{HGHeader}P95 & \cellcolor{HGHeader}H-GATE & Acc & \shortstack{Acc@\\Human} & Time & P95 & H-GATE & Acc & \shortstack{Acc@\\Human} & Time & P95 & H-GATE \\
\noalign{\color{HGRuleStrong}\hrule height 0.7pt}
\textbf{Human} & \textemdash & \cellcolor{HGBody}100.0 & \cellcolor{HGBody}99.0 & \cellcolor{HGBody}\textemdash & \cellcolor{HGBody}2.35 & \cellcolor{HGBody}\textemdash & \cellcolor{HGBody}66.2 & 99.2 & \textemdash & 1.46 & \textemdash & 66.2 & 100.0 & \textemdash & 0.97 & \textemdash & 66.7 \\
\hline
\multirow{1}{*}{UI-TARS-1.5} & 7B & \cellcolor{HGBody}24.2 & \cellcolor{HGBody}18.4 & \cellcolor{HGBody}9.8 & \cellcolor{HGBody}2.24 & \cellcolor{HGBody}9.52 & \cellcolor{HGBody}27.6 & 41.6 & 25.5 & 1.70 & 3.96 & 37.1 & 89.4 & 45.9 & 0.91 & 1.18 & 63.1 \\
\hline
\multirow{2}{*}{Jedi} & 3B & \cellcolor{HGBody}21.1 & \cellcolor{HGBody}13.2 & \cellcolor{HGBody}7.4 & \cellcolor{HGBody}1.89 & \cellcolor{HGBody}6.62 & \cellcolor{HGBody}24.9 & 36.2 & 36.1 & 0.89 & 1.11 & 40.5 & 88.9 & 86.8 & 0.72 & 0.82 & 66.0 \\
 & 7B & \cellcolor{HGBody}17.5 & \cellcolor{HGBody}11.2 & \cellcolor{HGBody}3.4 & \cellcolor{HGBody}2.20 & \cellcolor{HGBody}7.57 & \cellcolor{HGBody}23.0 & 34.1 & 33.4 & 1.00 & 1.13 & 37.9 & 91.2 & 68.0 & 0.85 & 0.94 & 64.8 \\
\hline
\multirow{2}{*}{SE-GUI} & 3B & \cellcolor{HGBody}17.5 & \cellcolor{HGBody}11.9 & \cellcolor{HGBody}3.0 & \cellcolor{HGBody}2.91 & \cellcolor{HGBody}8.68 & \cellcolor{HGBody}20.8 & 34.3 & 9.6 & 1.94 & 2.75 & 31.2 & 86.9 & 17.8 & 1.25 & 1.70 & 57.0 \\
 & 7B & \cellcolor{HGBody}20.5 & \cellcolor{HGBody}14.4 & \cellcolor{HGBody}2.9 & \cellcolor{HGBody}2.71 & \cellcolor{HGBody}5.53 & \cellcolor{HGBody}21.8 & 46.5 & 4.6 & 2.19 & 3.06 & 35.5 & 89.9 & 14.5 & 1.32 & 1.62 & 57.4 \\
\hline
\multirow{1}{*}{OpenCUA} & 7B & \cellcolor{HGBody}20.8 & \cellcolor{HGBody}13.8 & \cellcolor{HGBody}4.4 & \cellcolor{HGBody}1.98 & \cellcolor{HGBody}6.69 & \cellcolor{HGBody}24.3 & 46.0 & 25.5 & 2.07 & 5.46 & 38.7 & 93.2 & 54.6 & 0.87 & 1.13 & 65.8 \\
\hline
\multirow{4}{*}{ScaleCUA} & 3B & \cellcolor{HGBody}19.1 & \cellcolor{HGBody}12.9 & \cellcolor{HGBody}10.0 & \cellcolor{HGBody}1.60 & \cellcolor{HGBody}3.11 & \cellcolor{HGBody}25.6 & 39.7 & 24.8 & 1.43 & 2.00 & 36.8 & 88.0 & 72.2 & 0.82 & 0.93 & 63.5 \\
 & 3B-T & \cellcolor{HGBody}22.0 & \cellcolor{HGBody}15.9 & \cellcolor{HGBody}3.1 & \cellcolor{HGBody}5.14 & \cellcolor{HGBody}11.97 & \cellcolor{HGBody}18.3 & 29.8 & 0.1 & 2.66 & 4.11 & 25.7 & 79.7 & 0.0 & 2.99 & 5.51 & 42.8 \\
 & 7B & \cellcolor{HGBody}21.3 & \cellcolor{HGBody}14.5 & \cellcolor{HGBody}11.1 & \cellcolor{HGBody}1.66 & \cellcolor{HGBody}3.10 & \cellcolor{HGBody}26.0 & 47.8 & 17.6 & 1.52 & 1.89 & 39.9 & 92.6 & 69.2 & 0.83 & 0.95 & 65.8 \\
 & 7B-T & \cellcolor{HGBody}24.2 & \cellcolor{HGBody}18.4 & \cellcolor{HGBody}2.5 & \cellcolor{HGBody}4.20 & \cellcolor{HGBody}10.08 & \cellcolor{HGBody}20.8 & 29.2 & 0.3 & 2.32 & 2.80 & 26.8 & 85.5 & 0.3 & 2.13 & 2.51 & 49.0 \\
\hline
\multirow{1}{*}{GroundNext} & 7B & \cellcolor{HGBody}18.3 & \cellcolor{HGBody}13.1 & \cellcolor{HGBody}2.7 & \cellcolor{HGBody}3.36 & \cellcolor{HGBody}3.28 & \cellcolor{HGBody}19.7 & 53.0 & 3.4 & 2.50 & 5.37 & 37.6 & 90.4 & 15.5 & 1.38 & 1.76 & 57.1 \\
\hline
\multirow{4}{*}{Holo2} & 4B & \cellcolor{HGBody}24.3 & \cellcolor{HGBody}17.8 & \cellcolor{HGBody}16.5 & \cellcolor{HGBody}8.13 & \cellcolor{HGBody}1.33 & \cellcolor{HGBody}34.7 & 57.8 & 46.2 & 1.03 & 3.27 & 51.6 & 93.4 & 93.4 & 0.45 & 0.60 & 75.4 \\
 & 4B-T & \cellcolor{HGBody}24.9 & \cellcolor{HGBody}18.0 & \cellcolor{HGBody}16.5 & \cellcolor{HGBody}6.07 & \cellcolor{HGBody}1.34 & \cellcolor{HGBody}34.0 & 58.0 & 46.0 & 1.03 & 3.32 & 51.6 & 93.3 & 93.3 & 0.45 & 0.60 & 75.4 \\
 & 8B & \cellcolor{HGBody}28.2 & \cellcolor{HGBody}20.8 & \cellcolor{HGBody}17.9 & \cellcolor{HGBody}6.37 & \cellcolor{HGBody}1.29 & \cellcolor{HGBody}32.7 & 60.5 & 37.6 & 1.51 & 4.99 & 48.6 & 94.3 & 87.8 & 0.67 & 0.90 & 70.5 \\
 & 8B-T & \cellcolor{HGBody}27.2 & \cellcolor{HGBody}19.2 & \cellcolor{HGBody}3.3 & \cellcolor{HGBody}20.91 & \cellcolor{HGBody}14.58 & \cellcolor{HGBody}21.1 & 57.4 & 4.8 & 2.66 & 4.36 & 38.7 & 94.0 & 49.6 & 1.30 & 2.88 & 62.4 \\
\hline
\multirow{1}{*}{POINTS-GUI} & 8B & \cellcolor{HGBody}31.7 & \cellcolor{HGBody}23.4 & \cellcolor{HGBody}2.5 & \cellcolor{HGBody}2.48 & \cellcolor{HGBody}6.03 & \cellcolor{HGBody}26.8 & 59.2 & 3.7 & 1.99 & 2.64 & 42.2 & 95.2 & 18.6 & 1.35 & 1.86 & 60.3 \\
\hline
\multirow{2}{*}{UI-Venus-1.5} & 2B & \cellcolor{HGBody}29.2 & \cellcolor{HGBody}15.8 & \cellcolor{HGBody}15.0 & \cellcolor{HGBody}1.10 & \cellcolor{HGBody}2.98 & \cellcolor{HGBody}31.7 & 58.8 & 53.3 & 0.87 & 2.91 & \cellcolor{HGBest}\textbf{54.5} & 93.3 & 93.3 & 0.35 & 0.48 & \cellcolor{HGBest}\textbf{78.8} \\
 & 8B & \cellcolor{HGBody}37.7 & \cellcolor{HGBody}\underline{28.3} & \cellcolor{HGBody}25.0 & \cellcolor{HGBody}1.37 & \cellcolor{HGBody}4.09 & \cellcolor{HGBest}\textbf{36.0} & 68.2 & 41.4 & 1.52 & 5.40 & 52.7 & \underline{96.2} & 93.5 & 0.62 & 0.85 & 72.7 \\
\hline
\multirow{3}{*}{GUI-Owl-1.5} & 2B & \cellcolor{HGBody}21.9 & \cellcolor{HGBody}15.0 & \cellcolor{HGBody}14.2 & \cellcolor{HGBody}0.82 & \cellcolor{HGBody}1.80 & \cellcolor{HGBody}32.1 & 58.2 & 43.9 & 1.04 & 1.91 & 51.1 & 90.0 & 89.5 & 0.56 & 0.71 & 70.6 \\
 & 4B & \cellcolor{HGBody}21.8 & \cellcolor{HGBody}14.5 & \cellcolor{HGBody}12.7 & \cellcolor{HGBody}1.12 & \cellcolor{HGBody}2.46 & \cellcolor{HGBody}29.5 & 67.0 & 41.3 & 1.31 & 2.40 & 52.4 & 93.0 & 79.5 & 0.76 & 0.95 & 67.5 \\
 & 8B & \cellcolor{HGBody}26.3 & \cellcolor{HGBody}18.3 & \cellcolor{HGBody}15.1 & \cellcolor{HGBody}1.34 & \cellcolor{HGBody}3.21 & \cellcolor{HGBody}29.8 & \underline{70.5} & 34.0 & 1.81 & 3.24 & 49.9 & 94.2 & 33.4 & 1.04 & 1.28 & 63.2 \\
\noalign{\color{HGRuleStrong}\hrule height 0.8pt}
\multirow{8}{*}{Qwen3.5} & 0.8B & \cellcolor{HGBody}17.5 & \cellcolor{HGBody}11.7 & \cellcolor{HGBody}10.4 & \cellcolor{HGBody}1.50 & \cellcolor{HGBody}4.89 & \cellcolor{HGBody}27.3 & 29.6 & 28.0 & 0.80 & 1.97 & 38.7 & 77.0 & 77.0 & 0.41 & 0.59 & 67.4 \\
 & 0.8B-T & \cellcolor{HGBody}13.1 & \cellcolor{HGBody}6.7 & \cellcolor{HGBody}1.7 & \cellcolor{HGBody}8.58 & \cellcolor{HGBody}32.61 & \cellcolor{HGBody}14.2 & 13.9 & 10.7 & 8.19 & 39.30 & 19.8 & 67.3 & 49.4 & 1.55 & 4.49 & 51.4 \\
 & 2B & \cellcolor{HGBody}23.7 & \cellcolor{HGBody}16.1 & \cellcolor{HGBody}14.4 & \cellcolor{HGBody}0.88 & \cellcolor{HGBody}1.30 & \cellcolor{HGBody}31.4 & 43.0 & 31.1 & 1.17 & 3.32 & 42.2 & 81.6 & 80.7 & 0.56 & 0.77 & 65.9 \\
 & 2B-T & \cellcolor{HGBody}27.2 & \cellcolor{HGBody}17.0 & \cellcolor{HGBody}1.8 & \cellcolor{HGBody}11.79 & \cellcolor{HGBody}44.04 & \cellcolor{HGBody}16.8 & 36.9 & 1.3 & 7.69 & 20.62 & 22.7 & 80.2 & 18.4 & 7.26 & 10.82 & 45.2 \\
 & 4B & \cellcolor{HGBody}25.5 & \cellcolor{HGBody}19.3 & \cellcolor{HGBody}14.0 & \cellcolor{HGBody}1.75 & \cellcolor{HGBody}2.32 & \cellcolor{HGBody}30.4 & 47.6 & 29.8 & 1.38 & 3.60 & 42.5 & 89.0 & 78.5 & 0.73 & 0.98 & 66.1 \\
 & 4B-T & \cellcolor{HGBody}30.1 & \cellcolor{HGBody}16.9 & \cellcolor{HGBody}0.5 & \cellcolor{HGBody}60.19 & \cellcolor{HGBody}184.76 & \cellcolor{HGBody}9.5 & 40.4 & 0.3 & 36.61 & 69.15 & 16.4 & 87.0 & 0.1 & 23.06 & 23.06 & 31.8 \\
 & 9B & \cellcolor{HGBody}28.6 & \cellcolor{HGBody}21.9 & \cellcolor{HGBody}16.2 & \cellcolor{HGBody}1.43 & \cellcolor{HGBody}2.63 & \cellcolor{HGBody}30.8 & 50.0 & 26.5 & 1.82 & 5.12 & 40.7 & 91.3 & 56.1 & 0.84 & 1.10 & 65.3 \\
 & 9B-T & \cellcolor{HGBody}34.5 & \cellcolor{HGBody}19.4 & \cellcolor{HGBody}0.2 & \cellcolor{HGBody}103.47 & \cellcolor{HGBody}388.04 & \cellcolor{HGBody}9.2 & 47.0 & 0.0 & 62.89 & 126.60 & 17.2 & 90.1 & 0.1 & 25.70 & 97.91 & 35.5 \\
\hline
\multirow{1}{*}{MolmoPoint-GUI} & 8B & \cellcolor{HGBody}22.6 & \cellcolor{HGBody}15.8 & \cellcolor{HGBody}2.7 & \cellcolor{HGBody}4.53 & \cellcolor{HGBody}10.66 & \cellcolor{HGBody}20.5 & 61.2 & 1.5 & 2.27 & 2.76 & 41.5 & 93.9 & 16.3 & 1.50 & 2.08 & 58.0 \\
\hline
\multirow{2}{*}{VISTA} & 4B & \cellcolor{HGBody}31.5 & \cellcolor{HGBody}23.3 & \cellcolor{HGBody}19.6 & \cellcolor{HGBody}1.55 & \cellcolor{HGBody}3.11 & \cellcolor{HGBody}31.3 & 63.0 & 39.6 & 1.37 & 3.18 & 50.0 & 94.8 & 75.6 & 0.79 & 0.99 & 67.8 \\
 & 9B & \cellcolor{HGBody}32.3 & \cellcolor{HGBody}23.9 & \cellcolor{HGBody}14.3 & \cellcolor{HGBody}2.04 & \cellcolor{HGBody}4.77 & \cellcolor{HGBody}29.4 & 67.9 & 39.6 & 1.72 & 4.41 & 49.9 & 96.0 & 55.6 & 0.88 & 1.13 & 66.9 \\
\noalign{\global\arrayrulewidth=0.8pt}\arrayrulecolor{HGRuleStrong}\hline
\end{tabular}
\global\arrayrulewidth=0.4pt\arrayrulecolor{HGRule}
\endgroup
\end{table*}

\begin{table}
\centering
\caption{\textbf{Full results on RTX 5090: OSWorld-G.} Columns follow Table~\ref{tab:d_h200_screenspot}. Bold marks peak model H-GATE and underlining marks peak model Acc.}
\label{tab:d_5090_osworldg}
\begingroup
\footnotesize
\setlength{\tabcolsep}{1.32pt}
\renewcommand{\arraystretch}{1.08}
\arrayrulecolor{HGRule}
\begin{tabular}{c|c!{\color{HGRuleStrong}\vrule width 0.75pt}ccccc}
\noalign{\color{HGRuleStrong}\hrule height 0.8pt}
\multirow{2}{*}{Family} & \multirow{2}{*}{\shortstack{Model\\ /mode}} & \multicolumn{5}{c}{\textbf{OSWorld-G}} \\
\noalign{\global\arrayrulewidth=0.7pt}\arrayrulecolor{HGRuleStrong}\hhline{~~*{5}{-}}\arrayrulecolor{HGRule}\noalign{\global\arrayrulewidth=0.4pt}
 &  & Acc & \shortstack{Acc@\\Human} & Time & P95 & H-GATE \\
\noalign{\color{HGRuleStrong}\hrule height 0.7pt}
\textbf{Human} & \textemdash & 99.9 & \textemdash & 1.11 & \textemdash & 66.6 \\
\hline
\multirow{1}{*}{UI-TARS-1.5} & 7B & 53.2 & 53.0 & 0.65 & 0.76 & 50.0 \\
\hline
\multirow{2}{*}{Jedi} & 3B & 50.0 & 45.9 & 1.45 & 1.76 & 45.6 \\
 & 7B & 40.2 & 33.9 & 1.49 & 3.49 & 37.0 \\
\hline
\multirow{2}{*}{SE-GUI} & 3B & 37.8 & 28.5 & 0.95 & 1.05 & 37.4 \\
 & 7B & 47.5 & 34.0 & 0.99 & 1.11 & 41.9 \\
\hline
\multirow{1}{*}{OpenCUA} & 7B & 55.9 & 55.7 & 0.64 & 0.74 & 51.5 \\
\hline
\multirow{4}{*}{ScaleCUA} & 3B & 51.2 & 47.5 & 0.77 & 0.89 & 46.9 \\
 & 3B-T & 39.0 & 0.0 & 2.79 & 8.17 & 27.6 \\
 & 7B & 56.2 & 53.0 & 0.76 & 0.92 & 49.8 \\
 & 7B-T & 45.0 & 1.4 & 2.19 & 5.22 & 32.2 \\
\hline
\multirow{1}{*}{GroundNext} & 7B & 56.4 & 34.9 & 1.04 & 1.16 & 45.8 \\
\hline
\multirow{4}{*}{Holo2} & 4B & 62.4 & 62.4 & 0.31 & 0.36 & 63.0 \\
 & 4B-T & 62.6 & 62.6 & 0.32 & 0.36 & 62.9 \\
 & 8B & 63.3 & 63.3 & 0.45 & 0.51 & 59.8 \\
 & 8B-T & 61.9 & 14.5 & 1.44 & 2.12 & 44.5 \\
\hline
\multirow{1}{*}{POINTS-GUI} & 8B & 66.0 & 28.2 & 1.39 & 1.32 & 48.9 \\
\hline
\multirow{2}{*}{UI-Venus-1.5} & 2B & 58.3 & 58.3 & 0.22 & 0.26 & 63.2 \\
 & 8B & \underline{69.3} & 69.3 & 0.41 & 0.48 & \cellcolor{HGBest}\textbf{64.2} \\
\hline
\multirow{3}{*}{GUI-Owl-1.5} & 2B & 52.5 & 52.5 & 0.41 & 0.46 & 54.5 \\
 & 4B & 64.0 & 64.0 & 0.57 & 0.66 & 57.3 \\
 & 8B & 66.7 & 61.7 & 0.78 & 0.86 & 54.7 \\
\noalign{\color{HGRuleStrong}\hrule height 0.8pt}
\multirow{8}{*}{Qwen3.5} & 0.8B & 32.6 & 32.6 & 0.32 & 0.40 & 45.0 \\
 & 0.8B-T & 12.9 & 2.1 & 5.38 & 16.84 & 14.5 \\
 & 2B & 42.0 & 42.0 & 0.43 & 0.53 & 48.2 \\
 & 2B-T & 41.7 & 3.5 & 6.44 & 8.66 & 24.3 \\
 & 4B & 39.0 & 39.0 & 0.58 & 0.70 & 43.4 \\
 & 4B-T & 45.9 & 0.0 & 40.06 & 80.27 & 17.8 \\
 & 9B & 54.1 & 53.7 & 0.60 & 0.72 & 51.3 \\
 & 9B-T & 56.9 & 0.0 & 43.30 & 84.45 & 20.9 \\
\hline
\multirow{1}{*}{MolmoPoint-GUI} & 8B & 63.3 & 34.2 & 1.09 & 1.31 & 48.8 \\
\hline
\multirow{2}{*}{VISTA} & 4B & 59.4 & 59.4 & 0.60 & 0.66 & 54.1 \\
 & 9B & 67.4 & 66.3 & 0.69 & 0.77 & 56.8 \\
\noalign{\global\arrayrulewidth=0.8pt}\arrayrulecolor{HGRuleStrong}\hline
\end{tabular}
\global\arrayrulewidth=0.4pt\arrayrulecolor{HGRule}
\endgroup
\end{table}

\begin{table}
\centering
\caption{\textbf{Full results on RTX 5090: VisualWebBench.} Columns follow Table~\ref{tab:d_h200_screenspot}. Bold marks peak model H-GATE and underlining marks peak model Acc.}
\label{tab:d_5090_visualweb}
\begingroup
\footnotesize
\setlength{\tabcolsep}{1.32pt}
\renewcommand{\arraystretch}{1.08}
\arrayrulecolor{HGRule}
\begin{tabular}{c|c!{\color{HGRuleStrong}\vrule width 0.75pt}ccccc}
\noalign{\color{HGRuleStrong}\hrule height 0.8pt}
\multirow{2}{*}{Family} & \multirow{2}{*}{\shortstack{Model\\ /mode}} & \multicolumn{5}{c}{\textbf{VisualWebBench}} \\
\noalign{\global\arrayrulewidth=0.7pt}\arrayrulecolor{HGRuleStrong}\hhline{~~*{5}{-}}\arrayrulecolor{HGRule}\noalign{\global\arrayrulewidth=0.4pt}
 &  & Acc & \shortstack{Acc@\\Human} & Time & P95 & H-GATE \\
\noalign{\color{HGRuleStrong}\hrule height 0.7pt}
\textbf{Human} & \textemdash & 100.0 & \textemdash & 1.06 & \textemdash & 66.7 \\
\hline
\multirow{1}{*}{UI-TARS-1.5} & 7B & 88.2 & 88.2 & 0.55 & 0.58 & 70.5 \\
\hline
\multirow{2}{*}{Jedi} & 3B & 89.3 & 88.2 & 0.81 & 0.86 & 65.4 \\
 & 7B & 88.6 & 74.4 & 0.92 & 0.97 & 63.1 \\
\hline
\multirow{2}{*}{SE-GUI} & 3B & 86.4 & 84.1 & 0.82 & 0.93 & 63.7 \\
 & 7B & 89.1 & 85.1 & 0.86 & 0.92 & 64.5 \\
\hline
\multirow{1}{*}{OpenCUA} & 7B & 91.5 & 91.5 & 0.56 & 0.60 & 72.2 \\
\hline
\multirow{4}{*}{ScaleCUA} & 3B & 89.3 & 89.3 & 0.64 & 0.68 & 69.1 \\
 & 3B-T & 87.2 & 0.0 & 2.32 & 2.84 & 49.1 \\
 & 7B & 91.1 & 91.1 & 0.65 & 0.71 & 69.6 \\
 & 7B-T & 90.1 & 0.2 & 1.74 & 2.08 & 54.1 \\
\hline
\multirow{1}{*}{GroundNext} & 7B & 90.9 & 77.7 & 0.91 & 0.96 & 64.4 \\
\hline
\multirow{4}{*}{Holo2} & 4B & 91.1 & 91.1 & 0.26 & 0.28 & 81.5 \\
 & 4B-T & 90.7 & 90.7 & 0.27 & 0.29 & 80.9 \\
 & 8B & \underline{92.6} & 92.6 & 0.37 & 0.40 & 78.2 \\
 & 8B-T & 91.3 & 39.5 & 1.22 & 2.74 & 62.8 \\
\hline
\multirow{1}{*}{POINTS-GUI} & 8B & 92.1 & 92.1 & 0.69 & 0.77 & 69.5 \\
\hline
\multirow{2}{*}{UI-Venus-1.5} & 2B & 90.1 & 90.1 & 0.18 & 0.19 & \cellcolor{HGBest}\textbf{84.2} \\
 & 8B & 91.5 & 91.5 & 0.34 & 0.36 & 78.9 \\
\hline
\multirow{3}{*}{GUI-Owl-1.5} & 2B & 87.4 & 87.4 & 0.34 & 0.43 & 76.3 \\
 & 4B & 91.3 & 91.3 & 0.50 & 0.52 & 73.6 \\
 & 8B & 90.7 & 90.7 & 0.71 & 0.76 & 68.2 \\
\noalign{\color{HGRuleStrong}\hrule height 0.8pt}
\multirow{8}{*}{Qwen3.5} & 0.8B & 76.9 & 76.9 & 0.29 & 0.35 & 71.8 \\
 & 0.8B-T & 70.7 & 68.2 & 0.90 & 2.84 & 63.4 \\
 & 2B & 78.5 & 78.5 & 0.36 & 0.41 & 70.6 \\
 & 2B-T & 70.2 & 22.7 & 3.34 & 10.65 & 44.9 \\
 & 4B & 84.9 & 84.9 & 0.51 & 0.57 & 69.9 \\
 & 4B-T & 79.3 & 0.0 & 18.17 & 42.62 & 31.1 \\
 & 9B & 84.7 & 84.7 & 0.51 & 0.58 & 69.6 \\
 & 9B-T & 83.7 & 0.0 & 16.87 & 50.61 & 34.7 \\
\hline
\multirow{1}{*}{MolmoPoint-GUI} & 8B & 83.3 & 31.4 & 1.08 & 1.19 & 57.9 \\
\hline
\multirow{2}{*}{VISTA} & 4B & 91.9 & 91.9 & 0.55 & 0.57 & 72.5 \\
 & 9B & 92.1 & 92.1 & 0.63 & 0.66 & 70.7 \\
\noalign{\global\arrayrulewidth=0.8pt}\arrayrulecolor{HGRuleStrong}\hline
\end{tabular}
\global\arrayrulewidth=0.4pt\arrayrulecolor{HGRule}
\endgroup
\end{table}

\paragraph{SS-E and ScreenSpot results.}
Optimal model scale shifts across benchmarks on RTX 5090 in \cref{tab:d_5090_screenspot}.
UI-Venus-1.5-8B leads both Acc and H-GATE on SS-E and holds the top Acc on ScreenSpot-v2.
On ScreenSpot-v2, however, the 2B variant concedes 2.9 accuracy points while cutting mean latency from 0.62 to 0.35 seconds to claim the H-GATE lead.
Because compact accuracy is already near saturation on ScreenSpot-v2, lower latency dominates the efficiency trade-off.

\paragraph{OSWorld-G.}
On OSWorld-G, the larger UI-Venus checkpoint converts its extra spatial accuracy directly into timely completions in \cref{tab:d_5090_osworldg}.
UI-Venus-1.5-8B adds 11.0 accuracy points over the 2B variant, and every correct response from both models meets its expert reference.
This expanded coverage outweighs the added latency and yields a 0.95-point H-GATE advantage.

\paragraph{VisualWebBench.}
VisualWebBench favors the compact UI-Venus model whenever top-end accuracy gaps narrow in \cref{tab:d_5090_visualweb}.
Holo2-8B exceeds UI-Venus-1.5-2B by 2.5 accuracy points, yet the 2B model completes every correct response within its expert reference at a mean latency of 0.18 seconds.
This speed advantage secures a 6.0-point H-GATE lead.

\subsection{RTX 4090}
\begin{table*}
\centering
\caption{\textbf{Full results on RTX 4090: SS-E and the ScreenSpot benchmarks.} Columns follow Table~\ref{tab:d_h200_screenspot}. Bold marks peak model H-GATE and underlining marks peak model Acc.}
\label{tab:d_4090_screenspot}
\begingroup
\footnotesize
\setlength{\tabcolsep}{1.32pt}
\renewcommand{\arraystretch}{1.08}
\arrayrulecolor{HGRule}
\begin{tabular}{c|c!{\color{HGRuleStrong}\vrule width 0.75pt}cccccc@{\hspace{7pt}}ccccc@{\hspace{5pt}}ccccc}
\noalign{\color{HGRuleStrong}\hrule height 0.8pt}
\multirow{2}{*}{Family} & \multirow{2}{*}{\shortstack{Model\\ /mode}} & \multicolumn{6}{c@{\hspace{7pt}}}{\cellcolor{HGHeader}\textbf{ScreenSpot-Efficiency}} & \multicolumn{5}{c}{\textbf{ScreenSpot-Pro}} & \multicolumn{5}{c}{\textbf{ScreenSpot-v2}} \\
\noalign{\global\arrayrulewidth=0.7pt}\arrayrulecolor{HGRuleStrong}\hhline{~~*{16}{-}}\arrayrulecolor{HGRule}\noalign{\global\arrayrulewidth=0.4pt}
 &  & \cellcolor{HGHeader}Acc-S & \cellcolor{HGHeader}Acc & \cellcolor{HGHeader}\shortstack{Acc@\\Human} & \cellcolor{HGHeader}Time & \cellcolor{HGHeader}P95 & \cellcolor{HGHeader}H-GATE & Acc & \shortstack{Acc@\\Human} & Time & P95 & H-GATE & Acc & \shortstack{Acc@\\Human} & Time & P95 & H-GATE \\
\noalign{\color{HGRuleStrong}\hrule height 0.7pt}
\textbf{Human} & \textemdash & \cellcolor{HGBody}100.0 & \cellcolor{HGBody}99.0 & \cellcolor{HGBody}\textemdash & \cellcolor{HGBody}2.35 & \cellcolor{HGBody}\textemdash & \cellcolor{HGBody}66.2 & 99.2 & \textemdash & 1.46 & \textemdash & 66.2 & 100.0 & \textemdash & 0.97 & \textemdash & 66.7 \\
\hline
\multirow{1}{*}{UI-TARS-1.5} & 7B & \cellcolor{HGBody}24.2 & \cellcolor{HGBody}15.7 & \cellcolor{HGBody}3.1 & \cellcolor{HGBody}3.24 & \cellcolor{HGBody}14.07 & \cellcolor{HGBody}23.2 & 41.7 & 4.9 & 2.55 & 6.03 & 32.5 & 89.3 & 15.7 & 1.32 & 1.73 & 57.3 \\
\hline
\multirow{2}{*}{Jedi} & 3B & \cellcolor{HGBody}21.1 & \cellcolor{HGBody}13.9 & \cellcolor{HGBody}4.8 & \cellcolor{HGBody}2.58 & \cellcolor{HGBody}9.70 & \cellcolor{HGBody}22.7 & 36.4 & 33.9 & 1.07 & 1.49 & 38.3 & 89.0 & 55.9 & 0.89 & 1.00 & 62.9 \\
 & 7B & \cellcolor{HGBody}17.5 & \cellcolor{HGBody}11.4 & \cellcolor{HGBody}2.3 & \cellcolor{HGBody}2.80 & \cellcolor{HGBody}9.90 & \cellcolor{HGBody}21.3 & 34.4 & 29.6 & 1.22 & 1.53 & 35.9 & 91.0 & 28.0 & 1.00 & 1.14 & 62.1 \\
\hline
\multirow{2}{*}{SE-GUI} & 3B & \cellcolor{HGBody}17.6 & \cellcolor{HGBody}12.3 & \cellcolor{HGBody}2.2 & \cellcolor{HGBody}4.24 & \cellcolor{HGBody}17.10 & \cellcolor{HGBody}18.7 & 34.3 & 0.4 & 2.39 & 3.23 & 28.7 & 86.9 & 4.3 & 1.61 & 1.88 & 52.8 \\
 & 7B & \cellcolor{HGBody}20.5 & \cellcolor{HGBody}14.8 & \cellcolor{HGBody}3.1 & \cellcolor{HGBody}3.80 & \cellcolor{HGBody}6.53 & \cellcolor{HGBody}19.9 & 46.6 & 0.0 & 2.59 & 3.68 & 33.5 & 90.3 & 7.3 & 1.66 & 2.09 & 53.9 \\
\hline
\multirow{1}{*}{OpenCUA} & 7B & \cellcolor{HGBody}20.8 & \cellcolor{HGBody}13.8 & \cellcolor{HGBody}2.4 & \cellcolor{HGBody}2.60 & \cellcolor{HGBody}8.99 & \cellcolor{HGBody}22.2 & 46.1 & 4.4 & 3.12 & 7.72 & 34.2 & 93.2 & 17.1 & 1.27 & 1.65 & 59.9 \\
\hline
\multirow{4}{*}{ScaleCUA} & 3B & \cellcolor{HGBody}19.1 & \cellcolor{HGBody}12.9 & \cellcolor{HGBody}10.4 & \cellcolor{HGBody}1.49 & \cellcolor{HGBody}3.09 & \cellcolor{HGBody}26.3 & 40.0 & 29.5 & 1.32 & 1.66 & 37.8 & 87.7 & 82.6 & 0.76 & 0.86 & 64.5 \\
 & 3B-T & \cellcolor{HGBody}22.0 & \cellcolor{HGBody}15.9 & \cellcolor{HGBody}3.2 & \cellcolor{HGBody}5.46 & \cellcolor{HGBody}12.75 & \cellcolor{HGBody}17.9 & 30.1 & 0.0 & 2.71 & 3.69 & 25.5 & 79.4 & 0.0 & 3.07 & 5.09 & 42.4 \\
 & 7B & \cellcolor{HGBody}21.3 & \cellcolor{HGBody}14.5 & \cellcolor{HGBody}10.6 & \cellcolor{HGBody}1.77 & \cellcolor{HGBody}3.44 & \cellcolor{HGBody}25.6 & 47.8 & 15.4 & 1.57 & 1.98 & 39.7 & 92.6 & 44.9 & 0.92 & 1.01 & 64.0 \\
 & 7B-T & \cellcolor{HGBody}24.3 & \cellcolor{HGBody}17.4 & \cellcolor{HGBody}2.4 & \cellcolor{HGBody}5.52 & \cellcolor{HGBody}13.38 & \cellcolor{HGBody}18.2 & 29.3 & 0.0 & 3.05 & 3.61 & 24.2 & 85.4 & 0.0 & 2.80 & 3.34 & 45.4 \\
\hline
\multirow{1}{*}{GroundNext} & 7B & \cellcolor{HGBody}18.3 & \cellcolor{HGBody}13.3 & \cellcolor{HGBody}2.9 & \cellcolor{HGBody}3.73 & \cellcolor{HGBody}3.88 & \cellcolor{HGBody}18.9 & 53.2 & 0.5 & 2.66 & 5.58 & 36.7 & 90.5 & 9.0 & 1.60 & 2.06 & 54.6 \\
\hline
\multirow{4}{*}{Holo2} & 4B & \cellcolor{HGBody}24.3 & \cellcolor{HGBody}17.7 & \cellcolor{HGBody}15.2 & \cellcolor{HGBody}6.72 & \cellcolor{HGBody}2.21 & \cellcolor{HGBody}31.0 & 57.8 & 33.6 & 1.52 & 4.64 & 46.9 & 93.4 & 89.4 & 0.66 & 0.87 & 70.0 \\
 & 4B-T & \cellcolor{HGBody}24.9 & \cellcolor{HGBody}18.0 & \cellcolor{HGBody}15.6 & \cellcolor{HGBody}11.12 & \cellcolor{HGBody}2.64 & \cellcolor{HGBody}31.5 & 58.0 & 33.8 & 1.52 & 4.63 & 47.2 & 93.3 & 86.5 & 0.68 & 0.92 & 69.5 \\
 & 8B & \cellcolor{HGBody}28.2 & \cellcolor{HGBody}20.6 & \cellcolor{HGBody}10.1 & \cellcolor{HGBody}10.49 & \cellcolor{HGBody}1.96 & \cellcolor{HGBody}29.3 & 60.5 & 28.3 & 2.13 & 6.90 & 44.2 & 94.3 & 40.2 & 0.99 & 1.31 & 64.3 \\
 & 8B-T & \cellcolor{HGBody}27.2 & \cellcolor{HGBody}17.8 & \cellcolor{HGBody}2.0 & \cellcolor{HGBody}32.40 & \cellcolor{HGBody}23.12 & \cellcolor{HGBody}17.3 & 57.4 & 0.1 & 3.84 & 7.33 & 34.6 & 94.0 & 16.4 & 2.00 & 4.51 & 56.5 \\
\hline
\multirow{1}{*}{POINTS-GUI} & 8B & \cellcolor{HGBody}31.7 & \cellcolor{HGBody}23.5 & \cellcolor{HGBody}2.3 & \cellcolor{HGBody}2.62 & \cellcolor{HGBody}6.30 & \cellcolor{HGBody}26.3 & 59.2 & 2.3 & 2.07 & 2.72 & 41.8 & 95.2 & 18.0 & 1.39 & 1.96 & 59.8 \\
\hline
\multirow{2}{*}{UI-Venus-1.5} & 2B & \cellcolor{HGBody}29.1 & \cellcolor{HGBody}17.2 & \cellcolor{HGBody}15.8 & \cellcolor{HGBody}1.50 & \cellcolor{HGBody}4.02 & \cellcolor{HGBody}30.2 & 58.8 & 38.6 & 1.19 & 3.97 & \cellcolor{HGBest}\textbf{50.7} & 93.3 & 93.1 & 0.49 & 0.67 & \cellcolor{HGBest}\textbf{74.6} \\
 & 8B & \cellcolor{HGBody}37.6 & \cellcolor{HGBody}\underline{29.2} & \cellcolor{HGBody}16.3 & \cellcolor{HGBody}1.99 & \cellcolor{HGBody}5.82 & \cellcolor{HGBest}\textbf{32.9} & 68.2 & 31.5 & 2.19 & 6.67 & 47.7 & \underline{96.2} & 49.2 & 0.92 & 1.26 & 66.5 \\
\hline
\multirow{4}{*}{GUI-Owl-1.5} & 2B & \cellcolor{HGBody}21.8 & \cellcolor{HGBody}15.0 & \cellcolor{HGBody}14.3 & \cellcolor{HGBody}0.93 & \cellcolor{HGBody}1.83 & \cellcolor{HGBody}32.1 & 58.1 & 36.6 & 1.23 & 2.28 & 49.0 & 90.1 & 87.1 & 0.62 & 0.86 & 69.3 \\
 & 4B & \cellcolor{HGBody}21.8 & \cellcolor{HGBody}14.5 & \cellcolor{HGBody}12.3 & \cellcolor{HGBody}1.33 & \cellcolor{HGBody}3.90 & \cellcolor{HGBody}28.5 & 66.9 & 36.3 & 1.62 & 2.82 & 49.5 & 93.1 & 61.7 & 0.86 & 1.19 & 65.6 \\
 & 8B & \cellcolor{HGBody}26.3 & \cellcolor{HGBody}18.4 & \cellcolor{HGBody}13.8 & \cellcolor{HGBody}1.66 & \cellcolor{HGBody}4.04 & \cellcolor{HGBody}28.2 & \underline{70.5} & 8.0 & 2.39 & 4.24 & 46.1 & 94.2 & 14.4 & 1.35 & 1.67 & 58.8 \\
 & 8B-T & \cellcolor{HGBody}26.0 & \cellcolor{HGBody}17.8 & \cellcolor{HGBody}2.5 & \cellcolor{HGBody}4.39 & \cellcolor{HGBody}8.70 & \cellcolor{HGBody}20.2 & 60.1 & 0.0 & 3.73 & 5.36 & 35.4 & 92.7 & 0.0 & 2.68 & 3.53 & 48.4 \\
\noalign{\color{HGRuleStrong}\hrule height 0.8pt}
\multirow{8}{*}{Qwen3.5} & 0.8B & \cellcolor{HGBody}17.4 & \cellcolor{HGBody}11.7 & \cellcolor{HGBody}10.2 & \cellcolor{HGBody}1.73 & \cellcolor{HGBody}6.13 & \cellcolor{HGBody}26.6 & 29.8 & 27.5 & 0.85 & 2.12 & 38.2 & 76.9 & 76.7 & 0.41 & 0.56 & 67.3 \\
 & 0.8B-T & \cellcolor{HGBody}13.1 & \cellcolor{HGBody}6.0 & \cellcolor{HGBody}1.3 & \cellcolor{HGBody}11.41 & \cellcolor{HGBody}44.05 & \cellcolor{HGBody}12.4 & 15.2 & 8.5 & 9.04 & 30.83 & 17.2 & 67.5 & 46.8 & 2.63 & 8.59 & 49.4 \\
 & 2B & \cellcolor{HGBody}23.7 & \cellcolor{HGBody}16.1 & \cellcolor{HGBody}13.0 & \cellcolor{HGBody}1.04 & \cellcolor{HGBody}1.69 & \cellcolor{HGBody}30.2 & 42.9 & 28.6 & 1.31 & 3.55 & 41.1 & 81.7 & 77.8 & 0.63 & 0.98 & 64.3 \\
 & 2B-T & \cellcolor{HGBody}27.3 & \cellcolor{HGBody}15.9 & \cellcolor{HGBody}1.1 & \cellcolor{HGBody}16.77 & \cellcolor{HGBody}62.86 & \cellcolor{HGBody}14.4 & 37.0 & 0.1 & 8.66 & 23.45 & 21.6 & 80.0 & 0.3 & 4.65 & 10.41 & 41.4 \\
 & 4B & \cellcolor{HGBody}25.6 & \cellcolor{HGBody}19.3 & \cellcolor{HGBody}8.2 & \cellcolor{HGBody}2.41 & \cellcolor{HGBody}3.21 & \cellcolor{HGBody}28.1 & 48.5 & 26.5 & 1.76 & 4.60 & 40.0 & 89.9 & 47.7 & 0.93 & 1.23 & 62.7 \\
 & 4B-T & \cellcolor{HGBody}30.1 & \cellcolor{HGBody}16.3 & \cellcolor{HGBody}0.3 & \cellcolor{HGBody}95.18 & \cellcolor{HGBody}291.98 & \cellcolor{HGBody}8.0 & 40.6 & 0.0 & 62.34 & 62.34 & 14.7 & 86.2 & 0.0 & 21.83 & 38.81 & 33.0 \\
 & 9B & \cellcolor{HGBody}28.6 & \cellcolor{HGBody}21.9 & \cellcolor{HGBody}8.3 & \cellcolor{HGBody}1.97 & \cellcolor{HGBody}3.96 & \cellcolor{HGBody}28.1 & 50.0 & 18.0 & 2.27 & 6.36 & 38.1 & 91.3 & 34.7 & 1.02 & 1.37 & 62.1 \\
 & 9B-T & \cellcolor{HGBody}34.6 & \cellcolor{HGBody}19.1 & \cellcolor{HGBody}0.0 & \cellcolor{HGBody}169.88 & \cellcolor{HGBody}637.73 & \cellcolor{HGBody}8.2 & 46.9 & 0.0 & 88.29 & 213.83 & 16.9 & 89.9 & 0.0 & 39.68 & 105.08 & 32.3 \\
\hline
\multirow{1}{*}{MolmoPoint-GUI} & 8B & \cellcolor{HGBody}22.6 & \cellcolor{HGBody}16.4 & \cellcolor{HGBody}3.0 & \cellcolor{HGBody}5.92 & \cellcolor{HGBody}13.74 & \cellcolor{HGBody}19.0 & 61.0 & 0.0 & 2.69 & 3.23 & 39.3 & 93.9 & 11.4 & 1.96 & 2.67 & 54.1 \\
\hline
\multirow{2}{*}{VISTA} & 4B & \cellcolor{HGBody}31.5 & \cellcolor{HGBody}23.3 & \cellcolor{HGBody}18.7 & \cellcolor{HGBody}1.64 & \cellcolor{HGBody}3.33 & \cellcolor{HGBody}30.8 & 63.0 & 38.5 & 1.53 & 3.80 & 48.9 & 94.7 & 69.4 & 0.81 & 1.04 & 67.6 \\
 & 9B & \cellcolor{HGBody}32.3 & \cellcolor{HGBody}23.9 & \cellcolor{HGBody}10.0 & \cellcolor{HGBody}2.24 & \cellcolor{HGBody}5.43 & \cellcolor{HGBody}28.5 & 67.8 & 33.5 & 2.26 & 5.57 & 48.0 & 96.0 & 34.8 & 1.02 & 1.33 & 64.7 \\
\noalign{\global\arrayrulewidth=0.8pt}\arrayrulecolor{HGRuleStrong}\hline
\end{tabular}
\global\arrayrulewidth=0.4pt\arrayrulecolor{HGRule}
\endgroup
\end{table*}

\begin{table}
\centering
\caption{\textbf{Full results on RTX 4090: OSWorld-G.} Columns follow Table~\ref{tab:d_h200_screenspot}. Bold marks peak model H-GATE and underlining marks peak model Acc.}
\label{tab:d_4090_osworldg}
\begingroup
\footnotesize
\setlength{\tabcolsep}{1.32pt}
\renewcommand{\arraystretch}{1.08}
\arrayrulecolor{HGRule}
\begin{tabular}{c|c!{\color{HGRuleStrong}\vrule width 0.75pt}ccccc}
\noalign{\color{HGRuleStrong}\hrule height 0.8pt}
\multirow{2}{*}{Family} & \multirow{2}{*}{\shortstack{Model\\ /mode}} & \multicolumn{5}{c}{\textbf{OSWorld-G}} \\
\noalign{\global\arrayrulewidth=0.7pt}\arrayrulecolor{HGRuleStrong}\hhline{~~*{5}{-}}\arrayrulecolor{HGRule}\noalign{\global\arrayrulewidth=0.4pt}
 &  & Acc & \shortstack{Acc@\\Human} & Time & P95 & H-GATE \\
\noalign{\color{HGRuleStrong}\hrule height 0.7pt}
\textbf{Human} & \textemdash & 99.9 & \textemdash & 1.11 & \textemdash & 66.6 \\
\hline
\multirow{1}{*}{UI-TARS-1.5} & 7B & 53.5 & 37.2 & 0.97 & 1.12 & 45.3 \\
\hline
\multirow{2}{*}{Jedi} & 3B & 49.3 & 37.2 & 1.97 & 2.46 & 42.5 \\
 & 7B & 40.1 & 24.3 & 1.84 & 4.35 & 34.9 \\
\hline
\multirow{2}{*}{SE-GUI} & 3B & 37.4 & 9.6 & 1.33 & 1.48 & 33.3 \\
 & 7B & 48.2 & 12.1 & 1.38 & 1.73 & 38.2 \\
\hline
\multirow{1}{*}{OpenCUA} & 7B & 55.9 & 39.7 & 1.00 & 1.13 & 46.1 \\
\hline
\multirow{4}{*}{ScaleCUA} & 3B & 51.8 & 50.7 & 0.70 & 0.85 & 48.3 \\
 & 3B-T & 40.6 & 0.0 & 2.80 & 4.65 & 27.7 \\
 & 7B & 56.2 & 46.8 & 0.88 & 1.03 & 47.9 \\
 & 7B-T & 44.7 & 0.0 & 2.93 & 4.54 & 28.6 \\
\hline
\multirow{1}{*}{GroundNext} & 7B & 56.2 & 17.2 & 1.32 & 1.67 & 42.6 \\
\hline
\multirow{4}{*}{Holo2} & 4B & 62.4 & 62.4 & 0.46 & 0.52 & 59.0 \\
 & 4B-T & 62.6 & 62.6 & 0.49 & 0.55 & 58.5 \\
 & 8B & 63.3 & 62.4 & 0.68 & 0.76 & 54.9 \\
 & 8B-T & 61.9 & 2.7 & 2.37 & 3.83 & 38.9 \\
\hline
\multirow{1}{*}{POINTS-GUI} & 8B & 66.0 & 17.0 & 1.75 & 1.55 & 46.5 \\
\hline
\multirow{2}{*}{UI-Venus-1.5} & 2B & 58.3 & 58.3 & 0.31 & 0.37 & \cellcolor{HGBest}\textbf{60.7} \\
 & 8B & \underline{69.3} & 68.8 & 0.62 & 0.72 & 59.3 \\
\hline
\multirow{4}{*}{GUI-Owl-1.5} & 2B & 52.5 & 52.5 & 0.44 & 0.51 & 54.1 \\
 & 4B & 64.4 & 61.2 & 0.69 & 0.98 & 55.2 \\
 & 8B & 66.7 & 38.1 & 1.06 & 1.17 & 50.5 \\
 & 8B-T & 57.4 & 0.0 & 2.43 & 2.99 & 35.8 \\
\noalign{\color{HGRuleStrong}\hrule height 0.8pt}
\multirow{8}{*}{Qwen3.5} & 0.8B & 32.8 & 32.8 & 0.34 & 0.51 & 44.8 \\
 & 0.8B-T & 12.8 & 2.0 & 5.73 & 25.74 & 14.1 \\
 & 2B & 41.8 & 41.7 & 0.49 & 0.73 & 46.8 \\
 & 2B-T & 41.7 & 1.1 & 9.94 & 30.53 & 23.4 \\
 & 4B & 40.6 & 36.7 & 0.81 & 1.14 & 40.7 \\
 & 4B-T & 45.2 & 0.0 & 63.07 & 136.35 & 16.4 \\
 & 9B & 54.1 & 50.5 & 0.76 & 0.87 & 48.5 \\
 & 9B-T & 56.2 & 0.0 & 130.79 & 350.38 & 19.6 \\
\hline
\multirow{1}{*}{MolmoPoint-GUI} & 8B & 63.3 & 15.2 & 1.39 & 1.66 & 45.4 \\
\hline
\multirow{2}{*}{VISTA} & 4B & 59.6 & 59.6 & 0.59 & 0.68 & 54.4 \\
 & 9B & 67.2 & 65.6 & 0.71 & 0.83 & 56.2 \\
\noalign{\global\arrayrulewidth=0.8pt}\arrayrulecolor{HGRuleStrong}\hline
\end{tabular}
\global\arrayrulewidth=0.4pt\arrayrulecolor{HGRule}
\endgroup
\end{table}

\begin{table}
\centering
\caption{\textbf{Full results on RTX 4090: VisualWebBench.} Columns follow Table~\ref{tab:d_h200_screenspot}. Bold marks peak model H-GATE and underlining marks peak model Acc.}
\label{tab:d_4090_visualweb}
\begingroup
\footnotesize
\setlength{\tabcolsep}{1.32pt}
\renewcommand{\arraystretch}{1.08}
\arrayrulecolor{HGRule}
\begin{tabular}{c|c!{\color{HGRuleStrong}\vrule width 0.75pt}ccccc}
\noalign{\color{HGRuleStrong}\hrule height 0.8pt}
\multirow{2}{*}{Family} & \multirow{2}{*}{\shortstack{Model\\ /mode}} & \multicolumn{5}{c}{\textbf{VisualWebBench}} \\
\noalign{\global\arrayrulewidth=0.7pt}\arrayrulecolor{HGRuleStrong}\hhline{~~*{5}{-}}\arrayrulecolor{HGRule}\noalign{\global\arrayrulewidth=0.4pt}
 &  & Acc & \shortstack{Acc@\\Human} & Time & P95 & H-GATE \\
\noalign{\color{HGRuleStrong}\hrule height 0.7pt}
\textbf{Human} & \textemdash & 100.0 & \textemdash & 1.06 & \textemdash & 66.7 \\
\hline
\multirow{1}{*}{UI-TARS-1.5} & 7B & 87.8 & 86.0 & 0.82 & 0.88 & 64.6 \\
\hline
\multirow{2}{*}{Jedi} & 3B & 89.3 & 17.6 & 1.16 & 1.23 & 59.8 \\
 & 7B & 88.6 & 14.1 & 1.18 & 1.25 & 59.2 \\
\hline
\multirow{2}{*}{SE-GUI} & 3B & 86.0 & 8.3 & 1.23 & 1.30 & 57.3 \\
 & 7B & 89.1 & 8.3 & 1.26 & 1.53 & 58.4 \\
\hline
\multirow{1}{*}{OpenCUA} & 7B & 91.5 & 87.8 & 0.86 & 0.92 & 65.6 \\
\hline
\multirow{4}{*}{ScaleCUA} & 3B & 88.8 & 88.8 & 0.61 & 0.69 & 69.3 \\
 & 3B-T & 86.8 & 0.0 & 2.53 & 3.03 & 47.8 \\
 & 7B & 91.3 & 91.3 & 0.74 & 0.80 & 68.0 \\
 & 7B-T & 90.1 & 0.0 & 2.43 & 2.98 & 49.4 \\
\hline
\multirow{1}{*}{GroundNext} & 7B & 91.1 & 14.7 & 1.24 & 1.58 & 59.8 \\
\hline
\multirow{4}{*}{Holo2} & 4B & 91.1 & 91.1 & 0.39 & 0.42 & 76.7 \\
 & 4B-T & 90.7 & 90.7 & 0.42 & 0.45 & 75.6 \\
 & 8B & \underline{92.6} & 92.6 & 0.59 & 0.63 & 72.0 \\
 & 8B-T & 91.3 & 29.5 & 1.80 & 3.98 & 57.3 \\
\hline
\multirow{1}{*}{POINTS-GUI} & 8B & 92.1 & 91.9 & 0.74 & 0.83 & 68.3 \\
\hline
\multirow{2}{*}{UI-Venus-1.5} & 2B & 90.1 & 90.1 & 0.25 & 0.29 & \cellcolor{HGBest}\textbf{81.2} \\
 & 8B & 91.5 & 91.5 & 0.52 & 0.56 & 73.1 \\
\hline
\multirow{4}{*}{GUI-Owl-1.5} & 2B & 87.8 & 87.8 & 0.39 & 0.43 & 75.0 \\
 & 4B & 91.3 & 88.4 & 0.68 & 0.98 & 69.5 \\
 & 8B & 90.5 & 66.1 & 0.97 & 1.01 & 63.3 \\
 & 8B-T & 88.0 & 0.0 & 2.37 & 2.90 & 48.8 \\
\noalign{\color{HGRuleStrong}\hrule height 0.8pt}
\multirow{8}{*}{Qwen3.5} & 0.8B & 76.7 & 76.7 & 0.31 & 0.48 & 71.2 \\
 & 0.8B-T & 69.6 & 65.7 & 0.91 & 2.54 & 58.7 \\
 & 2B & 78.7 & 78.7 & 0.44 & 0.64 & 68.4 \\
 & 2B-T & 70.2 & 17.2 & 4.54 & 12.84 & 42.2 \\
 & 4B & 87.2 & 84.1 & 0.71 & 1.00 & 66.6 \\
 & 4B-T & 78.9 & 0.0 & 43.94 & 93.37 & 28.5 \\
 & 9B & 84.7 & 84.7 & 0.66 & 0.73 & 66.1 \\
 & 9B-T & 83.3 & 0.0 & 17.83 & 40.81 & 34.6 \\
\hline
\multirow{1}{*}{MolmoPoint-GUI} & 8B & 83.3 & 2.1 & 1.44 & 1.57 & 53.7 \\
\hline
\multirow{2}{*}{VISTA} & 4B & 92.2 & 92.2 & 0.53 & 0.58 & 73.3 \\
 & 9B & 92.1 & 92.1 & 0.62 & 0.65 & 71.0 \\
\noalign{\global\arrayrulewidth=0.8pt}\arrayrulecolor{HGRuleStrong}\hline
\end{tabular}
\global\arrayrulewidth=0.4pt\arrayrulecolor{HGRule}
\endgroup
\end{table}

\paragraph{SS-E and ScreenSpot results.}
Leading SS-E timely completions does not guarantee the highest overall efficiency in \cref{tab:d_4090_screenspot}.
VISTA-4B reaches 18.7\% Acc@Human compared with 16.3\% for UI-Venus-1.5-8B, yet UI-Venus-1.5-8B posts 5.9 points higher Acc and leads H-GATE.
The two metrics impose distinct timing structures.
Acc@Human enforces a hard cutoff at expert speed, whereas H-GATE grants latency-discounted credit to slower valid successes and penalizes failed-attempt time.

\paragraph{OSWorld-G.}
Faster execution gives UI-Venus-1.5-2B a narrow OSWorld-G efficiency lead in \cref{tab:d_4090_osworldg}.
It trails UI-Venus-1.5-8B by 11.0 accuracy points yet responds in roughly half the mean time and leads H-GATE by 1.4 points.
Although the 8B variant completes more tasks within expert references, H-GATE captures cumulative time saved across all attempts beyond binary timely counts.

\paragraph{VisualWebBench.}
When top models satisfy expert references on every correct attempt, raw latency separates accuracy and efficiency leaders on VisualWebBench in \cref{tab:d_4090_visualweb}.
Holo2-8B tops Acc at 92.6\%, whereas UI-Venus-1.5-2B lowers mean latency from 0.59 to 0.25 seconds.
Both meet expert references on all successes, so the faster model leads H-GATE by delivering comparable accuracy in far less time.

\subsection{RTX 3090}
\begin{table*}
\centering
\caption{\textbf{Full results on RTX 3090: SS-E and the ScreenSpot benchmarks.} Columns follow Table~\ref{tab:d_h200_screenspot}. Bold marks peak model H-GATE and underlining marks peak model Acc.}
\label{tab:d_3090_screenspot}
\begingroup
\footnotesize
\setlength{\tabcolsep}{1.32pt}
\renewcommand{\arraystretch}{1.08}
\arrayrulecolor{HGRule}
\begin{tabular}{c|c!{\color{HGRuleStrong}\vrule width 0.75pt}cccccc@{\hspace{7pt}}ccccc@{\hspace{5pt}}ccccc}
\noalign{\color{HGRuleStrong}\hrule height 0.8pt}
\multirow{2}{*}{Family} & \multirow{2}{*}{\shortstack{Model\\ /mode}} & \multicolumn{6}{c@{\hspace{7pt}}}{\cellcolor{HGHeader}\textbf{ScreenSpot-Efficiency}} & \multicolumn{5}{c}{\textbf{ScreenSpot-Pro}} & \multicolumn{5}{c}{\textbf{ScreenSpot-v2}} \\
\noalign{\global\arrayrulewidth=0.7pt}\arrayrulecolor{HGRuleStrong}\hhline{~~*{16}{-}}\arrayrulecolor{HGRule}\noalign{\global\arrayrulewidth=0.4pt}
 &  & \cellcolor{HGHeader}Acc-S & \cellcolor{HGHeader}Acc & \cellcolor{HGHeader}\shortstack{Acc@\\Human} & \cellcolor{HGHeader}Time & \cellcolor{HGHeader}P95 & \cellcolor{HGHeader}H-GATE & Acc & \shortstack{Acc@\\Human} & Time & P95 & H-GATE & Acc & \shortstack{Acc@\\Human} & Time & P95 & H-GATE \\
\noalign{\color{HGRuleStrong}\hrule height 0.7pt}
\textbf{Human} & \textemdash & \cellcolor{HGBody}100.0 & \cellcolor{HGBody}99.0 & \cellcolor{HGBody}\textemdash & \cellcolor{HGBody}2.35 & \cellcolor{HGBody}\textemdash & \cellcolor{HGBody}66.2 & 99.2 & \textemdash & 1.46 & \textemdash & 66.2 & 100.0 & \textemdash & 0.97 & \textemdash & 66.7 \\
\hline
\multirow{1}{*}{UI-TARS-1.5} & 7B & \cellcolor{HGBody}24.2 & \cellcolor{HGBody}14.1 & \cellcolor{HGBody}2.4 & \cellcolor{HGBody}5.06 & \cellcolor{HGBody}22.40 & \cellcolor{HGBody}19.0 & 41.4 & 0.0 & 4.30 & 9.63 & 27.2 & 89.4 & 10.1 & 2.07 & 2.76 & 51.2 \\
\hline
\multirow{2}{*}{Jedi} & 3B & \cellcolor{HGBody}21.1 & \cellcolor{HGBody}14.2 & \cellcolor{HGBody}3.1 & \cellcolor{HGBody}3.72 & \cellcolor{HGBody}13.95 & \cellcolor{HGBody}20.0 & 36.2 & 12.8 & 1.53 & 1.93 & 34.2 & 89.1 & 11.5 & 1.33 & 1.50 & 56.7 \\
 & 7B & \cellcolor{HGBody}17.5 & \cellcolor{HGBody}12.1 & \cellcolor{HGBody}2.6 & \cellcolor{HGBody}4.06 & \cellcolor{HGBody}14.17 & \cellcolor{HGBody}18.8 & 34.3 & 1.6 & 1.90 & 2.22 & 30.9 & 91.1 & 9.8 & 1.66 & 1.90 & 54.4 \\
\hline
\multirow{2}{*}{SE-GUI} & 3B & \cellcolor{HGBody}17.6 & \cellcolor{HGBody}13.1 & \cellcolor{HGBody}2.5 & \cellcolor{HGBody}5.72 & \cellcolor{HGBody}22.99 & \cellcolor{HGBody}16.9 & 34.3 & 0.0 & 3.51 & 4.77 & 25.0 & 86.9 & 0.0 & 2.32 & 2.75 & 47.8 \\
 & 7B & \cellcolor{HGBody}20.5 & \cellcolor{HGBody}14.6 & \cellcolor{HGBody}1.5 & \cellcolor{HGBody}5.41 & \cellcolor{HGBody}9.51 & \cellcolor{HGBody}17.1 & 46.6 & 0.0 & 4.22 & 5.81 & 28.5 & 90.3 & 0.0 & 2.83 & 3.53 & 47.0 \\
\hline
\multirow{1}{*}{OpenCUA} & 7B & \cellcolor{HGBody}20.8 & \cellcolor{HGBody}14.3 & \cellcolor{HGBody}3.2 & \cellcolor{HGBody}4.01 & \cellcolor{HGBody}17.50 & \cellcolor{HGBody}19.6 & 46.1 & 0.0 & 4.63 & 10.85 & 29.7 & 93.3 & 13.4 & 1.88 & 2.52 & 54.3 \\
\hline
\multirow{4}{*}{ScaleCUA} & 3B & \cellcolor{HGBody}19.1 & \cellcolor{HGBody}13.3 & \cellcolor{HGBody}5.0 & \cellcolor{HGBody}2.27 & \cellcolor{HGBody}4.52 & \cellcolor{HGBody}22.9 & 39.8 & 1.1 & 1.99 & 2.30 & 33.0 & 87.8 & 13.9 & 1.19 & 1.34 & 57.7 \\
 & 3B-T & \cellcolor{HGBody}22.0 & \cellcolor{HGBody}14.2 & \cellcolor{HGBody}0.9 & \cellcolor{HGBody}8.44 & \cellcolor{HGBody}19.03 & \cellcolor{HGBody}14.0 & 29.6 & 0.0 & 4.43 & 7.17 & 21.1 & 79.6 & 0.0 & 4.77 & 7.37 & 37.8 \\
 & 7B & \cellcolor{HGBody}21.3 & \cellcolor{HGBody}14.4 & \cellcolor{HGBody}3.0 & \cellcolor{HGBody}2.76 & \cellcolor{HGBody}5.30 & \cellcolor{HGBody}21.8 & 47.8 & 0.0 & 2.51 & 2.93 & 34.0 & 92.4 & 13.0 & 1.41 & 1.60 & 57.4 \\
 & 7B-T & \cellcolor{HGBody}24.3 & \cellcolor{HGBody}15.8 & \cellcolor{HGBody}1.5 & \cellcolor{HGBody}7.55 & \cellcolor{HGBody}18.34 & \cellcolor{HGBody}15.3 & 29.4 & 0.0 & 4.20 & 5.17 & 21.4 & 85.2 & 0.0 & 3.84 & 4.55 & 41.7 \\
\hline
\multirow{1}{*}{GroundNext} & 7B & \cellcolor{HGBody}18.3 & \cellcolor{HGBody}13.7 & \cellcolor{HGBody}1.0 & \cellcolor{HGBody}5.35 & \cellcolor{HGBody}5.42 & \cellcolor{HGBody}16.4 & 53.2 & 0.0 & 4.50 & 9.76 & 31.3 & 90.4 & 0.1 & 2.55 & 3.20 & 48.3 \\
\hline
\multirow{4}{*}{Holo2} & 4B & \cellcolor{HGBody}24.3 & \cellcolor{HGBody}17.0 & \cellcolor{HGBody}6.7 & \cellcolor{HGBody}8.44 & \cellcolor{HGBody}4.13 & \cellcolor{HGBody}26.2 & 57.8 & 17.3 & 2.61 & 8.97 & 40.6 & 93.4 & 33.2 & 1.05 & 1.45 & 63.3 \\
 & 4B-T & \cellcolor{HGBody}24.9 & \cellcolor{HGBody}17.4 & \cellcolor{HGBody}7.4 & \cellcolor{HGBody}13.16 & \cellcolor{HGBody}3.97 & \cellcolor{HGBody}26.7 & 58.0 & 15.7 & 2.63 & 8.97 & 40.6 & 93.3 & 30.1 & 1.08 & 1.48 & 62.6 \\
 & 8B & \cellcolor{HGBody}28.2 & \cellcolor{HGBody}19.9 & \cellcolor{HGBody}5.3 & \cellcolor{HGBody}14.26 & \cellcolor{HGBody}2.83 & \cellcolor{HGBody}25.3 & 60.5 & 0.6 & 4.01 & 13.52 & 37.0 & 94.3 & 15.0 & 1.64 & 2.23 & 56.8 \\
 & 8B-T & \cellcolor{HGBody}27.2 & \cellcolor{HGBody}16.5 & \cellcolor{HGBody}1.3 & \cellcolor{HGBody}42.24 & \cellcolor{HGBody}28.68 & \cellcolor{HGBody}14.6 & 57.4 & 0.0 & 6.13 & 10.17 & 30.2 & 94.0 & 9.9 & 2.97 & 6.28 & 50.6 \\
\hline
\multirow{1}{*}{POINTS-GUI} & 8B & \cellcolor{HGBody}31.7 & \cellcolor{HGBody}25.2 & \cellcolor{HGBody}4.6 & \cellcolor{HGBody}4.09 & \cellcolor{HGBody}9.49 & \cellcolor{HGBody}23.2 & 59.3 & 0.0 & 3.55 & 4.32 & 35.3 & 95.2 & 11.6 & 2.38 & 3.38 & 52.5 \\
\hline
\multirow{2}{*}{UI-Venus-1.5} & 2B & \cellcolor{HGBody}29.1 & \cellcolor{HGBody}20.8 & \cellcolor{HGBody}14.2 & \cellcolor{HGBody}2.45 & \cellcolor{HGBody}6.56 & \cellcolor{HGBody}28.0 & 58.8 & 30.4 & 2.17 & 7.87 & \cellcolor{HGBest}\textbf{43.9} & 93.3 & 60.4 & 0.79 & 1.13 & \cellcolor{HGBest}\textbf{67.7} \\
 & 8B & \cellcolor{HGBody}37.7 & \cellcolor{HGBody}\underline{25.5} & \cellcolor{HGBody}2.0 & \cellcolor{HGBody}6.44 & \cellcolor{HGBody}18.84 & \cellcolor{HGBody}21.2 & 68.2 & 0.0 & 6.63 & 20.15 & 35.0 & \underline{96.2} & 0.9 & 3.20 & 4.24 & 48.6 \\
\hline
\multirow{4}{*}{GUI-Owl-1.5} & 2B & \cellcolor{HGBody}21.8 & \cellcolor{HGBody}15.0 & \cellcolor{HGBody}13.6 & \cellcolor{HGBody}1.16 & \cellcolor{HGBody}2.25 & \cellcolor{HGBest}\textbf{30.4} & 58.1 & 29.8 & 1.98 & 3.82 & 43.4 & 90.0 & 53.1 & 0.85 & 1.17 & 64.6 \\
 & 4B & \cellcolor{HGBody}21.8 & \cellcolor{HGBody}14.6 & \cellcolor{HGBody}11.5 & \cellcolor{HGBody}1.48 & \cellcolor{HGBody}3.86 & \cellcolor{HGBody}27.2 & 66.9 & 4.7 & 2.74 & 5.29 & 43.1 & 93.1 & 17.1 & 1.34 & 1.75 & 58.9 \\
 & 8B & \cellcolor{HGBody}26.3 & \cellcolor{HGBody}18.9 & \cellcolor{HGBody}7.8 & \cellcolor{HGBody}2.19 & \cellcolor{HGBody}5.40 & \cellcolor{HGBody}26.0 & \underline{70.5} & 0.0 & 4.04 & 7.99 & 40.1 & 94.2 & 5.2 & 2.06 & 2.64 & 52.9 \\
 & 8B-T & \cellcolor{HGBody}26.0 & \cellcolor{HGBody}17.3 & \cellcolor{HGBody}2.6 & \cellcolor{HGBody}5.47 & \cellcolor{HGBody}10.81 & \cellcolor{HGBody}18.3 & 60.1 & 0.0 & 5.83 & 10.01 & 31.5 & 92.7 & 0.0 & 3.68 & 4.78 & 44.8 \\
\noalign{\color{HGRuleStrong}\hrule height 0.8pt}
\multirow{8}{*}{Qwen3.5} & 0.8B & \cellcolor{HGBody}17.5 & \cellcolor{HGBody}11.7 & \cellcolor{HGBody}9.6 & \cellcolor{HGBody}2.08 & \cellcolor{HGBody}7.05 & \cellcolor{HGBody}25.2 & 29.7 & 22.1 & 1.11 & 3.32 & 35.8 & 76.9 & 76.3 & 0.51 & 0.69 & 64.5 \\
 & 0.8B-T & \cellcolor{HGBody}13.1 & \cellcolor{HGBody}5.4 & \cellcolor{HGBody}1.1 & \cellcolor{HGBody}13.59 & \cellcolor{HGBody}52.68 & \cellcolor{HGBody}11.1 & 15.3 & 7.5 & 6.02 & 28.54 & 19.4 & 68.1 & 48.1 & 2.08 & 6.15 & 50.7 \\
 & 2B & \cellcolor{HGBody}23.7 & \cellcolor{HGBody}16.3 & \cellcolor{HGBody}8.3 & \cellcolor{HGBody}1.42 & \cellcolor{HGBody}1.97 & \cellcolor{HGBody}27.5 & 42.9 & 21.1 & 2.22 & 7.36 & 35.5 & 81.5 & 43.3 & 0.92 & 1.26 & 59.0 \\
 & 2B-T & \cellcolor{HGBody}27.3 & \cellcolor{HGBody}15.8 & \cellcolor{HGBody}0.9 & \cellcolor{HGBody}20.27 & \cellcolor{HGBody}76.05 & \cellcolor{HGBody}13.1 & 37.3 & 0.3 & 15.65 & 48.81 & 19.1 & 80.4 & 0.2 & 6.13 & 16.34 & 39.0 \\
 & 4B & \cellcolor{HGBody}25.5 & \cellcolor{HGBody}18.8 & \cellcolor{HGBody}3.9 & \cellcolor{HGBody}3.29 & \cellcolor{HGBody}4.46 & \cellcolor{HGBody}24.8 & 48.6 & 3.1 & 2.89 & 8.71 & 34.6 & 89.8 & 14.1 & 1.37 & 1.73 & 56.8 \\
 & 4B-T & \cellcolor{HGBody}30.1 & \cellcolor{HGBody}16.4 & \cellcolor{HGBody}0.1 & \cellcolor{HGBody}118.13 & \cellcolor{HGBody}359.86 & \cellcolor{HGBody}7.6 & 39.8 & 0.0 & 89.73 & 119.45 & 14.3 & 86.5 & 0.0 & 37.46 & 87.99 & 31.3 \\
 & 9B & \cellcolor{HGBody}28.6 & \cellcolor{HGBody}20.6 & \cellcolor{HGBody}1.9 & \cellcolor{HGBody}2.95 & \cellcolor{HGBody}5.35 & \cellcolor{HGBody}23.6 & 50.0 & 0.3 & 4.11 & 12.88 & 31.9 & 91.3 & 12.7 & 1.74 & 2.34 & 54.5 \\
 & 9B-T & \cellcolor{HGBody}34.5 & \cellcolor{HGBody}18.9 & \cellcolor{HGBody}0.0 & \cellcolor{HGBody}203.65 & \cellcolor{HGBody}760.58 & \cellcolor{HGBody}7.8 & 47.2 & 0.0 & 120.28 & 407.73 & 17.2 & 89.9 & 0.0 & 54.02 & 247.28 & 32.5 \\
\hline
\multirow{1}{*}{MolmoPoint-GUI} & 8B & \cellcolor{HGBody}22.6 & \cellcolor{HGBody}16.1 & \cellcolor{HGBody}1.8 & \cellcolor{HGBody}8.25 & \cellcolor{HGBody}19.37 & \cellcolor{HGBody}16.3 & 61.1 & 0.0 & 4.19 & 4.87 & 34.4 & 93.9 & 2.8 & 2.99 & 3.96 & 48.4 \\
\hline
\multirow{2}{*}{VISTA} & 4B & \cellcolor{HGBody}31.5 & \cellcolor{HGBody}22.6 & \cellcolor{HGBody}4.1 & \cellcolor{HGBody}2.56 & \cellcolor{HGBody}5.23 & \cellcolor{HGBody}26.5 & 63.2 & 11.2 & 2.58 & 7.06 & 42.7 & 94.7 & 22.9 & 1.22 & 1.62 & 61.3 \\
 & 9B & \cellcolor{HGBody}32.3 & \cellcolor{HGBody}22.4 & \cellcolor{HGBody}2.5 & \cellcolor{HGBody}3.77 & \cellcolor{HGBody}9.14 & \cellcolor{HGBody}23.5 & 67.9 & 0.8 & 3.78 & 10.77 & 40.5 & 96.0 & 14.2 & 1.62 & 2.11 & 57.6 \\
\noalign{\global\arrayrulewidth=0.8pt}\arrayrulecolor{HGRuleStrong}\hline
\end{tabular}
\global\arrayrulewidth=0.4pt\arrayrulecolor{HGRule}
\endgroup
\end{table*}

\begin{table}
\centering
\caption{\textbf{Full results on RTX 3090: OSWorld-G.} Columns follow Table~\ref{tab:d_h200_screenspot}. Bold marks peak model H-GATE and underlining marks peak model Acc.}
\label{tab:d_3090_osworldg}
\begingroup
\footnotesize
\setlength{\tabcolsep}{1.32pt}
\renewcommand{\arraystretch}{1.08}
\arrayrulecolor{HGRule}
\begin{tabular}{c|c!{\color{HGRuleStrong}\vrule width 0.75pt}ccccc}
\noalign{\color{HGRuleStrong}\hrule height 0.8pt}
\multirow{2}{*}{Family} & \multirow{2}{*}{\shortstack{Model\\ /mode}} & \multicolumn{5}{c}{\textbf{OSWorld-G}} \\
\noalign{\global\arrayrulewidth=0.7pt}\arrayrulecolor{HGRuleStrong}\hhline{~~*{5}{-}}\arrayrulecolor{HGRule}\noalign{\global\arrayrulewidth=0.4pt}
 &  & Acc & \shortstack{Acc@\\Human} & Time & P95 & H-GATE \\
\noalign{\color{HGRuleStrong}\hrule height 0.7pt}
\textbf{Human} & \textemdash & 99.9 & \textemdash & 1.11 & \textemdash & 66.6 \\
\hline
\multirow{1}{*}{UI-TARS-1.5} & 7B & 53.9 & 14.5 & 1.46 & 1.71 & 40.4 \\
\hline
\multirow{2}{*}{Jedi} & 3B & 49.5 & 8.9 & 2.69 & 3.45 & 37.3 \\
 & 7B & 40.2 & 5.1 & 2.99 & 7.26 & 30.0 \\
\hline
\multirow{2}{*}{SE-GUI} & 3B & 37.9 & 0.7 & 1.94 & 2.22 & 29.6 \\
 & 7B & 48.0 & 0.0 & 2.17 & 2.43 & 32.9 \\
\hline
\multirow{1}{*}{OpenCUA} & 7B & 55.7 & 18.1 & 1.37 & 1.70 & 42.2 \\
\hline
\multirow{4}{*}{ScaleCUA} & 3B & 51.8 & 25.4 & 1.12 & 1.33 & 42.7 \\
 & 3B-T & 40.1 & 0.0 & 4.36 & 8.07 & 23.6 \\
 & 7B & 56.2 & 19.3 & 1.34 & 1.67 & 42.8 \\
 & 7B-T & 44.1 & 0.0 & 4.13 & 5.66 & 25.5 \\
\hline
\multirow{1}{*}{GroundNext} & 7B & 56.2 & 0.4 & 2.01 & 2.28 & 37.4 \\
\hline
\multirow{4}{*}{Holo2} & 4B & 62.4 & 60.8 & 0.67 & 0.88 & 54.8 \\
 & 4B-T & 62.6 & 59.8 & 0.71 & 0.94 & 54.1 \\
 & 8B & 63.3 & 34.2 & 1.07 & 1.22 & 49.0 \\
 & 8B-T & 61.9 & 0.2 & 3.02 & 4.54 & 35.6 \\
\hline
\multirow{1}{*}{POINTS-GUI} & 8B & 66.0 & 1.4 & 2.55 & 2.34 & 41.6 \\
\hline
\multirow{2}{*}{UI-Venus-1.5} & 2B & 58.3 & 58.3 & 0.49 & 0.58 & \cellcolor{HGBest}\textbf{56.2} \\
 & 8B & \underline{69.3} & 0.5 & 2.28 & 2.63 & 42.1 \\
\hline
\multirow{4}{*}{GUI-Owl-1.5} & 2B & 52.7 & 52.1 & 0.59 & 0.72 & 51.1 \\
 & 4B & 64.5 & 47.0 & 0.93 & 1.11 & 51.5 \\
 & 8B & 67.0 & 10.5 & 1.49 & 1.67 & 46.1 \\
 & 8B-T & 57.3 & 0.0 & 3.21 & 3.82 & 32.7 \\
\noalign{\color{HGRuleStrong}\hrule height 0.8pt}
\multirow{8}{*}{Qwen3.5} & 0.8B & 32.8 & 32.6 & 0.38 & 0.49 & 43.8 \\
 & 0.8B-T & 11.5 & 0.2 & 7.05 & 30.57 & 12.0 \\
 & 2B & 41.0 & 40.2 & 0.66 & 0.87 & 43.2 \\
 & 2B-T & 41.5 & 1.6 & 10.16 & 29.85 & 24.0 \\
 & 4B & 41.1 & 23.0 & 1.09 & 1.34 & 37.4 \\
 & 4B-T & 46.3 & 0.0 & 78.98 & 189.29 & 16.8 \\
 & 9B & 53.9 & 21.6 & 1.19 & 1.41 & 43.0 \\
 & 9B-T & 56.4 & 0.0 & 192.39 & 485.90 & 19.4 \\
\hline
\multirow{1}{*}{MolmoPoint-GUI} & 8B & 63.1 & 0.7 & 2.10 & 2.50 & 40.2 \\
\hline
\multirow{2}{*}{VISTA} & 4B & 59.4 & 50.4 & 0.84 & 0.99 & 50.0 \\
 & 9B & 67.4 & 31.4 & 1.11 & 1.28 & 50.5 \\
\noalign{\global\arrayrulewidth=0.8pt}\arrayrulecolor{HGRuleStrong}\hline
\end{tabular}
\global\arrayrulewidth=0.4pt\arrayrulecolor{HGRule}
\endgroup
\end{table}

\begin{table}
\centering
\caption{\textbf{Full results on RTX 3090: VisualWebBench.} Columns follow Table~\ref{tab:d_h200_screenspot}. Bold marks peak model H-GATE and underlining marks peak model Acc.}
\label{tab:d_3090_visualweb}
\begingroup
\footnotesize
\setlength{\tabcolsep}{1.32pt}
\renewcommand{\arraystretch}{1.08}
\arrayrulecolor{HGRule}
\begin{tabular}{c|c!{\color{HGRuleStrong}\vrule width 0.75pt}ccccc}
\noalign{\color{HGRuleStrong}\hrule height 0.8pt}
\multirow{2}{*}{Family} & \multirow{2}{*}{\shortstack{Model\\ /mode}} & \multicolumn{5}{c}{\textbf{VisualWebBench}} \\
\noalign{\global\arrayrulewidth=0.7pt}\arrayrulecolor{HGRuleStrong}\hhline{~~*{5}{-}}\arrayrulecolor{HGRule}\noalign{\global\arrayrulewidth=0.4pt}
 &  & Acc & \shortstack{Acc@\\Human} & Time & P95 & H-GATE \\
\noalign{\color{HGRuleStrong}\hrule height 0.7pt}
\textbf{Human} & \textemdash & 100.0 & \textemdash & 1.06 & \textemdash & 66.7 \\
\hline
\multirow{1}{*}{UI-TARS-1.5} & 7B & 88.6 & 14.0 & 1.22 & 1.35 & 58.7 \\
\hline
\multirow{2}{*}{Jedi} & 3B & 89.1 & 0.0 & 1.72 & 1.82 & 53.7 \\
 & 7B & 88.6 & 0.0 & 1.82 & 1.91 & 52.6 \\
\hline
\multirow{2}{*}{SE-GUI} & 3B & 86.4 & 0.0 & 1.79 & 2.25 & 52.0 \\
 & 7B & 89.1 & 0.0 & 1.92 & 2.15 & 52.1 \\
\hline
\multirow{1}{*}{OpenCUA} & 7B & 91.5 & 14.9 & 1.22 & 1.42 & 60.2 \\
\hline
\multirow{4}{*}{ScaleCUA} & 3B & 89.0 & 60.9 & 0.96 & 1.05 & 62.6 \\
 & 3B-T & 86.6 & 0.0 & 3.93 & 4.71 & 42.7 \\
 & 7B & 91.1 & 23.4 & 1.14 & 1.31 & 61.1 \\
 & 7B-T & 89.9 & 0.0 & 3.31 & 3.95 & 45.5 \\
\hline
\multirow{1}{*}{GroundNext} & 7B & 90.9 & 0.2 & 1.77 & 1.92 & 54.1 \\
\hline
\multirow{4}{*}{Holo2} & 4B & 91.1 & 91.1 & 0.58 & 0.71 & 71.7 \\
 & 4B-T & 90.7 & 90.7 & 0.60 & 0.71 & 70.8 \\
 & 8B & \underline{92.6} & 81.8 & 0.89 & 0.96 & 65.6 \\
 & 8B-T & 91.3 & 12.8 & 2.34 & 4.85 & 52.9 \\
\hline
\multirow{1}{*}{POINTS-GUI} & 8B & 92.1 & 18.0 & 1.18 & 1.26 & 61.0 \\
\hline
\multirow{2}{*}{UI-Venus-1.5} & 2B & 90.1 & 90.1 & 0.38 & 0.42 & \cellcolor{HGBest}\textbf{76.5} \\
 & 8B & 91.5 & 0.2 & 1.91 & 2.05 & 53.3 \\
\hline
\multirow{4}{*}{GUI-Owl-1.5} & 2B & 87.6 & 87.6 & 0.51 & 0.58 & 71.4 \\
 & 4B & 91.3 & 87.0 & 0.83 & 0.92 & 66.1 \\
 & 8B & 90.7 & 4.5 & 1.32 & 1.39 & 58.5 \\
 & 8B-T & 87.8 & 0.0 & 3.06 & 3.82 & 45.5 \\
\noalign{\color{HGRuleStrong}\hrule height 0.8pt}
\multirow{8}{*}{Qwen3.5} & 0.8B & 76.4 & 76.4 & 0.34 & 0.45 & 70.0 \\
 & 0.8B-T & 72.1 & 69.0 & 0.87 & 1.49 & 60.2 \\
 & 2B & 78.9 & 78.9 & 0.56 & 0.69 & 65.2 \\
 & 2B-T & 68.8 & 4.5 & 5.37 & 15.94 & 39.1 \\
 & 4B & 86.8 & 66.1 & 0.94 & 1.13 & 61.9 \\
 & 4B-T & 76.4 & 0.0 & 42.15 & 126.09 & 28.1 \\
 & 9B & 84.5 & 40.5 & 1.02 & 1.10 & 59.5 \\
 & 9B-T & 83.9 & 0.0 & 21.83 & 55.01 & 33.8 \\
\hline
\multirow{1}{*}{MolmoPoint-GUI} & 8B & 83.3 & 0.0 & 2.03 & 2.20 & 48.7 \\
\hline
\multirow{2}{*}{VISTA} & 4B & 92.2 & 92.2 & 0.75 & 0.83 & 68.2 \\
 & 9B & 92.1 & 67.8 & 0.95 & 1.01 & 64.4 \\
\noalign{\global\arrayrulewidth=0.8pt}\arrayrulecolor{HGRuleStrong}\hline
\end{tabular}
\global\arrayrulewidth=0.4pt\arrayrulecolor{HGRule}
\endgroup
\end{table}

\paragraph{SS-E and ScreenSpot results.}
SS-E efficiency shifts decisively toward compact models on RTX 3090 in \cref{tab:d_3090_screenspot}.
UI-Venus-1.5-8B leads Acc, yet its 6.44-second mean latency compresses Acc@Human to 2.0\%.
GUI-Owl-1.5-2B achieves 13.6\% Acc@Human and leads H-GATE despite lower final accuracy, because it delivers more expert-speed successes and spends far less time per attempt.

\paragraph{OSWorld-G.}
High final accuracy can mask near-zero timely completion on RTX 3090 in \cref{tab:d_3090_osworldg}.
UI-Venus-1.5-8B reaches 69.3\% Acc yet only 0.5\% Acc@Human, whereas the 2B variant solves 58.3\% of tasks within the expert reference.
This steep latency divergence inverts the accuracy ranking under H-GATE even though OSWorld-G imposes no dynamic action window.

\paragraph{VisualWebBench.}
On VisualWebBench, extra accuracy from the largest contender fails to translate into more timely completions in \cref{tab:d_3090_visualweb}.
Holo2-8B attains 92.6\% Acc yet only 81.8\% Acc@Human, whereas UI-Venus-1.5-2B completes 90.1\% of tasks correctly within the expert reference.
The compact model therefore leads both timely completion and H-GATE over the accuracy leader.

\subsection{Ascend 910C}
\begin{table*}
\centering
\caption{\textbf{Full results on Ascend 910C: SS-E and the ScreenSpot benchmarks.} Columns follow Table~\ref{tab:d_h200_screenspot}. Bold marks peak model H-GATE and underlining marks peak model Acc.}
\label{tab:d_910c_screenspot}
\begingroup
\footnotesize
\setlength{\tabcolsep}{1.32pt}
\renewcommand{\arraystretch}{1.08}
\arrayrulecolor{HGRule}
\begin{tabular}{c|c!{\color{HGRuleStrong}\vrule width 0.75pt}cccccc@{\hspace{7pt}}ccccc@{\hspace{5pt}}ccccc}
\noalign{\color{HGRuleStrong}\hrule height 0.8pt}
\multirow{2}{*}{Family} & \multirow{2}{*}{\shortstack{Model\\ /mode}} & \multicolumn{6}{c@{\hspace{7pt}}}{\cellcolor{HGHeader}\textbf{ScreenSpot-Efficiency}} & \multicolumn{5}{c}{\textbf{ScreenSpot-Pro}} & \multicolumn{5}{c}{\textbf{ScreenSpot-v2}} \\
\noalign{\global\arrayrulewidth=0.7pt}\arrayrulecolor{HGRuleStrong}\hhline{~~*{16}{-}}\arrayrulecolor{HGRule}\noalign{\global\arrayrulewidth=0.4pt}
 &  & \cellcolor{HGHeader}Acc-S & \cellcolor{HGHeader}Acc & \cellcolor{HGHeader}\shortstack{Acc@\\Human} & \cellcolor{HGHeader}Time & \cellcolor{HGHeader}P95 & \cellcolor{HGHeader}H-GATE & Acc & \shortstack{Acc@\\Human} & Time & P95 & H-GATE & Acc & \shortstack{Acc@\\Human} & Time & P95 & H-GATE \\
\noalign{\color{HGRuleStrong}\hrule height 0.7pt}
\textbf{Human} & \textemdash & \cellcolor{HGBody}100.0 & \cellcolor{HGBody}99.0 & \cellcolor{HGBody}\textemdash & \cellcolor{HGBody}2.35 & \cellcolor{HGBody}\textemdash & \cellcolor{HGBody}66.2 & 99.2 & \textemdash & 1.46 & \textemdash & 66.2 & 100.0 & \textemdash & 0.97 & \textemdash & 66.7 \\
\hline
\multirow{1}{*}{UI-TARS-1.5} & 7B & \cellcolor{HGBody}24.2 & \cellcolor{HGBody}18.5 & \cellcolor{HGBody}7.8 & \cellcolor{HGBody}2.30 & \cellcolor{HGBody}9.08 & \cellcolor{HGBody}27.0 & 41.7 & 19.3 & 1.93 & 4.58 & 35.4 & 89.3 & 27.0 & 1.07 & 1.32 & 60.2 \\
\hline
\multirow{2}{*}{Jedi} & 3B & \cellcolor{HGBody}21.1 & \cellcolor{HGBody}13.9 & \cellcolor{HGBody}3.4 & \cellcolor{HGBody}3.05 & \cellcolor{HGBody}11.63 & \cellcolor{HGBody}21.7 & 36.2 & 35.9 & 0.89 & 1.10 & 40.3 & 88.9 & 69.6 & 0.87 & 0.92 & 62.8 \\
 & 7B & \cellcolor{HGBody}17.4 & \cellcolor{HGBody}11.4 & \cellcolor{HGBody}2.4 & \cellcolor{HGBody}2.67 & \cellcolor{HGBody}9.68 & \cellcolor{HGBody}21.8 & 34.2 & 33.2 & 0.97 & 1.12 & 38.4 & 91.2 & 67.1 & 0.89 & 0.94 & 63.6 \\
\hline
\multirow{2}{*}{SE-GUI} & 3B & \cellcolor{HGBody}17.5 & \cellcolor{HGBody}12.4 & \cellcolor{HGBody}2.2 & \cellcolor{HGBody}4.45 & \cellcolor{HGBody}14.77 & \cellcolor{HGBody}19.1 & 34.3 & 1.5 & 2.12 & 2.73 & 30.0 & 86.9 & 0.4 & 1.66 & 1.86 & 52.2 \\
 & 7B & \cellcolor{HGBody}20.5 & \cellcolor{HGBody}14.7 & \cellcolor{HGBody}2.5 & \cellcolor{HGBody}3.13 & \cellcolor{HGBody}7.33 & \cellcolor{HGBody}21.2 & 46.5 & 3.0 & 2.13 & 2.85 & 35.7 & 89.9 & 3.5 & 1.55 & 1.78 & 54.6 \\
\hline
\multirow{1}{*}{OpenCUA} & 7B & \cellcolor{HGBody}20.8 & \cellcolor{HGBody}14.0 & \cellcolor{HGBody}2.7 & \cellcolor{HGBody}2.61 & \cellcolor{HGBody}9.32 & \cellcolor{HGBody}22.5 & 46.1 & 10.2 & 3.29 & 11.27 & 35.7 & 93.3 & 15.0 & 1.28 & 1.60 & 59.4 \\
\hline
\multirow{4}{*}{ScaleCUA} & 3B & \cellcolor{HGBody}19.1 & \cellcolor{HGBody}13.2 & \cellcolor{HGBody}7.5 & \cellcolor{HGBody}2.05 & \cellcolor{HGBody}3.91 & \cellcolor{HGBody}23.8 & 39.8 & 14.5 & 1.62 & 2.23 & 35.6 & 87.8 & 2.8 & 1.18 & 1.18 & 57.4 \\
 & 3B-T & \cellcolor{HGBody}22.0 & \cellcolor{HGBody}12.1 & \cellcolor{HGBody}0.4 & \cellcolor{HGBody}10.99 & \cellcolor{HGBody}24.65 & \cellcolor{HGBody}11.6 & 29.7 & 0.0 & 4.99 & 5.54 & 19.7 & 79.7 & 0.0 & 6.50 & 12.27 & 35.4 \\
 & 7B & \cellcolor{HGBody}21.3 & \cellcolor{HGBody}14.6 & \cellcolor{HGBody}9.6 & \cellcolor{HGBody}1.87 & \cellcolor{HGBody}3.61 & \cellcolor{HGBody}25.1 & 47.8 & 18.1 & 1.60 & 2.28 & 39.6 & 92.6 & 24.4 & 1.03 & 1.12 & 62.1 \\
 & 7B-T & \cellcolor{HGBody}24.3 & \cellcolor{HGBody}15.6 & \cellcolor{HGBody}1.2 & \cellcolor{HGBody}8.02 & \cellcolor{HGBody}19.58 & \cellcolor{HGBody}14.9 & 29.2 & 0.0 & 4.25 & 4.94 & 21.2 & 85.5 & 0.0 & 4.03 & 4.57 & 41.4 \\
\hline
\multirow{1}{*}{GroundNext} & 7B & \cellcolor{HGBody}18.3 & \cellcolor{HGBody}12.5 & \cellcolor{HGBody}2.2 & \cellcolor{HGBody}3.08 & \cellcolor{HGBody}3.54 & \cellcolor{HGBody}20.1 & 53.0 & 2.8 & 2.31 & 4.73 & 38.2 & 90.4 & 3.9 & 1.50 & 1.75 & 55.3 \\
\hline
\multirow{4}{*}{Holo2} & 4B & \cellcolor{HGBody}24.3 & \cellcolor{HGBody}17.8 & \cellcolor{HGBody}10.8 & \cellcolor{HGBody}21.12 & \cellcolor{HGBody}3.22 & \cellcolor{HGBody}28.0 & 57.8 & 27.4 & 2.06 & 7.59 & 43.2 & 93.4 & 29.6 & 1.04 & 1.23 & 62.5 \\
 & 4B-T & \cellcolor{HGBody}24.9 & \cellcolor{HGBody}18.7 & \cellcolor{HGBody}11.5 & \cellcolor{HGBody}28.51 & \cellcolor{HGBody}3.69 & \cellcolor{HGBody}28.2 & 58.0 & 27.6 & 2.12 & 7.56 & 43.0 & 93.3 & 30.3 & 1.03 & 1.21 & 62.5 \\
 & 8B & \cellcolor{HGBody}28.2 & \cellcolor{HGBody}20.4 & \cellcolor{HGBody}8.5 & \cellcolor{HGBody}17.28 & \cellcolor{HGBody}2.67 & \cellcolor{HGBody}28.4 & 60.5 & 14.5 & 2.61 & 9.08 & 41.9 & 94.3 & 19.4 & 1.18 & 1.47 & 61.0 \\
 & 8B-T & \cellcolor{HGBody}27.2 & \cellcolor{HGBody}15.4 & \cellcolor{HGBody}1.0 & \cellcolor{HGBody}56.75 & \cellcolor{HGBody}41.62 & \cellcolor{HGBody}13.4 & 57.4 & 0.2 & 5.39 & 8.88 & 31.1 & 94.0 & 5.2 & 2.59 & 4.70 & 51.4 \\
\hline
\multirow{1}{*}{POINTS-GUI} & 8B & \cellcolor{HGBody}31.6 & \cellcolor{HGBody}23.4 & \cellcolor{HGBody}10.5 & \cellcolor{HGBody}1.86 & \cellcolor{HGBody}3.99 & \cellcolor{HGBody}29.2 & 58.1 & 57.3 & 1.00 & 1.21 & \cellcolor{HGBest}\textbf{50.5} & 95.2 & 65.9 & 0.91 & 0.96 & 65.2 \\
\hline
\multirow{2}{*}{UI-Venus-1.5} & 2B & \cellcolor{HGBody}29.1 & \cellcolor{HGBody}19.0 & \cellcolor{HGBody}16.3 & \cellcolor{HGBody}2.41 & \cellcolor{HGBody}4.59 & \cellcolor{HGBody}29.0 & 58.8 & 34.2 & 1.75 & 6.88 & 46.5 & 93.3 & 86.0 & 0.74 & 0.92 & \cellcolor{HGBest}\textbf{67.9} \\
 & 8B & \cellcolor{HGBody}37.6 & \cellcolor{HGBody}\underline{29.0} & \cellcolor{HGBody}17.2 & \cellcolor{HGBody}2.05 & \cellcolor{HGBody}5.28 & \cellcolor{HGBest}\textbf{32.8} & 68.2 & 24.9 & 2.41 & 7.32 & 46.5 & \underline{96.2} & 45.8 & 0.96 & 1.22 & 65.4 \\
\hline
\multirow{4}{*}{GUI-Owl-1.5} & 2B & \cellcolor{HGBody}21.8 & \cellcolor{HGBody}15.6 & \cellcolor{HGBody}8.4 & \cellcolor{HGBody}3.05 & \cellcolor{HGBody}5.15 & \cellcolor{HGBody}24.6 & 58.3 & 9.2 & 2.07 & 3.52 & 41.8 & 90.2 & 1.4 & 1.41 & 1.56 & 56.0 \\
 & 4B & \cellcolor{HGBody}21.8 & \cellcolor{HGBody}15.3 & \cellcolor{HGBody}4.5 & \cellcolor{HGBody}2.50 & \cellcolor{HGBody}5.31 & \cellcolor{HGBody}23.1 & 67.0 & 1.4 & 2.45 & 4.19 & 43.7 & 93.1 & 0.0 & 1.73 & 1.95 & 54.3 \\
 & 8B & \cellcolor{HGBody}26.3 & \cellcolor{HGBody}18.7 & \cellcolor{HGBody}4.3 & \cellcolor{HGBody}2.56 & \cellcolor{HGBody}6.39 & \cellcolor{HGBody}24.6 & \underline{70.5} & 0.4 & 3.02 & 5.31 & 43.1 & 94.2 & 0.0 & 1.83 & 2.04 & 53.9 \\
 & 8B-T & \cellcolor{HGBody}26.0 & \cellcolor{HGBody}15.9 & \cellcolor{HGBody}1.3 & \cellcolor{HGBody}8.05 & \cellcolor{HGBody}16.11 & \cellcolor{HGBody}15.1 & 60.1 & 0.0 & 5.73 & 7.90 & 31.1 & 92.7 & 0.0 & 4.46 & 6.16 & 42.8 \\
\noalign{\color{HGRuleStrong}\hrule height 0.8pt}
\multirow{10}{*}{Qwen3.5} & 0.8B & \cellcolor{HGBody}17.4 & \cellcolor{HGBody}11.7 & \cellcolor{HGBody}2.3 & \cellcolor{HGBody}12.44 & \cellcolor{HGBody}49.69 & \cellcolor{HGBody}16.3 & 29.7 & 0.0 & 3.43 & 7.03 & 23.6 & 77.0 & 0.6 & 2.30 & 4.81 & 45.2 \\
 & 0.8B-T & \cellcolor{HGBody}13.1 & \cellcolor{HGBody}3.7 & \cellcolor{HGBody}0.0 & \cellcolor{HGBody}143.82 & \cellcolor{HGBody}578.27 & \cellcolor{HGBody}3.2 & 14.0 & 0.0 & 157.16 & 157.16 & 5.1 & 65.8 & 0.0 & 32.53 & 102.77 & 26.9 \\
 & 2B & \cellcolor{HGBody}23.7 & \cellcolor{HGBody}16.3 & \cellcolor{HGBody}13.6 & \cellcolor{HGBody}1.10 & \cellcolor{HGBody}1.64 & \cellcolor{HGBody}29.7 & 42.8 & 26.0 & 1.81 & 4.41 & 38.2 & 81.1 & 53.8 & 1.13 & 2.77 & 57.0 \\
 & 2B-T & \cellcolor{HGBody}27.3 & \cellcolor{HGBody}15.5 & \cellcolor{HGBody}0.8 & \cellcolor{HGBody}21.95 & \cellcolor{HGBody}81.87 & \cellcolor{HGBody}12.9 & 37.2 & 0.0 & 22.81 & 22.81 & 15.3 & 80.5 & 0.1 & 3.78 & 7.67 & 41.7 \\
 & 4B & \cellcolor{HGBody}25.6 & \cellcolor{HGBody}18.4 & \cellcolor{HGBody}2.6 & \cellcolor{HGBody}8.17 & \cellcolor{HGBody}11.34 & \cellcolor{HGBody}20.3 & 47.7 & 0.0 & 3.79 & 11.29 & 30.7 & 88.9 & 0.0 & 2.87 & 4.05 & 45.9 \\
 & 4B-T & \cellcolor{HGBody}30.1 & \cellcolor{HGBody}16.0 & \cellcolor{HGBody}0.0 & \cellcolor{HGBody}426.75 & \cellcolor{HGBody}1322.53 & \cellcolor{HGBody}6.1 & 40.4 & 0.0 & 175.71 & 470.77 & 14.8 & 86.8 & 0.0 & 35.66 & 103.71 & 32.0 \\
 & 9B & \cellcolor{HGBody}28.6 & \cellcolor{HGBody}20.9 & \cellcolor{HGBody}2.0 & \cellcolor{HGBody}3.16 & \cellcolor{HGBody}7.23 & \cellcolor{HGBody}24.3 & 50.0 & 0.2 & 3.45 & 9.11 & 33.0 & 91.3 & 0.2 & 2.53 & 2.53 & 48.1 \\
 & 9B-T & \cellcolor{HGBody}34.6 & \cellcolor{HGBody}18.9 & \cellcolor{HGBody}0.0 & \cellcolor{HGBody}431.52 & \cellcolor{HGBody}1659.88 & \cellcolor{HGBody}7.1 & 47.0 & 0.0 & 210.95 & 673.44 & 16.4 & 89.9 & 0.0 & 105.46 & 404.73 & 31.6 \\
 & 27B & \cellcolor{HGBody}38.0 & \cellcolor{HGBody}21.6 & \cellcolor{HGBody}1.8 & \cellcolor{HGBody}8.10 & \cellcolor{HGBody}23.22 & \cellcolor{HGBody}18.3 & 55.0 & 0.0 & 5.49 & 13.16 & 30.3 & 90.2 & 0.0 & 3.84 & 5.51 & 43.1 \\
 & 27B-T & \cellcolor{HGBody}46.3 & \cellcolor{HGBody}28.2 & \cellcolor{HGBody}0.0 & \cellcolor{HGBody}638.02 & \cellcolor{HGBody}2809.69 & \cellcolor{HGBody}10.1 & 64.6 & 0.0 & 307.67 & 901.27 & 22.2 & 95.7 & 0.0 & 36.21 & 74.44 & 34.1 \\
\hline
\multirow{2}{*}{VISTA} & 4B & \cellcolor{HGBody}31.5 & \cellcolor{HGBody}23.4 & \cellcolor{HGBody}13.9 & \cellcolor{HGBody}1.98 & \cellcolor{HGBody}4.12 & \cellcolor{HGBody}29.2 & 62.9 & 21.7 & 2.45 & 6.58 & 43.7 & 94.7 & 0.8 & 1.97 & 1.97 & 53.1 \\
 & 9B & \cellcolor{HGBody}32.3 & \cellcolor{HGBody}24.2 & \cellcolor{HGBody}8.1 & \cellcolor{HGBody}2.49 & \cellcolor{HGBody}5.86 & \cellcolor{HGBody}28.0 & 67.8 & 1.2 & 3.62 & 7.54 & 41.5 & 96.0 & 6.2 & 1.68 & 3.72 & 57.5 \\
\noalign{\global\arrayrulewidth=0.8pt}\arrayrulecolor{HGRuleStrong}\hline
\end{tabular}
\global\arrayrulewidth=0.4pt\arrayrulecolor{HGRule}
\endgroup
\end{table*}

\begin{table}
\centering
\caption{\textbf{Full results on Ascend 910C: OSWorld-G.} Columns follow Table~\ref{tab:d_h200_screenspot}. Bold marks peak model H-GATE and underlining marks peak model Acc.}
\label{tab:d_910c_osworldg}
\begingroup
\footnotesize
\setlength{\tabcolsep}{1.32pt}
\renewcommand{\arraystretch}{1.08}
\arrayrulecolor{HGRule}
\begin{tabular}{c|c!{\color{HGRuleStrong}\vrule width 0.75pt}ccccc}
\noalign{\color{HGRuleStrong}\hrule height 0.8pt}
\multirow{2}{*}{Family} & \multirow{2}{*}{\shortstack{Model\\ /mode}} & \multicolumn{5}{c}{\textbf{OSWorld-G}} \\
\noalign{\global\arrayrulewidth=0.7pt}\arrayrulecolor{HGRuleStrong}\hhline{~~*{5}{-}}\arrayrulecolor{HGRule}\noalign{\global\arrayrulewidth=0.4pt}
 &  & Acc & \shortstack{Acc@\\Human} & Time & P95 & H-GATE \\
\noalign{\color{HGRuleStrong}\hrule height 0.7pt}
\textbf{Human} & \textemdash & 99.9 & \textemdash & 1.11 & \textemdash & 66.6 \\
\hline
\multirow{1}{*}{UI-TARS-1.5} & 7B & 53.4 & 47.9 & 0.85 & 0.95 & 46.7 \\
\hline
\multirow{2}{*}{Jedi} & 3B & 49.8 & 40.1 & 2.89 & 2.88 & 42.5 \\
 & 7B & 40.1 & 32.6 & 1.55 & 3.32 & 36.2 \\
\hline
\multirow{2}{*}{SE-GUI} & 3B & 37.9 & 4.8 & 1.40 & 1.47 & 32.8 \\
 & 7B & 47.9 & 14.0 & 1.27 & 1.34 & 38.9 \\
\hline
\multirow{1}{*}{OpenCUA} & 7B & 56.0 & 33.7 & 1.09 & 1.21 & 44.9 \\
\hline
\multirow{4}{*}{ScaleCUA} & 3B & 51.4 & 30.1 & 1.09 & 1.20 & 42.6 \\
 & 3B-T & 38.5 & 0.0 & 6.43 & 12.26 & 20.2 \\
 & 7B & 56.0 & 40.4 & 0.99 & 1.11 & 46.2 \\
 & 7B-T & 44.7 & 0.0 & 4.56 & 9.53 & 25.0 \\
\hline
\multirow{1}{*}{GroundNext} & 7B & 56.2 & 14.7 & 1.29 & 1.37 & 42.8 \\
\hline
\multirow{4}{*}{Holo2} & 4B & 62.4 & 57.8 & 0.82 & 0.88 & 51.8 \\
 & 4B-T & 62.6 & 57.1 & 0.85 & 0.90 & 51.5 \\
 & 8B & 63.3 & 52.8 & 0.92 & 0.99 & 50.8 \\
 & 8B-T & 61.9 & 0.4 & 3.84 & 3.84 & 32.8 \\
\hline
\multirow{1}{*}{POINTS-GUI} & 8B & 65.8 & 0.0 & 2.28 & 2.01 & 41.7 \\
\hline
\multirow{2}{*}{UI-Venus-1.5} & 2B & 58.3 & 58.3 & 0.56 & 0.62 & 54.3 \\
 & 8B & 69.3 & 68.1 & 0.70 & 0.78 & \cellcolor{HGBest}\textbf{57.7} \\
\hline
\multirow{4}{*}{GUI-Owl-1.5} & 2B & 52.7 & 14.4 & 1.24 & 1.31 & 41.5 \\
 & 4B & 63.8 & 3.5 & 1.48 & 1.55 & 44.5 \\
 & 8B & 66.7 & 1.8 & 1.55 & 1.70 & 45.2 \\
 & 8B-T & 58.0 & 0.0 & 4.29 & 5.45 & 30.4 \\
\noalign{\color{HGRuleStrong}\hrule height 0.8pt}
\multirow{10}{*}{Qwen3.5} & 0.8B & 33.5 & 0.0 & 1.83 & 1.83 & 27.8 \\
 & 0.8B-T & 11.5 & 0.0 & 49.67 & 142.31 & 5.3 \\
 & 2B & 41.3 & 41.3 & 0.56 & 0.63 & 44.9 \\
 & 2B-T & 41.0 & 0.0 & 12.21 & 31.11 & 20.3 \\
 & 4B & 39.4 & 0.0 & 2.39 & 2.96 & 28.0 \\
 & 4B-T & 46.1 & 0.0 & 295.25 & 295.25 & 15.6 \\
 & 9B & 54.4 & 9.9 & 1.43 & 1.72 & 40.7 \\
 & 9B-T & 56.4 & 0.0 & 180.76 & 484.85 & 19.4 \\
 & 27B & 58.7 & 0.0 & 2.90 & 3.14 & 34.3 \\
 & 27B-T & \underline{70.9} & 0.0 & 201.09 & 898.08 & 25.5 \\
\hline
\multirow{2}{*}{VISTA} & 4B & 59.6 & 56.2 & 0.81 & 0.85 & 50.5 \\
 & 9B & 67.4 & 55.7 & 0.95 & 0.95 & 52.4 \\
\noalign{\global\arrayrulewidth=0.8pt}\arrayrulecolor{HGRuleStrong}\hline
\end{tabular}
\global\arrayrulewidth=0.4pt\arrayrulecolor{HGRule}
\endgroup
\end{table}

\begin{table}
\centering
\caption{\textbf{Full results on Ascend 910C: VisualWebBench.} Columns follow Table~\ref{tab:d_h200_screenspot}. Bold marks peak model H-GATE and underlining marks peak model Acc.}
\label{tab:d_910c_visualweb}
\begingroup
\footnotesize
\setlength{\tabcolsep}{1.32pt}
\renewcommand{\arraystretch}{1.08}
\arrayrulecolor{HGRule}
\begin{tabular}{c|c!{\color{HGRuleStrong}\vrule width 0.75pt}ccccc}
\noalign{\color{HGRuleStrong}\hrule height 0.8pt}
\multirow{2}{*}{Family} & \multirow{2}{*}{\shortstack{Model\\ /mode}} & \multicolumn{5}{c}{\textbf{VisualWebBench}} \\
\noalign{\global\arrayrulewidth=0.7pt}\arrayrulecolor{HGRuleStrong}\hhline{~~*{5}{-}}\arrayrulecolor{HGRule}\noalign{\global\arrayrulewidth=0.4pt}
 &  & Acc & \shortstack{Acc@\\Human} & Time & P95 & H-GATE \\
\noalign{\color{HGRuleStrong}\hrule height 0.7pt}
\textbf{Human} & \textemdash & 100.0 & \textemdash & 1.06 & \textemdash & 66.7 \\
\hline
\multirow{1}{*}{UI-TARS-1.5} & 7B & 88.2 & 88.2 & 0.75 & 0.80 & 66.1 \\
\hline
\multirow{2}{*}{Jedi} & 3B & 89.3 & 1.2 & 1.36 & 1.41 & 57.3 \\
 & 7B & 88.6 & 8.1 & 1.22 & 1.26 & 58.6 \\
\hline
\multirow{2}{*}{SE-GUI} & 3B & 86.2 & 2.1 & 1.32 & 1.36 & 56.3 \\
 & 7B & 89.1 & 11.8 & 1.18 & 1.22 & 59.4 \\
\hline
\multirow{1}{*}{OpenCUA} & 7B & 91.3 & 55.2 & 1.02 & 1.11 & 62.9 \\
\hline
\multirow{4}{*}{ScaleCUA} & 3B & 89.3 & 53.7 & 1.02 & 1.07 & 61.9 \\
 & 3B-T & 87.2 & 0.0 & 5.33 & 6.07 & 39.7 \\
 & 7B & 91.1 & 83.7 & 0.89 & 0.95 & 65.0 \\
 & 7B-T & 89.9 & 0.0 & 3.65 & 4.43 & 44.3 \\
\hline
\multirow{1}{*}{GroundNext} & 7B & 90.7 & 10.1 & 1.19 & 1.23 & 60.1 \\
\hline
\multirow{4}{*}{Holo2} & 4B & 91.1 & 90.9 & 0.76 & 0.78 & 67.4 \\
 & 4B-T & 90.7 & 90.5 & 0.76 & 0.78 & 67.1 \\
 & 8B & \underline{92.6} & 90.1 & 0.84 & 0.90 & 66.5 \\
 & 8B-T & 91.3 & 29.3 & 1.40 & 2.26 & 59.1 \\
\hline
\multirow{1}{*}{POINTS-GUI} & 8B & 92.1 & 89.0 & 0.87 & 0.88 & 65.8 \\
\hline
\multirow{2}{*}{UI-Venus-1.5} & 2B & 90.1 & 90.1 & 0.50 & 0.54 & \cellcolor{HGBest}\textbf{73.0} \\
 & 8B & 91.5 & 91.5 & 0.60 & 0.63 & 71.1 \\
\hline
\multirow{4}{*}{GUI-Owl-1.5} & 2B & 87.4 & 14.1 & 1.17 & 1.19 & 58.7 \\
 & 4B & 91.1 & 0.8 & 1.42 & 1.43 & 57.6 \\
 & 8B & 90.7 & 0.2 & 1.51 & 1.56 & 56.4 \\
 & 8B-T & 87.8 & 0.0 & 4.26 & 5.36 & 41.9 \\
\noalign{\color{HGRuleStrong}\hrule height 0.8pt}
\multirow{10}{*}{Qwen3.5} & 0.8B & 76.7 & 1.9 & 1.76 & 2.26 & 48.0 \\
 & 0.8B-T & 69.2 & 0.0 & 14.89 & 40.92 & 29.7 \\
 & 2B & 79.7 & 79.7 & 0.55 & 0.61 & 65.9 \\
 & 2B-T & 69.0 & 1.0 & 5.16 & 13.42 & 36.5 \\
 & 4B & 85.1 & 0.0 & 2.14 & 2.47 & 48.9 \\
 & 4B-T & 80.0 & 0.0 & 86.64 & 86.64 & 27.7 \\
 & 9B & 84.7 & 27.7 & 1.10 & 1.09 & 58.3 \\
 & 9B-T & 83.9 & 0.0 & 69.43 & 189.79 & 29.6 \\
 & 27B & 90.3 & 0.0 & 2.70 & 2.83 & 48.0 \\
 & 27B-T & 92.1 & 0.0 & 46.73 & 141.34 & 32.8 \\
\hline
\multirow{2}{*}{VISTA} & 4B & 91.9 & 90.9 & 0.83 & 0.87 & 66.5 \\
 & 9B & 92.1 & 84.3 & 0.90 & 0.97 & 65.3 \\
\noalign{\global\arrayrulewidth=0.8pt}\arrayrulecolor{HGRuleStrong}\hline
\end{tabular}
\global\arrayrulewidth=0.4pt\arrayrulecolor{HGRule}
\endgroup
\end{table}

\paragraph{SS-E and ScreenSpot results.}
ScreenSpot-Pro rewards low-variance, rapid execution on Ascend 910C in \cref{tab:d_910c_screenspot}.
POINTS-GUI-8B leads H-GATE and completes 57.3\% of tasks within expert references, compared with 0.4\% for accuracy leader GUI-Owl-1.5-8B, while holding P95 to roughly one quarter of its rival's value.
Tighter upper-tail latency thus drives a clear efficiency gain at equal model scale despite lower final accuracy.

\paragraph{OSWorld-G.}
Severe upper-tail latency negates the accuracy edge of thinking configurations in \cref{tab:d_910c_osworldg}.
Qwen3.5-27B-T leads Acc at 70.9\%, yet zero correct responses meet their expert reference and P95 exceeds 898 seconds.
UI-Venus-1.5-8B approaches the same accuracy with P95 under one second, so rapid responses preserve the efficiency value of its localizations.

\paragraph{VisualWebBench.}
Within the UI-Venus family, optimal checkpoints diverge between timely completion counts and H-GATE in \cref{tab:d_910c_visualweb}.
The 8B variant solves 91.5\% of tasks within the expert reference versus 90.1\% for the 2B variant, yet the 2B model leads H-GATE through lower mean latency across all attempts.
The larger model's modest gain in timely successes therefore fails to offset its added inference cost.

\subsection{MetaX C550}
\begin{table*}
\centering
\caption{\textbf{Full results on MetaX C550: SS-E and the ScreenSpot benchmarks.} Columns follow Table~\ref{tab:d_h200_screenspot}. Bold marks peak model H-GATE and underlining marks peak model Acc.}
\label{tab:d_c550_screenspot}
\begingroup
\footnotesize
\setlength{\tabcolsep}{1.0pt}
\renewcommand{\arraystretch}{1.08}
\arrayrulecolor{HGRule}
\begin{tabular}{c|c!{\color{HGRuleStrong}\vrule width 0.75pt}cccccc@{\hspace{7pt}}ccccc@{\hspace{5pt}}ccccc}
\noalign{\color{HGRuleStrong}\hrule height 0.8pt}
\multirow{2}{*}{Family} & \multirow{2}{*}{\shortstack{Model\\ /mode}} & \multicolumn{6}{c@{\hspace{7pt}}}{\cellcolor{HGHeader}\textbf{ScreenSpot-Efficiency}} & \multicolumn{5}{c}{\textbf{ScreenSpot-Pro}} & \multicolumn{5}{c}{\textbf{ScreenSpot-v2}} \\
\noalign{\global\arrayrulewidth=0.7pt}\arrayrulecolor{HGRuleStrong}\hhline{~~*{16}{-}}\arrayrulecolor{HGRule}\noalign{\global\arrayrulewidth=0.4pt}
 &  & \cellcolor{HGHeader}Acc-S & \cellcolor{HGHeader}Acc & \cellcolor{HGHeader}\shortstack{Acc@\\Human} & \cellcolor{HGHeader}Time & \cellcolor{HGHeader}P95 & \cellcolor{HGHeader}H-GATE & Acc & \shortstack{Acc@\\Human} & Time & P95 & H-GATE & Acc & \shortstack{Acc@\\Human} & Time & P95 & H-GATE \\
\noalign{\color{HGRuleStrong}\hrule height 0.7pt}
\textbf{Human} & \textemdash & \cellcolor{HGBody}100.0 & \cellcolor{HGBody}99.0 & \cellcolor{HGBody}\textemdash & \cellcolor{HGBody}2.35 & \cellcolor{HGBody}\textemdash & \cellcolor{HGBody}66.2 & 99.2 & \textemdash & 1.46 & \textemdash & 66.2 & 100.0 & \textemdash & 0.97 & \textemdash & 66.7 \\
\hline
\multirow{1}{*}{UI-TARS-1.5} & 7B & \cellcolor{HGBody}22.9 & \cellcolor{HGBody}8.8 & \cellcolor{HGBody}0.0 & \cellcolor{HGBody}24.29 & \cellcolor{HGBody}115.30 & \cellcolor{HGBody}7.7 & 41.7 & 0.0 & 24.80 & 75.28 & 17.6 & 89.5 & 0.1 & 8.53 & 12.18 & 38.2 \\
\hline
\multirow{2}{*}{Jedi} & 3B & \cellcolor{HGBody}21.1 & \cellcolor{HGBody}14.8 & \cellcolor{HGBody}3.5 & \cellcolor{HGBody}5.32 & \cellcolor{HGBody}20.59 & \cellcolor{HGBody}18.0 & 36.2 & 0.4 & 2.04 & 2.50 & 31.0 & 88.9 & 2.5 & 1.81 & 1.99 & 52.0 \\
 & 7B & \cellcolor{HGBody}17.5 & \cellcolor{HGBody}12.0 & \cellcolor{HGBody}2.8 & \cellcolor{HGBody}4.81 & \cellcolor{HGBody}17.53 & \cellcolor{HGBody}17.6 & 34.1 & 0.8 & 2.00 & 2.53 & 30.3 & 91.2 & 6.3 & 1.73 & 1.88 & 53.7 \\
\hline
\multirow{2}{*}{SE-GUI} & 3B & \cellcolor{HGBody}17.5 & \cellcolor{HGBody}12.7 & \cellcolor{HGBody}1.0 & \cellcolor{HGBody}6.95 & \cellcolor{HGBody}22.70 & \cellcolor{HGBody}15.1 & 34.3 & 0.0 & 4.41 & 5.80 & 22.9 & 86.9 & 0.0 & 3.01 & 3.50 & 44.5 \\
 & 7B & \cellcolor{HGBody}20.5 & \cellcolor{HGBody}13.9 & \cellcolor{HGBody}0.9 & \cellcolor{HGBody}5.84 & \cellcolor{HGBody}12.37 & \cellcolor{HGBody}15.9 & 46.5 & 0.0 & 4.42 & 5.99 & 28.0 & 89.9 & 0.0 & 2.85 & 3.36 & 46.5 \\
\hline
\multirow{1}{*}{OpenCUA} & 7B & \cellcolor{HGBody}20.8 & \cellcolor{HGBody}14.1 & \cellcolor{HGBody}2.7 & \cellcolor{HGBody}4.00 & \cellcolor{HGBody}13.85 & \cellcolor{HGBody}19.0 & 46.1 & 0.0 & 4.98 & 12.33 & 28.6 & 93.3 & 10.2 & 2.17 & 2.82 & 52.2 \\
\hline
\multirow{4}{*}{ScaleCUA} & 3B & \cellcolor{HGBody}19.1 & \cellcolor{HGBody}12.9 & \cellcolor{HGBody}2.5 & \cellcolor{HGBody}3.88 & \cellcolor{HGBody}7.72 & \cellcolor{HGBody}18.8 & 39.7 & 0.0 & 3.24 & 3.82 & 27.7 & 87.8 & 0.6 & 2.12 & 2.39 & 49.2 \\
 & 3B-T & \cellcolor{HGBody}22.0 & \cellcolor{HGBody}10.0 & \cellcolor{HGBody}0.0 & \cellcolor{HGBody}14.78 & \cellcolor{HGBody}31.64 & \cellcolor{HGBody}9.1 & 29.6 & 0.0 & 8.23 & 13.99 & 16.8 & 79.7 & 0.0 & 8.93 & 15.33 & 33.3 \\
 & 7B & \cellcolor{HGBody}21.3 & \cellcolor{HGBody}13.2 & \cellcolor{HGBody}1.7 & \cellcolor{HGBody}3.68 & \cellcolor{HGBody}7.26 & \cellcolor{HGBody}19.2 & 47.8 & 0.0 & 3.33 & 3.82 & 30.9 & 92.6 & 0.5 & 2.14 & 2.37 & 51.2 \\
 & 7B-T & \cellcolor{HGBody}24.2 & \cellcolor{HGBody}13.5 & \cellcolor{HGBody}0.1 & \cellcolor{HGBody}11.92 & \cellcolor{HGBody}29.17 & \cellcolor{HGBody}11.5 & 29.2 & 0.0 & 6.57 & 7.61 & 17.9 & 85.5 & 0.0 & 6.09 & 7.75 & 37.4 \\
\hline
\multirow{1}{*}{GroundNext} & 7B & \cellcolor{HGBody}18.4 & \cellcolor{HGBody}13.6 & \cellcolor{HGBody}0.9 & \cellcolor{HGBody}6.20 & \cellcolor{HGBody}6.42 & \cellcolor{HGBody}15.4 & 53.1 & 0.0 & 4.69 & 9.46 & 30.6 & 90.5 & 0.0 & 2.83 & 3.45 & 46.8 \\
\hline
\multirow{4}{*}{Holo2} & 4B & \cellcolor{HGBody}24.3 & \cellcolor{HGBody}17.5 & \cellcolor{HGBody}6.6 & \cellcolor{HGBody}20.67 & \cellcolor{HGBody}3.46 & \cellcolor{HGBody}26.0 & 57.8 & 21.5 & 2.12 & 5.63 & 42.2 & 93.4 & 23.0 & 1.12 & 1.38 & 61.5 \\
 & 4B-T & \cellcolor{HGBody}24.9 & \cellcolor{HGBody}18.0 & \cellcolor{HGBody}7.0 & \cellcolor{HGBody}24.96 & \cellcolor{HGBody}4.36 & \cellcolor{HGBody}26.4 & 58.0 & 15.5 & 2.24 & 5.60 & 41.4 & 93.3 & 18.6 & 1.21 & 1.49 & 60.2 \\
 & 8B & \cellcolor{HGBody}28.2 & \cellcolor{HGBody}20.0 & \cellcolor{HGBody}5.4 & \cellcolor{HGBody}13.01 & \cellcolor{HGBody}2.86 & \cellcolor{HGBody}26.6 & 60.5 & 8.6 & 2.73 & 8.85 & 41.1 & 94.3 & 17.5 & 1.31 & 1.70 & 59.8 \\
 & 8B-T & \cellcolor{HGBody}27.2 & \cellcolor{HGBody}15.6 & \cellcolor{HGBody}1.1 & \cellcolor{HGBody}55.59 & \cellcolor{HGBody}39.14 & \cellcolor{HGBody}13.7 & 57.4 & 0.0 & 5.44 & 9.22 & 31.4 & 94.0 & 11.2 & 3.13 & 8.30 & 52.4 \\
\hline
\multirow{1}{*}{POINTS-GUI} & 8B & \cellcolor{HGBody}31.4 & \cellcolor{HGBody}23.3 & \cellcolor{HGBody}2.3 & \cellcolor{HGBody}2.64 & \cellcolor{HGBody}6.45 & \cellcolor{HGBody}26.0 & 59.2 & 1.4 & 2.16 & 2.71 & 41.2 & 95.2 & 16.7 & 1.47 & 1.97 & 58.7 \\
\hline
\multirow{2}{*}{UI-Venus-1.5} & 2B & \cellcolor{HGBody}29.2 & \cellcolor{HGBody}19.6 & \cellcolor{HGBody}16.6 & \cellcolor{HGBody}2.43 & \cellcolor{HGBody}5.03 & \cellcolor{HGBody}28.9 & 58.8 & 34.3 & 1.55 & 4.45 & \cellcolor{HGBest}\textbf{46.8} & 93.3 & 77.6 & 0.75 & 0.96 & \cellcolor{HGBest}\textbf{68.0} \\
 & 8B & \cellcolor{HGBody}37.7 & \cellcolor{HGBody}\underline{28.6} & \cellcolor{HGBody}10.8 & \cellcolor{HGBody}2.43 & \cellcolor{HGBody}6.81 & \cellcolor{HGBest}\textbf{31.0} & 68.2 & 14.9 & 2.71 & 8.25 & 45.0 & \underline{96.2} & 22.2 & 1.18 & 1.54 & 62.5 \\
\hline
\multirow{4}{*}{GUI-Owl-1.5} & 2B & \cellcolor{HGBody}21.8 & \cellcolor{HGBody}15.5 & \cellcolor{HGBody}7.5 & \cellcolor{HGBody}2.79 & \cellcolor{HGBody}5.23 & \cellcolor{HGBody}24.4 & 58.2 & 1.6 & 2.28 & 3.48 & 40.3 & 90.2 & 3.9 & 1.47 & 1.68 & 55.5 \\
 & 4B & \cellcolor{HGBody}21.8 & \cellcolor{HGBody}15.5 & \cellcolor{HGBody}5.1 & \cellcolor{HGBody}2.44 & \cellcolor{HGBody}6.06 & \cellcolor{HGBody}23.5 & 67.0 & 0.9 & 2.46 & 3.70 & 43.6 & 93.1 & 0.6 & 1.69 & 1.93 & 54.6 \\
 & 8B & \cellcolor{HGBody}26.2 & \cellcolor{HGBody}18.8 & \cellcolor{HGBody}4.4 & \cellcolor{HGBody}2.52 & \cellcolor{HGBody}6.44 & \cellcolor{HGBody}24.8 & \underline{70.5} & 0.3 & 3.00 & 5.19 & 42.9 & 94.2 & 0.2 & 1.89 & 2.24 & 53.7 \\
 & 8B-T & \cellcolor{HGBody}26.0 & \cellcolor{HGBody}15.8 & \cellcolor{HGBody}1.3 & \cellcolor{HGBody}8.01 & \cellcolor{HGBody}15.94 & \cellcolor{HGBody}15.1 & 60.1 & 0.0 & 5.94 & 8.38 & 30.9 & 92.6 & 0.0 & 4.37 & 5.97 & 42.9 \\
\noalign{\color{HGRuleStrong}\hrule height 0.8pt}
\multirow{10}{*}{Qwen3.5} & 0.8B & \cellcolor{HGBody}17.4 & \cellcolor{HGBody}11.7 & \cellcolor{HGBody}2.4 & \cellcolor{HGBody}11.76 & \cellcolor{HGBody}46.86 & \cellcolor{HGBody}16.5 & 29.8 & 1.6 & 2.13 & 3.67 & 27.8 & 77.0 & 0.6 & 1.59 & 1.89 & 48.1 \\
 & 0.8B-T & \cellcolor{HGBody}13.1 & \cellcolor{HGBody}3.7 & \cellcolor{HGBody}0.0 & \cellcolor{HGBody}136.88 & \cellcolor{HGBody}545.15 & \cellcolor{HGBody}3.3 & 14.2 & 0.0 & 73.67 & 297.75 & 7.4 & 66.2 & 0.0 & 22.51 & 72.52 & 28.9 \\
 & 2B & \cellcolor{HGBody}23.7 & \cellcolor{HGBody}16.7 & \cellcolor{HGBody}2.5 & \cellcolor{HGBody}2.91 & \cellcolor{HGBody}5.26 & \cellcolor{HGBody}22.2 & 42.7 & 0.1 & 2.68 & 4.44 & 31.6 & 81.3 & 0.2 & 1.83 & 2.10 & 48.2 \\
 & 2B-T & \cellcolor{HGBody}27.3 & \cellcolor{HGBody}14.2 & \cellcolor{HGBody}0.0 & \cellcolor{HGBody}106.79 & \cellcolor{HGBody}395.07 & \cellcolor{HGBody}7.1 & 37.0 & 0.0 & 110.12 & 110.12 & 13.0 & 80.7 & 0.0 & 21.73 & 45.37 & 30.3 \\
 & 4B & \cellcolor{HGBody}25.5 & \cellcolor{HGBody}18.4 & \cellcolor{HGBody}2.2 & \cellcolor{HGBody}8.24 & \cellcolor{HGBody}11.72 & \cellcolor{HGBody}19.7 & 47.8 & 0.0 & 3.35 & 6.76 & 31.3 & 89.0 & 0.0 & 2.36 & 2.68 & 48.2 \\
 & 4B-T & \cellcolor{HGBody}30.1 & \cellcolor{HGBody}15.9 & \cellcolor{HGBody}0.0 & \cellcolor{HGBody}420.49 & \cellcolor{HGBody}1316.57 & \cellcolor{HGBody}6.0 & 40.5 & 0.0 & 170.76 & 346.76 & 14.2 & 86.8 & 0.0 & 82.39 & 212.07 & 30.2 \\
 & 9B & \cellcolor{HGBody}28.6 & \cellcolor{HGBody}20.4 & \cellcolor{HGBody}1.5 & \cellcolor{HGBody}3.74 & \cellcolor{HGBody}8.35 & \cellcolor{HGBody}22.5 & 50.0 & 0.1 & 3.33 & 8.00 & 33.1 & 91.2 & 3.6 & 1.77 & 2.51 & 53.4 \\
 & 9B-T & \cellcolor{HGBody}34.5 & \cellcolor{HGBody}18.9 & \cellcolor{HGBody}0.0 & \cellcolor{HGBody}477.07 & \cellcolor{HGBody}1785.15 & \cellcolor{HGBody}7.0 & 47.1 & 0.0 & 270.30 & 553.34 & 16.0 & 89.9 & 0.0 & 105.54 & 238.33 & 30.8 \\
 & 27B & \cellcolor{HGBody}38.0 & \cellcolor{HGBody}21.7 & \cellcolor{HGBody}1.2 & \cellcolor{HGBody}8.60 & \cellcolor{HGBody}22.88 & \cellcolor{HGBody}17.4 & 54.6 & 0.0 & 5.90 & 12.79 & 29.2 & 90.3 & 0.0 & 3.57 & 4.35 & 43.9 \\
 & 27B-T & \cellcolor{HGBody}46.3 & \cellcolor{HGBody}28.3 & \cellcolor{HGBody}0.0 & \cellcolor{HGBody}662.80 & \cellcolor{HGBody}2946.32 & \cellcolor{HGBody}10.1 & 64.6 & 0.0 & 383.60 & 1097.44 & 21.9 & 96.0 & 0.0 & 24.91 & 50.12 & 35.0 \\
\hline
\multirow{1}{*}{MolmoPoint-GUI} & 8B & \cellcolor{HGBody}22.6 & \cellcolor{HGBody}11.2 & \cellcolor{HGBody}0.0 & \cellcolor{HGBody}37.04 & \cellcolor{HGBody}95.95 & \cellcolor{HGBody}7.1 & 61.2 & 0.0 & 23.32 & 28.56 & 23.7 & 93.9 & 0.0 & 14.46 & 21.50 & 36.7 \\
\hline
\multirow{2}{*}{VISTA} & 4B & \cellcolor{HGBody}31.5 & \cellcolor{HGBody}23.8 & \cellcolor{HGBody}8.5 & \cellcolor{HGBody}2.26 & \cellcolor{HGBody}4.49 & \cellcolor{HGBody}28.3 & 63.0 & 26.8 & 1.95 & 4.52 & 45.4 & 94.7 & 25.0 & 1.11 & 1.39 & 62.5 \\
 & 9B & \cellcolor{HGBody}32.3 & \cellcolor{HGBody}22.7 & \cellcolor{HGBody}3.2 & \cellcolor{HGBody}2.95 & \cellcolor{HGBody}7.01 & \cellcolor{HGBody}25.8 & 67.8 & 7.4 & 2.56 & 6.40 & 44.6 & 96.0 & 13.9 & 1.47 & 1.84 & 58.7 \\
\noalign{\global\arrayrulewidth=0.8pt}\arrayrulecolor{HGRuleStrong}\hline
\end{tabular}
\global\arrayrulewidth=0.4pt\arrayrulecolor{HGRule}
\endgroup
\end{table*}

\begin{table}
\centering
\caption{\textbf{Full results on MetaX C550: OSWorld-G.} Columns follow Table~\ref{tab:d_h200_screenspot}. Bold marks peak model H-GATE and underlining marks peak model Acc.}
\label{tab:d_c550_osworldg}
\begingroup
\footnotesize
\setlength{\tabcolsep}{1.32pt}
\renewcommand{\arraystretch}{1.08}
\arrayrulecolor{HGRule}
\begin{tabular}{c|c!{\color{HGRuleStrong}\vrule width 0.75pt}ccccc}
\noalign{\color{HGRuleStrong}\hrule height 0.8pt}
\multirow{2}{*}{Family} & \multirow{2}{*}{\shortstack{Model\\ /mode}} & \multicolumn{5}{c}{\textbf{OSWorld-G}} \\
\noalign{\global\arrayrulewidth=0.7pt}\arrayrulecolor{HGRuleStrong}\hhline{~~*{5}{-}}\arrayrulecolor{HGRule}\noalign{\global\arrayrulewidth=0.4pt}
 &  & Acc & \shortstack{Acc@\\Human} & Time & P95 & H-GATE \\
\noalign{\color{HGRuleStrong}\hrule height 0.7pt}
\textbf{Human} & \textemdash & 99.9 & \textemdash & 1.11 & \textemdash & 66.6 \\
\hline
\multirow{1}{*}{UI-TARS-1.5} & 7B & 53.4 & 0.0 & 4.90 & 6.04 & 28.1 \\
\hline
\multirow{2}{*}{Jedi} & 3B & 49.5 & 0.5 & 4.51 & 5.01 & 33.9 \\
 & 7B & 40.1 & 0.5 & 3.18 & 7.23 & 28.8 \\
\hline
\multirow{2}{*}{SE-GUI} & 3B & 37.6 & 0.0 & 2.46 & 2.69 & 26.8 \\
 & 7B & 47.7 & 0.0 & 2.45 & 2.66 & 31.4 \\
\hline
\multirow{1}{*}{OpenCUA} & 7B & 55.9 & 7.6 & 1.61 & 1.83 & 40.0 \\
\hline
\multirow{4}{*}{ScaleCUA} & 3B & 51.1 & 0.2 & 2.05 & 2.30 & 34.9 \\
 & 3B-T & 39.2 & 0.0 & 8.29 & 17.81 & 19.1 \\
 & 7B & 56.0 & 1.1 & 1.95 & 2.20 & 37.8 \\
 & 7B-T & 44.5 & 0.0 & 6.18 & 8.96 & 22.4 \\
\hline
\multirow{1}{*}{GroundNext} & 7B & 56.4 & 0.0 & 2.30 & 2.50 & 35.8 \\
\hline
\multirow{4}{*}{Holo2} & 4B & 62.4 & 54.3 & 0.90 & 0.97 & 50.6 \\
 & 4B-T & 62.6 & 47.2 & 0.98 & 1.06 & 49.6 \\
 & 8B & 63.3 & 48.8 & 0.94 & 1.07 & 50.6 \\
 & 8B-T & 61.9 & 0.9 & 6.30 & 9.87 & 30.7 \\
\hline
\multirow{1}{*}{POINTS-GUI} & 8B & 66.0 & 4.4 & 2.17 & 1.86 & 43.8 \\
\hline
\multirow{2}{*}{UI-Venus-1.5} & 2B & 58.5 & 58.5 & 0.53 & 0.59 & 55.1 \\
 & 8B & 69.3 & 61.9 & 0.80 & 0.92 & \cellcolor{HGBest}\textbf{55.8} \\
\hline
\multirow{4}{*}{GUI-Owl-1.5} & 2B & 52.7 & 12.9 & 1.29 & 1.36 & 41.1 \\
 & 4B & 63.8 & 5.7 & 1.46 & 1.56 & 44.7 \\
 & 8B & 66.8 & 2.8 & 1.55 & 1.68 & 45.4 \\
 & 8B-T & 57.8 & 0.0 & 4.36 & 5.22 & 30.0 \\
\noalign{\color{HGRuleStrong}\hrule height 0.8pt}
\multirow{10}{*}{Qwen3.5} & 0.8B & 32.8 & 0.9 & 1.62 & 1.67 & 29.2 \\
 & 0.8B-T & 10.8 & 0.0 & 109.10 & 285.48 & 4.2 \\
 & 2B & 42.4 & 0.2 & 1.72 & 1.95 & 32.7 \\
 & 2B-T & 41.1 & 0.0 & 47.13 & 47.13 & 15.1 \\
 & 4B & 39.0 & 0.0 & 2.41 & 2.96 & 27.7 \\
 & 4B-T & 46.5 & 0.0 & 183.36 & 467.12 & 16.2 \\
 & 9B & 54.3 & 9.9 & 1.42 & 1.65 & 40.6 \\
 & 9B-T & 56.6 & 0.0 & 316.85 & 316.85 & 19.1 \\
 & 27B & 59.0 & 0.0 & 3.04 & 3.27 & 33.9 \\
 & 27B-T & \underline{70.7} & 0.0 & 192.83 & 437.97 & 24.2 \\
\hline
\multirow{1}{*}{MolmoPoint-GUI} & 8B & 63.3 & 0.0 & 8.46 & 10.89 & 27.9 \\
\hline
\multirow{2}{*}{VISTA} & 4B & 59.6 & 53.7 & 0.83 & 0.90 & 50.2 \\
 & 9B & 67.4 & 44.7 & 1.01 & 1.12 & 51.7 \\
\noalign{\global\arrayrulewidth=0.8pt}\arrayrulecolor{HGRuleStrong}\hline
\end{tabular}
\global\arrayrulewidth=0.4pt\arrayrulecolor{HGRule}
\endgroup
\end{table}

\begin{table}
\centering
\caption{\textbf{Full results on MetaX C550: VisualWebBench.} Columns follow Table~\ref{tab:d_h200_screenspot}. Bold marks peak model H-GATE and underlining marks peak model Acc.}
\label{tab:d_c550_visualweb}
\begingroup
\footnotesize
\setlength{\tabcolsep}{1.32pt}
\renewcommand{\arraystretch}{1.08}
\arrayrulecolor{HGRule}
\begin{tabular}{c|c!{\color{HGRuleStrong}\vrule width 0.75pt}ccccc}
\noalign{\color{HGRuleStrong}\hrule height 0.8pt}
\multirow{2}{*}{Family} & \multirow{2}{*}{\shortstack{Model\\ /mode}} & \multicolumn{5}{c}{\textbf{VisualWebBench}} \\
\noalign{\global\arrayrulewidth=0.7pt}\arrayrulecolor{HGRuleStrong}\hhline{~~*{5}{-}}\arrayrulecolor{HGRule}\noalign{\global\arrayrulewidth=0.4pt}
 &  & Acc & \shortstack{Acc@\\Human} & Time & P95 & H-GATE \\
\noalign{\color{HGRuleStrong}\hrule height 0.7pt}
\textbf{Human} & \textemdash & 100.0 & \textemdash & 1.06 & \textemdash & 66.7 \\
\hline
\multirow{1}{*}{UI-TARS-1.5} & 7B & 88.0 & 0.0 & 3.68 & 4.00 & 43.5 \\
\hline
\multirow{2}{*}{Jedi} & 3B & 89.3 & 0.0 & 2.42 & 2.52 & 48.9 \\
 & 7B & 89.0 & 0.0 & 2.15 & 2.28 & 50.4 \\
\hline
\multirow{2}{*}{SE-GUI} & 3B & 86.4 & 0.0 & 2.33 & 2.44 & 48.2 \\
 & 7B & 89.1 & 0.0 & 2.14 & 2.31 & 50.6 \\
\hline
\multirow{1}{*}{OpenCUA} & 7B & 91.5 & 4.8 & 1.37 & 1.46 & 58.3 \\
\hline
\multirow{4}{*}{ScaleCUA} & 3B & 89.1 & 0.2 & 1.80 & 1.92 & 53.0 \\
 & 3B-T & 87.2 & 0.0 & 7.13 & 8.50 & 37.4 \\
 & 7B & 91.3 & 0.4 & 1.65 & 1.74 & 55.4 \\
 & 7B-T & 90.1 & 0.4 & 5.37 & 6.39 & 40.6 \\
\hline
\multirow{1}{*}{GroundNext} & 7B & 90.7 & 0.0 & 2.04 & 2.12 & 51.9 \\
\hline
\multirow{4}{*}{Holo2} & 4B & 91.1 & 90.5 & 0.80 & 0.83 & 66.6 \\
 & 4B-T & 90.7 & 88.4 & 0.86 & 0.91 & 65.3 \\
 & 8B & \underline{92.6} & 91.7 & 0.82 & 0.85 & 67.1 \\
 & 8B-T & 91.3 & 2.5 & 3.58 & 7.48 & 47.7 \\
\hline
\multirow{1}{*}{POINTS-GUI} & 8B & 92.1 & 86.6 & 0.86 & 0.91 & 65.9 \\
\hline
\multirow{2}{*}{UI-Venus-1.5} & 2B & 90.1 & 90.1 & 0.46 & 0.49 & \cellcolor{HGBest}\textbf{74.0} \\
 & 8B & 91.5 & 91.5 & 0.69 & 0.72 & 69.0 \\
\hline
\multirow{4}{*}{GUI-Owl-1.5} & 2B & 87.4 & 11.8 & 1.19 & 1.25 & 58.5 \\
 & 4B & 91.1 & 2.1 & 1.34 & 1.37 & 58.5 \\
 & 8B & 90.7 & 0.6 & 1.45 & 1.50 & 57.0 \\
 & 8B-T & 87.8 & 0.0 & 4.44 & 5.15 & 41.7 \\
\noalign{\color{HGRuleStrong}\hrule height 0.8pt}
\multirow{10}{*}{Qwen3.5} & 0.8B & 76.9 & 6.6 & 1.46 & 1.80 & 50.6 \\
 & 0.8B-T & 69.4 & 0.2 & 16.39 & 60.72 & 29.4 \\
 & 2B & 78.5 & 0.0 & 1.69 & 2.01 & 49.2 \\
 & 2B-T & 68.2 & 0.0 & 32.70 & 82.07 & 26.2 \\
 & 4B & 85.3 & 0.0 & 2.18 & 2.63 & 48.7 \\
 & 4B-T & 79.5 & 0.0 & 59.54 & 141.31 & 28.2 \\
 & 9B & 84.5 & 3.7 & 1.28 & 1.31 & 56.0 \\
 & 9B-T & 83.3 & 0.0 & 81.43 & 354.19 & 31.2 \\
 & 27B & 90.5 & 0.0 & 2.89 & 2.98 & 47.2 \\
 & 27B-T & 92.2 & 0.0 & 14.90 & 27.75 & 35.6 \\
\hline
\multirow{1}{*}{MolmoPoint-GUI} & 8B & 83.3 & 0.0 & 7.31 & 8.26 & 35.6 \\
\hline
\multirow{2}{*}{VISTA} & 4B & 91.9 & 91.7 & 0.74 & 0.77 & 68.3 \\
 & 9B & 92.1 & 86.4 & 0.87 & 0.95 & 65.7 \\
\noalign{\global\arrayrulewidth=0.8pt}\arrayrulecolor{HGRuleStrong}\hline
\end{tabular}
\global\arrayrulewidth=0.4pt\arrayrulecolor{HGRule}
\endgroup
\end{table}

\paragraph{SS-E and ScreenSpot results.}
Matching mean latencies can still conceal divergent timely completion profiles on SS-E in \cref{tab:d_c550_screenspot}.
Both the 2B and 8B UI-Venus-1.5 variants average 2.43 seconds, yet the 2B variant leads Acc@Human by 5.8 percentage points while the 8B variant leads Acc and H-GATE.
Task-level pairing of latency and correctness explains why mean latency alone cannot recover either ordering.

\paragraph{OSWorld-G.}
On OSWorld-G, top final accuracy coincides with zero expert-speed completions in \cref{tab:d_c550_osworldg}.
Qwen3.5-27B-T leads Acc at 70.7\%, whereas UI-Venus-1.5-8B solves 61.9\% of tasks within the expert reference at a 0.80-second mean latency.
Prompt delivery of correct localizations secures a decisive efficiency advantage for UI-Venus despite its narrow accuracy deficit.

\paragraph{VisualWebBench.}
Even when most successes arrive within expert time, elevated latency steadily erodes efficiency credit on VisualWebBench in \cref{tab:d_c550_visualweb}.
Holo2-8B tops Acc and slightly surpasses UI-Venus-1.5-2B in Acc@Human, yet its mean latency is 0.82 seconds compared with 0.46 seconds.
UI-Venus therefore leads by 6.9 H-GATE points, as continuous latency scoring separates models that both pass binary timing thresholds on most tasks.

\subsection{Cross-GPU Analysis}
\label{sec:app_cross_gpu}

\paragraph{Configurations shared by all deployments.}
H200 maintains its SS-E efficiency lead when all seven deployments are compared across identical models and inference modes.
Over the 31 shared configurations, H200 outperforms alternative deployments in 181 of 186 paired H-GATE comparisons.
This controlled setup also quantifies single-model gains across consumer GPU generations.
For UI-Venus-1.5-8B, mean latency drops from 6.44 seconds on RTX 3090 to 1.37 seconds on RTX 5090 and lifts H-GATE by 14.8 points, while Acc-S remains within 0.02 percentage points across all three RTX tiers, so hardware acceleration drives the entire efficiency gain at nearly fixed spatial accuracy.

\paragraph{Optima among deployment-specific configurations.}
Allowing each deployment's full tested pool to compete extends the hardware optima in \cref{tab:hardware_choices}.
On SS-E, H-GATE selects GUI-Owl-1.5-2B on RTX 3090, UI-Venus-1.5-8B across five deployments, and Qwen3.5-397B-A17B on H200.
A100 illustrates how evaluation criteria alter model selection, as UI-Venus-1.5-30B-A3B leads accuracy while the 8B variant leads H-GATE through more expert-speed completions.
On H200, expanding candidates beyond the shared set raises the SS-E H-GATE optimum from 38.1 to 43.0.

\paragraph{Sparse capacity and response cost.}
The expanded H200 pool demonstrates how large sparse architectures pair superior grounding accuracy with lower inference latency.
On SS-E, Qwen3.5-397B-A17B surpasses the dense Qwen3.5-27B by 10.1 accuracy points and 8.0 H-GATE points while cutting mean latency by 22.0\%.
Because sparse routing activates 17B parameters per token while storing full weights in high-capacity memory, MoE designs decouple model capacity from per-token compute on memory-rich deployments.

\paragraph{Accuracy and efficiency optima diverge except where accuracy is scarce.}
Across all 35 deployment-benchmark pairs, the accuracy leader diverges from the H-GATE leader in 29 cases.
On the four static benchmarks, both optima coincide in only 1 of 28 cases, and UI-Venus-1.5-2B wins H-GATE in 21 of them, because near-saturated accuracy leaves latency to govern efficiency.
On SS-E, by contrast, the two optima coincide on five of seven deployments, because unsaturated accuracy allows localization gains to outweigh latency overhead.
Established benchmarks therefore reward the fastest adequate checkpoint, whereas SS-E still rewards core grounding capability in the open regime toward human efficiency.

\paragraph{Benchmark-dependent deployment choices.}
H200's dominance on SS-E coexists with highly competitive consumer-GPU efficiency on simpler interfaces.
On VisualWebBench, UI-Venus-1.5-2B on RTX 5090 achieves the highest H-GATE across all deployments at 84.2 compared with 83.9 on H200.
Because compact models already saturate accuracy on such interfaces, a fast consumer deployment suffices to reach the efficiency frontier.

\section{Experimental Setup and Additional Analyses}
\label{app:setup_analysis}

\subsection{Experimental Setup}
\label{sec:appendix_experimental_setup}

This subsection details the hardware deployments, geometric and temporal grounding criteria, model interfaces, and latency measurement protocol underlying \cref{sec:experimental_setup}.

\paragraph{Hardware.}
\Cref{tab:appendix_hardware} summarizes the seven evaluated hardware platforms. On H200, we evaluate all open-weight models across the 14 families. To limit computational cost on the other six GPUs, we evaluate only models that deploy on a single accelerator without quantization. \Cref{tab:appendix_compute} reports the total GPU hours consumed on each platform.

\begin{table*}
    \centering
    \caption{\textbf{Evaluation hardware.} Memory and CPU counts are reported per server or per accelerator as indicated.}
    \label{tab:appendix_hardware}
    \small
    \begin{tabular}{@{}llrrl@{}}
        \toprule
        Accelerator & Configuration per server & GB per accelerator & GB per server & CPUs per server \\
        \midrule
        NVIDIA H200 SXM & $8\times$ H200 SXM & 141 & 1,128 & $2\times$ Intel Xeon Platinum 8570 \\
        NVIDIA RTX 3090 & $8\times$ RTX 3090 & 24 & 192 & $2\times$ Intel Xeon Gold 6338 \\
        NVIDIA RTX 4090 & $8\times$ RTX 4090 & 24 & 192 & $2\times$ Intel Xeon Platinum 8570 \\
        NVIDIA RTX 5090 & $8\times$ RTX 5090 & 32 & 256 & $2\times$ Intel Xeon 6 6760P \\
        NVIDIA A100 PCIe & $8\times$ A100 80 GB PCIe & 80 & 640 & $2\times$ Intel Xeon Platinum 8380 \\
        Huawei Ascend 910C & $8\times$ Ascend 910C & 128 & 1,024 & $4\times$ Huawei Kunpeng 920 \\
        MetaX C550 & $8\times$ C550 & 64 & 512 & $2\times$ Intel Xeon Platinum 8570 \\
        \bottomrule
    \end{tabular}
\end{table*}

\begin{table}
    \centering
    \caption{\textbf{Compute cost.} Total GPU hours consumed on each platform, summed over all evaluation runs and excluding debugging.}
    \label{tab:appendix_compute}
    \small
    \begin{tabular}{@{}lr@{}}
        \toprule
        Accelerator & GPU hours \\
        \midrule
        NVIDIA H200 SXM & 15,301.77 \\
        NVIDIA RTX 3090 & 8,527.84 \\
        NVIDIA RTX 4090 & 6,735.96 \\
        NVIDIA RTX 5090 & 4,562.86 \\
        NVIDIA A100 PCIe & 10,171.55 \\
        Huawei Ascend 910C & 28,185.03 \\
        MetaX C550 & 31,480.11 \\
        \midrule
        Total & 104,965.12 \\
        \bottomrule
    \end{tabular}
\end{table}

\paragraph{Grounding criteria.}
Each model output is parsed in its native format and mapped to the coordinate system of the benchmark annotations, where identical geometric criteria govern every model family.

\textbf{Point in a box.} ScreenSpot-v2 and ScreenSpot-Pro accept any predicted point falling inside or on the boundary of the target bounding box. OSWorld-G applies the same containment test to box annotations, and SS-E enforces it on single-point box targets before checking time.

\textbf{Point in a polygon.} OSWorld-G evaluates polygon targets via a ray-crossing containment test and grades refusals separately on tasks annotated as infeasible.

\textbf{Candidate selection.} VisualWebBench assigns the first predicted point to the smallest enclosing candidate box, and the prediction is accepted when that selected candidate matches the ground-truth answer.

\textbf{Multiple points and ordered actions.} On SS-E, unordered multi-target tasks require a one-to-one match between all predicted points and annotated target boxes, whereas ordered clicks must hit their designated boxes in strict sequence. Dragging requires the start and end coordinates to fall inside the source and destination boxes, respectively, while sliding is accepted when the predicted segment intersects the annotated region at the relevant timestamp.

\textbf{Boxes and contours.} SS-E box predictions must align with the annotated corners within 10 pixels along each axis, and multi-box tasks require every predicted box to match a distinct target. For polygon contours, the predicted and annotated polygons must intersect over an area that covers strictly more than half of each region. This area-overlap criterion is fast to compute, avoids impractical pixel-by-pixel comparison of the polygons, and has little effect on the final experimental conclusions because contour tasks account for 3.93\% of SS-E.

\textbf{Action windows.} SS-E evaluates spatial accuracy independently of response timing. A spatially correct response is accepted on a single-window task when the valid interval covers at least 95\% of the clip or when completion occurs inside that interval. For multi-window tasks, completion must occur within the specific window paired with the spatially matched target, and any boundary shared between windows belongs to the later interval.

\begin{figure*}
  \centering
  \includegraphics[width=\linewidth]{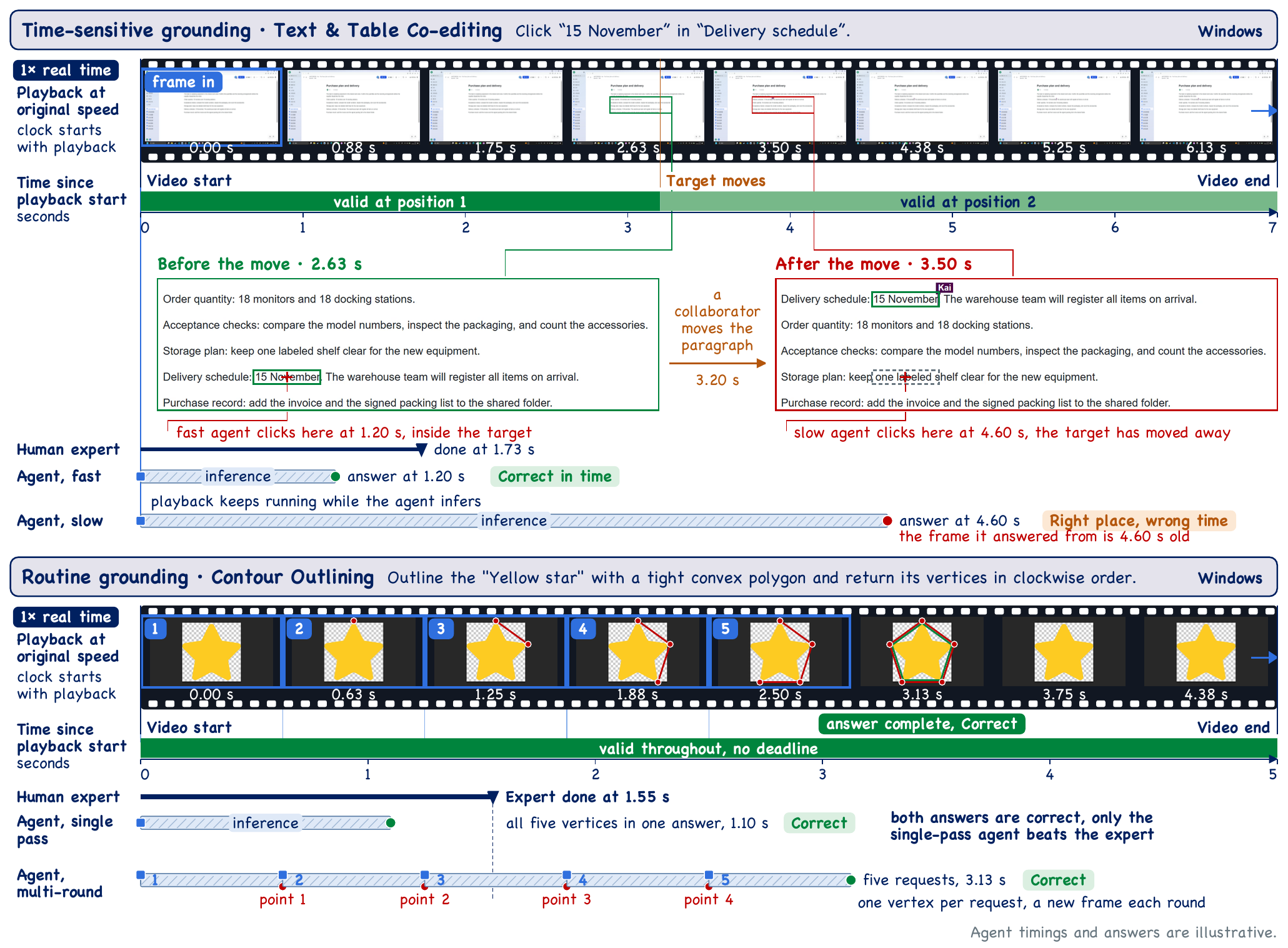}
  \caption{SS-E evaluation protocol under real-time playback. Clips play at original speed as the clock starts with playback, each request passes only the current single frame to the agent while playback advances during inference, and each answer is judged against the interface state at arrival. In the upper panel, a collaborator moves the target paragraph at 3.2 seconds, so a response derived from the first frame is accepted before the move yet lands on the vacated position afterward. In the lower panel, the routine task imposes no deadline, and an agent returning one vertex per request requires five requests and finishes after the expert, whereas a single-pass agent finishes before the expert. Triangles mark the mean expert completion time, agent timings and answers are illustrative, and lower panel frames are cropped around the object.}
  \label{fig:app_protocol}
\end{figure*}

\paragraph{Model settings.}
We preserve each family's released grounding interface and adapt it to the frame available at each SS-E request. Each observation supplies a single video frame, and a model may defer action to observe again as playback advances. \Cref{fig:app_protocol} illustrates this evaluation protocol on a time-sensitive task and a routine task. For tasks requiring a reference image, we place it beside the current frame and map the predicted point back to the original frame. On multi-point tasks, every model decides how many points to output. Models returning multiple points in one reply receive no additional instruction, and models returning one point per round are instructed in every round to append the marker \texttt{<FINAL>} if the current point is their last. The grounding criteria above apply identically to the points returned through both output paths, and this procedure ensures fair evaluation of multi-point tasks across model families.

\textbf{Qwen3.5.} We apply the released chat template across thinking and non-thinking modes~\cite{exp_qwen3_5}. On SS-E, the prompt requests structured points for all targets in the current observation, and returned coordinates are divided by 1,000.

\textbf{UI-TARS-1.5.} We adopt the released direct grounding action format and native image processor~\cite{exp_ui_tars_1_5}. On SS-E, we parse target points or drag endpoints from the action fields and map them from the processed image back to the original frame.

\textbf{UI-Venus-1.5.} We follow the released direct grounding format~\cite{exp_ui_venus_1_5}. On SS-E, the interface accepts a list of points within one reply and divides returned coordinates by 1,000.

\textbf{SE-GUI.} We adopt the released structured tool-call interface for direct grounding~\cite{exp_segui}. On SS-E, multi-location tasks request one tool call per coordinate within a single reply.

\textbf{Jedi.} We evaluate the released 1080p grounding checkpoints through their structured action interface~\cite{osworldg}. On SS-E, we parse individual action coordinates while preserving call order for sequential tasks.

\textbf{GroundNext.} We apply the released direct grounding template and tool-call syntax~\cite{exp_groundnext}. On SS-E, the prompt requests one coordinate-bearing call per required location in a single reply.

\textbf{Holo2.} We preserve the released grounding format and native thinking toggle~\cite{exp_holo2}. On SS-E, we parse either single points or full point lists and divide coordinates by 1,000.

\textbf{OpenCUA.} We adopt the released Python action syntax~\cite{exp_opencua}. On SS-E, we extract frame coordinates from predicted click calls and maintain their order on sequential tasks.

\textbf{GUI-Owl-1.5.} We evaluate both released Instruct and Think checkpoints via their native grounding interface~\cite{exp_gui_owl}. On SS-E, we extract one point per native tool response and issue iterative single-point requests for multi-coordinate tasks.

\textbf{MolmoPoint-GUI.} We adopt the released native pointing mechanism~\cite{exp_molmopoint}. On SS-E, we read coordinates from the service's pointing metadata and gather successive points across rounds for multi-location tasks.

\textbf{POINTS-GUI.} We follow the released direct point prediction format~\cite{exp_points_gui}. On SS-E, each round requests one point and allows a no-action response when task preconditions remain unmet.

\textbf{ScaleCUA.} We apply the released Direct Action and Reasoned Action prompts~\cite{exp_scalecua}. On SS-E, we extract the native point from each reply and collect additional points over successive rounds.

\textbf{VISTA.} We use the released point output format and checkpoint chat template~\cite{exp_vista}. On SS-E, parsed coordinates are divided by 1,000, and multi-location tasks proceed via successive single-point requests.

\textbf{UI-Venus-2.} We adopt the released direct grounding template and coordinate extractor for non-thinking inference~\cite{exp_ui_venus_2}. On SS-E, native infeasibility responses trigger continued observation, and multi-coordinate tasks use the shared point-list parser.

\paragraph{Latency.}
Latency follows the single-request protocol of \cref{sec:experimental_setup}, with one concurrent request per deployment replica. On SS-E, a client-side monotonic clock starts when real-time playback starts after video preloading and stops once the final answer or failed attempt finishes parsing, or when observation ends if playback concludes without an answer. Elapsed latency captures wait intervals, all inference rounds, and same-round retries, while geometric scoring executes outside the timed window. On static benchmarks, timing runs from the start of single-image request processing to the end of answer parsing. Measured model latency is an end-to-end duration, and H-GATE relates it as a whole to the expert reference time in \cref{app:expert_times} without aligning individual stages of the two processes.

For every experiment conducted on each GPU, we performed five runs, computed the performance metrics separately for each run, and averaged the resulting values.
We also computed the variances.
Under stable testing conditions on five GPUs, all variances were below 0.2 because of the large number of test samples.
To present more informative data within the available space, we omit the variances from the figures and tables.
These variances have little effect on the analysis of the experimental results.
Across all experimental tables, every entry requiring statistical computation reports the average of the metric computed separately for each of the five runs.
For analyses that require examining specific output behavior in per-sample logs, we manually analyze the logs from the first run by default.

\subsection{Specialized Post-Training}
\label{sec:e2_posttraining}

This subsection evaluates specialized post-training recipes on H200 to determine which localization gains translate into higher H-GATE. \Cref{tab:e2_posttraining} groups evaluated checkpoints by reported base initialization under default visual budgets and direct-grounding configurations.

\begin{table*}
\centering
\caption{\textbf{H200 grounding results for checkpoints grouped by initialization.} Each benchmark reports accuracy in percent, mean latency in seconds, and H-GATE. The final block averages each metric equally across the five benchmarks.}
\label{tab:e2_posttraining}
\footnotesize
\setlength{\tabcolsep}{1.55pt}
\renewcommand{\arraystretch}{1.08}
\begin{tabular}{@{}l*{6}{rrr}@{}}
\toprule
& \multicolumn{3}{c}{SS-E} & \multicolumn{3}{c}{ScreenSpot-Pro} & \multicolumn{3}{c}{ScreenSpot-v2} & \multicolumn{3}{c}{OSWorld-G} & \multicolumn{3}{c}{VisualWebBench} & \multicolumn{3}{c}{Five-benchmark mean} \\
\cmidrule(lr){2-4}\cmidrule(lr){5-7}\cmidrule(lr){8-10}\cmidrule(lr){11-13}\cmidrule(lr){14-16}\cmidrule(l){17-19}
Checkpoint & Acc. & Lat. & H-GATE & Acc. & Lat. & H-GATE & Acc. & Lat. & H-GATE & Acc. & Lat. & H-GATE & Acc. & Lat. & H-GATE & Acc. & Lat. & H-GATE \\
\midrule
\multicolumn{19}{@{}l}{\textit{Qwen3.5-4B}} \\
Qwen3.5-4B non-thinking & 19.0 & 1.01 & 35.1 & 47.4 & 0.72 & 49.5 & 89.0 & 0.37 & \textbf{75.3} & 37.9 & 0.55 & 48.7 & 84.5 & 0.25 & \textbf{77.7} & 55.6 & 0.58 & 57.3 \\
VISTA-4B & \underline{23.2} & 0.85 & \textbf{35.8} & \underline{63.0} & 0.78 & \textbf{57.0} & \underline{94.7} & 0.58 & 74.8 & \underline{59.6} & 0.40 & \textbf{58.6} & \underline{91.9} & 0.39 & 77.4 & \underline{66.5} & 0.60 & \textbf{60.7} \\
\midrule
\multicolumn{19}{@{}l}{\textit{Qwen3.5-9B}} \\
Qwen3.5-9B non-thinking & 21.7 & 0.74 & 36.4 & 49.8 & 0.79 & 50.1 & 91.2 & 0.37 & \textbf{77.0} & 54.3 & 0.25 & 59.9 & 84.7 & 0.21 & \textbf{79.4} & 60.3 & 0.47 & 60.5 \\
VISTA-9B & 23.4 & 1.04 & 34.9 & 67.9 & 1.01 & 58.1 & 96.0 & 0.55 & 74.6 & 67.2 & 0.42 & 62.7 & \underline{92.1} & 0.40 & 77.2 & 69.3 & 0.68 & 61.5 \\
UI-Venus-2-9B & \underline{33.3} & 1.15 & \textbf{41.4} & \underline{72.4} & 1.15 & \textbf{58.4} & \underline{96.3} & 0.58 & 74.0 & \underline{75.4} & 0.35 & \textbf{69.6} & 90.5 & 0.41 & 76.2 & \underline{73.6} & 0.73 & \textbf{64.0} \\
\midrule
\multicolumn{19}{@{}l}{\textit{Qwen2.5-VL-7B-Instruct}} \\
OpenCUA-7B & \underline{13.7} & 1.44 & \textbf{27.0} & 46.1 & 1.35 & 43.5 & \underline{93.3} & 0.55 & \textbf{72.8} & 55.9 & 0.41 & \textbf{56.6} & \underline{91.5} & 0.35 & \textbf{78.5} & 60.1 & 0.82 & \textbf{55.7} \\
GroundNext-7B & 11.8 & 2.63 & 20.8 & \underline{53.1} & 1.47 & \textbf{43.7} & 90.5 & 0.93 & 62.9 & \underline{56.4} & 0.77 & 49.4 & 90.7 & 0.70 & 68.4 & \underline{60.5} & 1.30 & 49.1 \\
\midrule
\multicolumn{19}{@{}l}{\textit{Qwen2.5-VL-7B family, exact initialization unspecified}} \\
ScaleCUA-7B & \underline{13.9} & 0.89 & \textbf{30.5} & \underline{47.8} & 0.59 & \textbf{51.0} & \underline{92.6} & 0.51 & \textbf{73.8} & \underline{55.5} & 0.44 & \textbf{55.5} & \underline{91.3} & 0.38 & \textbf{77.1} & \underline{60.2} & 0.56 & \textbf{57.6} \\
Jedi-7B & 10.9 & 1.87 & 24.5 & 34.2 & 0.69 & 41.8 & 91.2 & 0.58 & 70.9 & 40.1 & 1.49 & 37.6 & 88.6 & 0.74 & 66.5 & 53.0 & 1.07 & 48.2 \\
SE-GUI-7B & 13.8 & 2.28 & 23.2 & 46.5 & 1.52 & 39.7 & 89.9 & 1.05 & 60.6 & 47.7 & 0.90 & 43.4 & 89.1 & 0.80 & 65.6 & 57.4 & 1.31 & 46.5 \\
\bottomrule
\end{tabular}
\vspace{2pt}

\parbox{\textwidth}{\footnotesize All rows use the default visual budget and a direct grounding configuration. Bold H-GATE and underlined accuracy indicate the highest values within each group. The final group shares a backbone family with indeterminate exact initialization.}
\end{table*}

\paragraph{UI-Venus-2 improves efficiency on demanding benchmarks.}
Initialized from Qwen3.5-9B, UI-Venus-2-9B raises SS-E accuracy by 11.6 percentage points and lifts H-GATE by 5.1 points, alongside H-GATE gains of 8.4 points on ScreenSpot-Pro and 9.8 on OSWorld-G. On ScreenSpot-v2 and VisualWebBench, accuracy rises while H-GATE declines because marginal correctness gains cannot offset added inference latency. Its post-training pipeline integrates programmatically verified targets in cluttered interfaces, task-specific offline reinforcement learning, and action-aware distillation~\cite{exp_ui_venus_2}. On SS-E, this combined supervision translates directly into timely execution and lifts the fraction of tasks completed correctly within expert reference times from 20.1\% to 30.9\%.

\paragraph{Training on multiple views transfers to single-view localization.}
VISTA applies target-preserving crops to diversify supervision per training example while retaining single-point direct prediction at inference~\cite{exp_vista}. Across Qwen3.5-4B and Qwen3.5-9B initializations, this recipe boosts ScreenSpot-Pro H-GATE by 7.5 and 8.1 points, respectively. On SS-E, however, VISTA-9B adds 1.7 percentage points in accuracy yet loses 1.4 points in H-GATE due to higher latency. View-consistent training thus improves efficiency on complex static interfaces, whereas SS-E exposes latency as an independent constraint.

\paragraph{Current recipes optimize single points on static screenshots.}
Specialized recipes that disclose their training objectives supervise a single predicted point, rewarding either box containment as in POINTS-GUI, UI-Venus-1.5, and VISTA, or proximity to the target center as in SE-GUI~\cite{exp_points_gui,exp_ui_venus_1_5,exp_vista,exp_segui}.
None of these specialized recipes reports training on video streams or on responses constrained by valid time intervals.
These two properties mirror the dual bottlenecks identified in \cref{app:category_performance}.
From the shared Qwen3.5-9B base, VISTA-9B uses a point-in-box reward with a single-point interface and drops from 9.7\% to 2.2\% on Contour Outlining and from 12.3\% to 6.1\% on Region Cropping.
UI-Venus-2-9B elevates contour accuracy to 60.8\%, which suggests that structured answers survive specialization only when both recipe and interface retain multi-point outputs.
Meanwhile, at the median configuration, over half of spatially correct time-sensitive responses miss their valid window.
These findings motivate two concrete post-training extensions, namely structured spatial rewards built on corner error, polygon overlap, and one-to-one multi-target matching, alongside streaming supervision that conditions reward on timely arrival so models learn when to act as well as where.

\paragraph{Direct grounding preserves the time available for action.}
From the same Qwen2.5-VL-7B-Instruct base, GroundNext-7B outperforms OpenCUA-7B by 7.0 percentage points on ScreenSpot-Pro accuracy, yet OpenCUA-7B leads five-benchmark mean H-GATE by 6.6 points~\cite{exp_groundnext,exp_opencua}. Static accuracy gains can therefore accompany lower overall efficiency. Output decoding offers a complementary within-checkpoint contrast. On SS-E, switching ScaleCUA-7B from its native grounding configuration to Reasoned Action Mode expands mean output length from 32 to 216 tokens and cuts H-GATE by 7.7 points~\cite{exp_scalecua}. Together, these paired comparisons favor verifiable localization supervision, direct output formatting, and checkpoint selection based on end-to-end latency alongside accuracy.

\subsection{Performance across Task Categories}
\label{app:category_performance}
\Cref{tab:E3_category_L1L2,tab:E3_category_L3} disaggregate H200 SS-E performance across grounding modes, task categories, and individual scenarios for three GUI-specialized models, two of which are post-trained from Qwen3.5-9B, and four Qwen3.5 configurations.
This stratified view disentangles three distinct failure mechanisms conflated by the aggregate accuracy in \cref{tab:hgate_main}, namely response timing, required answer geometry, and preceding visual search.

\begin{table*}
\centering
\caption{\textbf{Scenario modes and categories on SS-E.} Rows report the two modes followed by their seven categories. $N$ counts scored tasks. Acc and Acc-S are percentages, and Time averages latency in seconds across all attempts. Underlining and bold mark the highest Acc and H-GATE within each unadjusted row among displayed configurations. The -T suffix denotes thinking.}
\label{tab:E3_category_L1L2}
\begingroup
\footnotesize
\setlength{\tabcolsep}{0.85pt}
\renewcommand{\arraystretch}{1.08}
\arrayrulecolor{HGRule}
\begin{tabular}{lr|*{6}{rr@{\hspace{4.6pt}}}rr}
\noalign{\color{HGRuleStrong}\hrule height 0.8pt}
\multirow{2}{*}{Category} & \multirow{2}{*}{$N$} & \multicolumn{6}{c}{Qwen3.5-9B base} & \multicolumn{2}{c}{GUI-Owl-1.5} & \multicolumn{6}{c}{Qwen3.5} \\
\cmidrule(lr){3-8}\cmidrule(lr){9-10}\cmidrule(lr){11-16}
 & & \multicolumn{2}{c}{Qwen3.5-9B} & \multicolumn{2}{c}{UI-Venus-2-9B} & \multicolumn{2}{c}{VISTA-9B} & \multicolumn{2}{c}{8B} & \multicolumn{2}{c}{27B} & \multicolumn{2}{c}{397B-A17B} & \multicolumn{2}{c}{397B-A17B-T} \\
\hline
 & & Acc & H-GATE & Acc & H-GATE & Acc & H-GATE & Acc & H-GATE & Acc & H-GATE & Acc & H-GATE & Acc & H-GATE \\
 & & Acc-S & Time & Acc-S & Time & Acc-S & Time & Acc-S & Time & Acc-S & Time & Acc-S & Time & Acc-S & Time \\
\noalign{\color{HGRuleStrong}\hrule height 0.8pt}
\multirow{2}{*}{Routine grounding} & \multirow{2}{*}{5,149} & 26.3 & 36.8 & 36.5 & 42.3 & 30.2 & 35.3 & 20.7 & 34.4 & 32.2 & 32.4 & 46.6 & \textbf{43.4} & \underline{52.9} & 21.4 \\
 & & 26.5 & 0.83 & 37.3 & 0.94 & 30.5 & 1.29 & 20.9 & 0.80 & 32.4 & 1.81 & 47.4 & 1.36 & 53.5 & 75.77 \\
\cline{3-16}
\multirow{2}{*}{Time-sensitive grounding} & \multirow{2}{*}{4,011} & 15.7 & 35.9 & 29.1 & 40.2 & 14.6 & 34.3 & 15.4 & 37.6 & 32.2 & 38.5 & \underline{36.5} & \textbf{42.4} & 7.4 & 10.9 \\
 & & 31.2 & 0.62 & 54.2 & 1.42 & 34.7 & 0.72 & 33.2 & 0.38 & 45.1 & 1.93 & 55.1 & 1.57 & 43.9 & 42.42 \\
\cline{3-16}
\hline
\multirow{2}{*}{\hspace{0.55em}Search \& Selection} & \multirow{2}{*}{1,595} & 52.2 & 51.5 & 67.6 & \textbf{60.6} & 70.1 & 59.3 & 47.8 & 52.7 & 60.8 & 48.8 & 70.4 & 56.8 & \underline{72.7} & 28.6 \\
 & & 52.9 & 0.66 & 70.3 & 0.68 & 71.0 & 0.80 & 48.7 & 0.43 & 61.8 & 1.26 & 73.2 & 0.99 & 74.6 & 33.02 \\
\cline{3-16}
\multirow{2}{*}{\hspace{0.55em}Editing \& Dragging} & \multirow{2}{*}{2,634} & 17.8 & 32.3 & 26.1 & 34.5 & 11.0 & 23.3 & 2.3 & 21.9 & 21.3 & 25.7 & 42.7 & \textbf{40.2} & \underline{49.6} & 17.3 \\
 & & 17.8 & 0.78 & 26.1 & 1.21 & 11.0 & 1.42 & 2.3 & 0.99 & 21.3 & 1.83 & 42.7 & 1.58 & 49.6 & 117.97 \\
\cline{3-16}
\multirow{2}{*}{\hspace{0.55em}Sequential Operations} & \multirow{2}{*}{220} & 0.9 & 23.3 & 5.5 & 29.9 & 3.2 & 21.6 & \underline{47.7} & \textbf{51.1} & 0.5 & 20.2 & 6.4 & 25.9 & 41.8 & 17.6 \\
 & & 0.9 & 1.22 & 5.5 & 0.67 & 3.2 & 2.02 & 47.7 & 1.04 & 0.5 & 2.14 & 6.4 & 1.29 & 41.8 & 103.36 \\
\cline{3-16}
\multirow{2}{*}{\hspace{0.55em}Visual Checks} & \multirow{2}{*}{700} & 7.6 & 24.7 & 14.7 & 33.7 & 20.0 & 30.3 & 19.6 & \textbf{34.8} & 17.6 & 24.0 & 19.7 & 30.4 & \underline{23.7} & 21.8 \\
 & & 7.6 & 1.27 & 14.7 & 0.64 & 20.0 & 1.64 & 19.6 & 0.87 & 17.6 & 2.91 & 19.7 & 1.42 & 23.7 & 5.72 \\
\cline{3-16}
\multirow{2}{*}{\hspace{0.55em}Event-Driven Interaction} & \multirow{2}{*}{2,205} & 4.8 & 29.9 & 23.7 & 36.1 & 5.6 & 30.3 & 5.1 & 32.3 & 23.3 & 34.9 & \underline{27.4} & \textbf{37.6} & 7.8 & 11.5 \\
 & & 21.1 & 0.64 & 51.6 & 1.67 & 27.3 & 0.62 & 25.3 & 0.26 & 40.7 & 1.85 & 48.5 & 1.71 & 38.3 & 50.33 \\
\cline{3-16}
\multirow{2}{*}{\hspace{0.55em}Collaboration \& Competition} & \multirow{2}{*}{1,384} & 27.8 & 43.1 & 35.6 & 45.1 & 30.8 & 42.8 & 28.4 & 45.4 & \underline{50.4} & 47.0 & 50.4 & \textbf{50.0} & 0.6 & 5.2 \\
 & & 42.8 & 0.52 & 58.6 & 1.25 & 50.4 & 0.73 & 46.1 & 0.36 & 60.2 & 2.03 & 66.3 & 1.45 & 50.8 & 40.46 \\
\cline{3-16}
\multirow{2}{*}{\hspace{0.55em}Real-Time Gameplay} & \multirow{2}{*}{422} & 33.2 & 43.3 & 36.3 & \textbf{46.1} & 9.0 & 27.4 & 26.7 & 39.1 & 19.1 & 29.7 & \underline{38.8} & 42.9 & 28.1 & 26.3 \\
 & & 46.8 & 0.93 & 53.2 & 0.59 & 21.5 & 1.29 & 31.9 & 1.12 & 19.1 & 1.97 & 53.2 & 1.17 & 51.1 & 7.54 \\
\noalign{\color{HGRuleStrong}\hrule height 0.8pt}
\end{tabular}
\endgroup
\end{table*}

\begin{table*}
\centering
\caption{\textbf{Performance by scenario type on SS-E.} $N$ counts scored tasks and Acc is a percentage. Underlining and bold mark the highest Acc and H-GATE within each unadjusted row among displayed configurations.}
\label{tab:E3_category_L3}
\begingroup
\footnotesize
\setlength{\tabcolsep}{0.82pt}
\renewcommand{\arraystretch}{1.08}
\arrayrulecolor{HGRule}
\begin{tabular}{lr|*{6}{rr@{\hspace{4.6pt}}}rr}
\noalign{\color{HGRuleStrong}\hrule height 0.8pt}
\multirow{2}{*}{Scenario} & \multirow{2}{*}{$N$} & \multicolumn{6}{c}{Qwen3.5-9B base} & \multicolumn{2}{c}{GUI-Owl-1.5} & \multicolumn{6}{c}{Qwen3.5} \\
\cmidrule(lr){3-8}\cmidrule(lr){9-10}\cmidrule(lr){11-16}
 & & \multicolumn{2}{c}{Qwen3.5-9B} & \multicolumn{2}{c}{UI-Venus-2-9B} & \multicolumn{2}{c}{VISTA-9B} & \multicolumn{2}{c}{8B} & \multicolumn{2}{c}{27B} & \multicolumn{2}{c}{397B-A17B} & \multicolumn{2}{c}{397B-A17B-T} \\
\hline
 & & Acc & H-GATE & Acc & H-GATE & Acc & H-GATE & Acc & H-GATE & Acc & H-GATE & Acc & H-GATE & Acc & H-GATE \\
\noalign{\color{HGRuleStrong}\hrule height 0.8pt}
Routine Control Selection & 1,002 & 41.1 & 47.4 & 56.7 & 56.9 & \underline{64.9} & \textbf{59.1} & 34.2 & 46.6 & 43.5 & 41.3 & 57.1 & 51.7 & \underline{64.9} & 25.6 \\
Dense Content Search & 88 & 54.5 & 42.2 & 79.5 & 56.1 & 77.3 & 52.2 & 85.2 & \textbf{64.8} & 68.2 & 41.3 & \underline{87.5} & 52.3 & 6.8 & 5.6 \\
Content-Based Retrieval & 255 & 93.3 & 76.3 & 94.9 & \textbf{78.5} & 98.4 & 75.7 & 87.8 & 76.8 & 96.5 & 70.3 & 98.8 & 75.4 & \underline{100.0} & 40.1 \\
Reference Image Matching & 250 & 53.6 & 46.1 & 79.6 & 58.7 & 59.6 & 45.7 & 48.4 & 48.3 & 91.2 & 59.2 & 88.8 & \textbf{60.0} & \underline{99.6} & 37.0 \\
\hline
Region Cropping & 1,994 & 12.3 & 27.6 & 11.0 & 22.5 & 6.1 & 19.0 & 0.5 & 20.0 & 7.7 & 17.1 & 29.4 & \textbf{31.6} & \underline{37.6} & 12.5 \\
Contour Outlining & 360 & 9.7 & 31.4 & 60.8 & \textbf{63.3} & 2.2 & 21.6 & 0.0 & 20.2 & 49.7 & 41.8 & 78.1 & 59.2 & \underline{87.5} & 30.8 \\
Cross-App Dragging & 250 & 75.2 & 71.6 & 99.6 & \textbf{89.4} & 63.6 & 59.3 & 20.0 & 39.2 & 91.2 & 71.8 & \underline{100.0} & 83.0 & 96.4 & 38.0 \\
Reordering \& Adjustment & 30 & 0.0 & 28.3 & 0.0 & 24.7 & 0.0 & 28.5 & 0.0 & 30.2 & 0.0 & 24.9 & \underline{20.0} & \textbf{30.3} & 0.0 & 4.6 \\
\hline
Sequential Tab Closing & 120 & 1.7 & 23.2 & 10.0 & 31.9 & 2.5 & 20.8 & \underline{45.8} & \textbf{49.5} & 0.8 & 17.2 & 9.2 & 27.5 & \underline{45.8} & 18.1 \\
Sequential Item Deletion & 100 & 0.0 & 23.4 & 0.0 & 27.6 & 4.0 & 22.6 & \underline{50.0} & \textbf{53.0} & 0.0 & 23.9 & 3.0 & 24.0 & 37.0 & 17.0 \\
\hline
Explicit Error Localization & 200 & 0.0 & 21.9 & \underline{12.0} & \textbf{33.0} & 0.0 & 17.0 & 0.0 & 22.1 & 0.0 & 15.6 & 0.0 & 22.5 & 5.5 & 16.8 \\
Missing Required Fields & 250 & 21.2 & 34.0 & 30.4 & 43.1 & 56.0 & 53.3 & 54.8 & \textbf{57.8} & 48.8 & 42.2 & 55.2 & 51.1 & \underline{57.6} & 39.3 \\
Formatting Inconsistencies & 250 & 0.0 & 17.6 & 1.2 & \textbf{24.9} & 0.0 & 17.9 & 0.0 & 21.8 & 0.4 & 12.6 & 0.0 & 16.0 & \underline{4.4} & 8.4 \\
\hline
Carousel Timing & 1,274 & 4.9 & 28.1 & 29.3 & \textbf{37.3} & 6.3 & 29.3 & 5.7 & 31.7 & 17.9 & 31.0 & \underline{29.5} & 37.2 & 4.7 & 8.2 \\
Send After Upload & 80 & 0.0 & 30.1 & 48.8 & 46.3 & 0.0 & 29.5 & 0.0 & 31.1 & 1.3 & 26.6 & 0.0 & 29.2 & \underline{71.3} & \textbf{48.5} \\
Ad Skip \& Close & 122 & 4.9 & 34.3 & 10.7 & 33.0 & 4.9 & 33.1 & 4.9 & 34.4 & \underline{58.2} & \textbf{51.9} & 36.1 & 42.2 & 23.8 & 25.9 \\
Autoplay Cancellation & 100 & 1.6 & 29.8 & 11.6 & 30.8 & 1.4 & 28.9 & 1.4 & 30.4 & \underline{29.0} & \textbf{37.8} & 23.6 & 35.6 & 7.2 & 13.8 \\
Incoming Requests & 150 & 22.7 & 41.2 & 25.5 & 39.2 & 22.8 & 39.9 & 22.1 & 42.4 & \underline{37.1} & 42.2 & 35.9 & \textbf{42.8} & 0.0 & 8.1 \\
Expiring Undo & 120 & 0.0 & 29.6 & 20.0 & 36.2 & 0.0 & 28.9 & 0.0 & 30.6 & 40.0 & 42.6 & \underline{44.2} & \textbf{46.6} & 0.0 & 9.0 \\
Event-Triggered Controls & 152 & \underline{0.0} & 29.4 & \underline{0.0} & 25.9 & \underline{0.0} & 28.6 & \underline{0.0} & \textbf{30.3} & \underline{0.0} & 26.7 & \underline{0.0} & 25.3 & \underline{0.0} & 9.2 \\
Post-Load Localization & 107 & 1.9 & 30.9 & 21.5 & 36.3 & 0.0 & 30.8 & 0.0 & 31.6 & \underline{75.7} & \textbf{62.6} & 50.5 & 51.2 & 16.8 & 16.1 \\
Chat List Reordering & 100 & 0.0 & 31.7 & 0.0 & \textbf{31.8} & \underline{1.0} & 31.1 & 0.0 & 31.4 & \underline{1.0} & 30.4 & 0.0 & 30.7 & 0.0 & 9.8 \\
\hline
Text \& Table Co-editing & 350 & 35.7 & 47.9 & 32.0 & 45.4 & 41.1 & 49.3 & 39.1 & \textbf{51.7} & 44.0 & 44.7 & \underline{47.1} & 49.0 & 0.6 & 4.1 \\
Slide Co-editing & 200 & 25.0 & 42.2 & 29.5 & 45.2 & 30.5 & 43.7 & 31.0 & \textbf{48.1} & \underline{40.5} & 45.1 & 38.0 & 46.7 & 2.0 & 6.9 \\
Collaborator Cues & 100 & 14.0 & 36.5 & 16.0 & 37.2 & 12.0 & 32.4 & 16.0 & 39.8 & 18.0 & 35.8 & \underline{22.0} & \textbf{40.1} & 0.0 & 3.3 \\
Live Review \& Comments & 150 & 14.7 & 36.4 & \underline{51.3} & \textbf{47.6} & 24.0 & 38.4 & 14.0 & 37.7 & 50.7 & 43.2 & 44.0 & 44.2 & 1.3 & 7.0 \\
Permissions \& Sync & 100 & 10.0 & 35.8 & 19.0 & 36.6 & 9.0 & 34.6 & 7.0 & 35.5 & \underline{61.0} & 53.8 & \underline{61.0} & \textbf{56.9} & 0.0 & 7.3 \\
Live Meeting Management & 210 & 50.5 & 55.7 & 52.8 & 52.4 & 50.5 & 53.8 & 49.5 & 57.9 & \underline{82.4} & 60.7 & 80.0 & \textbf{62.9} & 0.0 & 6.8 \\
Joint Ordering & 49 & 20.4 & 38.1 & 36.7 & 46.1 & 20.4 & 36.5 & 20.4 & 38.8 & \underline{63.3} & \textbf{52.6} & 49.0 & 48.9 & 0.0 & 2.7 \\
Shared Playlist Updates & 50 & 40.0 & 48.9 & 60.0 & 55.5 & 46.0 & 47.8 & 30.0 & 42.5 & \underline{70.0} & 55.1 & 68.0 & \textbf{58.4} & 0.0 & 3.3 \\
Seat Contention & 75 & 2.7 & 26.3 & \underline{6.7} & \textbf{28.3} & 2.7 & 23.5 & 1.3 & 25.5 & 2.7 & 22.9 & \underline{6.7} & 24.8 & 0.0 & 0.9 \\
Shared To-do Updates & 50 & 32.0 & 38.7 & 52.0 & 52.3 & 30.0 & 38.2 & 40.0 & 47.2 & 68.0 & 49.3 & \underline{70.0} & \textbf{55.2} & 0.0 & 3.0 \\
Human-Agent Handoffs & 50 & 20.0 & 36.5 & 40.0 & 43.7 & 16.0 & 34.1 & 0.0 & 29.6 & 66.0 & 53.9 & \underline{82.0} & \textbf{63.9} & 0.0 & 5.7 \\
\hline
Reaction Games & 147 & 44.6 & 52.0 & 44.6 & \textbf{54.7} & 2.2 & 25.4 & 29.8 & 41.9 & 17.0 & 30.1 & \underline{46.8} & 50.4 & 44.8 & 39.5 \\
Complex Game Controls & 175 & 27.2 & 37.9 & 32.5 & \textbf{40.5} & 17.0 & 30.3 & 28.5 & 39.4 & 20.3 & 29.0 & \underline{34.9} & 37.9 & 17.1 & 18.4 \\
Multiplayer Games & 100 & 27.0 & 39.9 & 30.8 & \textbf{43.1} & 5.2 & 25.3 & 19.0 & 34.6 & 20.2 & 30.3 & \underline{34.0} & 40.5 & 22.8 & 20.9 \\
\noalign{\color{HGRuleStrong}\hrule height 0.8pt}
\end{tabular}
\endgroup
\end{table*}

\paragraph{Time-sensitive tasks lose spatially correct answers to timing.}
Across the 53 H200 configurations, time-sensitive tasks underperform routine tasks in full accuracy in 49 cases, with a median decline of 6.90 points.
Spatial accuracy reverses this ordering.
The median gap between Acc-S and Acc expands from 0.43 points on routine tasks to 18.11 points on time-sensitive tasks.
In Event-Driven Interaction, the median configuration localizes 27.3\% of targets spatially yet delivers only 5.9\% within the valid interval.
At the median configuration, more than half of the spatially correct time-sensitive answers therefore fail on timing alone, and in Event-Driven Interaction the share exceeds three quarters, because the response misses the window in which the action remains valid.
\Cref{sec:overall_results} shows that spatially correct responses can arrive before the valid interval opens or after it closes, and both fail the same requirement of acting at the right time.
\Cref{app:failure_mechanisms} quantifies both sides across representative configurations, and the next paragraph and the case studies in \cref{fig:app_timing_cases} illustrate them.

\paragraph{When to act depends on incidental latency.}
Whether inference latency helps or harms temporal correctness depends on whether the valid interval opens late or closes early.
In Send After Upload, the send control becomes actionable only after the upload finishes.
Across the six Qwen3.5 models from 4B to 397B-A17B, enabling thinking raises full accuracy from at most 1.25\% to a range of 42.5\% to 71.3\%, while thinking-mode spatial accuracy holds between 93.8\% and 100\%, because extra reasoning time accidentally outlasts the upload.
In expiring-target scenarios, that same delay proves fatal.
Enabling thinking drops Qwen3.5-397B-A17B from 44.2\% to 0.0\% on Expiring Undo and from 80\% to 0\% on Live Meeting Management, as its mean latency climbs from about 2\,s to over 16\,s.
Enabling thinking similarly reduces its accuracy from 35.9\% to 0.0\% on Incoming Requests and from 23.6\% to 7.2\% on Autoplay Cancellation.
Temporal validity thus fluctuates with incidental execution duration, which suggests that current models do not condition response timing on live interface state.
Explicit action-timing control that waits for preconditions and executes before deadlines therefore offers a direct path to time-sensitive accuracy beyond scaling parameters or reasoning length.

\paragraph{Answer geometry is the main bottleneck of routine grounding.}
Routine accuracy spans more than an order of magnitude across answer geometries.
Content-Based Retrieval requires a single click on a described item and attains a median accuracy of 87.1\%.
Region Cropping demands two precise corners and drops to a 0.7\% median across 1,994 tasks, while Formatting Inconsistencies requires bounding boxes around every violation and scores zero in 40 of 53 configurations.
Contour Outlining, which requires three to ten vertices, separates models along both capacity and output interface.
Within Qwen3.5, contour accuracy rises from 9.7\% at 9B to 49.7\% at 27B and 78.1\% at 397B-A17B, and thinking pushes the largest model to 87.5\% against a 3.1\% configuration median.
Specialization shifts this capability in opposite directions from the same Qwen3.5-9B base, as UI-Venus-2-9B lifts contour accuracy to 60.8\% whereas VISTA-9B drops to 2.2\%.
The five families restricted to one point per request, namely GUI-Owl-1.5, MolmoPoint-GUI, POINTS-GUI, ScaleCUA, and VISTA, outline at most 8 of 360 contours, or 2.2\%, in each of their 16 configurations.
Those same interfaces nonetheless handle ordered clicks, with GUI-Owl-1.5-8B closing 45.8\% of tab sequences and completing 50.0\% of ordered deletions, because every step reduces to a single click.
Contour answers therefore require emitting the full geometry in one response, which suggests that grounding recipes and interfaces should preserve multi-point outputs rather than collapsing every task into a single point.

\paragraph{Collaboration and contention remain unsolved.}
Multi-participant scenarios where another user alters the target expose an independent failure mode.
Seat Contention, where another user can claim a seat before the model responds, remains at or below 12.0\% across every configuration, while Chat List Reordering exceeds 12\% solely for ScaleCUA-32B in reasoned-action mode.
Solving these tasks requires grounding the target that remains valid after external state changes, a dynamic capability that static-frame reasoning and spatial precision alone do not supply.

\begin{table}
\centering
\caption{\textbf{Scenario-balanced SS-E summaries on H200.} Pooled repeats the overall scores, Balanced averages the 36 scenarios with equal weight, and Reduced pools the 5,892 tasks outside Region Cropping and Carousel Timing. All values derive from the scenario-level entries of \cref{tab:E3_category_L3}. Acc is a percentage, and the -T suffix denotes thinking.}
\label{tab:E3_balanced}
\footnotesize
\setlength{\tabcolsep}{2.6pt}
\begin{tabular}{@{}lrrrrrr@{}}
\toprule
 & \multicolumn{2}{c}{Pooled} & \multicolumn{2}{c}{Balanced} & \multicolumn{2}{c}{Reduced} \\
\cmidrule(lr){2-3}\cmidrule(lr){4-5}\cmidrule(l){6-7}
Configuration & Acc & H-GATE & Acc & H-GATE & Acc & H-GATE \\
\midrule
Qwen3.5-9B & 21.7 & 36.4 & 21.2 & 37.5 & 28.5 & 41.2 \\
UI-Venus-2-9B & 33.3 & 41.4 & 33.5 & 43.2 & 41.7 & 48.6 \\
VISTA-9B & 23.4 & 34.9 & 21.6 & 35.7 & 32.9 & 41.5 \\
GUI-Owl-1.5-8B & 18.4 & 35.8 & 21.8 & 39.3 & 27.2 & 42.1 \\
Qwen3.5-27B & 32.2 & 35.1 & 38.4 & 39.9 & 43.6 & 42.1 \\
Qwen3.5-397B-A17B & 42.2 & 43.0 & 42.5 & 44.6 & 49.2 & 48.1 \\
Qwen3.5-397B-A17B-T & 33.0 & 16.8 & 23.8 & 15.9 & 37.5 & 20.1 \\
\bottomrule
\end{tabular}
\end{table}

\paragraph{Scenario-balanced summaries.}
Pooled SS-E scores weight each scenario by its task count, and Region Cropping and Carousel Timing together hold 3,268 of the 9,160 tasks, or 35.7\%.
\Cref{tab:E3_balanced} therefore recomputes Acc and H-GATE for the seven displayed configurations with equal weight per scenario and with these two scenarios removed.
Equal scenario weights shift H-GATE by $-0.9$ to $+4.8$ points and leave Qwen3.5-397B-A17B and UI-Venus-2-9B as the two highest-ranked configurations.
Removing the two scenarios raises H-GATE for every configuration by 3.3 to 7.2 points, which confirms that these scenarios lower pooled scores.
UI-Venus-2-9B then leads H-GATE at 48.6 against 48.1 for Qwen3.5-397B-A17B, while the latter keeps the highest Acc at 49.2\%.
The thinking configuration ranks last on H-GATE under all three weightings.
The highest H-GATE on the remaining 5,892 tasks stays 18.1 points below the expert-speed reference of 66.67, so the gap to expert efficiency on SS-E does not originate from these two scenarios alone.

\paragraph{Leading configurations remain stable, and the preferred choice depends on application conditions.}
Together, the failure-cost sensitivity analysis in \cref{app:hgate_sensitivity} and the scenario weightings in \cref{tab:E3_balanced} show that the leading configurations remain stable across the tested settings and that the preferred choice depends on task composition and failure cost.
On SS-E with H200, a failure cost of $\lambda=1$ preserves the three leading configurations and yields a ranking with Spearman $\rho=0.958$ relative to the original H-GATE ordering.
At $\lambda=2$, the top two remain unchanged and the correlation falls to 0.896, as the order among some of the remaining configurations shifts with the failure cost.
Among the seven configurations in \cref{tab:E3_balanced}, Qwen3.5-397B-A17B and UI-Venus-2-9B form the top two under task pooling, equal scenario weights, and removal of the two largest scenarios.
After those scenarios are removed, UI-Venus-2-9B moves ahead at 48.6 against 48.1 H-GATE.
The overall leaderboard therefore summarizes performance under the evaluated task distribution, and model selection for a specific application should draw on scenario-level results and the corresponding failure cost.

\subsection{Performance across UI Languages}
\label{app:language_performance}
\begin{table*}
\centering
\caption{\textbf{Performance by UI language on SS-E.} Adjusted differences weight shared scenario and platform cells with at least five tasks per group by their smaller count. $N$ counts scored tasks. Acc and Acc-S are percentages, and Time averages latency in seconds across all attempts. For adjusted rows, $N$ is the sum of cell weights and accuracy differences are in percentage points. Underlining and bold mark the highest Acc and H-GATE within each unadjusted row among displayed configurations.}
\label{tab:E4_language}
\begingroup
\footnotesize
\setlength{\tabcolsep}{1.10pt}
\renewcommand{\arraystretch}{1.08}
\arrayrulecolor{HGRule}
\begin{tabular}{lr|*{6}{rr@{\hspace{4.6pt}}}rr}
\noalign{\color{HGRuleStrong}\hrule height 0.8pt}
\multirow{2}{*}{UI language} & \multirow{2}{*}{$N$} & \multicolumn{4}{c}{Holo2} & \multicolumn{4}{c}{UI-Venus-1.5} & \multicolumn{6}{c}{Qwen3.5} \\
\cmidrule(lr){3-6}\cmidrule(lr){7-10}\cmidrule(lr){11-16}
 & & \multicolumn{2}{c}{8B} & \multicolumn{2}{c}{30B-A3B} & \multicolumn{2}{c}{8B} & \multicolumn{2}{c}{30B-A3B} & \multicolumn{2}{c}{27B} & \multicolumn{2}{c}{397B-A17B} & \multicolumn{2}{c}{397B-A17B-T} \\
\hline
 & & Acc & H-GATE & Acc & H-GATE & Acc & H-GATE & Acc & H-GATE & Acc & H-GATE & Acc & H-GATE & Acc & H-GATE \\
 & & Acc-S & Time & Acc-S & Time & Acc-S & Time & Acc-S & Time & Acc-S & Time & Acc-S & Time & Acc-S & Time \\
\noalign{\color{HGRuleStrong}\hrule height 0.8pt}
\multirow{2}{*}{German} & \multirow{2}{*}{1,039} & 19.2 & 34.8 & 20.3 & 31.1 & 25.1 & 38.2 & 25.5 & 34.4 & 33.3 & 35.9 & \underline{39.2} & \textbf{41.7} & 21.6 & 13.1 \\
 & & 26.9 & 0.90 & 27.9 & 2.45 & 36.7 & 0.76 & 36.2 & 1.55 & 38.9 & 1.97 & 46.9 & 1.38 & 38.3 & 52.24 \\
\cline{3-16}
\multirow{2}{*}{English} & \multirow{2}{*}{2,062} & 19.7 & 36.1 & 20.9 & 32.8 & 26.3 & 39.5 & 27.2 & 36.1 & 40.9 & 40.1 & \underline{42.2} & \textbf{43.7} & 22.4 & 15.3 \\
 & & 33.7 & 2.28 & 35.6 & 1.49 & 46.3 & 0.92 & 47.7 & 1.64 & 50.8 & 2.34 & 58.6 & 1.70 & 53.7 & 44.12 \\
\cline{3-16}
\multirow{2}{*}{French} & \multirow{2}{*}{867} & 21.0 & 35.3 & 21.6 & 31.0 & 29.9 & 39.9 & 29.0 & 35.3 & 36.7 & 36.7 & \underline{44.9} & \textbf{44.1} & 27.3 & 15.5 \\
 & & 27.5 & 1.38 & 28.7 & 3.69 & 40.0 & 0.92 & 37.7 & 1.78 & 42.1 & 2.10 & 52.9 & 1.45 & 44.1 & 53.48 \\
\cline{3-16}
\multirow{2}{*}{Japanese} & \multirow{2}{*}{1,099} & 17.2 & 34.0 & 19.0 & 30.8 & 22.8 & 37.4 & 23.1 & 33.5 & 27.8 & 34.0 & \underline{34.1} & \textbf{39.4} & 23.2 & 14.8 \\
 & & 27.8 & 2.06 & 27.8 & 1.51 & 34.6 & 0.74 & 34.4 & 1.54 & 36.3 & 1.73 & 45.9 & 1.37 & 39.9 & 58.51 \\
\cline{3-16}
\multirow{2}{*}{Chinese} & \multirow{2}{*}{4,093} & 22.8 & 36.5 & 26.7 & 33.5 & 26.8 & 37.2 & 33.4 & 37.2 & 27.8 & 32.4 & 44.6 & \textbf{43.8} & \underline{45.0} & 19.2 \\
 & & 26.7 & 0.85 & 30.4 & 6.14 & 33.8 & 1.03 & 39.8 & 1.94 & 31.0 & 1.58 & 48.8 & 1.37 & 53.6 & 74.36 \\
\noalign{\color{HGRuleStrong}\hrule height 0.8pt}
\textbf{Adjusted differences} & & & & & & & & & & & & & & & \\
\multirow{2}{*}{English vs. Chinese} & \multirow{2}{*}{1,073} & +0.7 & +0.1 & +1.4 & +0.5 & +0.2 & +0.6 & -1.3 & -0.8 & +2.8 & +1.0 & +0.4 & -0.3 & -7.4 & -3.3 \\
 & & +2.2 & -0.17 & +4.1 & +0.08 & +2.0 & -0.12 & +0.5 & -0.03 & +1.8 & +0.17 & +0.6 & +0.18 & -4.2 & +1.87 \\
\cline{3-16}
\multirow{2}{*}{French vs. Japanese} & \multirow{2}{*}{863} & +0.8 & +0.1 & -0.8 & -0.8 & +2.9 & +0.6 & +1.5 & +0.4 & +4.0 & +1.1 & +6.4 & +2.7 & -0.5 & -0.7 \\
 & & -1.4 & -1.06 & -1.1 & +2.04 & +3.0 & +0.15 & +0.5 & +0.07 & +2.8 & +0.21 & +5.4 & +0.08 & +1.5 & -1.12 \\
\noalign{\color{HGRuleStrong}\hrule height 0.8pt}
\end{tabular}
\endgroup
\end{table*}

\Cref{tab:E4_language} disaggregates performance by UI language both before and after matching task composition.
Raw cross-lingual gaps stem primarily from unequal scenario mixes across languages.
Region Cropping represents 31.6\% of Chinese tasks yet only 8.5\% of English tasks, so the median English minus Chinese accuracy gap flips from $-1.19$ points in raw scores to $+0.60$ points once scenarios and platforms are matched.
Residual disparities remain model specific.
After matching, Qwen3.5-397B-A17B maintains a 6.39-point advantage on French over Japanese, whereas enabling thinking narrows yet preserves a 7.43-point deficit on English relative to Chinese.
For most models, task composition dominates UI language, which indicates that cross-lingual GUI evaluations must control for scenario mix before attributing performance gaps to language.

\subsection{Performance across Platforms}
\label{app:platform_performance}
\begin{table*}
\centering
\caption{\textbf{Performance by platform on SS-E.} Adjusted differences weight shared scenario and language cells with at least five tasks per group by their smaller count. $N$ counts scored tasks. Acc and Acc-S are percentages, and Time averages latency in seconds across all attempts. For adjusted rows, $N$ is the sum of cell weights and accuracy differences are in percentage points. Underlining and bold mark the highest Acc and H-GATE within each unadjusted row among displayed configurations.}
\label{tab:E5_platform}
\begingroup
\footnotesize
\setlength{\tabcolsep}{1.10pt}
\renewcommand{\arraystretch}{1.08}
\arrayrulecolor{HGRule}
\begin{tabular}{lr|*{6}{rr@{\hspace{4.6pt}}}rr}
\noalign{\color{HGRuleStrong}\hrule height 0.8pt}
\multirow{2}{*}{Platform} & \multirow{2}{*}{$N$} & \multicolumn{4}{c}{Holo2} & \multicolumn{4}{c}{UI-Venus-1.5} & \multicolumn{6}{c}{Qwen3.5} \\
\cmidrule(lr){3-6}\cmidrule(lr){7-10}\cmidrule(lr){11-16}
 & & \multicolumn{2}{c}{8B} & \multicolumn{2}{c}{30B-A3B} & \multicolumn{2}{c}{8B} & \multicolumn{2}{c}{30B-A3B} & \multicolumn{2}{c}{27B} & \multicolumn{2}{c}{397B-A17B} & \multicolumn{2}{c}{397B-A17B-T} \\
\hline
 & & Acc & H-GATE & Acc & H-GATE & Acc & H-GATE & Acc & H-GATE & Acc & H-GATE & Acc & H-GATE & Acc & H-GATE \\
 & & Acc-S & Time & Acc-S & Time & Acc-S & Time & Acc-S & Time & Acc-S & Time & Acc-S & Time & Acc-S & Time \\
\noalign{\color{HGRuleStrong}\hrule height 0.8pt}
\multirow{2}{*}{Android} & \multirow{2}{*}{163} & 4.5 & 33.3 & 4.5 & 31.3 & 4.5 & \textbf{33.8} & 4.5 & 31.7 & 6.8 & 32.2 & 5.3 & 32.3 & \underline{18.8} & 23.1 \\
 & & 45.1 & 0.34 & 42.9 & 0.65 & 43.6 & 0.27 & 42.9 & 0.59 & 43.6 & 0.86 & 45.1 & 0.64 & 48.9 & 13.98 \\
\cline{3-16}
\multirow{2}{*}{iOS} & \multirow{2}{*}{151} & 7.3 & 33.1 & 7.3 & 30.5 & 9.9 & 32.4 & 25.2 & 37.2 & 19.2 & 33.9 & \underline{31.1} & \textbf{39.2} & 11.9 & 14.7 \\
 & & 19.2 & 0.55 & 17.2 & 0.91 & 51.0 & 0.99 & 68.9 & 1.63 & 31.8 & 1.76 & 62.3 & 1.92 & 51.0 & 37.48 \\
\cline{3-16}
\multirow{2}{*}{iPadOS} & \multirow{2}{*}{274} & 20.4 & 37.1 & 20.1 & 32.4 & 22.3 & 38.1 & 42.0 & 42.2 & 49.3 & 44.6 & \underline{50.7} & \textbf{47.9} & 0.0 & 2.9 \\
 & & 40.5 & 0.37 & 41.6 & 0.65 & 60.6 & 0.45 & 66.4 & 0.97 & 62.8 & 1.29 & 65.3 & 1.06 & 43.4 & 57.41 \\
\cline{3-16}
\multirow{2}{*}{macOS} & \multirow{2}{*}{1,010} & 10.0 & 30.8 & 11.1 & 29.3 & 13.1 & 32.8 & 15.5 & 31.7 & 23.4 & 33.3 & \underline{35.6} & \textbf{39.9} & 10.3 & 10.2 \\
 & & 27.3 & 0.69 & 27.7 & 0.95 & 37.8 & 0.69 & 39.1 & 0.97 & 40.5 & 1.51 & 51.6 & 1.50 & 34.1 & 49.08 \\
\cline{3-16}
\multirow{2}{*}{Ubuntu} & \multirow{2}{*}{38} & 86.8 & 70.2 & \underline{94.7} & 67.7 & 89.5 & \textbf{72.7} & 84.2 & 64.1 & 78.9 & 59.0 & 81.6 & 62.9 & \underline{94.7} & 39.2 \\
 & & 86.8 & 0.63 & 94.7 & 0.98 & 89.5 & 0.58 & 84.2 & 0.87 & 78.9 & 1.04 & 81.6 & 0.84 & 94.7 & 30.24 \\
\cline{3-16}
\multirow{2}{*}{Windows} & \multirow{2}{*}{7,524} & 22.7 & 36.3 & 25.4 & 32.8 & 28.7 & 38.8 & 31.3 & 36.3 & 33.3 & 35.0 & \underline{43.6} & \textbf{43.4} & 37.7 & 18.0 \\
 & & 27.7 & 1.54 & 30.5 & 4.55 & 36.1 & 0.99 & 38.6 & 1.94 & 36.6 & 1.95 & 49.9 & 1.47 & 51.3 & 64.58 \\
\noalign{\color{HGRuleStrong}\hrule height 0.8pt}
\textbf{Adjusted differences} & & & & & & & & & & & & & & & \\
\multirow{2}{*}{macOS vs. Windows} & \multirow{2}{*}{367} & +1.9 & +0.5 & +0.1 & -0.4 & +4.2 & +2.0 & +2.2 & +1.1 & +2.1 & +1.2 & +4.2 & +1.6 & +1.2 & +1.0 \\
 & & -0.8 & +0.04 & -0.7 & +0.03 & +1.1 & +0.02 & +2.3 & -0.03 & +5.4 & -0.07 & +1.0 & +0.07 & +3.2 & +2.16 \\
\cline{3-16}
\multirow{2}{*}{iOS vs. Windows} & \multirow{2}{*}{144} & -46.9 & -20.1 & -70.1 & -29.6 & -42.4 & -20.0 & -20.7 & -7.1 & -25.0 & -7.1 & -28.8 & -13.9 & -61.5 & -15.5 \\
 & & -37.2 & -0.05 & -62.4 & -0.02 & -3.5 & +0.41 & +21.0 & +0.75 & -14.6 & +0.60 & +0.4 & +1.13 & -23.3 & -1.32 \\
\cline{3-16}
\multirow{2}{*}{Android vs. Windows} & \multirow{2}{*}{126} & -14.0 & -6.7 & -17.4 & -7.9 & -21.5 & -9.3 & -18.6 & -7.5 & -44.3 & -15.3 & -40.8 & -15.9 & -4.2 & -0.6 \\
 & & +28.3 & -12.71 & +24.0 & -0.42 & +13.5 & -0.92 & +13.8 & -0.81 & -4.6 & -2.72 & -14.7 & -2.10 & -13.7 & -0.53 \\
\cline{3-16}
\multirow{2}{*}{Ubuntu vs. Windows} & \multirow{2}{*}{38} & +32.3 & +16.7 & +17.0 & +7.4 & +36.7 & +20.0 & +37.8 & +19.4 & +33.8 & +17.7 & +20.8 & +9.4 & +20.7 & +8.5 \\
 & & +32.3 & +0.03 & +17.0 & +0.04 & +36.7 & +0.00 & +37.8 & +0.01 & +33.8 & -0.16 & +20.8 & +0.06 & +20.7 & -9.11 \\
\noalign{\color{HGRuleStrong}\hrule height 0.8pt}
\end{tabular}
\endgroup
\end{table*}

\Cref{tab:E5_platform} compares grounding performance across operating platforms.
Because Windows comprises 82.1\% of scored tasks, unadjusted platform averages likewise confound platform effects with scenario mix.
Matching scenarios and UI languages eliminates the apparent macOS deficit and shifts its median gap relative to Windows from $-7.47$ to $+0.03$ points.
Mobile platforms diverge sharply from this desktop parity.
On Chinese Routine Control Selection tasks shared between iOS and Windows, the iOS deficit deepens to $-37.37$ points, and all 53 configurations underperform on iOS.
Android exhibits a 38.35-point median gap between Acc-S and Acc, the widest across all evaluated platforms.

\subsection{Comparison with a Frontier GUI Agent System}
\label{app:frontier_system}
We further evaluate Codex with GPT-6.1 Sol High, a frontier GUI agent system, on SS-E.
All reported Codex metrics are averaged across three runs.
Because the reported Codex latency excludes the HTTPS round trip between the evaluation host and the OpenAI service, these timings represent a relatively ideal setting.
\Cref{tab:appendix_frontier} compares Codex against the open-weight model with the highest SS-E H-GATE in each of three widely used families.

\begin{table}
    \centering
    \caption{\textbf{Frontier GUI agent system on SS-E.} Acc-S and Acc are percentages, and Time is mean latency in seconds. Codex results average three runs, and the open-weight rows repeat the H200 results from \cref{tab:hgate_main}.}
    \label{tab:appendix_frontier}
    \footnotesize
    \setlength{\tabcolsep}{4pt}
    \begin{tabular}{@{}lrrrr@{}}
        \toprule
        System & Acc-S & Acc & Time & H-GATE \\
        \midrule
        Codex with GPT-6.1 Sol High & 94.7 & 53.1 & 23.80 & 22.8 \\
        \midrule
        Qwen3.5-397B-A17B & 50.8 & 42.2 & 1.45 & 43.0 \\
        UI-Venus-2-9B & 44.7 & 33.3 & 1.15 & 41.4 \\
        UI-TARS-1.5-7B & 24.2 & 18.4 & 1.54 & 30.6 \\
        \bottomrule
    \end{tabular}
\end{table}

\paragraph{Near-saturated localization does not yield efficient grounding.}
Codex achieves 94.7\% Acc-S on SS-E and exceeds Qwen3.5-397B-A17B, the most accurate open-weight model in the main text, by 43.9 points.
Its full accuracy nevertheless stalls at 53.1\%, so 41.6 points of spatially correct responses fail the timing condition, the widest such gap in \cref{tab:appendix_frontier}.
H-GATE captures this latency penalty even on solved tasks.
Despite leading accuracy across the table, Codex records the lowest H-GATE at 22.8, roughly half that of Qwen3.5-397B-A17B and below the 33.33 utility-neutral reference in \cref{app:hgate}.
At 16.4 times the mean latency of Qwen3.5-397B-A17B, each correct response earns little beyond correctness credit, whereas each failed attempt incurs substantial latency without credit.
Because the reported latency already excludes the HTTPS round trip, deployed latency would be longer.
For this frontier agent system, the GUI grounding bottleneck therefore lies in arrival time rather than spatial localization.
Accuracy ranks Codex first whereas H-GATE ranks it last, which shows that human-referenced evaluation separates systems that accuracy alone ranks in the opposite order.
These results suggest pairing a frontier planner with a fast specialized grounding model so that reasoning capacity and response speed need not trade off at every grounding step.

\subsection{Grounding under Maximum Concurrency}
\label{app:max_concurrency}
The results above follow the single-request protocol.
We additionally evaluate SS-E on H200 under concurrent load and select for each test batch the highest per-replica concurrency admission level that each model and mode actually ran.
At that level, we retain only samples for which every request has a vLLM queue record and no queue time exceeds 100\,ms.
\Cref{tab:appendix_concurrency} presents these concurrent results alongside the corresponding single-request baselines for the same model and mode.

\begin{table*}
    \centering
    \caption{\textbf{SS-E on H200 under maximum concurrency.} Conc. denotes the per-replica concurrency admission level, and rows at level 1 repeat the single-request results. Acc-S, Acc, and Acc@Human are percentages. Time and P95 report mean and 95th-percentile latency in seconds. Each concurrent row covers the queue-filtered samples of one test batch.}
    \label{tab:appendix_concurrency}
    \footnotesize
    \setlength{\tabcolsep}{6pt}
    \begin{tabular}{@{}llrrrrrr@{}}
        \toprule
        Model & Conc. & Acc-S & Acc & Time & P95 & H-GATE & Acc@Human \\
        \midrule
        \multirow{2}{*}{UI-Venus-1.5-8B} & 1 & 37.64 & 26.30 & 0.93 & 2.89 & 38.06 & 24.21 \\
         & 17 & 50.49 & 38.80 & 2.19 & 7.13 & 39.99 & 24.93 \\
        \midrule
        \multirow{2}{*}{Qwen3.5-397B-A17B} & 1 & 50.84 & 42.22 & 1.45 & 3.52 & 43.03 & 35.89 \\
         & 128 & 45.07 & 31.42 & 4.93 & 7.70 & 25.99 & 3.66 \\
        \midrule
        \multirow{2}{*}{Qwen3.5-397B-A17B-T} & 1 & 49.27 & 32.98 & 61.17 & 194.17 & 16.78 & 1.29 \\
         & 128 & 42.92 & 0.24 & 471.88 & 1265.06 & 0.98 & 0.03 \\
        \midrule
        \multirow{2}{*}{Holo2-4B-T} & 1 & 24.95 & 18.03 & 3.32 & 5.53 & 33.46 & 16.38 \\
         & 256 & 19.91 & 15.15 & 4.03 & 2.35 & 26.79 & 9.19 \\
        \midrule
        \multirow{2}{*}{GUI-Owl-1.5-8B} & 1 & 26.29 & 18.42 & 0.62 & 1.61 & 35.79 & 17.97 \\
         & 59 & 38.26 & 29.13 & 0.69 & 1.73 & 42.14 & 28.44 \\
        \bottomrule
    \end{tabular}
\end{table*}

\paragraph{Higher latency under load leaves fewer correct responses on time.}
Each concurrent row covers only the queue-filtered samples of one test batch, so it differs from its single-request row in both load and sample composition.
Acc-S ignores response time, so a change in it signals a change in the retained samples rather than an effect of load.
It rises by 12.85 and 11.97 points for UI-Venus-1.5-8B and GUI-Owl-1.5-8B and falls by 5.04 to 6.35 points for the other three configurations.
The two configurations whose Acc-S rises also score a higher H-GATE in the concurrent row, 39.99 against 38.06 for UI-Venus-1.5-8B and 42.14 against 35.79 for GUI-Owl-1.5-8B, even though their latency does not improve.
We attribute this increase to the retained samples, not to load.
Acc, Acc@Human, and H-GATE therefore cannot be compared across the two rows as effects of load, and our analysis relies on latency and on comparisons within a row, where all metrics share the same samples.
Mean latency is higher in every concurrent row.
It reaches 1.1 times the single-request value for GUI-Owl-1.5-8B at 59 concurrent requests, 1.2 times for Holo2-4B-T at 256, 2.4 times for UI-Venus-1.5-8B at 17, 3.4 times for Qwen3.5-397B-A17B at 128, and 7.7 times for its thinking mode.
Since the queue filter caps recorded queue time at 100\,ms per request, queueing cannot account for the multi-second increases of the two Qwen3.5-397B-A17B configurations.
Within a row, the ratio of Acc@Human to Acc gives the share of correct responses delivered within the expert reference time.
This share falls from 92.1\% to 64.3\% for UI-Venus-1.5-8B, from 90.8\% to 60.7\% for Holo2-4B-T, and from 85.0\% to 11.6\% for Qwen3.5-397B-A17B, whereas it stays at 97.6\% for GUI-Owl-1.5-8B, whose mean latency changes by only 0.07 seconds.
In the thinking mode of Qwen3.5-397B-A17B, mean latency reaches 471.88 seconds and the 95th percentile exceeds 1,265 seconds, so only 0.24\% of responses remain correct although 42.92\% are spatially correct, compared with 32.98\% and 49.27\% under the single-request protocol.
Higher latency under load thus shifts correct responses past the expert reference time and, at the extreme, past their valid intervals.
Efficiency measured under the single-request protocol therefore reflects dedicated-replica performance and should not be assumed under shared serving.
Because the two rows cover different samples, the table does not quantify the H-GATE cost of concurrency.
For interactive GUI agents, deployment capacity should be sized to the latency that valid intervals tolerate rather than to aggregate throughput alone.

\subsection{Single-Point Grounding}
\label{app:single_point}
\Cref{app:category_performance} identifies multi-element answer geometry as the primary bottleneck of routine grounding, which raises the question of whether the gap to human experts stems more from output format than from grounding itself.
\Cref{tab:app-e-single-point} therefore evaluates 4,789 single-point tasks from SS-E, split into 3,422 in time-sensitive grounding and 1,367 in routine grounding.
It compares human experts against Qwen3.5-397B-A17B, the top H200 configuration in both SS-E accuracy and H-GATE, alongside two paired controls, namely Qwen3.5-27B with and without thinking and POINTS-GUI-8B under two relative budgets.

\begin{table*}
\centering
\caption{\textbf{Single-point grounding on SS-E with H200.} Results cover 4,789 single-point tasks from SS-E, with 3,422 in time-sensitive grounding and 1,367 in routine grounding. Acc-S and Acc report percentages, Time reports mean latency in seconds, -T denotes thinking, and $r$ denotes the relative visual-token budget in \cref{app:visual_budget}.}
\label{tab:app-e-single-point}
\begingroup
\footnotesize
\setlength{\tabcolsep}{2.5pt}
\renewcommand{\arraystretch}{1.09}
\begin{tabular*}{\textwidth}{@{\extracolsep{\fill}}lrrrrrrrrrrrr@{}}
\arrayrulecolor{HGRuleStrong}\toprule[0.8pt]
 & \multicolumn{4}{c}{Both modes} & \multicolumn{4}{c}{Time-sensitive grounding} & \multicolumn{4}{c}{Routine grounding} \\
\arrayrulecolor{HGRule}\cmidrule(lr){2-5}\cmidrule(lr){6-9}\cmidrule(lr){10-13}
Model & Acc-S & Acc & Time & H-GATE & Acc-S & Acc & Time & H-GATE & Acc-S & Acc & Time & H-GATE \\
\arrayrulecolor{HGRuleStrong}\midrule[0.7pt]\arrayrulecolor{HGRule}
Human & 100.0 & 99.2 & 2.8 & 66.3 & 100.0 & 99.0 & 3.4 & 66.2 & 99.9 & 99.9 & 1.3 & 66.6 \\
\midrule[0.4pt]
Qwen3.5-397B-A17B & 63.8 & 48.9 & 1.4 & 47.7 & 60.0 & 39.1 & 1.6 & 43.2 & 73.4 & 73.4 & 0.9 & 59.1 \\
\midrule[0.4pt]
Qwen3.5-27B & 53.9 & 43.7 & 1.7 & 42.9 & 49.5 & 35.3 & 1.9 & 39.6 & 64.7 & 64.7 & 1.2 & 50.9 \\
Qwen3.5-27B-T & 57.5 & 27.5 & 73.3 & 13.1 & 43.8 & 1.8 & 86.3 & 4.3 & 91.9 & 91.9 & 40.6 & 34.9 \\
\midrule[0.4pt]
POINTS-GUI-8B, $r=0.125$ & 30.2 & 17.1 & 0.4 & 37.9 & 31.5 & 13.1 & 0.4 & 36.6 & 27.1 & 27.1 & 0.4 & 41.3 \\
POINTS-GUI-8B, $r=0.75$ & 40.7 & 25.9 & 0.9 & 37.4 & 36.2 & 15.5 & 0.8 & 33.5 & 51.8 & 51.8 & 0.9 & 47.1 \\
\arrayrulecolor{HGRuleStrong}\bottomrule[0.8pt]
\end{tabular*}
\endgroup
\end{table*}

\paragraph{The gap to experts persists when a single point suffices.}
On these tasks, experts attain 99.2\% Acc whereas Qwen3.5-397B-A17B achieves 48.9\%, and its 47.7 H-GATE still trails the 66.3 expert score by 18.6 points.
The H-GATE gap across both modes is only about one fifth smaller than the full-benchmark gap of 23.2 points, so answer geometry accounts for only a minor share of the distance to experts.
The remaining deficit splits into two distinct shortfalls.
On routine grounding, the model already responds faster than experts at 0.9 versus 1.3 seconds yet solves 73.4\% of tasks versus 99.9\%, so its 7.5-point H-GATE gap stems from accuracy rather than speed.
On time-sensitive grounding, the gap triples to 23.0 points as localization and timing compound, with the model reaching 60.0\% Acc-S and only 39.1\% Acc. The 14.9-point drop from Acc-S to Acc across both modes comes entirely from time-sensitive grounding.
The SS-E efficiency gap therefore resides in grounding itself rather than in output format, and it widens most where a single point must also arrive on time.

\paragraph{Computation buys accuracy but not expert-level efficiency.}
Enabling thinking demonstrates that routine single-point errors are recoverable, as it lifts Qwen3.5-27B from 64.7\% to 91.9\% Acc, well above the 73.4\% of the far larger Qwen3.5-397B-A17B, so added computation per response recovers more of these errors than added model scale.
Yet this recovery demands 40.6 seconds per task versus 1.2, so H-GATE drops from 50.9 to 34.9.
Direct grounding thus attains expert speed without expert accuracy, thinking narrows the routine accuracy deficit to 8.0 points without expert speed, and no configuration in \cref{tab:app-e-single-point} reaches the expert regime of 99.9\% Acc at 1.3 seconds.
Under a valid interval, that same extra computation turns from costly to harmful, as thinking reduces time-sensitive Acc from 35.3\% to 1.8\% at an 86.3-second mean latency while its 43.8\% Acc-S shows that nearly all spatially correct responses miss the valid interval.
Expanding the visual budget reproduces this asymmetry at a much lower time cost.
Raising POINTS-GUI-8B's relative budget from 0.125 to 0.75 boosts routine Acc by 24.7 points and routine H-GATE from 41.3 to 47.1, whereas on time-sensitive grounding Acc rises by only 2.4 points despite a 4.7-point Acc-S gain and H-GATE declines from 36.6 to 33.5 as mean latency doubles from 0.4 to 0.8 seconds.
Across both modes, an 8.8-point Acc gain therefore accompanies a 0.5-point H-GATE decline from 37.9 to 37.4.
Additional reasoning or visual input thus yields efficiency only when accuracy returns outweigh latency costs, a condition that neither control meets on time-sensitive grounding, consistent with \cref{sec:inference_budgets}.

\subsection{Robustness to Task Subset and Scenario Weighting}
\label{app:robustness}
The main text's diagnoses rely on task-pooled scores over all SS-E tasks, with scenarios weighted by task count and answers spanning single-point and multi-element formats.
\Cref{tab:app-e-robustness} therefore repeats the underlying comparisons across four scoring choices that pair all tasks and the single-point tasks of \cref{tab:app-e-single-point} with task-pooled and scenario-balanced weighting.
It reports Acc@Human, Acc, and mean latency for the leading H200 configuration alongside the two paired controls from \cref{app:single_point}.

\begin{table*}
\centering
\caption{\textbf{Robustness of SS-E results to task subset and scenario weighting on H200.} Task-pooled scoring gives each task equal weight, and scenario-balanced scoring gives each scenario equal weight within its task set. All tasks span 36 scenarios, and the single-point tasks of Table~\ref{tab:app-e-single-point} span 25. Acc@Human and Acc are percentages, Time reports mean latency in seconds, -T denotes thinking, and $r$ denotes the relative visual-token budget in \cref{app:visual_budget}.}
\label{tab:app-e-robustness}
\begingroup
\footnotesize
\setlength{\tabcolsep}{2.5pt}
\renewcommand{\arraystretch}{1.09}
\begin{tabular*}{\textwidth}{@{\extracolsep{\fill}}lrrrrrrrrrrrr@{}}
\arrayrulecolor{HGRuleStrong}\toprule[0.8pt]
 & \multicolumn{6}{c}{All tasks} & \multicolumn{6}{c}{Single-point tasks} \\
\arrayrulecolor{HGRule}\cmidrule(lr){2-7}\cmidrule(lr){8-13}
 & \multicolumn{3}{c}{Task-pooled} & \multicolumn{3}{c}{Scenario-balanced} & \multicolumn{3}{c}{Task-pooled} & \multicolumn{3}{c}{Scenario-balanced} \\
\cmidrule(lr){2-4}\cmidrule(lr){5-7}\cmidrule(lr){8-10}\cmidrule(lr){11-13}
Model & Acc@Human & Acc & Time & Acc@Human & Acc & Time & Acc@Human & Acc & Time & Acc@Human & Acc & Time \\
\arrayrulecolor{HGRuleStrong}\midrule[0.7pt]\arrayrulecolor{HGRule}
Qwen3.5-397B-A17B & 35.9 & 42.2 & 1.5 & 36.4 & 42.5 & 1.5 & 41.3 & 48.9 & 1.4 & 38.5 & 46.6 & 1.4 \\
\midrule[0.4pt]
Qwen3.5-27B & 21.5 & 32.2 & 1.9 & 25.8 & 38.4 & 2.0 & 29.6 & 43.7 & 1.7 & 29.8 & 46.6 & 1.8 \\
Qwen3.5-27B-T & 0.1 & 29.0 & 108.9 & 0.2 & 21.4 & 79.7 & 0.2 & 27.5 & 73.3 & 0.3 & 14.6 & 57.8 \\
\midrule[0.4pt]
POINTS-GUI-8B, $r=0.125$ & 13.1 & 13.5 & 0.5 & 15.0 & 15.3 & 0.5 & 16.4 & 17.1 & 0.4 & 15.8 & 16.2 & 0.4 \\
POINTS-GUI-8B, $r=0.75$ & 20.1 & 24.1 & 1.1 & 16.9 & 21.0 & 1.0 & 21.4 & 25.9 & 0.9 & 17.3 & 22.0 & 0.8 \\
\arrayrulecolor{HGRuleStrong}\bottomrule[0.8pt]
\end{tabular*}
\endgroup
\end{table*}

\paragraph{The diagnoses hold under every scoring choice.}
Extended reasoning reduces timely correctness under each scoring choice.
Enabling thinking on Qwen3.5-27B increases mean latency from about 2 seconds to between 57.8 and 108.9 seconds and cuts Acc@Human to at most 0.3\% in every setting, from a range of 21.5\% to 29.8\% without thinking.
Thinking also lowers Acc across all four settings, by 3.2 points under task-pooled scoring over all tasks and by 16.2 to 32.0 points under the other three, so the pooled benchmark shows the smallest drop.
The visual-budget trade-off between accuracy and latency is equally stable.
Raising POINTS-GUI-8B's relative budget from 0.125 to 0.75 increases Acc by 5.7 to 10.6 points and roughly doubles mean latency in every setting, while Acc@Human rises by only 1.5 to 7.0 points, so the share of correct responses arriving within the expert reference time falls from at least 95.9\% to at most 83.4\%.
Acc@Human ranks the five configurations identically in all four settings, with Qwen3.5-397B-A17B first and the thinking configuration last.
The ranking by Acc changes under scenario-balanced weighting on single-point tasks, so Acc@Human gives the more stable ranking.
Absolute scores vary with the scoring choice.
The Acc of Qwen3.5-397B-A17B ranges from 42.2\% to 48.9\%, and under scenario-balanced weighting on single-point tasks Qwen3.5-27B ties it at 46.6\% Acc while still trailing by 8.7 points in Acc@Human.
Both diagnoses, that extended reasoning lowers timely correctness and that a larger visual budget trades latency for accuracy, therefore hold under all four scoring choices and reflect model behavior rather than the composition or weighting of SS-E.

\subsection{Failure Mechanisms on SS-E}
\label{app:failure_mechanisms}
The gap between Acc-S and Acc captures the loss of spatially correct responses due to timing.
\Cref{tab:app-e-failure-mechanisms} examines how failures arise by assigning each failed SS-E task for eight H200 configurations to one of six mutually exclusive primary mechanisms.
Premature and Overdue both require sufficient information in the observed frame and an answer that hits the target in that frame, with arrival before the valid interval opens for Premature and after it closes for Overdue.
Relocated target applies when the answer hits the target in the observed frame and the target remains actionable at a new location on arrival, invalidating the answer.
Insufficient evidence covers observed frames lacking the cues needed to identify the target or judge the task state, such as when the target has not appeared yet, has left the screen, or is occluded.
No valid output covers logs with no scorable action, including no answer, a refusal, or invalid coordinates.
Localization error applies when the observed frame provides enough information and the answer misses the target in that frame or gets the required position, path, or order wrong.
Valid then invalid separately reports the percentage of failures in which the same answer satisfied both spatial and temporal requirements in the observed frame and became invalid on arrival because the valid interval closed or the target relocated.
Mechanism classification checks whether an answer hits the target in the observed frame, whereas Acc-S checks for a target hit in at least one annotated state as described in \cref{sec:appendix_experimental_setup}, so the combined share of timing mechanisms can differ from the gap between Acc-S and Acc.

\begin{table*}
\centering
\caption{\textbf{Failure mechanisms on SS-E with H200.} Failures gives the failed-task count out of 9,160 based on the reported Acc, and the six mutually exclusive primary mechanism columns report percentages of failures summing to 100\% per row. Valid then invalid reports a separate diagnostic percentage of failures whose answer satisfied both spatial and temporal requirements in the observed frame and became invalid on arrival because the valid interval closed or the target relocated. -T denotes thinking.}
\label{tab:app-e-failure-mechanisms}
\begingroup
\footnotesize
\setlength{\tabcolsep}{2.5pt}
\renewcommand{\arraystretch}{1.09}
\begin{tabular*}{\textwidth}{@{\extracolsep{\fill}}lrrrrrrrr@{}}
\arrayrulecolor{HGRuleStrong}\toprule[0.8pt]
Model & Failures & Premature & Overdue & Relocated target & Insufficient evidence & No valid output & Localization error & Valid then invalid \\
\midrule[0.7pt]\arrayrulecolor{HGRule}
Holo2-8B & 7,246 & 8.67 & 0.24 & 0.06 & 30.88 & 2.03 & 58.12 & 0.24 \\
Holo2-8B-T & 7,677 & 4.55 & 5.20 & 4.95 & 25.07 & 0.72 & 59.51 & 7.12 \\
Holo2-235B-A22B & 5,826 & 10.42 & 1.67 & 2.08 & 27.92 & 9.86 & 48.05 & 2.65 \\
\midrule[0.4pt]
Qwen3.5-9B & 7,173 & 1.47 & 0.46 & 0.10 & 33.53 & 0.02 & 64.42 & 0.46 \\
Qwen3.5-27B & 6,211 & 1.32 & 1.12 & 0.91 & 27.88 & 1.32 & 67.45 & 1.04 \\
Qwen3.5-27B-T & 6,504 & 0.03 & 14.89 & 0.22 & 17.83 & 14.11 & 52.92 & 7.80 \\
Qwen3.5-397B-A17B & 5,295 & 5.18 & 1.69 & 1.92 & 26.27 & 2.82 & 62.12 & 1.69 \\
\midrule[0.4pt]
UI-Venus-2-9B & 6,110 & 10.82 & 0.40 & 1.80 & 22.45 & 30.66 & 33.87 & 0.40 \\
\arrayrulecolor{HGRuleStrong}\bottomrule[0.8pt]
\end{tabular*}
\endgroup
\end{table*}

\paragraph{Timing failures are mostly decided at observation.}
Under direct grounding, insufficient evidence in the observed frame and arrivals before the valid interval opens account for far more failures than answers becoming invalid during inference.
Insufficient evidence accounts for 17.83\% to 33.53\% of failures and is the second largest mechanism for seven of the eight configurations.
Together, Insufficient evidence and Premature account for 29.20\% to 39.55\% of failures across the six configurations without thinking, compared with at most 2.65\% for Valid then invalid.
Premature answers also outnumber Overdue answers in all six of these configurations, and \cref{app:explicit_waiting} evaluates explicit waiting as a remedy.

\paragraph{Extended reasoning shifts timing failures from premature to overdue answers.}
Enabling thinking reverses the ordering of premature and overdue failures.
From Holo2-8B to Holo2-8B-T, Premature answers drop from 8.67\% to 4.55\% of failures, Overdue answers climb from 0.24\% to 5.20\%, and Relocated target failures rise from 0.06\% to 4.95\%.
From Qwen3.5-27B to Qwen3.5-27B-T, Premature answers fall to 0.03\%, Overdue answers rise from 1.12\% to 14.89\%, and No valid output failures rise from 1.32\% to 14.11\%.
Valid then invalid accounts for 7.12\% and 7.80\% of failures in the two thinking configurations.
The loss after the valid interval closes is thus most pronounced in long-latency configurations, and under direct grounding early misses of the valid interval occur more frequently than late misses.

\paragraph{Spatial localization remains the largest source of failure.}
Localization error is the largest mechanism for every configuration at 33.87\% to 67.45\% of failures, so spatial localization remains a central difficulty on SS-E.
No valid output accounts for at most 2.82\% of failures for five configurations, and the 30.66\% for UI-Venus-2-9B, whose native infeasibility responses trigger continued observation, suggests that a model that keeps observing can exhaust the clip without committing to an answer.

\subsection{Explicit Waiting on Delayed Targets}
\label{app:explicit_waiting}
\Cref{app:category_performance} shows that inference latency helps or harms depending on whether the valid interval opens late or closes early, and points to explicit action-timing control as a remedy.
This subsection evaluates that remedy directly.
Under the SS-E protocol in \cref{sec:appendix_experimental_setup}, a model can issue a wait or no-action response and re-observe as playback advances.
We rerun five H200 configurations with this option disabled and compare both settings across identical tasks drawn from three groups of the full benchmark.
Specifically, the 106 post-load tasks expose visible or actionable targets only after loading, the 80 send-after-upload tasks enable valid sending once the upload finishes, and the 390 short-window tasks feature valid intervals starting at clip onset, lasting at most 3 seconds each, and closing before the clip ends.
The first two groups penalize premature responses, the third penalizes overdue responses, and Acc-S separates spatial errors from both timing failures.

\begin{table*}
\centering
\caption{\textbf{Explicit waiting on SS-E with H200.} Waiting specifies whether the wait or no-action response is enabled. The three groups cover 106 post-load tasks, 80 send-after-upload tasks, and 390 short-window tasks. Acc-S and Acc are percentages, Time is mean latency in seconds, and -T denotes thinking.}
\label{tab:app-e-waiting}
\begingroup
\footnotesize
\setlength{\tabcolsep}{2.5pt}
\renewcommand{\arraystretch}{1.09}
\begin{tabular*}{\textwidth}{@{\extracolsep{\fill}}llrrrrrrrrrrrr@{}}
\arrayrulecolor{HGRuleStrong}\toprule[0.8pt]
 & & \multicolumn{4}{c}{Post-load} & \multicolumn{4}{c}{Send after upload} & \multicolumn{4}{c}{Short window} \\
\arrayrulecolor{HGRule}\cmidrule(lr){3-6}\cmidrule(lr){7-10}\cmidrule(lr){11-14}
Model & Waiting & Acc-S & Acc & Time & H-GATE & Acc-S & Acc & Time & H-GATE & Acc-S & Acc & Time & H-GATE \\
\arrayrulecolor{HGRuleStrong}\midrule[0.7pt]\arrayrulecolor{HGRule}
\multirow{2}{*}{Holo2-8B} & On & 9.4 & 2.8 & 0.8 & 31.2 & 100.0 & 0.0 & 0.6 & 29.8 & 81.0 & 76.9 & 0.6 & 57.1 \\
 & Off & 9.4 & 0.9 & 0.8 & 30.9 & 100.0 & 0.0 & 0.6 & 29.8 & 81.3 & 76.9 & 0.6 & 57.1 \\
\midrule[0.4pt]
\multirow{2}{*}{Holo2-8B-T} & On & 44.3 & 37.7 & 4.5 & 38.9 & 100.0 & 10.0 & 2.8 & 27.6 & 67.2 & 12.6 & 2.9 & 13.4 \\
 & Off & 39.6 & 19.8 & 4.2 & 33.4 & 100.0 & 3.8 & 2.7 & 25.8 & 66.7 & 12.6 & 2.7 & 13.9 \\
\midrule[0.4pt]
\multirow{2}{*}{UI-Venus-2-9B} & On & 80.2 & 20.8 & 2.5 & 35.7 & 100.0 & 48.8 & 3.5 & 46.3 & 78.7 & 75.1 & 0.5 & 58.5 \\
 & Off & 75.5 & 7.5 & 2.5 & 28.7 & 100.0 & 0.0 & 0.5 & 30.2 & 77.4 & 74.4 & 0.5 & 58.1 \\
\midrule[0.4pt]
\multirow{2}{*}{Qwen3.5-9B} & On & 15.1 & 1.9 & 0.8 & 30.9 & 100.0 & 0.0 & 0.5 & 30.1 & 71.8 & 67.2 & 0.6 & 53.2 \\
 & Off & 17.0 & 0.0 & 0.8 & 30.0 & 100.0 & 0.0 & 0.5 & 30.1 & 71.8 & 67.2 & 0.6 & 53.2 \\
\midrule[0.4pt]
\multirow{2}{*}{Qwen3.5-9B-T} & On & 77.4 & 17.0 & 28.2 & 16.1 & 100.0 & 61.3 & 12.7 & 43.8 & 50.8 & 0.3 & 31.6 & 2.5 \\
 & Off & 75.5 & 4.7 & 23.8 & 11.5 & 100.0 & 43.8 & 12.7 & 39.9 & 50.8 & 0.0 & 30.9 & 2.5 \\
\arrayrulecolor{HGRuleStrong}\bottomrule[0.8pt]
\end{tabular*}
\endgroup
\end{table*}

\paragraph{Waiting turns spatially correct answers into valid ones on delayed targets.}
When waiting is enabled, post-load Acc rises across all five configurations in \cref{tab:app-e-waiting} by 1.9 to 17.9 points while Acc-S changes by at most 4.7 points, so the gain stems primarily from timing rather than localization.
Send-after-upload tasks isolate this timing effect because every configuration reaches perfect spatial accuracy on all tasks under both settings, which leaves timing as the sole source of failure.
On these tasks, explicit waiting lifts UI-Venus-2-9B from 0.0\% to 48.8\% Acc as mean latency rises from 0.5 to 3.5 seconds, while H-GATE climbs from 30.2 to 46.3.
H-GATE therefore rewards a slower response when added time turns a failure into a success, whereas the same model's rapid failures without waiting yield a score below the 33.33 utility-neutral reference of \cref{app:hgate}, so the metric does not reward speed for its own sake.

\paragraph{Waiting by observation is more efficient than waiting by reasoning.}
Thinking modes already wait incidentally, as analyzed in \cref{app:category_performance}, so even with waiting disabled Qwen3.5-9B-T reaches 43.8\% Acc on send-after-upload tasks at a 12.7-second mean latency.
Through explicit waiting, UI-Venus-2-9B, which is initialized from Qwen3.5-9B, achieves a higher 48.8\% Acc in 3.5 seconds, roughly a quarter of that latency, and attains 46.3 H-GATE versus 39.9.
Even when Qwen3.5-9B-T also waits explicitly and posts the group's highest Acc at 61.3\%, its 43.8 H-GATE still trails that of UI-Venus-2-9B because its mean latency remains 12.7 seconds.
Yet the option benefits only models that use it.
Without thinking, Holo2-8B and Qwen3.5-9B remain at 0.0\% under both settings with mean latencies of at most 0.6 seconds and identical metrics, which indicates that these configurations leave the waiting option unused on these tasks, whereas UI-Venus-2-9B, whose native infeasibility responses trigger continued observation, benefits most.
A prompt-level wait option thus already helps models that use it, while configurations that leave it unused suggest that waiting should become an explicit target of GUI grounding post-training.

\paragraph{Short-window responses are preserved.}
Across the 390 short-window tasks, enabling waiting alters Acc by at most 0.7 points and H-GATE by at most 0.5 points, while the three non-thinking configurations maintain identical mean latency and the two thinking modes slow by 0.2 and 0.7 seconds.
Offering a wait option therefore largely preserves responses when the valid interval is already open.
Thinking modes fail these tasks under both settings at 12.6\% and at most 0.3\% Acc as their mean latencies of 2.7 to 31.6 seconds approach or exceed the 3-second bound on these windows, so overdue responses require faster inference rather than waiting.
Even with waiting enabled, delayed targets remain far from solved, as UI-Venus-2-9B achieves 80.2\% Acc-S on post-load tasks yet only 20.8\% Acc, which leaves about three quarters of its spatially correct answers outside the valid interval.
Deciding when to act thus remains an open capability distinct from deciding where to act.

\paragraph{Faster inference and waiting address different timing failures.}
Elapsed latency on SS-E runs from playback start through answer parsing and includes waiting, so this duration determines whether a response arrives before, within, or after the valid interval.
The explicit-wait intervention and thinking comparison in \cref{sec:inference_budgets} show why the appropriate remedy depends on the timing failure.
For configurations without thinking, shorter inference alone cannot resolve early misses where the target has yet to appear or the valid interval has yet to open.
When the valid interval is already open and about to close, additional waiting cannot compensate for long inference latency.

\section{Extended Reasoning and Local Zooming}
\label{app:reasoning_zoom}
This section details the per-model results behind \cref{sec:inference_budgets} for SS-E on H200.
Acc-TS measures accuracy on tasks whose valid intervals constrain response timing.
Acc@Human measures the share of tasks answered correctly no later than their expert reference time.

\subsection{Extended Reasoning}
\label{app:reasoning}
\begin{table*}
\centering
\caption{\textbf{Paired reasoning configurations on SS-E with H200.} Acc-TS measures joint accuracy on tasks whose valid intervals constrain response timing and Acc@Human counts correct responses within expert time over all tasks. Accuracy uses percentages. Time and P95 are mean and 95th-percentile latency in seconds. Changes and ratios use each pair's first row. Bold marks the best H-GATE and underlining the best Acc across all rows.}
\label{tab:app-f-thinking}
\begingroup
\footnotesize
\setlength{\tabcolsep}{2.5pt}
\renewcommand{\arraystretch}{1.09}
\begin{tabular*}{\textwidth}{@{\extracolsep{\fill}}llrrrrrrrrr@{}}
\arrayrulecolor{HGRuleStrong}\toprule[0.8pt]
Model & Setting & Acc-S & Acc & Acc-TS & Acc@Human & Time & P95 & H-GATE & Time ratio & $\Delta$ H-GATE \\
\midrule[0.7pt]\arrayrulecolor{HGRule}
\multicolumn{11}{@{}l}{\hspace{2pt}\textbf{Shared weights}} \\
\multirow{2}{*}{Qwen3.5-0.8B} & Non-thinking & 17.4 & 11.7 & 11.0 & 11.1 & 1.17 & 4.07 & 29.6 & \textemdash & \textemdash \\
 & Thinking & 13.1 & 6.9 & 8.6 & 1.9 & 8.32 & 32.34 & 14.8 & \ensuremath{7.12\times} & \ensuremath{-14.79} \\
\midrule[0.4pt]
\multirow{2}{*}{Qwen3.5-2B} & Non-thinking & 23.7 & 16.1 & 11.1 & 15.3 & 0.54 & 0.90 & 35.1 & \textemdash & \textemdash \\
 & Thinking & 27.3 & 18.6 & 12.0 & 4.2 & 8.16 & 30.61 & 20.2 & \ensuremath{14.99\times} & \ensuremath{-14.94} \\
\midrule[0.4pt]
\multirow{2}{*}{Qwen3.5-4B} & Non-thinking & 25.6 & 19.0 & 12.7 & 17.3 & 1.01 & 1.42 & 35.1 & \textemdash & \textemdash \\
 & Thinking & 30.1 & 17.7 & 4.3 & 0.9 & 36.08 & 110.52 & 11.6 & \ensuremath{35.87\times} & \ensuremath{-23.42} \\
\midrule[0.4pt]
\multirow{2}{*}{Qwen3.5-9B} & Non-thinking & 28.6 & 21.7 & 14.8 & 20.1 & 0.74 & 1.37 & 36.4 & \textemdash & \textemdash \\
 & Thinking & 34.5 & 20.2 & 3.2 & 0.7 & 53.90 & 202.05 & 11.4 & \ensuremath{72.95\times} & \ensuremath{-24.96} \\
\midrule[0.4pt]
\multirow{2}{*}{Qwen3.5-27B} & Non-thinking & 38.0 & 32.2 & 35.1 & 21.5 & 1.86 & 4.92 & 35.1 & \textemdash & \textemdash \\
 & Thinking & 46.3 & 29.0 & 1.8 & 0.1 & 108.93 & 481.93 & 12.9 & \ensuremath{58.49\times} & \ensuremath{-22.15} \\
\midrule[0.4pt]
\multirow{2}{*}{Qwen3.5-35B-A3B} & Non-thinking & 31.7 & 25.7 & 22.5 & 19.4 & 1.31 & 3.30 & 35.6 & \textemdash & \textemdash \\
 & Thinking & 41.8 & 30.2 & 21.8 & 9.3 & 15.95 & 64.79 & 23.5 & \ensuremath{12.18\times} & \ensuremath{-12.18} \\
\midrule[0.4pt]
\multirow{2}{*}{Qwen3.5-122B-A10B} & Non-thinking & 33.2 & 26.0 & 18.6 & 22.7 & 1.06 & 2.26 & 36.9 & \textemdash & \textemdash \\
 & Thinking & 47.3 & 36.2 & 28.7 & 8.8 & 21.47 & 98.72 & 26.1 & \ensuremath{20.21\times} & \ensuremath{-10.79} \\
\midrule[0.4pt]
\multirow{2}{*}{Qwen3.5-397B-A17B} & Non-thinking & 50.8 & \underline{42.2} & 39.0 & 35.9 & 1.45 & 3.52 & \textbf{43.0} & \textemdash & \textemdash \\
 & Thinking & 49.3 & 33.0 & 5.6 & 1.3 & 61.17 & 194.17 & 16.8 & \ensuremath{42.13\times} & \ensuremath{-26.25} \\
\midrule[0.4pt]
\multirow{2}{*}{Holo2-4B} & Non-thinking & 24.2 & 17.5 & 14.4 & 15.9 & 2.64 & 5.51 & 33.4 & \textemdash & \textemdash \\
 & Thinking & 25.0 & 18.0 & 14.7 & 16.4 & 3.32 & 5.53 & 33.5 & \ensuremath{1.26\times} & \ensuremath{+0.01} \\
\midrule[0.4pt]
\multirow{2}{*}{Holo2-8B} & Non-thinking & 28.5 & 20.9 & 15.7 & 18.9 & 1.37 & 1.05 & 35.8 & \textemdash & \textemdash \\
 & Thinking & 26.8 & 16.2 & 13.3 & 3.4 & 13.86 & 14.48 & 20.3 & \ensuremath{10.13\times} & \ensuremath{-15.50} \\
\midrule[0.4pt]
\multirow{2}{*}{Holo2-30B-A3B} & Non-thinking & 30.8 & 23.3 & 16.8 & 19.5 & 3.89 & 5.54 & 32.5 & \textemdash & \textemdash \\
 & Thinking & 27.7 & 19.7 & 16.5 & 12.5 & 5.26 & 5.58 & 27.6 & \ensuremath{1.35\times} & \ensuremath{-4.98} \\
\midrule[0.4pt]
\multirow{2}{*}{Holo2-235B-A22B} & Non-thinking & 42.9 & 36.4 & 27.6 & 8.3 & 3.49 & 6.44 & 33.3 & \textemdash & \textemdash \\
 & Thinking & 39.5 & 33.4 & 26.7 & 5.1 & 12.27 & 18.36 & 30.2 & \ensuremath{3.52\times} & \ensuremath{-3.13} \\
\midrule[0.4pt]
\multicolumn{11}{@{}l}{\hspace{2pt}\textbf{Reasoned Action Mode}} \\
\multirow{2}{*}{ScaleCUA-3B} & Native & 19.1 & 12.7 & 12.5 & 11.4 & 0.98 & 1.89 & 29.3 & \textemdash & \textemdash \\
 & RAM & 21.9 & 15.7 & 16.1 & 2.7 & 4.29 & 9.21 & 19.4 & \ensuremath{4.37\times} & \ensuremath{-9.94} \\
\midrule[0.4pt]
\multirow{2}{*}{ScaleCUA-7B} & Native & 21.3 & 13.9 & 13.4 & 12.7 & 0.89 & 1.62 & 30.5 & \textemdash & \textemdash \\
 & RAM & 24.2 & 18.7 & 17.4 & 4.1 & 3.28 & 7.82 & 22.8 & \ensuremath{3.71\times} & \ensuremath{-7.67} \\
\midrule[0.4pt]
\multirow{2}{*}{ScaleCUA-32B} & Native & 22.4 & 15.6 & 11.2 & 11.7 & 1.61 & 3.04 & 26.7 & \textemdash & \textemdash \\
 & RAM & 25.9 & 19.9 & 16.1 & 3.4 & 4.90 & 9.60 & 19.8 & \ensuremath{3.04\times} & \ensuremath{-6.92} \\
\midrule[0.4pt]
\multicolumn{11}{@{}l}{\hspace{2pt}\textbf{Distinct checkpoints}} \\
\multirow{2}{*}{GUI-Owl-1.5-8B} & Instruct & 26.3 & 18.4 & 16.1 & 18.0 & 0.62 & 1.61 & 35.8 & \textemdash & \textemdash \\
 & Think & 26.0 & 18.2 & 15.9 & 14.6 & 1.44 & 2.78 & 29.3 & \ensuremath{2.33\times} & \ensuremath{-6.46} \\
\midrule[0.4pt]
\multirow{2}{*}{GUI-Owl-1.5-32B} & Instruct & 26.9 & 18.7 & 15.1 & 15.1 & 1.39 & 2.70 & 29.5 & \textemdash & \textemdash \\
 & Think & 24.0 & 18.0 & 14.9 & 2.6 & 4.58 & 8.32 & 19.6 & \ensuremath{3.29\times} & \ensuremath{-9.90} \\
\arrayrulecolor{HGRuleStrong}\bottomrule[0.8pt]
\end{tabular*}
\endgroup
\end{table*}

\paragraph{Reasoning improves what to select but delays when it is selected.}
\Cref{tab:app-f-thinking} pairs each thinking or reasoned-action configuration with its direct-grounding counterpart.
H-GATE declines in 16 of 17 pairs by 3.13 to 26.25 points as mean latency expands by factors of 1.26 to 72.95.
The sole exception is a 0.01-point gain for Holo2-4B, which coincides with the smallest latency increase.
Across the 15 same-weight and reasoned-action pairs, spatial accuracy increases by 2.82 points on average while Acc-TS drops by 4.65 points, because additional computation sharpens localization yet consumes the window during which targets remain valid.
Qwen3.5-122B-A10B exemplifies this trade-off, as a twentyfold latency increase delivers 10.25 points of accuracy yet forfeits 13.94 points of Acc@Human and 10.79 points of H-GATE.
Relative to their Instruct counterparts, both dedicated GUI-Owl-1.5 Think checkpoints lose accuracy alongside speed, which indicates that their reasoning post-training fails to transfer to single-step grounding.
By contrast, ScaleCUA's reasoned-action mode improves Acc-TS across all three sizes under moderate latency inflation of 3.04 to 4.37 times, which suggests that compact structured reasoning fits inside broader valid intervals whereas extended free-form reasoning does not.
As shown in \cref{app:setup_analysis}, this extra latency aids scenarios whose valid interval opens late yet severely degrades scenarios whose interval closes early.

\subsection{Local Zooming}
\label{app:zoom}
\Cref{tab:app-f-zoom} pairs each zoom-in configuration with its default counterpart.
Each zoom-in runs only two visual passes, and because the model decides per task whether to trigger zoom-in, only a subset of SS-E tasks undergoes zoom-in during evaluation.
The 87 inference configurations in the main text also include a zoom-in configuration for UI-TARS-1.5-7B that is omitted from \cref{fig:visual_budget}d and \cref{tab:app-f-zoom}.

\begin{table*}
\centering
\caption{\textbf{Paired zoom-in configurations on SS-E with H200.} Acc@Human counts correct responses within expert time over all tasks. Accuracy uses percentages. Time and P95 follow Table~\ref{tab:app-f-thinking}. Changes and ratios use each pair's first row. Bold marks the best H-GATE and underlining the best Acc across all rows.}
\label{tab:app-f-zoom}
\begingroup
\footnotesize
\setlength{\tabcolsep}{2.5pt}
\renewcommand{\arraystretch}{1.09}
\begin{tabular*}{\textwidth}{@{\extracolsep{\fill}}llrrrrrrrr@{}}
\arrayrulecolor{HGRuleStrong}\toprule[0.8pt]
Model & Setting & Acc-S & Acc & Acc@Human & Time & P95 & H-GATE & Time ratio & $\Delta$ H-GATE \\
\midrule[0.7pt]\arrayrulecolor{HGRule}
\multirow{2}{*}{GUI-Owl-1.5-2B} & Default & 21.84 & 15.12 & 14.93 & 0.47 & 0.85 & 36.51 & \textemdash & \textemdash \\
 & Zoom-in & 30.32 & 22.27 & 19.46 & 1.18 & 2.26 & 35.28 & \ensuremath{2.51\times} & \ensuremath{-1.23} \\
\midrule[0.4pt]
\multirow{2}{*}{GUI-Owl-1.5-4B} & Default & 21.79 & 14.63 & 14.34 & 0.49 & 1.10 & 34.94 & \textemdash & \textemdash \\
 & Zoom-in & 27.20 & 19.26 & 14.59 & 0.98 & 2.46 & 32.09 & \ensuremath{2.00\times} & \ensuremath{-2.85} \\
\midrule[0.4pt]
\multirow{2}{*}{GUI-Owl-1.5-8B} & Default & 26.29 & 18.42 & 17.97 & 0.62 & 1.61 & 35.79 & \textemdash & \textemdash \\
 & Zoom-in & 29.09 & 21.07 & 13.47 & 1.65 & 3.81 & 30.27 & \ensuremath{2.66\times} & \ensuremath{-5.52} \\
\midrule[0.4pt]
\multirow{2}{*}{GUI-Owl-1.5-8B-T} & Default & 26.04 & 18.23 & 14.56 & 1.44 & 2.78 & 29.33 & \textemdash & \textemdash \\
 & Zoom-in & 37.30 & 27.98 & 6.95 & 3.60 & 7.89 & 26.83 & \ensuremath{2.50\times} & \ensuremath{-2.50} \\
\midrule[0.4pt]
\multirow{2}{*}{GUI-Owl-1.5-32B} & Default & 26.95 & 18.70 & 15.13 & 1.39 & 2.70 & 29.52 & \textemdash & \textemdash \\
 & Zoom-in & 31.16 & 22.10 & 7.21 & 3.51 & 8.28 & 24.50 & \ensuremath{2.53\times} & \ensuremath{-5.02} \\
\midrule[0.4pt]
\multirow{2}{*}{GUI-Owl-1.5-32B-T} & Default & 24.01 & 18.04 & 2.58 & 4.58 & 8.32 & 19.62 & \textemdash & \textemdash \\
 & Zoom-in & 32.16 & 22.96 & 1.35 & 11.63 & 25.30 & 16.39 & \ensuremath{2.54\times} & \ensuremath{-3.23} \\
\midrule[0.4pt]
\multirow{2}{*}{UI-Venus-1.5-2B} & Default & 29.16 & 15.74 & 14.88 & 1.09 & 2.93 & 31.59 & \textemdash & \textemdash \\
 & Zoom-in & 33.93 & 22.39 & 13.95 & 2.33 & 6.86 & 29.91 & \ensuremath{2.14\times} & \ensuremath{-1.68} \\
\midrule[0.4pt]
\multirow{2}{*}{UI-Venus-1.5-8B} & Default & 37.64 & 26.30 & 24.21 & 0.93 & 2.89 & \textbf{38.06} & \textemdash & \textemdash \\
 & Zoom-in & 41.27 & 30.18 & 15.57 & 2.04 & 6.94 & 34.17 & \ensuremath{2.19\times} & \ensuremath{-3.89} \\
\midrule[0.4pt]
\multirow{2}{*}{UI-Venus-1.5-30B-A3B} & Default & 40.28 & 29.55 & 23.13 & 1.77 & 4.26 & 36.04 & \textemdash & \textemdash \\
 & Zoom-in & 42.71 & \underline{33.53} & 10.17 & 3.82 & 9.71 & 31.81 & \ensuremath{2.16\times} & \ensuremath{-4.23} \\
\arrayrulecolor{HGRuleStrong}\bottomrule[0.8pt]
\end{tabular*}
\endgroup
\end{table*}

\paragraph{Zoom-in delivers accuracy gains past the expert reference time.}
Across all nine pairs, zoom-in raises Acc by 2.65 to 9.75 points, yet inflates mean latency to 2.00 to 2.66 times the default value and lowers H-GATE in every pair by 1.23 to 5.52 points.
Mean latency therefore at least doubles even though only part of the tasks undergo zoom-in, so delegating the zoom-in decision to the model fails to contain the latency overhead.
Averaged across all nine pairs, Acc increases from 19.41\% to 24.64\% while Acc@Human falls from 15.75\% to 11.41\%, so the share of all tasks answered correctly after the expert reference time rises from 3.67\% to 13.22\%.
Only GUI-Owl-1.5-2B and GUI-Owl-1.5-4B raise Acc@Human, by 4.53 and 0.25 points.
They are the two fastest default configurations and the only ones whose mean latency with zoom-in remains near 1\,s.
Even these two configurations forfeit 1.23 and 2.85 H-GATE points, because H-GATE accounts for response time on every attempt, including failures, whereas Acc@Human checks only correct responses against the expert reference time.

\paragraph{Larger accuracy gains mitigate but never offset the latency penalty.}
Larger accuracy gains generally narrow the H-GATE drop, yet no gain offsets the added latency.
The three pairs gaining over 6.5 points of Acc lose 1.23 to 2.50 H-GATE points, whereas GUI-Owl-1.5-8B and GUI-Owl-1.5-32B post the smallest Acc gains of 2.65 and 3.40 points alongside the steepest H-GATE losses of 5.52 and 5.02 points.
Applying zoom-in to a Think checkpoint compounds both sources of delay, so GUI-Owl-1.5-32B-T reaches an 11.63\,s mean latency alongside the table's lowest Acc@Human and H-GATE at 1.35\% and 16.39.
Consequently, the highest-Acc configuration, UI-Venus-1.5-30B-A3B with zoom-in at 33.53\% Acc, trails the default UI-Venus-1.5-8B by 6.25 H-GATE points.
The latter attains the best H-GATE of 38.06 at 26.30\% Acc.
On SS-E, zoom-in therefore fits applications that prioritize final accuracy over response time, whereas human-referenced efficiency favors the default configuration across all nine pairs.

\section{Visual-Budget Reduction}
\label{app:visual_budget}
\Cref{tab:app-g-budgets} compiles the four complete visual-token budget sweeps on SS-E behind \cref{fig:visual_budget}a--c, which span 20 reduced-budget configurations.
Each reduced-budget setting of a model counts as one inference configuration.
The 87 inference configurations in the main text include 24 reduced-budget configurations, namely these 20 alongside one half-budget setting each for Jedi-7B, GroundNext-7B, ScaleCUA-32B, and Holo2-8B, which were evaluated without a complete sweep.
In each sweep, the relative budget $r$ denotes the configured fraction of the default visual-token budget, while mean input-token count serves as a proxy for visual-token volume.

\begin{table*}
\centering
\caption{\textbf{Visual-budget sweeps on SS-E with H200.} Input tokens give the mean prompt length as a proxy for visual tokens. Accuracy uses percentages. Time and P95 follow Table~\ref{tab:app-f-thinking}. Changes use $r=1$ as reference. Bold marks the best H-GATE and underlining the best Acc across all rows.}
\label{tab:app-g-budgets}
\begingroup
\footnotesize
\setlength{\tabcolsep}{2.5pt}
\renewcommand{\arraystretch}{1.09}
\begin{tabular*}{\textwidth}{@{\extracolsep{\fill}}lrrrrrrrrr@{}}
\arrayrulecolor{HGRuleStrong}\toprule[0.8pt]
Model & $r$ & Input tokens & Acc-S & Acc & Acc@Human & Time & P95 & H-GATE & $\Delta$ H-GATE \\
\midrule[0.7pt]\arrayrulecolor{HGRule}
\multirow{6}{*}{GUI-Owl-1.5-8B} & 1 & 2469 & 26.3 & 18.4 & 18.0 & 0.62 & 1.61 & 35.8 & \textemdash \\
 & 0.75 & 2010 & 24.2 & 16.3 & 15.5 & 0.75 & 1.60 & 33.3 & \ensuremath{-2.46} \\
 & 0.5 & 1601 & 21.0 & 13.4 & 12.8 & 0.68 & 1.54 & 32.3 & \ensuremath{-3.46} \\
 & 0.375 & 1415 & 19.0 & 11.4 & 10.9 & 0.65 & 1.44 & 31.4 & \ensuremath{-4.38} \\
 & 0.25 & 1221 & 16.3 & 9.4 & 9.0 & 0.61 & 1.39 & 30.7 & \ensuremath{-5.13} \\
 & 0.125 & 1025 & 12.3 & 5.7 & 5.5 & 0.58 & 1.33 & 28.7 & \ensuremath{-7.06} \\
\midrule[0.4pt]
\multirow{6}{*}{POINTS-GUI-8B} & 1 & 6267 & 31.7 & 23.2 & 20.0 & 0.99 & 1.61 & 34.2 & \textemdash \\
 & 0.75 & 4806 & 32.6 & \underline{24.1} & 20.1 & 1.08 & 2.06 & 34.3 & \ensuremath{+0.02} \\
 & 0.5 & 3354 & 30.3 & 21.9 & 20.3 & 0.84 & 1.64 & 35.4 & \ensuremath{+1.14} \\
 & 0.375 & 2641 & 28.4 & 20.2 & 19.0 & 0.76 & 1.81 & 35.3 & \ensuremath{+1.09} \\
 & 0.25 & 1932 & 25.1 & 17.3 & 16.5 & 0.63 & 1.47 & 35.1 & \ensuremath{+0.91} \\
 & 0.125 & 1148 & 21.1 & 13.5 & 13.1 & 0.53 & 1.26 & 33.9 & \ensuremath{-0.31} \\
\midrule[0.4pt]
\multirow{6}{*}{Qwen3.5-9B} & 1 & 3773 & 28.6 & 21.7 & 20.1 & 0.74 & 1.37 & \textbf{36.4} & \textemdash \\
 & 0.75 & 2803 & 27.8 & 20.8 & 18.8 & 0.85 & 1.49 & 34.8 & \ensuremath{-1.56} \\
 & 0.5 & 1919 & 26.2 & 19.3 & 18.0 & 0.72 & 1.33 & 35.4 & \ensuremath{-0.91} \\
 & 0.375 & 1479 & 25.3 & 18.3 & 17.5 & 0.65 & 1.23 & 35.7 & \ensuremath{-0.66} \\
 & 0.25 & 1056 & 24.0 & 17.0 & 16.3 & 0.59 & 1.19 & 35.6 & \ensuremath{-0.75} \\
 & 0.125 & 595 & 17.0 & 11.0 & 10.7 & 0.52 & 1.14 & 32.9 & \ensuremath{-3.43} \\
\midrule[0.4pt]
\multirow{6}{*}{UI-TARS-1.5-7B} & 1 & 9455 & 24.2 & 18.4 & 14.8 & 1.54 & 6.19 & 30.6 & \textemdash \\
 & 0.75 & 5996 & 24.1 & 18.4 & 14.7 & 1.51 & 4.11 & 29.8 & \ensuremath{-0.80} \\
 & 0.5 & 5262 & 23.2 & 18.1 & 15.8 & 1.32 & 5.33 & 31.7 & \ensuremath{+1.17} \\
 & 0.375 & 3537 & 22.2 & 15.3 & 13.8 & 0.94 & 3.05 & 32.2 & \ensuremath{+1.61} \\
 & 0.25 & 2531 & 20.4 & 12.9 & 12.1 & 0.76 & 2.73 & 32.4 & \ensuremath{+1.78} \\
 & 0.125 & 1434 & 14.7 & 7.6 & 7.1 & 0.55 & 0.91 & 31.0 & \ensuremath{+0.42} \\
\arrayrulecolor{HGRuleStrong}\bottomrule[0.8pt]
\end{tabular*}
\endgroup
\end{table*}

\paragraph{Moderate reductions help models with large default budgets.}
H-GATE improves in 8 of the 20 reduced-budget configurations, and every gain occurs in the two models with the largest default inputs, namely UI-TARS-1.5-7B at about 9,455 input tokens and POINTS-GUI-8B at about 6,267.
Specifically, UI-TARS-1.5-7B improves across every budget from one half down to one eighth and peaks at a quarter of its default budget with a 1.78-point H-GATE gain, whereas POINTS-GUI-8B improves from three quarters down to one quarter and peaks at half budget with a 1.14-point gain.
At both peaks, latency savings from shorter visual inputs outweigh the accompanying drop in accuracy.
By contrast, GUI-Owl-1.5-8B and Qwen3.5-9B already operate near 2,469 and 3,773 default tokens and lose H-GATE at every tested reduction.

\paragraph{Fewer tokens do not guarantee lower latency.}
Mean latency nevertheless rises at five reduced-budget settings.
Three of these slowdowns occur at three quarters of the default budget, where GUI-Owl-1.5-8B, POINTS-GUI-8B, and Qwen3.5-9B all run slower than at their default settings.
For GUI-Owl-1.5-8B, this overhead persists down to 37.5\% of the default budget at 5.41\% to 20.88\% higher latency, while output length and interaction rounds stay unchanged at half budget, which places the slowdown outside decoding.
At the opposite extreme, aggressive compression to one eighth of the default budget cuts accuracy by 9.69 to 12.72 points across all four sweeps.
Effective visual-token budget selection therefore requires empirical latency and accuracy profiling, because neither token count nor default resolution predicts where compression stops improving efficiency.

\paragraph{Single-point controls show that the return on computation depends on task timing.}
\Cref{sec:inference_budgets} connects efficiency to task timing through the waiting intervention, and the paired controls for thinking and visual budget in \cref{tab:app-e-single-point} and \cref{tab:app-e-robustness} show when extra computation helps or hurts on single-point tasks.
On single-point tasks, raising the relative budget of POINTS-GUI-8B from 0.125 to 0.75 lifts routine Acc from 27.1\% to 51.8\% and H-GATE from 41.3 to 47.1, whereas time-sensitive Acc rises only from 13.1\% to 15.5\% and H-GATE declines from 36.6 to 33.5.
Likewise, on single-point tasks, thinking raises routine Acc for Qwen3.5-27B from 64.7\% to 91.9\% yet lowers time-sensitive Acc from 35.3\% to 1.8\%.
Both trade-offs thus occur even when the task requires only a single point, so multi-coordinate output is not a necessary condition for them.
\Cref{tab:app-e-robustness} further demonstrates that, for the paired configurations evaluated there, thinking lowers Acc@Human and a larger visual budget raises both Acc and latency across all four settings that pair all tasks or single-point tasks with task-pooled or scenario-balanced weighting.
Routine tasks can gain efficiency from added visual input, and on time-sensitive tasks the response produced with that extra computation must still be correct and arrive within the valid interval.
These findings support choosing the computation strategy according to each task's temporal constraints rather than applying a uniform increase or reduction of computation to all tasks.

\section{Agreement of H-GATE with Human Judgments}
\label{app:human_agreement}
This section presents the human preference study referenced in \cref{sec:metric_diagnostics}, a second preference study comparing H-GATE against Acc@Human, and the timing statistics of the experts who supply H-GATE's reference times.

\subsection{Pairwise Preference Study}
\label{app:preference_study}
This subsection evaluates whether H-GATE ranks configurations in alignment with human rater preferences and complements the ranking ablations in \cref{sec:metric_diagnostics}.

\paragraph{Protocol.}
Each evaluation sample pairs two H200 configurations on the same six SS-E tasks.
With left and right positions randomized and no access to metric scores or expert reference times, each rater inspects both configurations' responses and selects the configuration they would prefer to use.
Five raters participate, three judge every sample, and the majority vote determines human preference.
We construct 120,000 candidate samples by drawing a pair uniformly from the 1,275 unordered pairs of 51 H200 configurations and drawing six of the 9,160 tasks uniformly without replacement.
We compute every metric solely on those six tasks.
From this candidate pool, we draw 200 random samples and four discordant sets in which a comparison metric orders the pair strictly opposite to H-GATE.
The five sets do not overlap.
The four comparison metrics are Acc, mean latency, H-GATE with the benchmark-mean reference, and H-GATE without failed-attempt time, whose discordant candidates span 11.2\%, 22.1\%, 1.8\%, and 10.1\% of the pool and yield 200, 200, 300, and 200 samples, respectively.
Across all sets, 1,100 samples collect 3,300 judgments at 660 per rater.

\paragraph{Measures.}
On random samples, a metric receives 1 when its ordering matches the majority vote, 0 when it opposes the vote, and 0.5 on a tie, with overall agreement defined as the mean score.
On discordant samples, we report the share whose majority vote selects the higher-H-GATE configuration and test that proportion against 50\% via two-sided exact binomial tests under Holm correction.

\begin{table*}
\centering
\caption{\textbf{Agreement between evaluation metrics and rater preference on SS-E with H200.} Agreement is the mean score across 200 random samples, expressed as a percentage. Matching the majority vote scores one, opposing it scores zero, and a tie scores one half. The Discordant samples column counts cases where the comparison metric orders the pair strictly opposite to H-GATE. Favors H-GATE is the percentage of those cases in which the majority vote selects the configuration with higher H-GATE. Both interval columns report 95\% bootstrap intervals over samples. The $p$ column reports two-sided exact binomial tests against 50\% after Holm correction. Bold marks the highest agreement.}
\label{tab:app-h-preference}
\begingroup
\footnotesize
\setlength{\tabcolsep}{2.5pt}
\renewcommand{\arraystretch}{1.09}
\begin{tabular*}{\textwidth}{@{\extracolsep{\fill}}lrrrrrr@{}}
\arrayrulecolor{HGRuleStrong}\toprule[0.8pt]
Metric & Agreement (\%) & 95\% CI & Discordant samples & Favors H-GATE (\%) & 95\% CI & $p$ \\
\midrule[0.7pt]\arrayrulecolor{HGRule}
Acc & 60.75 & [55.3, 66.0] & 200 & 66.5 & [60.0, 73.0] & \ensuremath{7.1\times10^{-6}} \\
Mean latency & 77.00 & [71.0, 82.5] & 200 & 69.0 & [62.5, 75.5] & \ensuremath{3.2\times10^{-7}} \\
H-GATE with benchmark-mean reference & 82.50 & [77.0, 87.5] & 300 & 64.0 & [58.7, 69.3] & \ensuremath{4.3\times10^{-6}} \\
H-GATE without failed-attempt time & 75.25 & [69.5, 80.8] & 200 & 65.0 & [58.5, 71.5] & \ensuremath{2.7\times10^{-5}} \\
\midrule[0.4pt]
H-GATE & \textbf{84.00} & [79.0, 89.0] & \textemdash & \textemdash & \textemdash & \textemdash \\
\arrayrulecolor{HGRuleStrong}\bottomrule[0.8pt]
\end{tabular*}
\endgroup
\end{table*}

\paragraph{H-GATE aligns with rater preference more closely than accuracy or latency.}
On the random samples in \cref{tab:app-h-preference}, H-GATE matches the majority vote in 84.00\% of cases, compared with 77.00\% for mean latency and 60.75\% for Acc.
Acc ties on 71 of the 200 samples, yet the majority follows the H-GATE ordering in 64 of those ties, which suggests that raters distinguish equally accurate configurations by response speed.
Whenever H-GATE and a single metric disagree strictly, the majority sides with H-GATE in 26 of 34 cases against Acc and in 27 of 40 against mean latency, where two-sided exact sign tests yield $p=0.0029$ and $p=0.0385$.
Mean latency alone already agrees with most majority votes on these samples, and H-GATE adds 7.00 percentage points by pairing speed with correctness.

\paragraph{Raters favor H-GATE when metrics disagree.}
On discordant samples, the majority selects the higher-H-GATE configuration in 66.5\% of pairs against Acc and 69.0\% against mean latency.
Both design components analyzed in \cref{sec:metric_diagnostics} receive parallel support, as the majority prefers H-GATE in 64.0\% of 300 samples against the benchmark-mean reference and in 65.0\% of 200 samples against the variant without failed-attempt time.
All four proportions exceed 50\% significantly after Holm correction.
On random samples, the benchmark-mean variant reverses only 3 of 200 orderings, each decided in favor of H-GATE, so the discordant set directly separates per-task from benchmark-mean references.
The variant without failed-attempt time ties on 21 random samples and attains 75.25\% agreement.
Inter-rater agreement reaches 92.3\% across pairwise comparisons, with a 95\% interval of 89.3\% to 95.0\% and a Krippendorff's $\alpha$ of 0.850.
Within the preferences of these five raters, H-GATE and both of its design components therefore align with human judgment more closely than accuracy, mean latency, or either ablated variant.

\subsection{Preference Study against Acc@Human}
\label{app:preference_acchuman}
Acc@Human, reported in \cref{sec:app_full_gpu_results}, is the metric closest to H-GATE because it likewise combines correctness with the expert reference time.
The two summaries nevertheless rank 129 of the 1,378 H200 configuration pairs oppositely on SS-E, as reported in \cref{tab:app_hgate_rank}.
This subsection examines which ordering raters prefer across such pairs and which underlying trade-offs drive their choices.

\paragraph{Protocol.}
We select 48 H200 configuration pairs, namely 24 discordant pairs that H-GATE and Acc@Human rank oppositely on SS-E and 24 concordant pairs that both rank identically.
The discordant pairs split into three groups of eight by the primary source of disagreement.
On-time speed covers pairs differing in how quickly correct responses arrive within the expert reference time, late success covers pairs where one configuration delivers correct responses after the expert reference time, and failure time covers pairs differing in latency spent on failed attempts.
For each pair, we draw 100 of the 9,160 SS-E tasks uniformly at random and display both configurations on identical tasks.
Each sample presents the responses of the two configurations on one task.
We use the same interface as in \cref{app:preference_study} and randomize the left and right positions. Raters see neither metric scores nor expert reference times and select the configuration they would prefer to use.
Five raters each evaluate all 4,800 samples to yield 24,000 votes, and the majority of five votes determines each sample's outcome.
We report the percentage of samples whose majority vote chooses the configuration ranked higher on SS-E by each metric.

\paragraph{Pair selection.}
We index model rows in \cref{tab:d_h200_screenspot} and \cref{tab:d_h200_screenspot_part2} starting from the first configuration below the Human row and denote these two tables as D1 and D2.
The on-time speed group pairs D2 rows 1 to 3 with D1 row 24, D2 row 1 with D2 rows 3 and 9, D2 row 2 with D1 row 25, and D2 row 13 with D2 rows 24 and 25.
The late success group pairs D1 rows 22 and 23 with D1 rows 2 and 6, D1 row 22 with D1 row 20, D1 row 23 with D1 row 21, D2 row 20 with D2 row 18, and D2 row 22 with D2 row 8.
The failure time group pairs D1 row 8 with D1 row 14 and with D2 rows 6 and 10, D1 row 12 with D2 row 10, D1 rows 22 and 23 with D2 row 20, and D1 rows 4 and 5 with D1 row 19.
The 24 concordant pairs include 17 comparisons between the two inference modes of one model, pairing consecutive rows in order within D1 rows 9 to 14, D1 rows 16 to 23, D2 rows 3 to 6, and D2 rows 7 to 22, alongside seven within-family model-size comparisons using D1 rows 2 and 3, 4 and 5, 6 and 8, 7 and 8, 25 and 26, and 26 and 27, plus D2 rows 24 and 25.

\begin{table}
\centering
\caption{\textbf{Rater choices on configuration pairs ranked by H-GATE and Acc@Human.} Entries report the percentage of samples where the majority of five raters selects the configuration that the indicated metric ranks higher on SS-E. Both metrics rank the same configuration higher on concordant pairs. We group discordant pairs by the primary source of disagreement.}
\label{tab:app-h-acchuman}
\begingroup
\footnotesize
\setlength{\tabcolsep}{2.5pt}
\renewcommand{\arraystretch}{1.09}
\begin{tabular*}{\linewidth}{@{\extracolsep{\fill}}lrrr@{}}
\arrayrulecolor{HGRuleStrong}\toprule[0.8pt]
Pair group & Pairs & H-GATE & Acc@Human \\
\midrule[0.7pt]\arrayrulecolor{HGRule}
Concordant & 24 & \multicolumn{2}{c}{96.7} \\
\midrule[0.4pt]
Discordant & 24 & 92.5 & 7.5 \\
\quad On-time speed & 8 & 98.5 & 1.5 \\
\quad Late success & 8 & 91.5 & 8.5 \\
\quad Failure time & 8 & 87.4 & 12.6 \\
\arrayrulecolor{HGRuleStrong}\bottomrule[0.8pt]
\end{tabular*}
\endgroup
\end{table}

\paragraph{Raters follow H-GATE where it departs from Acc@Human.}
Across the concordant pairs in \cref{tab:app-h-acchuman}, the majority vote selects the configuration ranked higher by both metrics in 96.7\% of samples.
Across the discordant pairs, the majority chooses the higher-H-GATE configuration in 92.5\% of samples versus only 7.5\% for Acc@Human, and H-GATE's preferred configuration wins the majority of samples in every one of the 24 discordant pairs.
The two groups are not matched, as the median H-GATE gap is 7.3 points on concordant pairs versus 2.6 on discordant pairs and 17 concordant pairs each contrast two modes of the same model, yet agreement with H-GATE across these narrower discordant margins declines by only 4.2 percentage points.
H-GATE margins of a few points therefore track differences that raters perceive consistently, whereas Acc@Human orders the same pairs opposite to rater choice.

\paragraph{Each departure of H-GATE from Acc@Human matches a distinction that raters draw.}
Acc@Human awards full credit to every correct response within the expert reference time and zero credit to late successes or failed attempts, so it cannot separate configurations differing solely within those regimes.
The three discordant groups isolate the three ways H-GATE departs from this step function, because the score for a correct response falls continuously with latency, late correct responses retain partial credit, and the score for a failed attempt falls as its duration increases.
Raters favor H-GATE across all three groups, in 98.5\% of samples for on-time speed, 91.5\% for late success, and 87.4\% for failure time.
The failure time group compares configurations with similar on-time accuracy within each pair, with one keeping the user waiting longer before failing, and its result extends to Acc@Human the finding of \cref{app:preference_study} that raters prefer H-GATE over its variant without failed-attempt time.
Agreement decreases from on-time successes to late successes to failures, which suggests that raters weigh latency most consistently when a response is useful and that some regard two failed responses as equally unusable regardless of duration.
Because the selected pairs concentrate on few models, with 8 of the 24 discordant pairs involving one of the two Holo2-235B-A22B configurations and 6 of the 8 on-time speed pairs involving a GUI-Owl-1.5 model, these group percentages describe the selected pairs rather than all 129.
Within the preferences of these five raters, the continuous treatment of time separating H-GATE from Acc@Human therefore tracks choices that raters make rather than a property of the formula alone.

\paragraph{H-GATE adds per-task timing information that affects practical model selection.}
H-GATE adds per-task timing information that can change which configuration is selected by distinguishing faster and slower correct responses within the expert reference time, retaining partial credit for correct responses that exceed that time yet still satisfy the task's success condition, and charging the time spent on failed attempts.
On SS-E, the H-GATE and Acc@Human rankings have a Spearman correlation of $\rho=0.940$ and order 129 of the 1,378 configuration pairs oppositely, as shown in \cref{tab:app_hgate_rank}.
We therefore report both metrics, with H-GATE describing the continuous trade-off between correctness and response time and Acc@Human reporting the share of tasks answered correctly within the expert reference time.

\subsection{Expert Operation Times}
\label{app:expert_times}
\begin{figure}
  \centering
  \includegraphics[width=\linewidth]{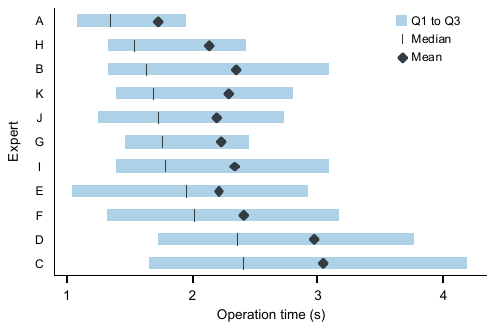}
  \caption{\textbf{Operation times of the 11 human experts.} Bars span the first to third quartiles of each expert's operation times, ticks denote medians, and diamonds denote means, with experts ordered by median.}
  \label{fig:app-h-expert-times}
\end{figure}

H-GATE references every task to the mean operation time of 11 human experts.
\Cref{fig:app-h-expert-times} summarizes the mean, median, and interquartile range of operation times for each expert, anonymized as A to K.

\paragraph{Role of the expert reference.}
\Cref{sec:human_timing} explains why the expert reference is demanding and how it guides comparisons across model configurations.
This subsection explains how the expert reference time relates to model latency and describes the expert timing protocol.
As an efficiency target, the expert reference time defines how rapidly an agent must complete a grounding step to replace a proficient human at that step.
The favorable protocol described below lets this reference approach the limit of expert proficiency, which makes the target as stringent as our design allows and helps advance the field.
Its common time scale lets model efficiency be compared across tasks that differ in difficulty and response time requirements.
Because humans and agents execute a grounding step through distinct processes, their intermediate stages are neither aligned nor compared.
H-GATE relates only their end-to-end durations at the task level, including the time required to complete their respective answers on multi-point tasks.

\paragraph{Timing protocol.}
Before timing, every expert becomes thoroughly familiar with the operations that the tasks require, the problems that may arise, and the operations that the interfaces afford, which is why we describe them as highly proficient experts.
This familiarity covers the applications and the tasks themselves alike, since the tasks draw on everyday work and life scenarios familiar to the experts, and we draw no distinction between the two.
Each expert is timed while highly focused and takes a break after every 20 tasks.
Before switching to a different application, the expert refreshes their familiarity with the applications that the upcoming tasks may involve.
Across all tasks, including those recorded on touch platforms, experts respond by clicking directly on the screen with a mouse, on everyday devices of the kinds described in \cref{sec:appendix-b-coverage}.
For each task, the instruction and the clip or screenshot appear together, the clock starts at this first exposure, and it stops when the expert clicks the target.
On tasks whose answer comprises multiple points, the clock stops only after the expert clicks every required point.
Model latency on these tasks likewise runs until parsing of the full answer finishes, so both the expert reference time and model latency cover their respective complete answers.
The reference time therefore covers reading the instruction, locating the target, and executing every required pointer action, and on SS-E it also covers any wait before a valid interval opens.

\paragraph{Failed records and the Human row.}
Expert and model answers are evaluated under identical grounding criteria and tolerances, including on box and multi-point tasks.
We retain incorrect or mistimed attempts as failures during annotation, although such attempts are very rare.
The time of a failed attempt remains in the average, so each task's reference time is the mean across all 11 experts.
In \cref{tab:hgate_main}, Human accuracy is computed per expert and averaged over the 11 experts, whereas Human H-GATE is computed directly from the definition in \cref{app:hgate} using each task's mean expert time.
These recorded failures explain why Human accuracy falls slightly below 100\% on some benchmarks, and spatial misses on SS-E are rare enough that Human Acc-S rounds to 100.0 after averaging.

\paragraph{Experts align more closely on fast operations than on slow ones.}
Across the 11 experts, median operation times span 1.35 to 2.41\,s, with nine experts at or below 2.01\,s and experts D and C reaching 2.36 and 2.41\,s.
The first quartile spans 1.05 to 1.72\,s, whereas the third quartile ranges from 1.95 to 4.19\,s, so experts vary considerably less at the fast end of their distributions than at the slow end.
Every expert's mean exceeds the median by 14\% to 45\%, which indicates right-skewed distributions where slow operations pull the mean upward.
The sample variance of operation times spans 0.92 to 3.01\,s$^2$ and correlates with the mean at $r=0.93$, so slower experts are also less consistent.
Mean operation time differs by a factor of 1.77 between the fastest expert at 1.72\,s and the slowest at 3.04\,s, so anchoring the reference to any single expert would propagate that individual pace into every score.
Averaging operation times across all 11 experts for each task, as H-GATE does, mitigates this dependence on individual pace.

\paragraph{H-GATE rankings are stable under alternative expert references.}
We evaluate the sensitivity of H-GATE rankings to the expert reference time while holding model predictions, correctness labels, model latencies, and the handling of failed expert records fixed.
For each task, we use the median time of the 11 experts as one alternative and construct 11 leave-one-expert-out references by removing one expert at a time and taking the mean time of the remaining 10.
Under all 12 alternative references, the ranking of all 53 H200 configurations is exactly identical to the original mean-reference ranking.
Comparisons among these configurations are therefore stable to the choice between mean and median and to the removal of any single expert.
Because this analysis yields a single clear conclusion, we report it in prose without a separate table, which conserves space and keeps the definition of H-GATE distinct from the alternative statistics.

\section{Benchmark Samples and Evaluation Cases}
\label{app:examples}

This section presents representative SS-E tasks alongside concrete model responses.
\Cref{app:samples} displays three tasks from each of the seven task categories, and \cref{app:cases} examines 33 model responses that illustrate the efficiency effects analyzed in \cref{sec:experiments}.

\subsection{Benchmark Samples}
\label{app:samples}

\begin{figure*}
  \centering
  \includegraphics[width=\linewidth]{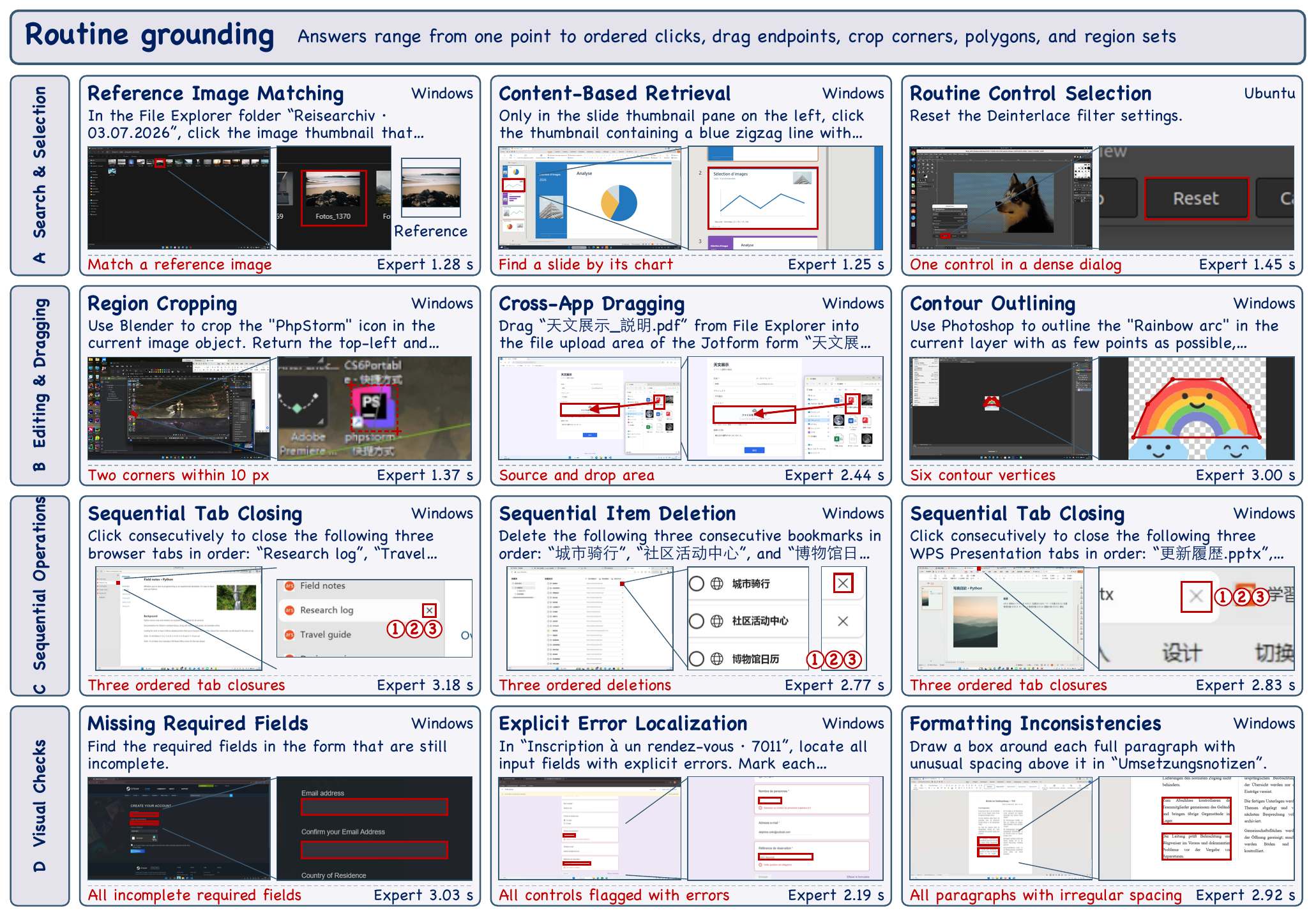}
  \caption{Routine grounding samples from SS-E across the Search and Selection, Editing and Dragging, Sequential Operations, and Visual Checks categories. Red overlays mark annotated ground-truth answers, and each card reports the mean expert completion time.}
  \label{fig:app_routine_samples}
\end{figure*}

\begin{figure*}
  \centering
  \includegraphics[width=\linewidth]{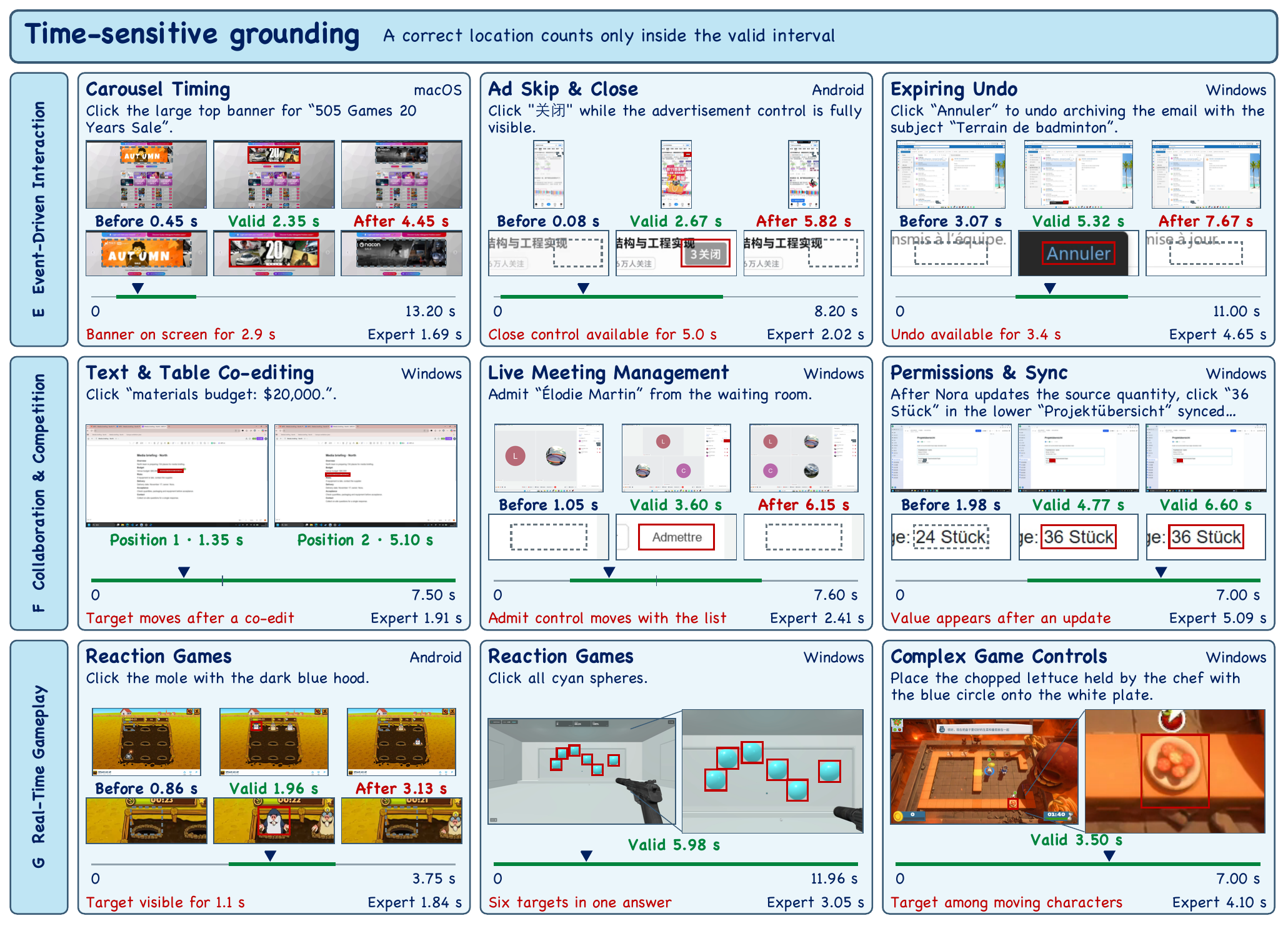}
  \caption{Time-sensitive grounding samples from SS-E across the Event-Driven Interaction, Collaboration and Competition, and Real-Time Gameplay categories. Each frame label shows the timestamp and the target state at that time, while the timeline marks the valid interval in green and the mean expert completion time with a triangle.}
  \label{fig:app_timesensitive_samples}
\end{figure*}

\paragraph{Routine grounding.}
\Cref{fig:app_routine_samples} demonstrates that routine SS-E tasks diverge primarily in required answer geometry.
Search and Selection demands a single point, yet targets reside among look-alike thumbnails or dense tool dialogs.
Editing and Dragging requires two crop corners, paired source and destination endpoints, or polygon vertices, whereas Sequential Operations requires three targets in prescribed order and Visual Checks requires every rule-violating region.
Human experts complete these twelve tasks in 1.25 to 3.18 seconds on average, yet the 53 H200 base configurations in \cref{sec:app_full_gpu_results} diverge starkly across them.
All but one configuration solve the single-control GIMP task, whereas none solve the Google Forms error-field task or the three-tab WPS Presentation sequence, and only 2 of 53 solve the Blender cropping task.
Among these samples, tasks that experts finish within about three seconds thus span the full range from nearly solved to completely unsolved, and every task solved by fewer than 10\% of configurations demands more than one answer element.

\paragraph{Time-sensitive grounding.}
\Cref{fig:app_timesensitive_samples} illustrates SS-E's second axis, namely the temporal interval during which a spatially accurate localization remains valid.
In Event-Driven Interaction, a carousel banner remains visible for 2.9 seconds, an advertisement close control stays active for 5.0 seconds, and an undo control persists for 3.4 seconds after an email is archived.
In Collaboration and Competition, external participant actions modify target state, as co-editing shifts a line of text, waiting-room updates relocate the admit control, and source edits reveal a value only afterward.
In Real-Time Gameplay, a valid target can appear for as little as 1.1 seconds.
The same instruction therefore maps to different valid targets, or to none at all, depending on response arrival time, a dynamic dependency that static screenshot benchmarks cannot represent.

\subsection{Evaluation Cases}
\label{app:cases}
The evaluation cases below sample H200 responses across nine model families.
Each card displays the frame at model completion, or the final clip frame when completion occurs after playback ends, alongside the concurrently valid ground-truth target, the model prediction, and a timeline marking the valid interval, expert reference time, and completion time.
Every card reports the per-task score $s_i$ defined in \cref{app:hgate}, where 33.33 separates failures from correct responses and 66.67 marks a correct response delivered at expert speed.

\begin{figure*}
  \centering
  \includegraphics[width=\linewidth]{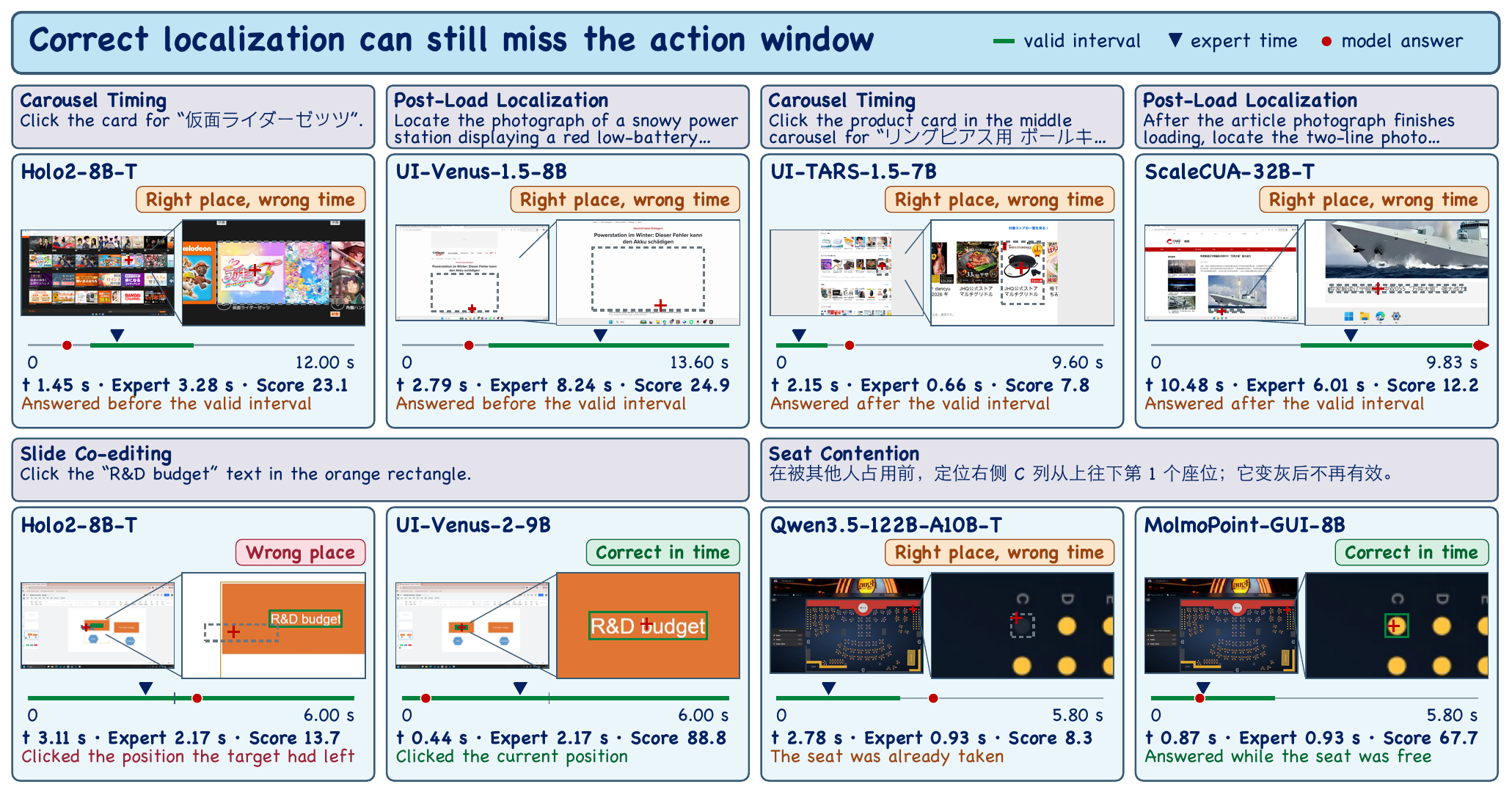}
  \caption{Spatially accurate responses that miss the valid action window and collaborative-task failures paired with timely successes on the same task. Green boxes indicate the target valid at completion time, gray dashed boxes indicate an invalid target, and red crosses mark predictions, while each timeline denotes the valid interval in green, the expert time with a triangle, and model completion with a dot.}
  \label{fig:app_timing_cases}
\end{figure*}

\paragraph{Correct localization can still miss the action window.}
In the first row of \cref{fig:app_timing_cases}, all four responses localize the correct region yet still fail.
Holo2-8B-T responds prematurely at 1.45 seconds before the carousel card becomes valid at 2.3 seconds, and UI-Venus-1.5-8B clicks at 2.79 seconds on a placeholder populated only at 3.6 seconds.
Conversely, UI-TARS-1.5-7B responds at 2.15 seconds after the card departs at 1.5 seconds, while ScaleCUA-32B-T finishes at 10.48 seconds after the valid interval closes at 9.83 seconds.
Timing failures therefore occur in both directions, and their task scores of 7.8 to 24.9 fall below the 33.33 reference because each attempt incurs latency without earning completion credit.
Both directions fail the same timing requirement of arriving within the valid interval, and \cref{app:setup_analysis} reports how often each occurs.
The second row exposes the same mechanism when another participant alters interface state.
Holo2-8B-T clicks a slide position vacated at 2.7 seconds and scores 13.7, whereas UI-Venus-2-9B solves the same task at 0.44 seconds and scores 88.8.
Similarly, Qwen3.5-122B-A10B-T selects a seat at 2.78 seconds after another user claims it at 2.2 seconds, whereas MolmoPoint-GUI-8B selects it at 0.87 seconds.
In dynamic interfaces, response latency determines which coordinate remains valid.

\begin{figure*}
  \centering
  \includegraphics[width=\linewidth]{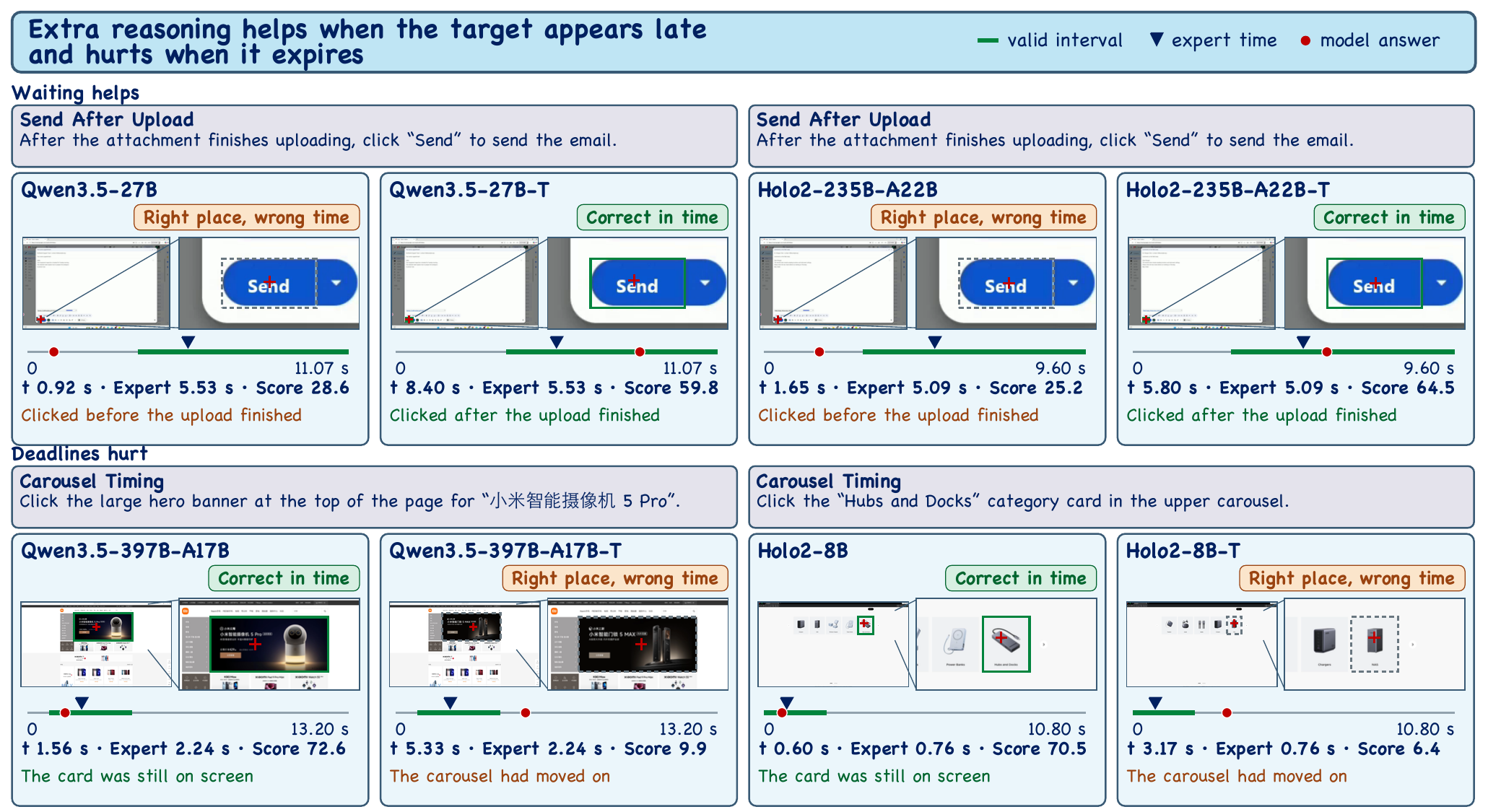}
  \caption{Same-model comparisons with and without thinking on tasks whose target appears late and on tasks whose target expires.}
  \label{fig:app_thinking_cases}
\end{figure*}

\paragraph{Extra reasoning helps when the target appears late and hurts when it expires.}
\Cref{fig:app_thinking_cases} pairs non-thinking and thinking inference from identical weights on the same task.
In Send After Upload, Qwen3.5-27B clicks the send control at 0.92 seconds before the upload finishes at 3.8 seconds and fails, whereas thinking mode answers at 8.40 seconds and succeeds.
Holo2-235B-A22B exhibits the same reversal at 1.65 and 5.80 seconds.
In Carousel Timing, this effect inverts completely.
Qwen3.5-397B-A17B responds at 1.56 seconds and scores 72.6, whereas thinking mode finishes at 5.33 seconds after the card departs at 4.3 seconds and falls to 9.9.
Holo2-8B drops from 70.5 to 6.4 for the same reason.
Because all eight responses localize the correct target, arrival time alone governs the outcome reversal, which matches the scenario-level pattern in \cref{app:setup_analysis}.

\begin{figure*}
  \centering
  \includegraphics[width=\linewidth]{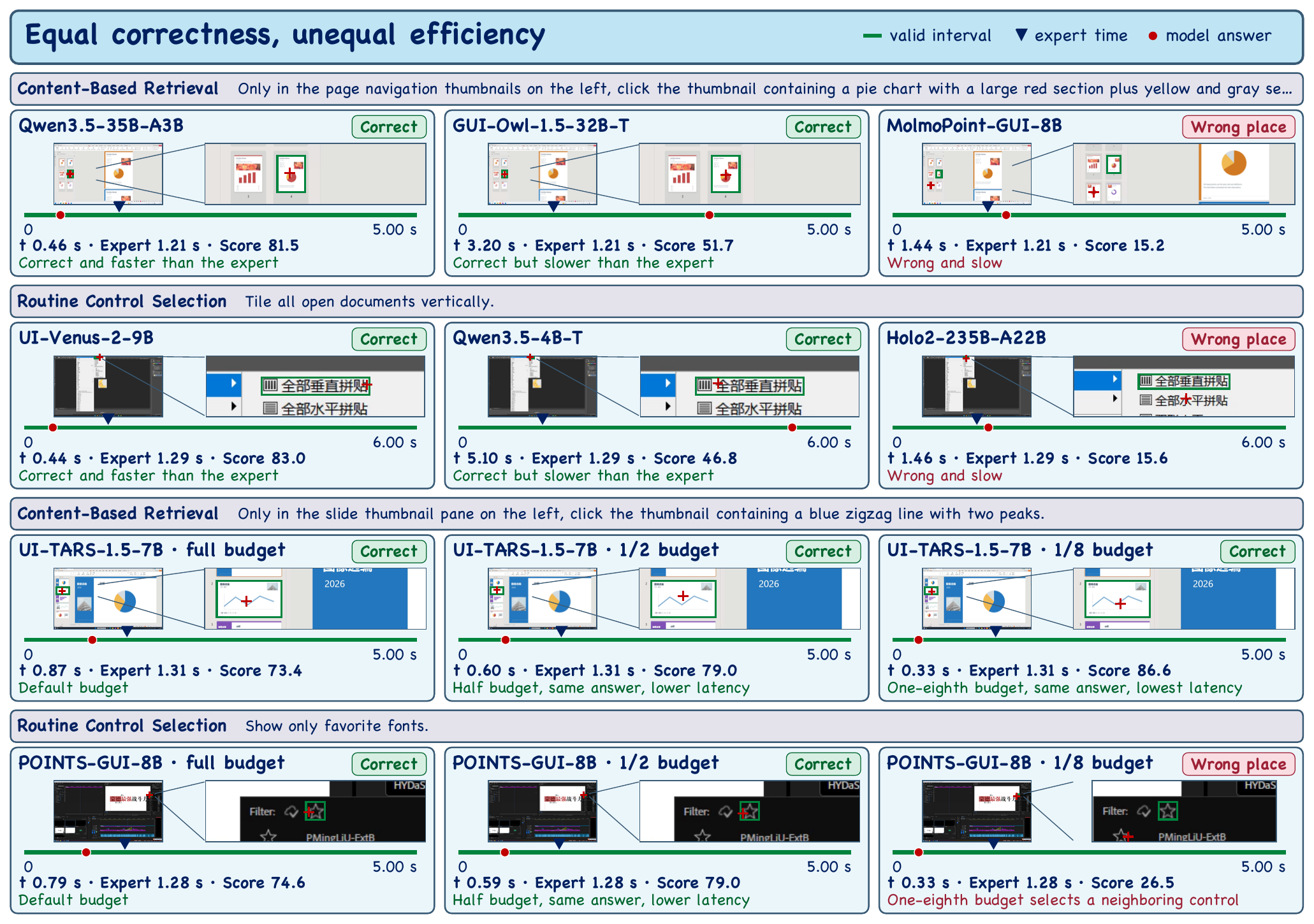}
  \caption{Matched-task responses comparing distinct models within each of the first two rows and a single model under three visual-token budgets within each of the last two rows.}
  \label{fig:app_efficiency_cases}
\end{figure*}

\paragraph{Equal correctness, unequal efficiency.}
Each of the first two rows of \cref{fig:app_efficiency_cases} contrasts two correct responses and one failure on a routine task.
On a thumbnail retrieval task completed by experts in 1.21 seconds, Qwen3.5-35B-A3B succeeds at 0.46 seconds with a score of 81.5, GUI-Owl-1.5-32B-T succeeds at 3.20 seconds with 51.7, and MolmoPoint-GUI-8B fails at 1.44 seconds with 15.2.
Because routine tasks impose no temporal deadline, both correct localizations pass.
Accuracy grants identical credit to both successes, whereas H-GATE separates them by about 30 points while ranking both above the failure.
Each of the bottom two rows tracks one model across three visual-token budgets.
UI-TARS-1.5-7B preserves correctness as latency drops from 0.87 to 0.60 and 0.33 seconds, which raises its score from 73.4 to 86.6.
POINTS-GUI-8B gains similarly at half budget, yet at one eighth of the budget it clicks a neighboring control and drops from 79.0 to 26.5.
Together, these cases illustrate at the task level how a moderate budget reduction can raise efficiency whereas aggressive compression can sacrifice correctness.

\begin{figure*}
  \centering
  \includegraphics[width=\linewidth]{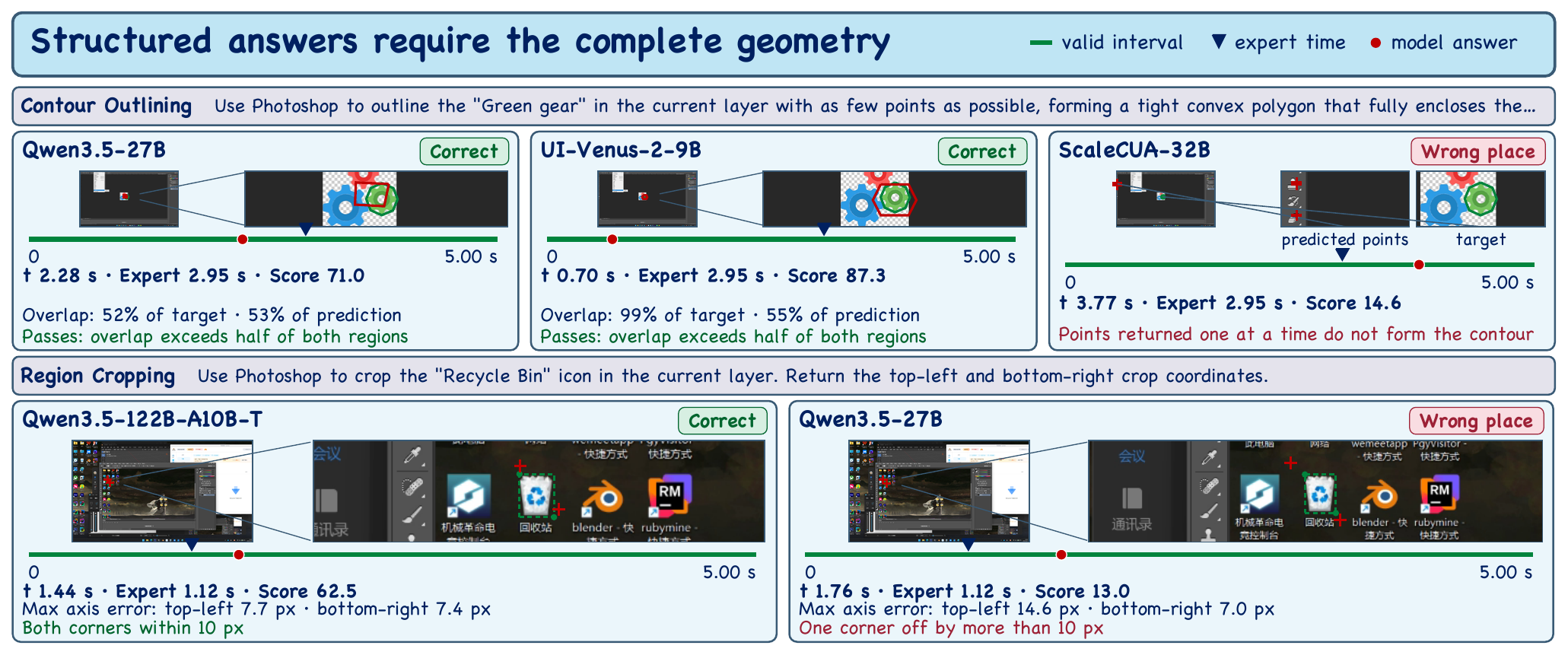}
  \caption{Model responses on contour outlining and region cropping tasks, where red polygons and crosses mark predictions and green outlines denote ground-truth annotations. Overlap ratios and corner errors are evaluated under the criteria in \cref{app:setup_analysis}.}
  \label{fig:app_structured_cases}
\end{figure*}

\paragraph{Structured answers require the complete geometry.}
In \cref{fig:app_structured_cases}, Qwen3.5-27B and UI-Venus-2-9B each emit a passing polygon on the contour task in 2.28 and 0.70 seconds, respectively.
Under the contour criterion, polygon overlap must exceed half of both the annotated and predicted regions.
UI-Venus-2-9B's hexagon covers 99\% of the annotated region, whereas Qwen3.5-27B's quadrilateral is accepted only marginally with overlap covering 52\% of the annotated region and 53\% of its own area.
ScaleCUA-32B, constrained to one point per request, outputs three points that fail to trace the object in 3.77 seconds, which reflects the interface constraint reported in \cref{app:setup_analysis}.
On the cropping task, both corners predicted by Qwen3.5-122B-A10B-T fall within 10 pixels of ground truth, whereas one corner from Qwen3.5-27B misses by about 15 pixels and invalidates the attempt.
Because cropping requires both corners to meet the threshold, a single misplaced corner forfeits the entire response.

\section{Additional Related Work}
\label{app:related_work}
This section extends \cref{sec:related_work} with additional grounding benchmarks, efficiency evaluation in complete GUI tasks, GUI grounding methods, and broader advances in language agents.

\subsection{Additional Grounding Benchmarks}
Beyond the coverage-oriented suites in Sec.~2, recent benchmarks isolate distinct failure modes on static screenshots.
GUI-HalluBench provides both parsing and grounding-hallucination annotations for each of 2,000 instances across web, desktop, and mobile interfaces, with hallucination queries that include absent targets~\cite{guihallubench}.
ScreenHaystack relocates target icons across dense screenshots while holding appearance and instruction fixed, and reports a 16.1-point accuracy gap for Qwen3-VL between targets inside and outside identified blind zones on unseen ScreenSpot-Pro examples~\cite{screenhaystack}.
On the language side, Jandial et al. evaluate 12 grounding models under alternative references to identical desktop elements~\cite{uiref}, GUI-Primitives builds 994 contrastive instruction pairs across seven spatial relations where 19 vision-language models achieve at most 32\% strict accuracy~\cite{guiprimitives}, and GUI-Perturbed reports accuracy drops of 27 to 56 points for three 7B models when instructions require spatial reasoning rather than direct naming~\cite{guiperturbed}.
Argus benchmarks 27 post-hoc uncertainty estimators for single-step grounding across four open-weight agents and four datasets~\cite{argus}.

A second line of benchmarks expands what a grounding prediction must satisfy.
FineState-Bench specifies the exact state each control must attain across 2,209 instances covering 23 component types, and separately evaluates localization, interaction, and final state achievement~\cite{finestate}.
DragOn evaluates drag grounding for text highlighting, cell selection, element resizing, and slider manipulation across a 2,000-example suite~\cite{dragon}.
MedSPOT chains two or three dependent grounding steps across 597 keyframes from 216 clinical software videos and halts scoring at the first error~\cite{medspot}.
GUI-CEval spans Chinese mobile interfaces from atomic operations to online tasks across 201 applications and four device types~\cite{guiceval}, while UI-Hawk contributes FunUI with about 32,000 annotations across four screenshot tasks~\cite{uihawk}.

Where temporal content enters prior evaluations, it serves primarily as context for comprehension.
GUI-World assesses video understanding of dynamic and sequential GUI content across six scenarios~\cite{guiworld}, and VideoGUI derives tasks from instructional videos of professional software and separately scores high-level planning, mid-level planning, and atomic actions~\cite{videogui}.
DynamicGUIBench evaluates online agents across ten application domains where substantive interface changes occur between actions~\cite{dynamicguibench}, while LivingScreen studies continuously playing short-video feeds and reports task success, tool calls per episode, and the recorded fraction of feed runtime~\cite{livingscreen}.
Outside GUIs, RefEgo grounds referring expressions across over 12,000 first-person video clips~\cite{refego}, GroundingME and RefCOCOs probe fine discrimination, absent targets, and objects spanning roughly one fiftieth of the image~\cite{groundingme,smallground}, while OVO-Bench evaluates across 644 videos whether video models respond at the right timestamp, including deferring until sufficient evidence arrives~\cite{ovobench}.
SS-E evaluates a different unit.
Each grounding response in a real screen recording must achieve spatial accuracy within its valid interval, and per-task expert completion times measure how long a human expert requires.

\subsection{Efficiency in Complete GUI Tasks}
Interactive environments first established human baselines for end-to-end task success.
OSWorld reports 72.36\% human success versus 12.24\% for its top model across 369 tasks in real computer environments~\cite{osworld}, VisualWebArena stratifies task difficulty by estimated human action counts and reports 88.7\% human success on a 230-task sample~\cite{visualwebarena}, and Windows Agent Arena records 74.5\% for humans against 19.5\% for its strongest agent~\cite{windowsagentarena}.
AndroidWorld validates 116 tasks across 20 applications via system state~\cite{androidworld}, while UI-CUBE partitions 226 enterprise tasks into two difficulty tiers where humans attain 97.9\% and 61.2\%~\cite{uicube}.
Windows Agent Arena also runs a full evaluation in parallel in as little as 20 minutes, which reflects suite turnaround rather than per-task agent time.

Subsequent benchmarks quantify the operational cost behind each success.
For mobile agents, SPA-Bench measures the ratio of agent steps to human golden steps on successful tasks alongside per-step execution time and token cost across 340 English and Chinese tasks~\cite{spabench}, AndroidLab tracks a Reversed Redundancy Ratio and a Reasonable Operation Ratio alongside success rates across 138 tasks in nine applications~\cite{androidlab}, and MobileAgentBench logs step-wise efficiency, latency, and estimated token counts across 100 tasks in 10 applications~\cite{mobileagentbench}.
Under one GPT-5 framework with a 50-step cap, MobileWorld measures mean completion lengths of 27.8 steps versus 14.3 on AndroidWorld~\cite{mobileworld,androidworld}, while the online track of GUI-CEval logs total steps, per-step latency, and token usage~\cite{guiceval}.
For web agents, Online-Mind2Web compares agent step counts against reference trajectories on successful tasks within 300 tasks across 136 websites~\cite{onlinemind2web}, while WebRetriever stratifies live web tasks by the action counts of human annotators~\cite{webretriever}.
Across broader workflows, TheAgentCompany records 30.3\% full completion, 27.2 model calls, and an estimated \$4.20 per task for Gemini 2.5 Pro~\cite{theagentcompany}, while the Holistic Agent Leaderboard derives accuracy and cost Pareto frontiers from 21,730 rollouts of nine models across nine benchmarks~\cite{hal}.

Human time serves as an explicit reference in a smaller set of studies.
ProSoftArena logs human baselines of 83.1 seconds and 20.4 steps on software-level tasks alongside 506.8 seconds and 86.9 steps on pipeline-level tasks~\cite{prosoftarena}.
OSWorld 2.0 estimates a median human operation time of roughly 1.6 hours per task from annotator-provided ranges, and Claude Opus 4.7 averages 318 tool calls~\cite{osworldtwo}.
OSExpert-Eval adopts human experts as a baseline for execution time on professional software tasks~\cite{osexpert}, and Kwa et al. characterize each agent by the typical human completion time of software tasks it solves with 50\% probability~\cite{metr}.
These suites aggregate planning, grounding, execution, and recovery, so their human references capture complete workflows, as do the efficiency comparisons in OSWorld-Human, cua-speedrun, and ComponentBench in Sec.~2.
H-GATE instead places the human reference at the individual grounding attempt, where per-task expert completion times across five benchmarks isolate localization cost from other stages.

\subsection{Additional GUI Grounding Methods}
Pixel-only interface understanding predates current GUI grounders.
Pix2Struct pretrains by parsing masked web screenshots into simplified HTML~\cite{pix2struct}, Spotlight addresses mobile UI tasks without view hierarchies~\cite{spotlight}, and Pix2Act follows instructions from pixels using generic mouse and keyboard actions~\cite{pix2act}.
For dense screens, CogAgent adds a high-resolution pathway supporting 1120$\times$1120 inputs~\cite{cogagent}, Ferret-UI partitions each screen into sub-images by aspect ratio~\cite{ferretui}, Ferret-UI 2 extends this design across iPhone, Android, iPad, web, and AppleTV interfaces~\cite{ferretui2}, and ScreenAI combines PaLI with the flexible patching of Pix2Struct~\cite{screenai}.

Later methods scale grounding supervision and unify localization with action.
UGround trains a coordinate predictor on 10 million elements from 1.3 million screenshots~\cite{uground}, and UIPro pretrains its SLiME variant on 20.6 million GUI understanding task samples~\cite{uipro}.
AutoGUI labels element functionality at scale by comparing interface states before and after simulated interactions~\cite{autogui}.
Aria-UI conditions grounding on interaction histories~\cite{ariaui}, while Aguvis trains grounding prior to planning and reasoning within a pure vision interface~\cite{aguvis}.
OmniParser instead parses interactable elements for a general-purpose model~\cite{omniparser}, and UI-TARS trains a native end-to-end agent operating solely from screenshots~\cite{uitars}.
Magma, SpiritSight, MP-GUI, and Iris further explore agentic pretraining, single-look inference, modality-specific perception, and adaptive focus across complex interfaces~\cite{magma,spiritsight,mpgui,iris}.

Reinforcement learning has since been adopted as a post-training stage.
UI-R1 applies rule-based rewards to GUI action prediction~\cite{uir1}, GUI-G$^2$ replaces binary hit-or-miss rewards with target-centered Gaussian rewards~\cite{guig2}, InfiGUI-G1 guides exploration across multiple candidate answers~\cite{infiguig1}, and LPO optimizes location preferences~\cite{lpo}.
UI-Ins identifies instruction flaws in 23.3\% of 1,909 entries sampled from existing grounding datasets and frames instruction perspectives as reasoning pathways~\cite{uiins}.
GTA1 samples multiple action proposals, employs a judge model to select one, and forwards it to a grounder trained with reinforcement learning~\cite{gta1}, and UI-AGILE adds decomposed grounding with candidate selection at inference~\cite{uiagile}.
\Cref{sec:e2_posttraining} evaluates which specialized post-training gains transfer to H-GATE.

A parallel direction targets grounding at inference time.
TAG extracts target locations from the attention maps of a pretrained model~\cite{tag}, GUI-AIMA instead learns such attention alignment through supervised fine-tuning~\cite{guiaima}, and DiMo-GUI executes training-free dynamic visual grounding via modality-aware optimization~\cite{dimogui}.
MVP aggregates predictions across multiple cropped views to mitigate coordinate instability~\cite{mvp}, DRS-GUI narrows, shifts, or expands the region passed to an existing grounder~\cite{drsgui}, and BAMI identifies precision and ambiguity biases in ScreenSpot-Pro errors and counters them via training-free coarse-to-fine focusing and candidate selection~\cite{bami}.
Several training-free methods apply zoom-in on screenshots to resolve small targets.
ZoomClick characterizes pre-zoom, depth, shrink size, and minimal crop size as four properties of zoom and, at a default depth of three iterations, raises UI-Venus-72B's accuracy to 73.1\% on ScreenSpot-Pro~\cite{zoomclick}.
UI-Zoomer samples eight candidates and triggers zoom-in only when spatial consensus and token-level confidence signal uncertainty, following its observation on ScreenSpot-v2 that unconditional zoom-in lowers accuracy below direct prediction while increasing latency~\cite{uizoomer}.
AdaGround sets magnification, field of view, and zoom depth from the model's own target-size estimate and verifies each zoom, reporting accuracy gains of 7.4 to 20.6 points over its Qwen3-VL backbones on OSWorld-G and ScreenSpot-Pro and latency 3.4 to 4.2 times that of a fixed crop~\cite{adaground}.
GUI-RC aggregates spatial consensus across sampled predictions~\cite{guirc}, and confidence-anchored test-time adaptation uses coordinate-token confidence to select reliable negatives for test-time reinforcement learning~\cite{reliableneg}.
In the reverse direction, FocusUI prunes visual tokens while maintaining positional continuity and reduces latency for its 7B variant on ScreenSpot-Pro from 1.75 to 1.22 seconds per sample as accuracy moves from 48.3\% to 45.1\%~\cite{focusui}.
STP-GUI selects spatially coherent windows of visual tokens through a lightweight learned selector and, at 21\% token retention, reports 77.7\% of the full-token task success rate on Online-Mind2Web and, in separate efficiency measurements, a 31\% reduction in inference latency and a 57\% reduction in FLOPs~\cite{stpgui}.
The zoom studies report grounding accuracy on static benchmarks, and those that report cost show that adding views, regions, or samples incurs extra forward passes and increases latency, while the token-pruning studies report latency savings at the cost of accuracy or task success.
\Cref{app:zoom} measures the trade-off against expert reference times on SS-E, where zoom-in raises Acc and lowers H-GATE in all nine pairs, and \cref{app:visual_budget} shows that reduced visual-token budgets improve H-GATE only for the two models with the largest default inputs in the budget sweeps.

Non-autoregressive decoding provides a further alternative.
Kumbhar et al. adapt the discrete diffusion model LLaDA-V to GUI grounding, introduce a hybrid masking schedule that raises Step Success Rate by up to 6.1 points over linear masking, and find on Mind2Web that jointly increasing diffusion steps, generation length, and block length raises latency while accuracy levels off~\cite{guidiffusion}.
As these three settings scale from 32 to 64 and 128, Step Success Rate shifts from 78.15\% to 80.67\% and 80.63\% while mean latency climbs from 2.56 to 4.84 and 5.01 seconds.
Motivated by the decoding efficiency of block-parallel generation, LLaDA-UI scales block-wise diffusion to a 16.7B-parameter mixture-of-experts agent that surpasses Qwen3-VL-8B on four of six reported GUI benchmarks~\cite{lladaui}.
ShowUI-$\pi$ applies flow-based generation to continuous cursor trajectories such as drags~\cite{showuip}.
As decoding paradigms diversify, H-GATE provides a common human-referenced scale for comparing their accuracy and latency with those of autoregressive grounders.

\subsection{Broader Advances in Language Agents}
\label{app:agent_frontier}
GUI grounding supports language agents whose capabilities are advancing across several fronts beyond interface perception.
Interleaving reasoning with actions enables language models to plan, invoke tools, and adapt from environment feedback~\cite{react}.
Verbal reflection and growing libraries of reusable skills enable agents to adapt across trials without weight updates~\cite{reflexion,voyager}.
Related work focuses on reusable skill extraction for small language models~\cite{skiller}.
Memory, reflection, and planning sustain believable behavior over extended interactions, and research on role-playing agents addresses persona stance and style~\cite{generativeagents,rolecraft}.
Multi-agent frameworks coordinate collaboration through specialized roles and structured workflows, and multi-agent systems also support paper-to-poster generation~\cite{metagpt,postermeld}.
Domain-specific applications include full-chain Earth observation with agents and agentic verification for multimodal misinformation detection~\cite{earthagentpro,remmd}.
As agent tasks grow longer and more interactive, the efficiency of each perception and action step matters more, and SS-E and H-GATE assess that efficiency at the GUI grounding step.

\section{Limitations}
\label{app:limitations}
This section states the scope within which the benchmark, the metric, and the findings of the main text should be read.

\paragraph{Open-loop replay.}
SS-E replays recorded clips, so an agent's answer does not change the interface, and the clips contain no state transitions caused by the target action.
The benchmark therefore measures where and when an agent would act, and it leaves closed-loop execution, error recovery, and multi-step task completion to interactive benchmarks.
Each request observes a single frame without history, so an event that is visible only across frames must be inferred from the current interface state.

\paragraph{Timing depends on the response protocol.}
As \cref{app:category_performance} shows, added latency raises accuracy when a valid interval opens late and lowers it when the interval closes early.
The gap between Acc-S and Acc thus combines premature and late responses and should not be read as a latency cost alone.
Results also follow each family's released interface, and the five families restricted to one point per request need several requests for a multi-point answer, which lengthens their latency by construction.
Measured latency is end-to-end at the client and includes frame encoding and the request round trip to the serving replica alongside model inference.

\paragraph{Benchmark composition.}
Seven experienced users identified the 36 scenario types and five collectors recorded the clips, so task counts per scenario follow their selection and were not sampled from usage statistics.
Windows accounts for 82.14\% of tasks, a single QHD display class for 78.36\%, Chinese interfaces for 44.68\%, and phones and tablets for 6.42\%, while the swipe action has three tasks.
Region Cropping and Carousel Timing together hold 35.7\% of tasks, and \cref{tab:E3_balanced} reports how scenario-balanced weighting changes the scores of seven configurations.
Pooled SS-E scores should be read with this composition in mind.

\paragraph{Human reference.}
The reference times from highly proficient experts set a deliberately demanding target for H-GATE, as stated in \cref{sec:human_timing} and detailed in \cref{app:expert_times}, with a smaller efficiency gap expected relative to typical or first-time users.
Because expert reference times and model latencies are end-to-end durations of different processes, H-GATE relates them at the task level without attributing their difference to any individual stage such as perception, reasoning, or motor execution.

\paragraph{Metric scope.}
H-GATE charges a failed attempt only its elapsed time, so an agent that fails instantly approaches the neutral reference of 33.33, as discussed in \cref{app:hgate_references}.
Scores near or below this reference must be read together with Acc and Acc@Human, and \cref{eq:app_failure_cost} lets readers add an explicit failure cost.
H-GATE also scores response time without normalizing for hardware cost.
On H200, an additional configuration serves Qwen3.5-397B-A17B across eight accelerators, whereas the other six platforms evaluate only single-accelerator models.
Comparisons between this model and single-accelerator models are therefore deployment-level comparisons and do not isolate the effect of sparse architectures.

\paragraph{Evaluation conditions.}
All main results use one concurrent request per replica, and \cref{app:max_concurrency} shows that latency and timely completion change under shared serving.
The preference study in \cref{app:preference_study} involves five raters and six tasks per sample, so it supports H-GATE within the preferences of these raters and this sample design.

\section{Ethics Statement}
\label{app:ethics}
All collectors who record and annotate SS-E, human experts who supply reference times, and raters in the preference study participated voluntarily and without compensation.
No activity conducted in this research raises ethical or moral concerns.
Every participant was informed of and consented to the public release and publication of the research results.

\end{document}